\documentclass[]{arxiv}
\usepackage{natbib}

\usepackage{fixltx2e}

\usepackage{amsmath}
\usepackage{amsfonts}
\usepackage{amssymb}

\usepackage{enumitem}

\usepackage{listings}

\usepackage{lstlangbugs}
\usepackage{caption}

\let\counterwithout\relax

\usepackage{chngcntr}

\usepackage{longtable}
\usepackage{tabulary}
\usepackage{rotating}
\usepackage{booktabs}
\usepackage{subcaption}
\usepackage{tikz}
\usetikzlibrary{bayesnet}

\usepackage{mwe}

\usepackage{listings}
\usepackage{color}
\usepackage{xcolor}
\definecolor{blue}{rgb}{0,0.3,0.7}
\definecolor{red}{rgb}{0.60,0.0,0.0}
\definecolor{purple}{rgb}{0.5,0,0.7}
\definecolor{cyan}{rgb}{0.0,0.6,0.5}
\definecolor{orange}{rgb}{0.7,0.5,0.0}
\definecolor{gray}{rgb}{0.4,0.4,0.4}

\lstdefinelanguage{scheme}
{sensitive, %
 alsoletter={:,-,+,*,?,/,!,>,<}, %
 morecomment=[l];, %
}[comments]

\lstdefinelanguage{anglican}%
{
 morekeywords=[1]{},
 morekeywords=[2]{%
   def, def-, defn, defn-, defmacro, defmulti, defmethod, %
   defstruct, defonce, declare, definline, definterface, %
   defprotocol, defrecord, defstruct, deftype, defproject, ns, %
 }, %
 morekeywords=[3]{->, ->>, .., amap, and, areduce, as->, assert, binding, %
   bound-fn, case, comment, cond, cond->, cond->>, condp, declare, definline, %
   definterface, defmacro, defmethod, defmulti, defn, defn-, defonce, %
   defprotocol, defrecord, defstruct, deftype, delay, doseq, dosync, dotimes, %
   doto, extend-protocol, extend-type, fn, for, foreach, future, gen-class, %
   gen-interface, if, if-let, if-not, if-some, import, io!, lazy-cat, lazy-seq, let, %
   letfn, locking, loop, memfn, ns, or, proxy, proxy-super, pvalues, %
   recur, refer-clojure, reify, some->, some->>, sync, time, when, when-first, %
   when-let, when-not, when-some, while, with-bindings, with-in-str, %
   with-loading-context, with-local-vars, with-open, with-out-str, %
   with-precision, with-redefs}, %
  morekeywords=[4]{*, *', +, +', -, -', ->ArrayChunk, ->Vec, ->VecNode, %
    ->VecSeq, -cache-protocol-fn, -reset-methods, /, <, <=, =, ==, >, >=, %
    accessor, aclone, add!, add-classpath, add-watch, agent, agent-error, %
    agent-errors, aget, alength, alias, all-ns, alter, alter-meta!, %
    alter-var-root, ancestors, append, apply, array-map, aset, aset-boolean, aset-byte, %
    aset-char, aset-double, aset-float, aset-int, aset-long, aset-short, assoc, %
    assoc!, assoc-in, associative?, atom, await, await-for, await1, bases, bean, %
    bigdec, bigint, biginteger, bit-and, bit-and-not, bit-clear, bit-flip, %
    bit-not, bit-or, bit-set, bit-shift-left, bit-shift-right, bit-test, %
    bit-xor, boolean, boolean-array, booleans, bound-fn*, bound?, butlast, byte, %
    byte-array, bytes, cast, char, char-array, char?, chars, chunk, %
    chunk-append, chunk-buffer, chunk-cons, chunk-first, chunk-next, chunk-rest, %
    chunked-seq?, class, class?, clear-agent-errors, clojure-version, coll?, %
    commute, comp, comparator, compare, compare-and-set!, compile, complement, %
    concat, conj, conj!, cons, constantly, construct-proxy, contains?, count, %
    counted?, create-ns, create-struct, cycle, dec, dec', decimal?, delay?, %
    deliver, denominator, deref, derive, descendants, destructure, disj, disj!, %
    dissoc, dissoc!, distinct, distinct?, doall, dorun, double, %
    double-array, doubles, drop, drop-last, drop-while, empty, empty?, ensure, %
    enumeration-seq, error-handler, error-mode, eval, even?, every-pred, every?, %
    ex-data, ex-info, extend, extenders, extends?, false?, ffirst, file-seq, %
    filter, filter-ns-publics, filterv, find, find-keyword, find-ns, %
    find-protocol-impl, find-protocol-method, find-var, first, flatten, float, %
    float-array, float?, floats, flush, fn?, fnext, fnil, force, format, %
    frequencies, future-call, future-cancel, future-cancelled?, future-done?, %
    future?, gensym, get, get-in, get-method, get-proxy-class, %
    get-thread-bindings, get-validator, group-by, hash, hash-combine, hash-map, %
    hash-ordered-coll, hash-set, hash-unordered-coll, identical?, identity, 
    ifn?, in-ns, inc, inc', init-proxy, instance?, int, int-array, integer?, %
    interleave, intern, interpose, intersection, into, into-array, ints, isa?, iterate, %
    iterator-seq, juxt, keep, keep-indexed, key, keys, keyword, keyword?, last, %
    line-seq, list, list*, list?, load, load-file, load-reader, load-string, %
    loaded-libs, long, long-array, longs, macroexpand, macroexpand-1, %
    make-array, make-hierarchy, map, map-indexed, map?, mapcat, mapv, max, %
    max-key, memoize, merge, merge-with, meta, method-sig, methods, min, %
    min-key, mix-collection-hash, mod, munge, 
    name, namespace, namespace-munge, %
    neg?, newline, next, nfirst, nil?, nnext, not, not-any?, not-empty, %
    not-every?, not=, ns-aliases, ns-functions, ns-imports, ns-interns, %
    ns-macros, ns-map, ns-name, ns-publics, ns-refers, ns-resolve, ns-unalias, %
    ns-unmap, nth, nthnext, nthrest, num, number?, numerator, object-array, %
    odd?, parents, partial, partition, partition-all, partition-by, pcalls, %
    peek, persistent!, pmap, pop, pop!, pop-thread-bindings, pos?, pr, pr-str, %
    prefer-method, prefers, prepend, print, print-ctor, print-simple, print-str, printf, %
    println, println-str, prn, prn-str, promise, proxy-call-with-super, %
    proxy-mappings, proxy-name, push-thread-bindings, quot, quote, rand, rand-int, %
    rand-nth, range, ratio?, rational?, rationalize, receive, re-find, re-groups, %
    re-matcher, re-matches, re-pattern, re-seq, read, read-line, read-string, %
    realized?, record?, reduce, reduce-kv, reduced, reduced?, reductions, ref, %
    ref-history-count, ref-max-history, ref-min-history, ref-set, refer, %
    release-pending-sends, rem, remove-all-methods, remove-method, %
    remove-ns, remove-watch, repeat, repeatedly, replace, replicate, require, %
    reset!, reset-meta!, resolve, rest, restart-agent, resultset-seq, reverse, %
    reversible?, rseq, rsubseq, satisfies?, second, select-keys, send, send-off, %
    send-via, seq, seq?, seque, sequence, sequential?, set, %
    set-agent-send-executor!, set-agent-send-off-executor!, set-error-handler!, %
    set-error-mode!, set-validator!, set?, short, short-array, shorts, shuffle, %
    shutdown-agents, slurp, some, some-fn, some?, sort, sort-by, sorted-map, %
    sorted-map-by, sorted-set, sorted-set-by, sorted?, special-symbol?, spit, %
    split-at, split-with, str, string?, struct, struct-map, subs, subseq, %
    subvec, supers, swap!, symbol, symbol?, take, take-last, take-nth, %
    take-while, test, the-ns, thread-bound?, to-array, to-array-2d, trampoline, %
    transient, tree-seq, true?, type, unchecked-add, unchecked-add-int, %
    unchecked-byte, unchecked-char, unchecked-dec, unchecked-dec-int, %
    unchecked-divide-int, unchecked-double, unchecked-float, unchecked-inc, %
    unchecked-inc-int, unchecked-int, unchecked-long, unchecked-multiply, %
    unchecked-multiply-int, unchecked-negate, unchecked-negate-int, %
    unchecked-remainder-int, unchecked-short, unchecked-subtract, %
    unchecked-subtract-int, underive, unsigned-bit-shift-right, update-in, %
    update-proxy, use, val, vals, var-get, var-set, var?, vary-meta, vec, %
    vector, vector-of, vector?, with-bindings*, with-meta, with-redefs-fn, %
    xml-seq, zero?, zipmap}, %
  morekeywords=[5]{def-cps-fn, defanglican, defm, defquery, defun}, %
  morekeywords=[6]{cps-fn, fm, lambda, mem, query, with-primitive-procedures, rep}, %
  morekeywords=[7]{%
    doquery, %
    conditional, %
    collect-by, equalize, exec, infer, log-marginal, print-predicts, %
    rand, rand-int, rand-nth, rand-roulette, stripdown, warmup, %
    ->CRP-process, ->DP-process, ->GP-process, %
    ->bernoulli-distribution, ->beta-distribution, ->binomial-distribution, %
    ->categorical-crp-distribution, ->categorical-distribution, %
    ->categorical-dp-distribution, ->chi-squared-distribution, %
    ->dirichlet-distribution, ->discrete-distribution, %
    ->exponential-distribution, ->flip-distribution, ->gamma-distribution, %
    ->mvn-distribution, ->normal-distribution, ->poisson-distribution, %
    ->uniform-continuous-distribution, ->uniform-discrete-distribution, %
    ->wishart-distribution, CRP, DP, GP, abs, absorb, acos, asin, atan, %
    bernoulli, beta, binomial, categorical, categorical-crp, categorical-dp, %
    cbrt, ceil, chi-squared, cos, cosh, cov, dirac, dirichlet, discrete, exp, %
    exponential, flip, floor, gamma, gen-matrix, log, log-gamma-fn, %
    log-mv-gamma-fn, log-sum-exp, map->CRP-process, map->DP-process, %
    map->GP-process, map->bernoulli-distribution, map->beta-distribution, %
    map->binomial-distribution, map->categorical-crp-distribution, %
    map->categorical-distribution, map->categorical-dp-distribution, %
    map->chi-squared-distribution, map->dirichlet-distribution, %
    map->discrete-distribution, map->exponential-distribution, %
    map->flip-distribution, map->gamma-distribution, map->mvn-distribution, %
    map->normal-distribution, map->poisson-distribution, %
    map->uniform-continuous-distribution, map->uniform-discrete-distribution, %
    map->wishart-distribution, mvn, diag-mvn, normal, negative-binomial, poisson, pow, produce, %
    rint, round, signum, sin, sinh, sqrt, sum, tag, tan, tanh, transform-sample, %
    uniform, uniform-continuous, uniform-discrete, wishart, %
    sample*, observe*, %
    add-log-weight, add-predict, clear-predicts, get-log-weight, %
    get-mem, get-predicts, in-mem?, set-log-weight, set-mem, %
    log-prob, dist-log-prob, dist-sample, dist-broadcast, event-size, mean, put, put!, remove, softmax, sum, unbox, multinomial, ones, zeros, shape, reshape, read-csv %
  }, %
  morekeywords=[8]{factor, propose, observe, predict, retrieve, return, sample, store, set!, immutable, mutable}, %
  sensitive, %
  alsoletter={:,-,+,*,?,/,!,>,<}, %
  morecomment=[l];, %
  morestring=[b]", %
  keywordsprefix=:, %
}[keywords,comments,strings]
 
\lstnewenvironment{anglican}[1][]{%
    \let\lstoldname\lstlistingname
    \renewcommand{\lstlistingname}{Program}
    \lstset{
        language=anglican, 
        mathescape=true,
        captionpos=b, 
        #1}
}{%
  \renewcommand{\lstlistingname}{\lstoldname}
}

\lstnewenvironment{hoppl}[1][]{%
    \let\lstoldname\lstlistingname
    \renewcommand{\lstlistingname}{Program}
    \lstset{
        language=anglican, 
        mathescape=true,
        captionpos=b, 
        #1}
}{%
  \renewcommand{\lstlistingname}{\lstoldname}
}

\lstnewenvironment{foppl}[1][]{%
    \let\lstoldname\lstlistingname
    \renewcommand{\lstlistingname}{Program}
    \lstset{
        language=anglican, 
        mathescape=true,
        captionpos=b, 
        morekeywords=[3]{map}
        #1}
}{%
  \renewcommand{\lstlistingname}{\lstoldname}
}

\lstnewenvironment{bugs}[1][]{%
    \let\lstoldname\lstlistingname
    \renewcommand{\lstlistingname}{Program}
    \lstset{
        language=STAN, 
        mathescape=true,
        captionpos=b, 
        #1}
}{%
  \renewcommand{\lstlistingname}{\lstoldname}
}

\newcounter{grammarcounter}
\lstnewenvironment{grammar}[1][]{%
    \let\lstoldname\lstlistingname
    \renewcommand{\lstlistingname}{Language}
    \setcounter{lstlisting}{\value{grammarcounter}}
    \lstset{
        language=anglican, 
        mathescape=true,
        captionpos=b, 
        numberbychapter=false,
        #1}
}{%
  \renewcommand{\lstlistingname}{\lstoldname}
   \addtocounter{grammarcounter}{1}
}

\counterwithout{grammarcounter}{chapter}

\usepackage{algorithm}

\newcommand{\argmax}{\operatornamewithlimits{argmax}}
\newcommand{\argmin}{\operatornamewithlimits{argmin}}

\newcommand{\map}[1]{\ensuremath{\mathcal{#1}}}
\newcommand{\m}[1]{\map{#1}}

\renewcommand{\l}{\ensuremath{\lambda}}

\newcommand{\KL}[2]{\ensuremath{D_{\rm KL}\left(#1 \:\middle\vert\middle\vert\: #2\right)}}

\newcommand*{\dom}{\ensuremath{\mathrm{dom}}}

\newcommand{\Normal}{\text{Normal}}

\newcommand{\Ev}{\mathbb{E}}

\newcommand{\R}{\mathbb{R}}
\newcommand{\X}{\mathcal{X}}
\newcommand{\Y}{\mathcal{Y}}

\usepackage{xspace}

\newcommand{\sample}{\mang{sample}\xspace}
\newcommand{\observe}{\mang{observe}\xspace}

\newcommand{\id}{\ensuremath{\sigma}}
\newcommand{\state}{\ensuremath{\sigma}}

\newcommand{\pa}{\ensuremath{\textsc{pa}}}

\newcommand{\ang}{\lstinline[mathescape,keepspaces,language=Anglican]}
\newcommand{\hop}{\lstinline[mathescape,keepspaces,language=Anglican]}
\newcommand{\fop}{\lstinline[mathescape,keepspaces,language=Anglican]}
\newcommand{\py}{\lstinline[mathescape,keepspaces,language=Python]}
\newcommand{\mfop}[1]{\mbox{\fop!#1!}}
\newcommand{\mhop}[1]{\mbox{\hop!#1!}}
\newcommand{\lsi}{\ang}
\newcommand{\lisp}{\lstinline[mathescape,language=Lisp]}
\newcommand{\mang}[1]{\mbox{\ang!#1!}}
\newcommand{\mlisp}[1]{\mbox{\lisp`#1`}}
\newcommand{\cps}[1]{\ensuremath{\overline{#1}}}

\newcommand{\lldots}{\hbox to 1em{.\hss.\hss.}}

\newcommand{\ad}[1]{\ensuremath{\tilde{#1}}}

\newcommand{\cmt}[2]{}

\usepackage{enumitem}
\newlist{todolist}{itemize}{2}
\setlist[todolist]{label=$\square$}
\usepackage{pifont}
\usepackage{proof}
\usepackage{algorithmicx}
\usepackage[noend]{algpseudocode}

\usepackage[]{hyperref}

\algnewcommand\algorithmicassert{\texttt{assert}}
\algnewcommand\Assert[1]{\State{} \algorithmicassert(#1)}

\algnewcommand\algorithmicswitch{\textbf{switch}}
\algnewcommand\algorithmiccase{\textbf{case}}
\algnewcommand\algorithmicdefault{\textbf{default}}
\algdef{SE}[SWITCH]{Switch}{EndSwitch}[1]{\algorithmicswitch\ #1\ \algorithmicdo}{\algorithmicend\ \algorithmicswitch}
\algdef{SE}[CASE]{Case}{EndCase}[1]{\algorithmiccase\ #1}{\algorithmicend\ \algorithmiccase}
\algdef{SE}[DEFAULT]{Default}{EndDefault}[1]{\algorithmicdefault\ #1}{\algorithmicend\ \algorithmicdefault}
\algtext*{EndSwitch}
\algtext*{EndCase}
\algtext*{EndDefault}

\algnewcommand\algorithmicmatch{\textbf{match}}
\algdef{SE}[MATCH]{Match}{EndMatch}[1]{\algorithmicmatch\ #1}{\algorithmicend\ \algorithmicmatch}
\algtext*{EndMatch}

\algnewcommand\algorithmictry{\textbf{try}}
\algnewcommand\algorithmiccatch{\textbf{catch}}
\algblockdefx[Try]{Try}{EndTry}{\algorithmictry}{\algorithmicend\ \algorithmictry}
\algtext*{EndTry}
\algcblockdefx[Catch]{Try}{Catch}{EndTry}[1]{\algorithmiccatch\ #1}{\algorithmicend\ \algorithmictry}%
\algtext*{EndTry}
\MakeRobust{\Call}

\title{An Introduction to Probabilistic Programming}

\maintitleauthorlist{Jan-Willem van de Meent  \\
        Institute of Informatics \\
        University of Amsterdam \\
        j.w.vandemeent@uva.nl
        \and
        Brooks Paige \\
        University College London \\ Alan Turing Institute \\ 
        b.paige@ucl.ac.uk
        \and
        Hongseok Yang \\
        School of Computing\\
	KAIST \\
	hongseok.yang@kaist.ac.kr
        \and
        Frank Wood \\
	Department of Computer Science \\
        University of British Columbia \\
        fwood@cs.ubc.ca}

\author[1]{van de Meent, Jan-Willem}
\author[2]{Paige, Brooks}
\author[3]{Yang, Hongseok}
\author[4]{Wood, Frank}

\affil[1]{Institute of Informatics, University of Amsterdam; j.w.vandemeent@uva.nl}
\affil[2]{University College London and Alan Turing Institute; b.paige@ucl.ac.uk}
\affil[3]{School of Computing, KAIST; hongseok.yang@kaist.ac.kr}
\affil[4]{Department of Computer Science, University of British Columbia; fwood@cs.ubc.ca}

\begin{document}

\makeabstracttitle

\begin{abstract}

This book is designed to be a first-year graduate-level introduction to probabilistic programming.  It not only provides a thorough background for anyone wishing to use a probabilistic programming system, but also introduces the techniques needed to design and build these systems.  It is aimed at people who have an undergraduate-level understanding of either or, ideally, both probabilistic machine learning and programming languages.

We start with a discussion of model-based reasoning and explain why conditioning is a foundational computation  central to the fields of probabilistic machine learning and artificial intelligence. We then introduce a simple first-order probabilistic programming language (PPL) whose programs correspond to static-computation-graph, finite-random-variable-cardinality graphical models.  In the context of this restricted PPL we introduce fundamental inference algorithms and describe how they can be implemented so as to apply to any PPL-denoted model.

In the second part of this book, we introduce a higher-order probabilistic programming language with language features that highlight the problems one will encounter if one were to design a PPL using an existing higher-order programming language as the model specification language.  Namely, such languages allow one to define models with dynamic computation graphs, which may not instantiate the same set of random variables in each execution. Inference in such languages requires methods that generate samples by repeatedly evaluating the program. Foundational inference algorithms for this kind of probabilistic programming language are explained in the context of an interface between program executions and an inference controller.

The last two chapters of this book consider approaches that combine probabilistic and differentiable programming. We begin with a discussion of gradient-based inference methods for higher-order programs that denote densities over a fixed set of variables. In this context we discuss automatic differentiation, and how it can be used to implement efficient inference methods based on Hamiltonian Monte Carlo. We then turn to connections between between probabilistic programming and deep learning. Specifically, we present how to use gradient-based methods to perform maximum likelihood estimation in partially-specified probabilistic programs that are parameterized using neural networks, how to amortize inference using by learning neural approximations to the program posterior, and how PPL language features impact the design of deep probabilistic programming systems.  

\end{abstract}

\begin{acknowledgements}

We would like to thank the very large number of people who have read through preliminary versions of this manuscript.  Comments from the reviewers have been particularly helpful, as well as general interactions with David Blei and Kevin Murphy in particular. Some people we would like to individually thank are Adam Scibior, Mitch Wand, Rif Saurous, Tobias Kohn, Rob Zinkov, Marcin Szymczak, Gunes Baydin, Andrew Warrington, Yuan Zhou, Celeste Hollenbeck, Babak Esmaeili, Hao Wu, Heiko Zimmermann, Sam Stites, and Bradley Gram-Hansen, as well as numerous group members of at Oxford, UBC, and Northeastern, who graciously answered the call to comment and contribute. 

We would also like to acknowledge colleagues who have contributed intellectually to our thinking about probabilistic programming. First among these is David Tolpin, whose work with us at Oxford decisively shaped the design of the Anglican probabilistic programming language, and forms the basis for the material in Chapter~\ref{ch:eval-two}. We would also like to thank Josh Tenenbaum, Dan Roy,  Vikash Mansinghka, Zoubin Ghahramani, and Noah Goodman for inspiration, periodic but important research interactions, and friendly competition over the years. Chris Heunen, Ohad Kammar and Sam Staton helped us to understand subtle issues about the semantics of higher-order probabilistic programming languages. Lastly we would like to thank Mike Jordan for asking us to do this, providing the impetus to collate everything we thought we learned while having put together a NeurIPS tutorial years ago.   

During the writing of this manuscript the authors received generous support from various granting agencies.  Most critically, while all of the authors were at Oxford together, three of them were explicitly supported at various times by the DARPA under its Probabilistic Programming for Advanced Machine Learning (PPAML) (FA8750-14-2-0006) program. Jan-Willem van de Meent was additionally supported by the NSF (2047253), 3M, Intel, and startup funds from Northeastern University. Brooks Paige and Frank Wood were additionally supported by the Alan Turing Institute under the EPSRC grant EP/N510129/1.  Frank Wood was also supported by Intel, DARPA via its D3M (FA8750-17-2-0093) program, NSERC via its Discovery grant program, and CIFAR via its AI chair program. Hongseok Yang was supported by the Engineering Research Center Program through the National Research Foundation of Korea (NRF) funded by the Korean Government MSIT (NRF-2018R1A5A1059921), and also by Next-Generation Information Computing Development Program through the National Research Foundation of Korea (NRF) funded by the Ministry of Science, ICT (2017M3C4A7068177). 

\end{acknowledgements}

\chapter*{Notation}
    \markboth{\sffamily\slshape Notation}
    	{\sffamily\slshape Notation}
\vspace{-96pt}

\section*{Grammars}

\begin{longtable}{ll}
    $c ::=$
    & A constant value or primitive function.
    \\
    $v ::=$
    & A variable.
    \\
    $f ::=$
    & A user-defined procedure.
    \\
    \\
    $e ::=$
    & \mang{$c$\ |\ $v$\ | (let [$v$\ $e_1$]\ $e_2$)\ |  (if\ $e_1$\ $e_2$\ $e_3$) | ($f$\ $e_1$\ $\lldots$\ $e_n$)}
    \\
    & \mang{| ($c$\ $e_1$\ $\lldots$\ $e_n$) | (sample\ $e$) | (observe\ $e_1$\ $e_2$)}
    \\
    & An expression in the first-order probabilistic \\
    & programming language (FOPPL).
    \\
    \\
    $E ::=$
    & \mang{$c$\ |\ $v$\ | (if\ $E_1$\ $E_2$\ $E_3$) | ($c$\ $E_1$\ $\lldots$\ $E_n$)}
    \\
    &
    An expression in the (purely deterministic) target language.
    \\
    \\
    $e ::=$
    & \mang{$c$\ |\ $v$\ |\ $f$\ | (if\ $e$\ $e$\ $e$)\ |\ ($e$\ $e_1$\ $\lldots$\ $e_n$)}
    \\
    & \mang{|\ (sample\ $e$)\ |\ (observe\ $e$\ $e$) | (fn [$v_1$\ $\lldots$\ $v_n$]\ $e$)}
    \\
    & An expression in the higher-order probabilistic \\
    & programming language (HOPPL).
    \\
    \\
    $q ::= $
    & \mang{$e$\ | (defn\ $f$\ [$v_1$\ $\lldots$\ $v_n$]\ $e$)\ $q$}
    \\
    & A program in the FOPPL or the HOPPL.
\end{longtable}

\section*{Sets, Lists, Maps, and Expressions}

\begin{longtable}{p{0.5\textwidth}p{0.5\textwidth}}
    $C = \{c_1, \lldots, c_n\}$
    & A set of constants \\
    & ($c_i \in C$ refers to elements).
    \\
    $C = (c_1, \lldots, c_n)$
    & A list of constants\\
    & ($C_i$ indexes elements $c_i$).
    \\
    $\mathcal{C} = [v_1 \mapsto c_1, \lldots, v_n \mapsto c_n]$
    & A map from variables to constants \\& 
    ($\mathcal{C}(v_i)$ indexes entries $c_i$).
    \\
    $\mathcal{C}' = \mathcal{C}[v_i \mapsto c'_i]$
    & A map update in which $\mathcal{C}'(v_i) = c'_i$ \\ 
    & replaces $\mathcal{C}(v_i)=c_i$.
    \\
    $\mathcal{C}(v_i) = c'_i$
    & An in-place update in which $\mathcal{C}(v_i) = c'_i$ replaces $\mathcal{C}(v_i)=c_i$.
    \\
    $C = \text{dom}(\mathcal{C}) = \{v_1, \lldots, v_n\}$
    & The set of keys in a map.
    \\
    \\
    $E = \mang{(*\ $v$\ $v$)}$
    & An expression literal.
    \\
    $E' = E[v := c]
    = \mang{(*\ $c$\ $c$)}$
    & An expression in which a constant \\
    & $c$ replaces the variable $v$.
    \\
    $\textsc{free-vars}(e)$ & The free variables in an expression.

\end{longtable}

\section*{Directed Graphical Models}

\begin{longtable}{p{0.45\textwidth}p{0.55\textwidth}}
    $G = (V, A, \mathcal{P}, \mathcal{Y})$
    &
    A directed graphical model.
    \\
    \\
    $V = \{v_1, \lldots, v_{|V|}\}$
    & The variable nodes in the graph.
    \\
    $Y = \text{dom}(\mathcal{Y}) \subseteq V$
    & The observed variable nodes.
    \\
    $X = V \setminus Y \subseteq V$
    & The unobserved variable nodes.
    \\
    $y \in Y$
    & An observed variable node.
    \\
    $x \in X$
    & An unobserved variable node.
    \\
    \\
    $A = \{(u_1,v_1), \lldots, (u_{|A|},v_{|A|})\}$
    & The directed edges $(u_i,v_i)$ between parents $u_i \in V$ and children $v_i \in V$.
    \\
    $\mathcal{P} = [v_1 \mapsto E_1, \lldots, v_{|V|} \mapsto E_{|V|}]$
    & The probability mass or density for each variable $v_i$, represented as a target language expression  $\mathcal{P}(v_i)=E_i$
    \\
    $\mathcal{Y} = [y_1 \mapsto c_1, \lldots, y_{|Y|} \mapsto c_{|Y|}]$
    & The observed values $\mathcal{Y}(y_i) = c_i$.
    \\
    \\
    $\pa(v) = \{u: (u,v) \in A\}$
    & The set of parents of a variable $v$.
\end{longtable}

\newpage
\section*{Factor Graphs}

\begin{longtable}{p{0.45\textwidth}p{0.55\textwidth}}
    $G = (V, F, A, \Psi)$
    &
    A factor graph.
    \\
    \\
    $V = \{v_1, \lldots, v_{|V|}\}$
    & The variable nodes in the graph.
    \\
    $F = \{f_1, \lldots, f_{|F|}\}$
    & The factor nodes in the graph.
    \\
    $A = \{(v_1,f_1), \lldots, (v_{|A|}, f_{|A|})\}$
    & The undirected edges between variables $v_i$ and factors $f_i$.
    \\
    $\Psi = [f_1 \mapsto E_1, \lldots, f_{|F|} \mapsto E_{|F|}]$
    & Potentials for factors $f_i$, represented as target language expressions $E_i$.
\end{longtable}

\section*{Probability Densities}

\begin{longtable}{p{0.35\textwidth}p{0.65\textwidth}}
    $p(Y,X) = p(V)$
    &
    The joint density over all variables.
    \\
    $p(X)$
    & 
    The prior density over unobserved variables. 
    \\
    $p(Y \mid X)$ 
    & 
    The likelihood of observed variables $Y$ given unobserved variables $X$.
    \\
    $p(X \mid Y)$ 
    & 
    The posterior density for unobserved variables $X$ given observed variables $Y$.
\end{longtable}

\begin{longtable}{p{0.5\textwidth}p{0.5\textwidth}}
    $\mathcal{X} = [x_1 \mapsto c_1, \lldots, x_n \mapsto c_n]$
    &
    A trace of values $\mathcal{X}(x_i) = c_i$ assocated with the instantiated set of variables $X = \text{dom}(\mathcal{X})$.
    \\
    $p(X \!=\! \mathcal{X}) = p(x_1\!=\!c_1, \lldots, x_n\!=\!c_n)$
    &
    The probability density $p(X)$ evaluated at a trace $\m{X}$.
    \\
    \\
    $p_0(v_0 \,;\, c_1, \lldots, c_n)$
    &
    A probability mass or density function for a variable $v_0$ with parameters $c_1, \lldots, c_n$.
    \\
    $P(v_0) = \mang{($p_0$\ $v_0$\ $c_1$\ $\ldots$\ $c_n$)}$
    &
    The language expression that evaluates to the probability mass or density
    $p_0(v_0 ; c_1, \lldots, c_n)$.
\end{longtable}



\chapter{Introduction}
\label{ch:introduction}

How do we engineer machines that reason?  This is a question that has long vexed humankind;
answering it would be incredibly valuable.
There exist various hypotheses.  One major division of hypothesis space delineates along lines of assertion: are random variables and probabilistic calculation more-or-less a requirement \citep{ghahramani2015probabilistic,tenenbaum2011grow}, or the opposite \citep{lecun2015deep,Goodfellow-et-al-2016}?
The field ascribed to the former camp is roughly known as Bayesian or probabilistic machine learning; the latter as deep learning.  
The first requires inference as a fundamental tool; the latter optimization, usually gradient-based, for classification and regression.

Probabilistic programming languages are to the former as automated differentiation tools are to the latter.  Probabilistic programming is fundamentally about developing languages that allow the denotation of inference problems and evaluators that ``solve'' those inference problems.  
The rapid exploration of the deep learning, big-data-regression approach to artificial intelligence has been triggered to a large degree by the emergence of programming language tools that automate the tedious and troublesome derivation and calculation of gradients for optimization.  
Probabilistic programming aims to build and deliver a toolchain that does the same for probabilistic machine learning; supporting supervised, unsupervised, and semi-supervised inference.  
Without such a toolchain, one could argue the complexity of inference-based approaches to artificial intelligence systems is too high to allow rapid exploration of the kind seen recently in deep learning. 

Probabilistic programming tools and techniques are already transforming the way Bayesian statistical analyses are performed.  Traditionally the majority of the effort required in a Bayesian statistical analysis was in iterating model design where  each iteration often involved a painful implementation of an inference algorithm specific to the current model.  Automating inference, as probabilistic programming systems do, significantly lowers the cost of iterating model design, leading to both a better overall model in a shorter period of time and  consequent benefits.    

This introduction to probabilistic programming covers the basics of probabilistic programming, from language design to evaluator implementation, with the dual aim of explaining existing systems at a deep enough level that readers of this text should have no trouble adopting and using any of the languages and systems that are currently out there, and making it possible for the next generation of probabilistic programming language designers and implementers to use this as a foundation upon which to build.

This introduction starts with an important, motivational look at what a model is and how model-based inference can be used to solve many interesting problems.  Like automated differentiation tools for gradient-based optimization, the utility of probabilistic programming systems is grounded in applications simpler and more immediately practical than futuristic artificial intelligence applications; building from this is how we will start.    

\section{Model-based Reasoning}
\label{sec:model-based-reasoning}

Model-building starts early.  Children build model airplanes then
blow them up with firecrackers just to see what happens.  Civil engineers build physical models  of bridges and dams then see what happens in scale-model wave pools and wind tunnels. Disease researchers use mice as model organisms to simulate how cancer tumors might respond to different drug dosages in humans.  

These examples show exactly what a model is: a stand-in, an imposter, an artificial construct designed to respond in the same way as the system you would like to understand. A mouse is not a human but it is often close enough to get a sense of what a particular drug will do at particular concentrations in humans.  A scale model of an earthen embankment dam has the wrong relative granularity of soil composition but studying overtopping in a wave pool still tells us something about how an actual dam might respond.

As computers have become faster and more capable, numerical models and computer simulations have replaced physical models.  Such simulations are by nature approximations. However, in many cases they can be as exacting as even the most highly sophisticated physical models --- consider that the US was happy to abandon physical testing of nuclear weapons. 

Numerical models emulate stochasticity, i.e.~using pseudorandom number generators, to simulate actually random phenomena and other uncertainties.  Running a simulator with stochastic value generation leads to a many-worlds-like explosion of possible simulation outcomes.  Every little kid knows that even the slightest variation in the placement of a firecracker or the most seemingly minor imperfection of a glue joint will lead to dramatically different model airplane explosions.  Effective stochastic modeling means writing a program that can produce all possible explosions, each corresponding to a particular set of random values, including for example the random final resting position of a rapidly dropped lit firecracker.  

Arguably this intrinsic variability of the real world is the most significant complication for modeling and understanding. Did the mouse die in two weeks because of a particular individual drug sensitivity, because of its particular phenotype, or because the drug regiment trial arm it was in was particularly aggressive? If we are interested in average effects, a single trial is never enough to learn anything for sure because random things almost always happen.  You need a population of mice to gain any kind of real knowledge.  You need to conduct several wind-tunnel bridge tests, numerical or physical, because of variability arising everywhere --- the particular stresses induced by a particular vortex, the particular frailty of an individual model bridge or component, etc.  Stocha stic numerical simulation aims to computationally encompass the complete distribution of possible outcomes.  


When we write ``model'' we generally will mean ``stochastic simulator'' and the measurable values it produces.  Note, however, that this is not the only notion of model that one can adopt.  An important related family of models is specified solely in terms of an unnormalized density or ``energy'' function; these are treated in Chapter~\ref{ch:graph-based}. 

Models produce values for things we can measure in the real world. We call such measured values {\em observations}.  What counts as an observation is model, experiment, and query specific --- you might measure the daily weight of mice in a drug trial or you might observe whether or not a particular bridge design fails under a particular load.  

Generally one does not observe every detail produced by a model, physical or numerical, and sometimes one simply cannot.  Consider the standard model of physics and the large hadron collider.  The standard model is arguably the most precise and predictive model ever conceived; it can be used to describe what can happen in  fundamental particle interactions.  At high energies these interactions can result in a particle jet that stochastically transitions between energy-equivalent decompositions with varying particle-type and momentum constituencies.  It is simply not possible to observe the initial particle products and their first transitions because of how fast they occur.  The energy 
of particles that make up the jet deposited into various detector elements constitute the observables.  

So how does one use models?  One way is to use them to falsify theories.  To this one needs encode the theory as a model then simulate from it many times.  If the population distribution of observations generated by the model is not in agreement with observations generated by the real world process then there is evidence that the theory can be falsified.  To a large extent, this describes the scientific process.  Good theories take the form of models that can be used to make testable predictions.  We can test those predictions and falsify model variants that fail to replicate observed statistics.


Models also can be used to make decisions.  For instance when playing a game you  either consciously or unconsciously use a model of how your opponent will play.  To use such a model to make decisions about what move to play next yourself, you could simulate taking a variety of different actions, then pick one amongst them by simulating your opponent's reaction according to your model of them, and so forth until reaching a game state whose value you know, for instance, the end of the game.  Choosing the action that maximizes your chances of winning is a rational strategy that can be framed as model-based reasoning. 
Abstracting this to broader life, as a game in which individuals each aim to maximize their own utility function under their own model of the entire world, draws a connection between model-based probabilistic machine learning and artificial intelligence.


%

A useful model can take a number of forms. 
One kind is a reusable, interpretable abstraction, which describes summary statistics or features extracted from raw observable data, and has a good associated inference algorithm. 
Another kind would be a reusable but non-interpretable (perhaps entirely black-box) model, that can accurately generate complex data that closely resembles what would be observed in the real world. 
Yet another kind of model, particularly in science and engineering, takes the form of a problem-specific simulator that describes a generative process very explicitly in engineering-like terms and precision. 
Over the course of this introduction it will become apparent how probabilistic programming addresses the complete spectrum of them all.

All model types have parameters.  Fitting these parameters, when few, can sometimes be performed manually, by intensive theory-based reasoning and a priori experimentation (the masses of particles in the standard model),  by measuring conditional subcomponents of a simulator (the compressive strength of various concrete types and their action under load), or by simply fiddling with parameters to see which values produce the most realistic outputs.

Automated model fitting describes the process of using algorithms to determine either point or distributional estimates for model parameters and structure.  Such automation is particularly useful when the parameters of a model are uninterpretable, or if there are too many to consider exhaustively. 
We will return to model fitting in Chapter \ref{sec:gbli}, however it is important to realize that inference can be used for model learning too, simply by lifting the inference problem to include uncertainty about the model itself (e.g.~see the neural network example in \ref{sec:foppl-examples} and the program induction example in \ref{sec:hoppl-examples}).
%
%
%
%

The key point for now is to understand that models come in many forms, from scientific and engineering simulators in which the results of every subcomputation are interpretable to abstract models in statistics and computer science which are, by design, significantly less interpretable but often are valuable for predictive inference nonetheless.  

\subsection{Model Denotation}
How do we denote such models, and how can models be manipulated to compute quantities of interest?
These are 
arguably the foundational questions that led to the field of probabilistic programming.


To make clear what we mean by model denotation, let us first look at the specification of simple statistical model.
Statistical models are typically denoted mathematically, subsequently manipulated algebraically, then ``solved''  computationally.   By ``solved'' we mean that an inference problem involving conditioning on the values of a subset of the variables in the model is answered. 
Such a model denotation stands in contrast to simulators, which are often denoted in terms of software source code that is directly executed.  
This also stands in contrast (though less so) to generative models in machine learning, which usually take the form of probability distributions whose factorization properties can be read from diagrams like graphical models or factor graphs.  

A simple textbook statistical model, for generating a coin flip from a potentially biased coin, is a beta-Bernoulli model.
This model is typically denoted
\begin{align}
x &\sim \mathrm{Beta}(\alpha, \beta) \nonumber \\ 
y &\sim \mathrm{Bernoulli}(x) \label{eqn:betabernoulli}
\end{align}
where $\alpha$ and $\beta$ are parameters, $x$ is a latent variable (the bias of the coin) and $y$ is the value of the flipped coin.  A trained statistician will also ascribe a learned, folk-meaning to the symbol $\sim$ and the keywords $\mathrm{Beta}$ and $\mathrm{Bernoulli}$.  
For example $\mathrm{Beta}(a,b)$ means that, given the value of arguments $a$ and $b$ we can construct what is effectively (from a computer scientist's point of view) an object with two methods.
The first method defines a probability density (or distribution) function, in this case computing
\[ 
    p(x|a,b) = \frac{\Gamma(a+b)}{\Gamma(a)\Gamma(b)}x^{a-1}(1-x)^{b-1},
\] 
and the second method draws exact samples from said distribution.
A statistician will also usually be able to intuit not only that some variables in a model are to be observed, here for instance $y$, but that there is an inference objective, here for instance to characterize $p(x|y)$.  This denotation is extremely compact, and being mathematical in nature means that we can use our learned mathematical algebraic skills to manipulate expressions to solve for quantities of interest.  We will return to this shortly.


In this book we will generally have conditioning as our goal, namely the characterization of some conditional distribution given a specification of a model in the form of a joint distribution.
This will involve the extensive use of Bayes' rule 
\begin{align}
p(X|Y)  &= \frac{p(Y|X)p(X)}{p(Y)} = \frac{p(X,Y)}{p(Y)} = \frac{p(X,Y)}{\int p(X,Y)dX}.
\end{align}
Bayes' rule tells us how to derive a conditional probability from a joint, conditioning tells us how to rationally update our beliefs, and updating beliefs is what learning and inference are all about.


The constituents of Bayes' rule have common names that are well known and will appear throughout this text: $p(Y | X)$ the likelihood, $p(X)$ the prior, $p(Y)$ the marginal likelihood (or evidence), and $p(X | Y)$ the posterior.  For our purposes a model is the joint distribution $p(Y,X) = p(Y | X)p(X)$ of the observations $Y$ and the random choices made in the generative model $X$, also called latent variables. 

\begin{table}[t]
\caption{Probabilistic Programming Models}
\begin{center}
\begin{tabular}{|c|c|}
\hline
$X$ & $Y$ \\
\hline
scene description & image  \\
simulation & simulator output\\
program source code & program return value\\
policy prior and world simulator & rewards  \\
cognitive decision making process & observed behavior \\
\hline
\end{tabular}
\end{center}
\label{table:newmodels}
\end{table}%

The subject of Bayesian inference, including both philosophical and methodological aspects, is in and of itself worthy of book length treatment.  There are a large number of excellent references available, foremost amongst them the excellent book by \citet{gelman2013bayesian}.  In the space of probabilistic programming arguably the recent books by \citet{davidson2015bayesian} and \citet{pfeffer2016practical} are the best current references.  They all aim to explain 
that conditioning a joint distribution --- the fundamental Bayesian update --- is a formalism that succinctly prescribes a way to express and solve a huge variety of problems.

Before continuing on to the special-case analytic solution to this particular Bayesian statistical model and inference problem, let us build some intuition about the power of both programming languages for model denotation and automated conditioning by considering  Table~\ref{table:newmodels}. In this table we list a number of pairs of domains $X$,$Y$ where denoting the joint distribution of $P(X,Y)$ is realistically only doable in a probabilistic programming language, and the posterior distribution $P(X|Y)$ is of interest.  Take the first, ``scene description'' and ``image.''  What would such a joint distribution look like?  
To imagine $P(X,Y)$, 
start by thinking about $P(X)$ as some distribution over a so-called scene graph --- the actual object geometries, textures, and poses  in a physical environment --- perhaps defined by 
a stochastic simulator that only needs to generate reasonably plausible scene graphs. 
Noting that $P(X,Y) = P(Y|X)P(X)$ then all we need is a way to go from scene graph to observable image and we have a complete description of a joint distribution.  There are many kinds of renderers that do just this and, although deterministic in general, they are perfectly fine to use when specifying a joint distribution because they map from some latent scene description to observable pixel space and, with the addition of some image-level pixel noise reflecting, for instance, sensor imperfections or Monte-Carlo ray-tracing artifacts, form a perfectly valid likelihood.   

An example of this ``vision as inverse graphics'' idea \citep{kulkarni2015deep}, appearing first in \citet{mansinghka2013approximate} and then subsequently in \citet{Le2017UsingSD,le2016inference}, took the image $Y$ to be a Captcha image and the scene description $X$ to include the obscured string.  In all three papers the point was not Captcha-breaking per se, but rather demonstrating both that such a model is denotable in a probabilistic programming language and that such a model can be solved by general purpose inference.  

Let us momentarily consider alternative ways to solve such a ``Captcha problem.''  A non-probabilistic programming approach would require gathering a very large number of Captchas, hand-labeling them all, then designing and training a neural network to regress from the image to a text string \citep{bursztein2014end}.  The probabilistic programming approach in contrast merely requires one to write a program that generates Captchas that are stylistically similar to the Captcha family one would like to break --- a {\em model} of Captchas --- in a probabilistic programming language.  Conditioning such a model on its observable output, the Captcha image, will yield a posterior distribution over text strings.  This kind of conditioning is what probabilistic programming evaluators do.

\begin{figure}[t]
\begin{center}
\includegraphics[width=\textwidth]{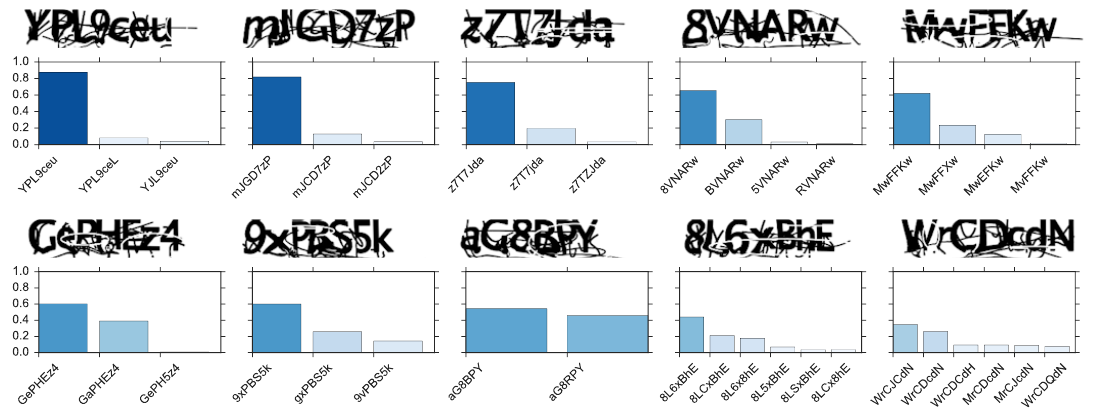}
\end{center}
\caption{Posterior uncertainties after inference in a probabilistic programming language model of 2017 Facebook Captchas (reproduced from \citet{le2016inference})}
\label{fig:captcha-posterior}
\end{figure}

Figure~\ref{fig:captcha-posterior} shows a representation of the output of such a conditioning computation.  Each Captcha/bar-plot pair consists of a held-out Captcha image and a truncated marginal posterior distribution over unique string interpretations.  Drawing your attention to the middle of the bottom row, notice that the noise on the Captcha makes it more-or-less impossible to tell if the string is ``aG8BPY'' or ``aG8RPY.''  The posterior distribution $P(X|Y)$ arrived at by conditioning reflects this uncertainty.

By this example, whose source code appears in Chapter~\ref{ch:hoppl} in a simplified form, we aim only to liberate your thinking in regards to what a model is (a joint distribution, potentially over richly structured objects, produced by adding stochastic choice to normal computer programs like Captcha generators) and what the output of a conditioning computation can be like.  What probabilistic programming languages do is to allow denotation of any such model.  What this book covers in great detail is how to develop inference algorithms that allow computational characterization of the posterior distribution of interest, increasingly very rapidly as well (see Chapter~\ref{sec:gbli}).

%



\subsection{Conditioning}

Returning to our simple coin-flip statistics example, let us continue and write out the joint probability density for the distribution on $X$ and $Y$.  The reason to do this is to paint a picture, by this simple example, of what the mathematical operations involved in conditioning are like and why the problem of conditioning is, in general, hard.

Assume that the symbol $Y$ denotes the observed outcome of the coin flip and that we encode the event ``comes up heads'' as the mathematical value of the integer 1 and 0 for the converse.  We will denote the bias of the coin, i.e.~the probability it comes up heads, by the symbol $x$ and encode it using a real positive number between 0 and 1 inclusive, i.e.~$x\in [0,1]$.  Then using standard definitions for the distributions indicated by the joint denotation in Equation~\eqref{eqn:betabernoulli} we can write
%
%
\begin{align}
p(x,y) = x^y (1 - x)^{1 - y} \frac{\Gamma(\alpha + \beta)}{\Gamma(\alpha)\Gamma(\beta)} x^{\alpha - 1} (1 - x)^{\beta - 1} \label{eqn:betabinpdfmessy}
\end{align}
and then use rules of algebra to simplify this expression to
\begin{align}
p(x,y) = \frac{\Gamma(\alpha + \beta)}{\Gamma(\alpha)\Gamma(\beta)} x^{y+\alpha - 1} (1 - x)^{\beta - y}\label{eqn:betabinpdfsimplified}.
\end{align}

Note that we have been extremely pedantic here, using words like ``symbol,'' ``denotes,'' ``encodes,'' and so forth, to try to get you, the reader, to think in advance about other ways one might denote such a  model and to realize if you don't already that there is a fundamental difference between the symbol or expression used to represent or denote a meaning and the meaning itself.  Where we haven't been pedantic here is probably the most interesting thing to think about:  What does it mean to use rules of algebra to manipulate Equation~\eqref{eqn:betabinpdfmessy} into Equation~\eqref{eqn:betabinpdfsimplified}?  To most reasonably trained mathematicians, applying expression transforming rules that obey the laws of associativity, commutativity, and the like are natural and are performed almost unconsciously.  To a reasonably trained programming languages person these manipulations are meta-programs, i.e. programs that consume and output programs, that perform semantics-preserving transformations on expressions.  Some probabilistic programming systems  operate in exactly this way \citep{narayanan2016probabilistic}.  What we mean by semantics-preserving in general is that, after evaluation, expressions in pre-simplified and post-simplified form have the same meaning; in other words, they evaluate to the same object, usually mathematical, in an underlying formal language whose meaning is well established and agreed.  In probabilistic programming, semantics-preserving generally means that the mathematical objects denoted correspond to the same distribution \citep{StatonYWHK16}. Here, after algebraic manipulation, we can agree that the expressions in Equations~\eqref{eqn:betabinpdfmessy} and \eqref{eqn:betabinpdfsimplified}, when evaluated on inputs $x$ and $y$, would evaluate to the same value and thus are semantically equivalent alternative denotations.  
 
That said, our implicit objective here is not to compute the value of the joint probability of some variables, but to do conditioning instead, for instance, to compute $p(x|y=\mathrm{``heads''})$.
Using Bayes' rule this is {\em theoretically} easy to do. It is just
\begin{align}
p(x|y)  &= \frac{p(x,y)}{\int p(x,y)dx} = \frac{\frac{\Gamma(\alpha + \beta)}{\Gamma(\alpha)\Gamma(\beta)} x^{y+\alpha - 1} (1 - x)^{\beta - y}}{\int \frac{\Gamma(\alpha + \beta)}{\Gamma(\alpha)\Gamma(\beta)} x^{y+\alpha - 1} (1 - x)^{\beta - y} dx}.
\end{align}

In this special case  the rules of algebra and semantics preserving transformations of integrals can be used to algebraically solve for an analytic form for this posterior distribution.  

To start the preceding expression can be simplified to 
\begin{align}
p(x|y)  &= \frac{ x^{y+\alpha - 1} (1 - x)^{\beta - y}}{\int x^{y+\alpha - 1} (1 - x)^{\beta - y} dx}.
\end{align}
which still leaves a nasty looking integral in the denominator.  This is the complicating crux of Bayesian inference.  The integral that appears in the denominator is in general intractable, as it involves integrating over the entire space of the latent variables.  Consider the Captcha example: simply summing over the latent character sequence itself would require an exponential-time operation.

This example has a very special property, called conjugacy, which means that this integral can be performed by inspection; by identifying that the integrand is the same as the non-constant part of the beta distribution and using the fact that the beta distribution must sum to one,
\begin{align}
\int x^{y+\alpha - 1} (1 - x)^{\beta - y} dx = \frac{\Gamma(\alpha+y)\Gamma(\beta-y+1)}{\Gamma(\alpha + \beta +1)}.
\end{align}
Consequently, 
\begin{align}
p(x|y)  &= \mathrm{Beta}(\alpha+y,\beta-y+1),
\end{align}
which is equivalent to
\begin{align}
x|y  &\sim \mathrm{Beta}(\alpha+y,\beta-y+1).
\end{align}

There are several things that can be learned about conditioning from even this simple example.  The result of the conditioning operation is a {\em distribution} parameterized by the observed or given quantity.  Unfortunately this  distribution will in general not have an analytic form; we usually won't be so lucky that the normalizing integral has an algebraic analytic solution nor will it usually be easily calculable numerically.

This does not mean that all is lost.  Remember that the $\sim$ operator is overloaded to mean two things, density evaluation and exact sampling.  Neither of these are possible in general. However the latter, in particular, can be approximated, and often consistently even without being able to do the former.
For this reason amongst others our focus will be on sampling-based characterizations of conditional distributions in general.

\subsection{Query}
Regardless of the characterization of the resulting posterior distribution, whether a method for drawing samples or an explicit normalized probability density, we can now use it to ask questions --- ``queries'' in general.
These are best expressed in integral form as well.  
For instance, we could ask: what is the probability that the bias of the coin is greater than $0.7$, given that the coin came up heads? This is mathematically denoted as
\begin{align}
p(x>0.7|y=1) = \int \mathbb{I}(x>0.7) p(x|y = 1) dx,
\end{align}
where $\mathbb{I}(\cdot)$ is an indicator function which evaluates to $1$ when its argument takes value true and $0$ otherwise, which in this instance can be directly calculated using the cumulative distribution function of the beta distribution.

Fortunately we can still answer queries when we only have the ability to sample from the posterior distribution owing to the Markov strong law of large numbers, which states under mild assumptions that 
\begin{align}
\lim_{L \rightarrow \infty} \frac{1}{L} \sum_{\ell=1}^L{f(X^\ell)} \rightarrow \int f(X) p(X) dX,
\qquad
X^\ell \sim p(X),
\end{align}
for general distributions $p$ and functions $f$.  We will exploit this technique repeatedly throughout.  Note that the distribution on the right hand side is approximated by a set of $L$ samples on the left and that different functions $f$ can be evaluated at the same sample points chosen to represent $p$ after the samples have been generated.

This more or less completes the small part of the computational statistics story we will tell, at least insofar as how models are denoted then algebraically manipulated.  We highly recommend that unfamiliar readers interested in the fundamental concepts of Bayesian analysis and common mathematical evaluation strategies the book ``Bayesian Data Analysis'' \citep{gelman2013bayesian}.  

The field of statistics long-ago recognized that computerized systemization of the denotation of models and evaluators for inference was essential, and so developed specialized languages for model writing and query answering, amongst them BUGS \citep{spiegelhalter_software_1995} and, more recently, STAN \citep{stan_software_2014}.  
From initial goals of automating computation for Bayesian statistics in finite-dimensional models, the field has grown in breadth and depth, expanding to tackle many different classes of models and with applications including modern approaches to artificial intelligence.
Common to all these languages and systems is the shared objective of inference via conditioning.

%
%
%
%
%


\section{Probabilistic Programming}

The Bayesian approach, in particular the theory and utility of conditioning, is remarkably general in its applicability.  One view of probabilistic programming is that it is about automating Bayesian inference. In this view probabilistic programming concerns the development of syntax and semantics for languages that denote conditional inference problems and the development of corresponding evaluators or ``solvers'' that computationally characterize the denoted conditional distribution.  For this reason probabilistic programming sits at the intersection of the fields of machine learning, statistics, and programming languages, drawing on the formal semantics, compilers, and other tools from programming languages to build efficient inference evaluators for models and applications from machine learning using the inference algorithms and theory from statistics.  

\begin{figure}[t]
\begin{center}
\includegraphics[width=8cm]{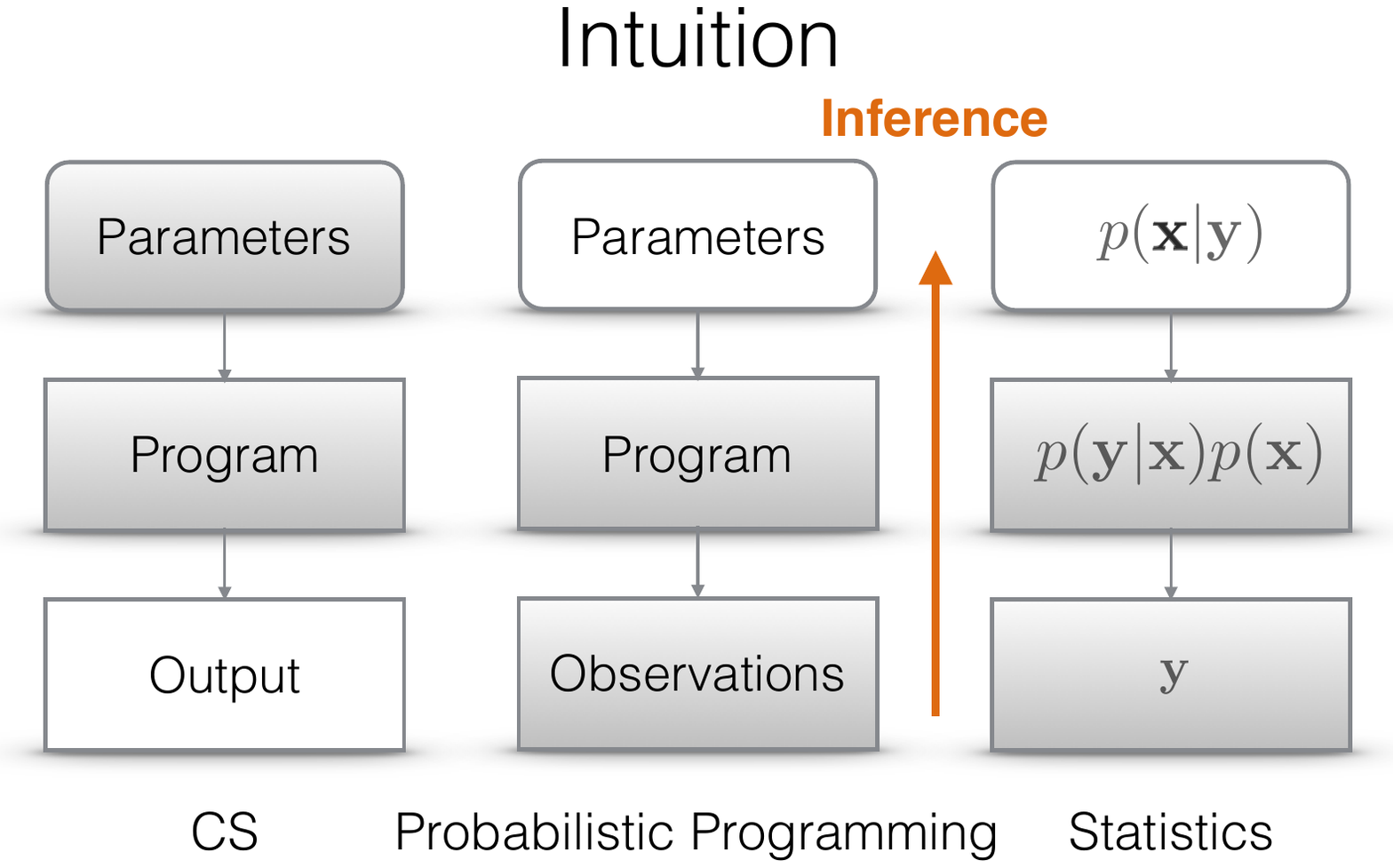}
\caption{Probabilistic programming, an intuitive view.}
\label{fig:probprogintuition}
\end{center}
\end{figure}

Probabilistic programming is about doing statistics using the tools of computer science.  Computer science, both the theoretical  and engineering discipline, has largely been about finding ways to efficiently evaluate programs, given parameter or argument values, to produce some output.  In Figure \ref{fig:probprogintuition} we show the typical computer science programming pipeline on the left hand side: write a program, specify the values of its arguments or situate it in an evaluation environment in which all free variables can be bound, then evaluate the program to produce an output.  The right hand side illustrates the approach taken to modeling in statistics: start with the output, the observations or data $Y$, then specify a usually abstract generative model $p(X, Y)$, often denoted mathematically, and finally use algebra and inference techniques to characterize the posterior distribution, $p(X \,|\, Y)$, of the unknown quantities in the model given the observed quantities.  Probabilistic programming is about performing Bayesian inference using the tools of computer science: programming language for model denotation and statistical inference algorithms for computing the conditional distribution of program inputs that could have given rise to the observed program output.

Thinking back to our earlier example, reasoning about the bias of a coin is an example of the kind of inference probabilistic programming systems do.  Our data is the outcome, heads or tails, of one coin flip.  Our model, specified in a forward direction, stipulates that a coin and its bias is generated according to the hand-specified model then the coin flip outcome is observed and analyzed under this model.  One challenge, the writing of the model, is a major focus of applied statistics research where ``useful'' models are painstakingly designed for every new important problem.  
The other challenge is computational: though Bayes' rule gives us a theoretical framework defining what to calculate, we need to select and implement an algorithm to computationally characterize the posterior distribution of the latent quantities (e.g.~bias) given the observed quantity (e.g.~``heads'' or ``tails'').  In the beta-Bernoulli problem we were able to analytically derive the form of the posterior distribution, in effect allowing us to transform the original inference problem denotation into a denotation of a program that completely characterizes the inverse computation.  

When performing inference in probabilistic programming systems, we need to design algorithms that are applicable to any program that a user could write in some language.  In probabilistic programming the language used to denote the generative model is critical, ranging from intentionally restrictive modeling languages, such as the one used in BUGS, to arbitrarily complex computer programming languages like C, C++, and Clojure. Any outputs generated from the forward computation can be considered observables.  The inference objective is to computationally characterize the posterior distribution of all of the random choices made during the forward execution of the program given that the program produces a particular output. 

There are subtleties, but that is a fairly robust intuitive definition of probabilistic programming.  Throughout most of this book we will assume that the program is fixed and that the primary objective is inference in the model specified by the program.  In the penultimate chapter we will discuss connections between probabilistic programming and deep learning, in particular through the lens of semi-supervised learning in the variational autoencoder family where parts of or the whole generative model itself, i.e.~the probabilistic program or ``decoder,'' is also learned from data. 

Before that, though, let us consider how one would recognize or distinguish a probabilistic program from a non-probabilistic program.  Quoting  \citet{gordon2014probabilistic}, 
``probabilistic programs are usual functional or imperative programs with two added constructs:  the ability to draw values at random from distributions, and the ability to {\em condition} values of variables in a program via observations.''  
We emphasize conditioning here. 
The meaning of a probabilistic program is that it simultaneously denotes a joint and conditional distribution, the latter by syntactically indicating where conditioning will occur, i.e.~which random variable values will be observed.  
Almost all languages have pseudo-random value generators or packages; what they lack in comparison to probabilistic programming languages is syntactic constructs for conditioning and evaluators that implement conditioning.  
We will call languages that include such constructs probabilistic programming languages.  We will call languages that do not but that are used for forward modeling stochastic simulation languages or, more simply, programming languages.  

There are many libraries for constructing graphical models and performing inference;
this software works by programmatically constructing a data structure which represents a model, and then, given observations, running graphical model inference.
What distinguishes between this kind of approach and probabilistic programming is that a program is used to construct a model as a data structure, rather than considering the ``model'' that arises implicitly from direct evaluation of the program expression itself.
In probabilistic programming systems, either a model data structure is constructed explicitly via a non-standard interpretation of the probabilistic program itself (if it can be, see Chapter~\ref{ch:graph-based}), 
or it is a general Markov model whose state is the evolving evaluation environment generated by the probabilistic programming language evaluator (see Chapter~\ref{ch:eval-one}). 
In the former case, we often perform inference by compiling the model data structure to a density function (see Chapter \ref{ch:graph-based}), whereas in the latter case, we employ methods that are fundamentally generative (see Chapters \ref{ch:eval-one} and \ref{ch:eval-two}).

\subsection{Existing Languages}


The design of any book on probabilistic programming will have to include a mix of programming languages and statistical inference material along with a smattering of models and ideas germane to machine learning.  In order to discuss modeling and programming languages one must choose a language to use in illustrating key concepts and for showing examples.  There are a very large number of languages from a number of research communities; programming languages: {Hakaru \citep{narayanan2016probabilistic}, Augur \citep{tristan2014augur}, R2 \citep{nori2014r2}, Figaro \citep{pfeffer_rep_2009}, IBAL \citep{pfeffer_ijcai_2001}), PSI \citep{gehr2016psi}; machine learning: Church \citep{goodman_uai_2008}, Anglican \citep{wood_aistats_2014} (updated syntax \citep{anglican_arxiv}), BLOG \citep{milch_ijcai_2005}, Turing.jl \citep{ge2016turing}, BayesDB \citep{mansinghka2015bayesdb}, Venture \citep{mansinghka_arxiv_2014}, Probabilistic-C \citep{paige2014compilation}, WebPPL \citep{goodman2014dippl}, CPProb \citep{lezcano2017cpprob}, \citep{koller_aaai_1997}, \citep{thrun_icra_2000}; and statistics: Biips \citep{todeschini2014biips}, LibBi \citep{murray2013}, Birch \citep{MurrayLKBS18}, STAN \citep{stan_software_2014}, JAGS \citep{plummer2003jags}, BUGS \citep{spiegelhalter_software_1995}\footnote{sincere apologies to the authors of any languages left off this list}.  

In this book we will not attempt to explain each of the languages and catalogue their numerous similarities and differences.  Instead we will focus on the concepts and implementation strategies that underlie most, if not all, of these languages.  We will highlight one extremely important distinction, namely, between languages in which all programs induce models with a finite number of random variables and languages for which this is not true.  
 The language we choose for the book has to be a language in which a coherent shift from the former to the latter is possible.  For this and other reasons we chose to write the book using an abstract language similar in syntax and semantics to Anglican. Anglican is similar to WebPPL, Church, and Venture. It is  a Lisp-like language which, by virtue of its syntactic simplicity, also makes for efficient and easy meta-programming, an approach many
implementors will take.  
That said, the real substance of this book is language agnostic and the main points should be understood in this light.

We have left off of the preceding extensive list of languages both one important class of language --- probabilistic logic languages (\citep{kimmig2011implementation,sato_ijcai_1997} --- and sophisticated, useful, and widely deployed libraries/embedded domain-specific languages for modeling and inference (Infer.NET \citep{minka_software_2010}, Factorie \citep{mccallum_nips_2009}, Edward \citep{tran2016edward}, PyMC3 \citep{salvatier2016probabilistic}).  One link between the material presented in this book and these additional languages and libraries is that the inference methodologies we will discuss apply to advanced forms of probabilistic logic programs \citep{alberti2016probabilistic,kimmig2017probabilistic} and, in general, to the graph representations constructed by such libraries.  In fact the libraries can be thought of as compilation targets for appropriately restricted languages.  In the latter case strong arguments can be made that these are also languages in the sense that there is an (implicit) grammar, a set of domain-specific values, and a library of primitives that can be applied to these values.  The more essential distinction is the one we have structured this book around, that being the difference between static languages in which the denoted model can be compiled to a finite-node graphical model and dynamic languages in which no such compilation can be performed.


\section{Example Applications}


%

Before diving into specifics, let us consider some motivating examples of what has been done with probabilistic programming languages and how phrasing things in terms of a model plus conditioning can lead to elegant solutions to otherwise extremely difficult tasks.

Besides the obvious benefits that derive from having an evaluator that implements inference automatically, the main benefit of probabilistic programming is having additional expressivity, significantly more compact and readable than mathematical notation, in the modeling language.  While it is possible to write down the mathematical formalism for a model of latents $X$ and observables $Y$ for each of the examples shown in Table~\ref{table:newmodels}, doing so is usually neither efficient nor helpful in terms of intuition and clarity.  
We have already given one example, generating and breaking Captchas, earlier in this chapter. Let us proceed to more.  

\subsubsection{Constrained Simulation}

\begin{figure}[htbp]
\begin{center}
\includegraphics[width=3.5cm]{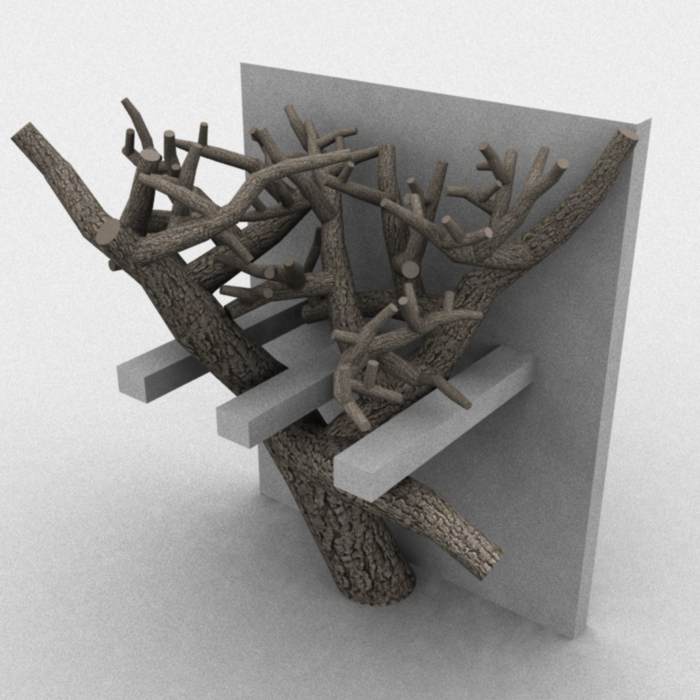}
~~
\includegraphics[width=3.5cm]{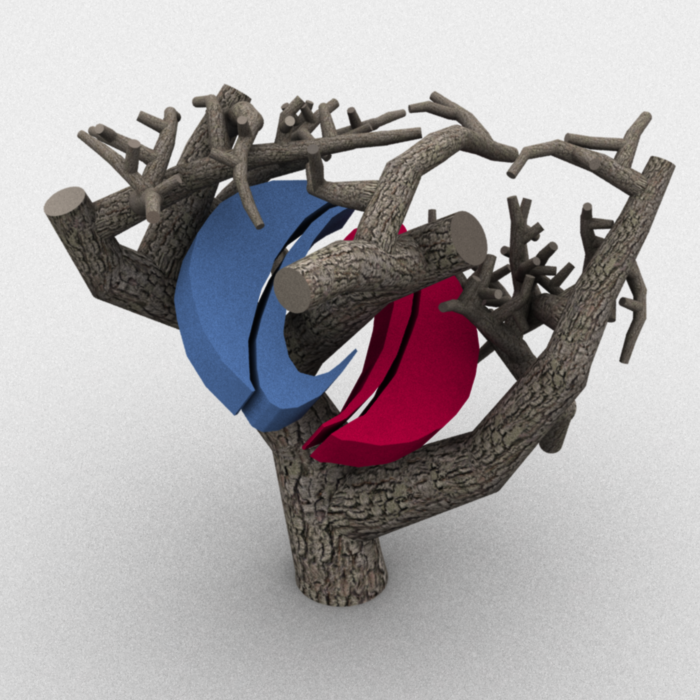}
\caption{Posterior samples of procedurally generated, constrained trees  (reproduced from \citet{ritchie2015controlling})}
\label{fig:ritchietrees}
\end{center}
\end{figure}

Constrained procedural graphics \citep{ritchie2015controlling}
is a visually compelling and elucidating application of probabilistic programming.
Consider how one makes a forest with computer graphics, e.g.\ for a movie or computer game.  One does not hire one thousand designers, each drawing a single tree by hand. Instead, one hires a procedural graphics programmer who writes what we call a generative model --- a stochastic simulator that generates a synthetic tree each time it is run.  A forest is then constructed by calling such a program many times and arranging the trees on a landscape.  What if, however, a director enters the design process and stipulates, for whatever reason, that the tree cannot touch some other elements in the scene, i.e.~in probabilistic programming lingo we ``observe'' that the tree cannot touch some elements?  Figure~\ref{fig:ritchietrees} shows examples of such a situation where the tree on the left must miss the back wall and grey bars and the tree on the right must miss the blue and red logo.  In these figures you can see, visually, what we will examine in a high level of detail throughout the book.  The random choices made by the generative procedural graphics model correspond to branch elongation lengths, how many branches diverge from the trunk and subsequent branch locations, the angles that the diverged branches take, the termination condition for branching and elongation, and so forth.  Each tree literally corresponds to one execution path or setting of the random variables of the generative program.  Conditioning with hard constraints like these transforms the prior distribution on trees into a posterior distribution in which all posterior trees conform to the constraint.  Valid program variable settings (those present in the posterior) have to make choices at all intermediate sampling points that allow all other sampling points to take at least one value that can result in a tree obeying the statistical regularities specified by the prior and the specified constraints as well.   

\subsubsection{Program Induction}

How do you automatically write a program that performs an operation you would like it to?  One approach is to use a probabilistic programming system and inference to invert a generative model that generates normal, regular, computer program code and conditions on its output, when run on examples, conforming to the observed specification.  This is the central idea in the work of \citet{perov-agi-2016} whose use of probabilistic programming is what distinguishes their work from the related  literature \citep{gulwani2017program,hwang2011inducing,liang2010learning}.  Examples such as this, even more than the preceding visually compelling examples, illustrate the denotational convenience of a rich and expressive programming language as the generative modeling language.  A program that writes programs is most naturally expressed as a recursive program with random choices that generates abstract syntax trees according to some learned prior on the same space.  While models from the natural language processing literature exist  that allow specification and generation of computer source code (e.g.~adaptor grammars \citep{johnson2007adaptor}), they are at best cumbersome to denote mathematically.

\subsubsection{Recursive Multi-Agent Reasoning}

Some of the most interesting uses for probabilistic programming systems derive from the rich body of work around the Church and WebPPL systems.  The latter, in particular, has been used to study the mutually-recurisive reasoning among multiple agents. A number of examples on this are detailed in an excellent online tutorial \citep{goodman2014dippl}.  

\vspace{.5cm}

The list goes on and could occupy a substantial part of a book itself.  The critical realization to make is that, of course, any traditional statistical model can be expressed in a probabilistic programming framework, but, more importantly, so too can many others and with significantly greater ease.  Models that take advantage of existing source code packages to do sophisticated nonlinear deterministic computations are particularly of interest.  One exciting example application under consideration at the time of writing is to instrument the stochastic simulators that simulate the standard model and the detectors employed by the large hadron collider \citep{baydin2018efficient}.  By ``observing'' the detector outputs, inference in the generative model specified by the simulation pipeline may prove to be able to produce the highest fidelity event reconstruction and science discoveries.

This last example highlights one of the principle promises of probabilistic programming.  There exist a large number of software simulation modeling efforts to simulate, stochastically and deterministically, engineering and science phenomena of interest.  Unlike in machine learning where often the true generative model is not well understood, in engineering situations (like building, engine, or other system modeling) the forward model is often incredibly well understood, and already exists as code.  Probabilistic programming techniques and evaluators that work within the framework of existing languages should prove to be very valuable in disciplines where significant effort has been put into modeling complex engineering or science phenomena of interest and the power of general purpose inverse reasoning has not yet been made available.


\section{A First Probabilistic Program}

Before we get started in earnest, it is worth considering at least one simple probabilistic program to informally introduce a bit of syntax, and relate a model denotation in a probabilistic programming language to the underlying mathematical denotation and inference objective.  There will be source code examples provided throughout, though not always with accompanying mathematical denotation.

Recall the simple beta-Bernoulli model from Section~\ref{sec:model-based-reasoning}.  This is one in which the probabilistic program denotation is actually longer than the mathematical denotation. (That will only be the case for such trivial simple models!) Here is a probabilistic program that represents the beta-Bernoulli model:

\begin{anglican}[mathescape,caption={The beta-Bernoulli model as a probabilistic program},label=example:beta-binomial-anglican]
(let [prior (beta a b)
      x (sample prior)
      likelihood (bernoulli x)
      y 1]
  (observe likelihood y)
  x))
\end{anglican}

This program is written in the Lisp dialect we will use throughout, and which we will explain in glorious detail in the next chapter.  For those completely new to functional programming languages this kind of syntax can be confusing at first.  We will discuss many reasons why we chose this kind of language anyway later in the book. 

\section{A First Probabilistic Program Evaluator}


It is also worth establishing at least a vague idea about how one might evaluate such a program so as to produce an {\em inference} result.  To reiterate, evaluating this program requires performing the same kind of inference we described mathematically earlier in this chapter, not just running the program forward.  Here specifically this means to characterize the distribution of the return value \ang{x} conditioned on the observed value \ang{y}.  

We have found it generally helpful when teaching probabilistic programming to beginners to describe, in words, right at the outset, the simplest way one can go about writing a PPL evaluator that does inference.  To start, consider simply ``running the program forward.''  When this is done, as in running any program, some kind of ``state of the computer'' is usually modified with every function invocation.  Here \ang{(beta a b)} is a function call that creates a distribution objection called \ang{prior}, \ang{(sample prior)} is a function call that creates \ang{x}, a number between zero and one, and so forth.  The return value of this program is the value of \ang{x}.  However you implement an evaluator (interpreter) for this kind of program it will maintain memory slots that keep track of the values of \ang{prior}, \ang{x}, \ang{likelihood}, and \ang{y}.  
If we ran this program while ignoring the \ang{observe} statement (i.e., supposing it did nothing), the program would simply return a sample from the marginal distribution over \ang{x}.

A big part of this book is about writing evaluators for PPLs of various kinds.  Since this is case we can pretend that you will be the one writing the PPL evaluator.  In this situation you could decide to track, in your interpreter that evaluates the program in the usual forward way, an additional piece of state, consisting of, say, one extra floating point memory slot.  In this context, a reasonable implementation of \ang{observe} statements is that they only affect this extra memory slot and have no other effect whatsoever.  Consider, in this program, that since \ang{y} is assigned value \ang{1} (``heads'') any \ang{x} that is more likely to generate a heads-up coin flip would be more likely.  You could, in those executions in which the value of \ang{x} preferentially generates heads-up coin flips, put a positive number in this extra memory slot whose value is large when that value of \ang{x} likes to generate heads (and vice versa otherwise).  If you then ran the program many times you would have return-value/extra-memory-slot pairs in which return values of \ang{x} that prefer to generate heads-up coins have higher ``weights'' than those that do not.  By making a sensible choice of what number to put in this slot, such weighted ``samples'' can be made to form an asymptotically exact representation of the true posterior.  This style of interpretation is the subject of Chapter~\ref{ch:eval-one}, particularly Section~\ref{sec:eval-likelihood}, Chapter~\ref{ch:eval-two}, particularly Section~\ref{sec:hoppl-likelihood}, and finally Chapter~\ref{sec:gbli}.  Note that such an interpreter no longer runs the program once but, instead, needs to run the program many times in order to characterize the posterior. 

We start the book, however, not with this kind of evaluator, but instead an evaluator that interprets the program as a specification of a graphical structure on which traditional inference algorithms can be run.  In fact, the first approach to PPL inference does not ``run'' the program at all! So, if you are already familiar with inference in graphical models and factor graphs --- but unfamiliar with programming language design and non-standard interpretations of programs --- we recommend proceeding linearly through the chapters, perhaps skipping inference algorithm implementation details you already know.  If you are relatively unfamiliar with with inference but are comfortable with standard functional program interpretation, you might wish to skim Chapter~\ref{ch:foppl}, then start with Chapter~\ref{ch:eval-one} where we discuss PPL evaluation of the ``run the program'' style just described in a way that will feel familiar and safe. 
If you already know probabilistic inference algorithms by heart, you might be able to read Chapter~\ref{ch:foppl}, Chapter~\ref{ch:hoppl},  Chapter~\ref{ch:eval-two}, particularly Section~\ref{sec:addressing}, and Chapter~\ref{sec:gbli} to extract just the programming language design considerations specific to PPLs and how they dictate what kind of inference algorithms can be used.  

In any event we hope that you enjoy the following and learn something useful from it.

\chapter{A Probabilistic Language Without Recursion}
\label{ch:foppl}

%
%


In this and the next two chapters of this introduction we will present the key ideas of probabilistic programming using a carefully designed first-order probabilistic programming language (FOPPL). The FOPPL includes most common features of programming languages, such as conditional statements (e.g.~\fop{if}), primitive operations (e.g.~\fop{+},\fop{-}, etc.), and user-defined functions. The restrictions that we impose are that functions must be first order, which is to say that functions cannot accept other functions as arguments, and that they cannot be recursive.

These two restrictions result in a language where models describe distributions over a finite number of random variables. In terms of expressivity, this places the FOPPL on even footing with many existing languages and libraries for automating inference in graphical models with finite graphs. As we will see in Chapter \ref{ch:graph-based}, we can compile any program in the FOPPL to a data structure that represents the corresponding graphical model.  This turns out to be a very useful property when reasoning about inference, since it allows us to make use of existing theories and algorithms for inference in graphical models.

A corollary to this characteristic is that the computation graph of any FOPPL program can be completely determined in advance.  This suggests a place for FOPPL programs in the spectrum between static and dynamic computation graph programs.  While in a FOPPL program conditional branching might dictate that not all of the nodes of its computation graph are active in the sense of being on the control-flow path, it is the case that all FOPPL programs can be unrolled to computation graphs where all possible control-flow paths are explicitly and completely enumerated at compile time.  FOPPL programs have static computation graphs.



Although we have endeavored to make this tutorial as self-contained as possible, readers unfamiliar with graphical models or wishing to brush up on them are encouraged to refer to the textbooks by \citet{bishop2006pattern}, \citet{murphy2012machine}, or \citet{koller2009probabilistic}, all of which contain a great deal of material on graphical models and associated inference algorithms.

\section{Syntax}

\begin{grammar}[mathescape,caption={First-order probabilistic programming language (FOPPL)},label=lang:foppl,float=tp,floatplacement=tbp]
  $v ::=$ $\textrm{variable}$
  $c ::=$ $\textrm{constant value or primitive operation}$ 
  $f ::=$ $\textrm{procedure}$
  $e ::=$ $c$ | $v$ | (let [$v$ $e_1$] $e_2$) |  (if $e_1$ $e_2$ $e_3$) 
     | ($f$ $e_1$ $\ldots$ $e_n$) | ($c$ $e_1$ $\ldots$ $e_n$) 
     | (sample $e$) | (observe $e_1$ $e_2$) 
  $q ::=$ $e$ | (defn $f$ [$v_1$ $\ldots$ $v_n$] $e$) $q$
\end{grammar}

The FOPPL is a Lisp variant that is based on Clojure
\citep{hickey2008clojure}. 
The syntax of the FOPPL is specified by the
grammar in Language~\ref{lang:foppl}. A grammar like this formulates a set of
production rules, which are recursive, from which all valid programs must be
constructed. 

We define the FOPPL in terms of two sets of production rules: one
for expressions $e$ and another for programs $q$. Each set of rules is shown
on the right hand side of \fop{$::=$} separated by a \fop{$|$}. We will here provide a very brief self-contained explanation of each of the production rules. For those who wish to read about programming languages essentials in further detail, we recommend the books by \citet{abelson1996structure} and \citet{friedman2008essentials}.

The rules for $q$ state that a program can either be a single expression $e$, or a function declaration \fop{(defn$\;f\;\ldots$)} followed by any valid program $q$. Because the second rule is recursive, these two rules together state that a program is a single expression $e$ that can optionally be preceded by one or more function declarations.

The rules for expressions $e$ are similarly defined recursively. For example, in the production rule \fop{(if $e_1$ $e_2$ $e_3$)}, each of the sub-expressions \fop{$e_1$}, \fop{$e_2$}, and \fop{$e_3$} can be expanded by choosing again from the matching rules on the left hand side. 
The FOPPL defines eight expression types. The first six are ``standard'' in the sense that they are commonly found in non-probabilistic Lisp variants:
\begin{itemize}
\item[1.] A constant $c$ can be a value of a primitive data type such as a number, a string, or a boolean, a built-in primitive function such as \fop{+}, or a value of any other data type that can be constructed using primitive procedures, such as lists, vectors, maps, and distributions, which we will briefly discuss below.
\item[2.] A variable $v$ is a symbol that references the value of another expression in the program.
\item[3.] A let form \fop{(let [$v$ $e_1$] $e_2$)} binds the value of  the expression $e_1$ to the variable $v$, which can then be referenced in the expression $e_2$, which is often referred to as the body of the let expression.
\item[4.] An if form \fop{(if $e_1$ $e_2$ $e_3$)} takes the value of $e_2$ when the value of $e_1$ is logically true and the value of $e_3$ when $e_1$ is logically false.
\item[5.] A function application \fop{($f$ $e_1$ $\ldots$ $e_n$)} calls the user-defined function $f$, which we also refer to as a procedure, with arguments $e_1$ through $e_n$. Here the notation $e_1 \ldots e_n$ refers to a variable-length sequence of arguments, which includes the case \fop{($f$)} for a procedure call with no arguments.
\item[6.] A primitive procedure applications \fop{($c$ $e_1$ $\ldots$ $e_n$)} calls a built-in function $c$, such as \fop{+}.
\end{itemize}
The remaining two forms are what makes the FOPPL a probabilistic programming language:
\begin{itemize}
\item[7.] A sample form \fop{(sample $e$)} represents an unobserved random variable. It accepts a single expression $e$, which must evaluate to a distribution object, and returns a value that is a sample from this distribution. Distributions are constructed using primitives provided by the FOPPL. For example, \fop{(normal 0.0 1.0)} evaluates to a standard normal distribution.
\item[8.] An observe form \fop{(observe $e_1$ $e_2$)} represents an observed random variable. It accepts an argument $e_1$, which must evaluate to a distribution, conditions on the next argument $e_2$, which is the value of the random variable, and returns the value of $e_2$.
\end{itemize}

This language is simple; the grammar only has a small number of special forms. It also has no input/output functionality, which means that all data must be inlined in the form of an expression. However, despite this relative simplicity, we will see that we can express any graphical model as a FOPPL program. At the same time, the relatively small number of expression forms makes it much easier to reason about implementations of compilation and evaluation strategies.

Relative to other Lisp variants, the property of the FOPPL that is most critical for our purposes is that it is a first-order language. Provided that all primitives halt on all possible inputs, potentially non-halting computations are disallowed; for any program, there is a finite upper bound on the number of computation steps and this upper bound can be determined at compilation time. This design choice has several consequences. The first is that all data needs to be inlined so that the number of data points is known at compile time. A second consequence is that FOPPL grammar precludes higher-order functions, which is to say that user-defined functions cannot accept other functions as arguments. The reason for this is that a reference to user-defined function $f$ is in itself not a valid expression type. Since arguments to a function call must be expressions, this means that we cannot pass a function $f'$ as an argument to another function $f$. 

Finally, the FOPPL does not allow recursive function calls, although the  syntax does not forbid them. This restriction can be enforced via the scoping rules in the language. In a program $q$ of the form
\begin{foppl}[mathescape]
    (defn $f_1$ $\ldots$) (defn $f_2$ $\ldots$) $e$
\end{foppl}
we can call $f_1$ inside of $f_2$, but not vice versa, since $f_2$ is defined after $f_1$. Similarly, we impose the restriction that we cannot call $f_1$ inside $f_1$, which we can intuitively think of as $f_1$ not having been defined yet. Enforcing this restriction can be done using a pre-processing step.

A second property that differentiates the FOPPL from most Lisps is that it includes vector and map data structures, analogous to the ones provided by Clojure.
\begin{itemize}
  \item[-] Vectors are similar to lists. Vectors are constructed by the 
  expression \fop{(vector $e_1$ $\ldots$ $e_n$)}, which we will abbriate as\fop{[$e_1$ $\ldots$ $e_n$]}. For example, we can use \fop{[1 2]} to represent a pair, rather than the more cumbersome  the expression \fop{(vector 1 2)} or \fop{(list 1 2)}.
  \item[-] Hash maps \fop{(hash-map $e_1$ $e'_1$ $\ldots$ $e_n$ $e'_n$)} are constructed from a sequence of key-value pairs $e_i~e'_i$. A hash-map can be represented with the literal \mfop{\{$e_1$\ $e^\prime_1$\ $\ldots$\ $e_n$\ $e^\prime_n$\}}.
\end{itemize}


Note that we have not explicitly enumerated primitive functions in the FOPPL. We will implicitly assume existence of arithmetic primitives like \fop{+}, \fop{-}, \fop{*}, and \fop{/}, as well as distribution primitives like \fop{normal} and \fop{discrete}. In addition we will assume the following functions for interacting with data structures 
\begin{itemize}
  \item \fop{(first $e$)} retrieves the first element of a list or vector $e$.
  \item \fop{(rest $e$)} returns a list or vector containing the second to last elements of a list or vector $e$.
  \item \fop{(last $e$)} retrieves the last element of a list or vector $e$.
  \item \fop{(append $e_1$ $e_2$)} appends $e_2$ to the end of a list or vector $e_1$.\footnote{Readers familiar with Lisp dialects will notice that \fop{append} differs somewhat from the semantics of primitives like \fop{cons}, which prepends to a list, or the Clojure primitive \fop{conj} which prepends to a list and appends to a vector.}
  \item \fop{(get $e_1$ $e_2$)} retrieves an element at index $e_2$ from a list or vector $e_1$, or the element at key $e_2$ from a hash map $e_1$. 
  \item \fop{(put $e_1$ $e_2$ $e_3$)} replaces the element at index/key $e_2$ with the value $e_3$ in a vector or hash-map $e_1$.  
  \item \fop{(remove $e_1$ $e_2$)} removes the element at index/key $e_2$ in a vector or hash-map $e_1$.  
\end{itemize}

Note that primitive procedures in the FOPPL are pure functions. In other words, the \fop{append}, \fop{put}, and \fop{remove} primitives do not modify $e_1$ in place, but instead return a modified copy of $e_1$. Efficient implementations of such functionality may be advantageously achieved via pure functional data structures \citep{okasaki1999purely}.

Finally we note that we have not specified any type system or specified exactly what values are allowable in the language. For example, \hop{(sample e)} will fail if at runtime \hop{e} does not evaluate to a distribution.

\begin{foppl}[mathescape,caption={Bayesian linear regression in the FOPPL.},label={foppl:linreg},float=t,floatplacement=t]
(defn observe-data [slope intercept x y]
  (let [fx (+ (* slope x) intercept)]
    (observe (normal fx 1.0) y)))

(let [slope (sample (normal 0.0 10.0))]
  (let [intercept  (sample (normal 0.0 10.0))]
    (let [y1 (observe-data slope intercept 1.0 2.1)]
    (let [y2 (observe-data slope intercept 2.0 3.9)]
    (let [y3 (observe-data slope intercept 3.0 5.3)]
    (let [y4 (observe-data slope intercept 4.0 7.7)]
    (let [y5 (observe-data slope intercept 5.0 10.2)]
      [slope intercept]))))))).
\end{foppl}

Now that we have defined our syntax, let us illustrate what a program in the FOPPL looks like. Program~\ref{foppl:linreg} shows a simple univariate linear regression model.
The program defines a distribution on lines expressed in terms of their slopes and intercepts by first defining a prior distribution on slope and intercept and then conditioning it using five observed data pairs. The procedure \fop{observe-data}  conditions the generative model given a pair (\fop{x},\fop{y}), by observing the value \fop{y} from a normal centered around the value \fop{(+ (* slope x) intercept)}. Using a procedure lets us avoid rewriting observation code for each observation pair. The procedure returns the observed value, which is ignored in our case.  The program defines a prior on \fop{slope} and \fop{intercept} using the primitive procedure \fop{normal} for creating an object for normal distribution. After conditioning this prior with data points, the program return a pair \fop{[slope intercept]}, which is a sample from the posterior distribution conditioned on the 5 observed values.

\section{Syntactic Sugar}

The fact that the FOPPL only provides a small number of expression types is a big advantage when building a probabilistic programming system. We will see this in Chapter~\ref{ch:graph-based}, where we will define a translation from any FOPPL program to a Bayesian network using only 8 rules (one for each expression type). At the same time, for the purposes of writing probabilistic programs, having a small number of expression types is not always convenient. For this reason we will provide a number of alternate expression forms, which are referred to as syntactic sugar, to aid readability and ease of use. 

We have already seen two very simple forms of syntactic sugar: \fop{[$\ldots$]} is a sugared form of \fop{(vector $\ldots$)} and \mang{\{$\ldots$\}} is a sugared form for \fop{(hash-map $\ldots$)}. In general, each sugared expression form can be desugared, which is to say that it can be reduced to an expression in the grammar in Language~\ref{lang:foppl}. This desugaring is done as a preprocessing step, often implemented as a macro rewrite rule that expands each sugared expression into the equivalent desugared form. 

\subsection{Let forms}

The base let form \fop{(let [$v$ $e_1$] $e_2$)} binds a single variable $v$ in the expression $e_2$. Very often, we will want to define multiple variables, which leads to nested let expressions like the ones in Program~\ref{foppl:linreg}. Another distracting piece of syntax in this program is that we define dummy variables \fop{y1} to \fop{y5} which are never used. The reason for this is that we are not interested in the values returned by calls to \fop{observe-data}; we are using this function in order to observe values, which is a side-effect of the procedure call. 

To accommodate both these use cases in let forms, we will make use of the following generalized let form
\begin{foppl}[mathescape]
(let [$v_1$ $e_1$
      $\vdots$
      $v_n$ $e_n$]
  $e_{n+1}$ $\ldots$ $e_{m-1}$ $e_m$).
\end{foppl}
This allows us to simplify the nested let forms in Program~\ref{foppl:linreg} to
\begin{foppl}[mathescape]
(let [slope (sample (normal 0.0 10.0))
      intercept (sample (normal 0.0 10.0))]
  (observe-data slope intercept 1.0 2.1)
  (observe-data slope intercept 2.0 3.9)
  (observe-data slope intercept 3.0 5.3)
  (observe-data slope intercept 4.0 7.7)
  (observe-data slope intercept 5.0 10.2)
  [slope intercept])
\end{foppl}
This form of \fop{let} is desugared to the following expression in the FOPPL
\begin{foppl}[mathescape]
(let [$v_1$ $e_1$]
  $\vdots$
  (let [$v_n$ $e_n$] 
    (let [_ $e_{n+1}$]
       $\vdots$
       (let [_ $e_{m-1}$]
         $e_m$)$\cdots$))).
\end{foppl}
Here the underscore \fop{_} is a second form of syntactic sugar that will be expanded to a fresh (i.e. previously unused) variable. For instance
\begin{foppl}[mathescape]
(let [_ (observe (normal 0 1) 2.0)] $\ldots$)
\end{foppl}
will be expanded by generating some {\em fresh variable} symbol, say \fop{x284xu},
\begin{foppl}[mathescape]
(let [x284xu (observe (normal 0 1) 2.0)] $\ldots$)
\end{foppl}
We will assume each instance of \fop{_} is a guaranteed-to-be-unique or fresh symbol that is generated by some \fop{gensym} primitive in the implementing language of the evaluator. We will use the concept of a fresh variable extensively throughout this tutorial, with the understanding that fresh variables are unique symbols in all cases.


\subsection{For loops}

A second syntactic inconvenience in Program~\ref{foppl:linreg} is that we have to repeat the expression \fop{(observe-data $\ldots$)} once for each data point. Just about any language provides looping constructs for this purpose. In the FOPPL we will make use of two such constructs. The first is the \fop{foreach} form, which has the following syntax

\begin{foppl}[mathescape]
(foreach $c$ 
  [$v_1$ $e_1$ $\ldots$ $v_n$ $e_n$]
  $e'_1$ $\ldots$ $e'_k$)
\end{foppl}
Where $c$ is a non-negative integer constant. This form desugars into a vector containing $c$ let forms
\begin{foppl}[mathescape]
(vector
  (let [$v_1$ (get $e_1$ 0) $\ldots$ $v_n$ (get $e_n$ 0)]
    $e'_1$ $\ldots$ $e'_k$)
  $\vdots$
  (let [$v_1$ (get $e_1$ (- $c$ 1)) $\ldots$ $v_n$ (get $e_n$ (- $c$ 1))]
    $e'_1$ $\ldots$ $e'_k$))
\end{foppl}
Note that this syntax looks very similar to that of the \fop{let} form. However, whereas \fop{let} binds each variable to a single value, the \fop{foreach} form associates each variable $v_i$ with a sequence $e_i$ and then maps over the values in this sequence for a total of $c$ steps, returning a vector of results. If the length of any of the bound sequences is less than $c$, then let form will result in a runtime error. 

With the foreach form, we can rewrite Program~\ref{foppl:linreg} without having to make use 
of the helper function \fop{observe-data}
\begin{foppl}[mathescape]
(let [y-values [2.1 3.9 5.3 7.7 10.2]
      slope (sample (normal 0.0 10.0))
      intercept (sample (normal 0.0 10.0))]
  (foreach 5 
    [x (range 1 6)
     y y-values]
    (let [fx (+ (* slope x) intercept)]
      (observe (normal fx 1.0) y)))
  [slope intercept])
\end{foppl}

There is a very specific reason why we defined the foreach syntax using a constant for the number of loop iterations \fop{(foreach $c$ [$\ldots$] $\ldots$)}. Suppose we were to define the syntax using an arbitrary expression \fop{(foreach $e$ [$\ldots$] $\ldots$)}. Then we could write programs such as
\begin{foppl}[mathescape]
(let [m (sample (poisson 10.0))]
  (foreach m []
    (sample (normal 0 1))))
\end{foppl}
This defines a program in which there is no upper bound on the number of times that the expression \fop{(sample (normal 0 1))} will be evaluated. By requiring  $c$ to be a  constant, we can guarantee that the number of iterations is known at compile time.

Note that there are less obtrusive mechanisms for achieving the functionality of \fop{foreach}, which is fundamentally a language feature that maps a function, here the body, over a sequence of arguments, here the \fop{let}-like bindings.  Such functionality is much easier to express and implement using higher-order language features like those discussed in Chapter~\ref{ch:hoppl}.



\subsection{Loop forms}

The second looping construct that we will use is the loop form, which has the following syntax.
\begin{foppl}[mathescape]
(loop $c$ $e$ $f$ $e_1$ $\ldots$ $e_n$)
\end{foppl}
Once again, $c$ must be a non-negative integer {\em constant} and $f$ a procedure, primitive or user-defined. 
This notation can be used to write most kinds of
 for loops. Desugaring this syntax rolls out a nested set of lets and function calls in the following manner
\begin{foppl}[mathescape]
(let [$a_1$ $e_1$
      $a_2$ $e_2$
         $\vdots$
      $a_n$ $e_n$]      
  (let [$v_0$ ($f$ 0 $e$ $a_1$ $\ldots$ $a_n$)]
    (let [$v_1$ ($f$ 1 $v_0$ $a_1$ $\ldots$ $a_n$)]
      (let [$v_2$ ($f$ 2 $v_1$ $a_1$ $\ldots$ $a_n$)]
        $\vdots$
          (let [$v_{c-1}$ ($f$ (- $c$ 1) $v_{c-2}$ $a_1$ $\ldots$ $a_n$)]
            $v_{c-1}$) $\cdots$ )))
\end{foppl}
where $v_0, \ldots, v_{c-1}$ and $a_0, \ldots, a_n$ are fresh variables. Note that the \fop{loop} sugar computes an iteration over a fixed set of indices. 

To illustrate how the \fop{loop} form differs from the \fop{foreach} form, we show a new variant of the linear regression example in Program~\ref{example:linear-regression-loop}. In this version of the program, we not only observe a sequence of values $y_n$ according to a normal centered at $f(x_n)$, but we also compute the sum of the squared residuals $r^2 = \sum_{n=1}^5 (y_n - f(x_n))^2$. To do this, we define a function \fop{regr-step}, which accepts an argument \fop{n}, the index of the loop iteration. It also accepts a second argument \fop{r2}, which represents the sum of squares for the preceding datapoints. Finally it accepts the arguments \fop{xs}, \fop{ys}, \fop{slope}, and \fop{intercept}, which we have also used in previous versions of the program. 

\begin{foppl}[mathescape,caption={The Bayesian linear regression model, written using the loop form.},label=example:linear-regression-loop,float=t,floatplacement=t]
(defn regr-step [n r2 xs ys slope intercept]
  (let [x (get xs n)
        y (get ys n)
        fx (+ (* slope x) intercept)
        r (- y fx)]
    (observe (normal fx 1.0) y)
    (+ r2 (* r r))))

(let [xs [1.0 2.0 3.0 4.0  5.0]
      ys [2.1 3.9 5.3 7.7 10.2]
      slope (sample (normal 0.0 10.0))
      bias  (sample (normal 0.0 10.0))
      r2 (loop 5 0.0 regr-step xs ys slope bias)]
  [slope bias r2])
\end{foppl}

At each loop iteration, the function \fop{regr-step} computes the residual $r = y_n - f(x_n)$ and returns the value \fop{(+ r2 (* r r))}, which becomes the new value for \fop{r2} at the next iteration. The value of the entire loop form is the value of the final call to \fop{regr-step}, which is the sum of squared residuals. 

The difference between \fop{loop} and \fop{foreach} is that \fop{loop} can be used to accumulate a result over the course of the iterations. This is useful when you want to compute some form of sufficient statistics, filter a list of values, or really perform any sort of computation that iteratively builds up a data structure. The \fop{foreach} form provides a much more specific loop type that evaluates a single expression repeatedly with different values for its variables. From a statistical point of view, we can think of \fop{loop} as defining a sequence of dependent variables, whereas \fop{foreach} creates variables that are conditionally independent given variables that are defined before the start of the loop.




\section{Examples}
\label{sec:foppl-examples}


Now that we have defined the fundamental expression forms in the FOPPL, along with syntactic sugar for variable bindings and loops, let us look at how we would use the FOPPL to define some models that are commonly used in statistics and machine learning.

\subsection{Gaussian mixture model}

We will begin with a three-component Gaussian mixture model \citep{mclachlan2004finite}. A Gaussian mixture model is a density estimation model often used for clustering, in which each data point $y_n$ is assigned to a latent class $z_n$. We will here consider the following generative model

\begin{foppl}[mathescape,label=model:gmm,caption={FOPPL - Gaussian mixture model with three components},float=t,floatplacement=t]
(let [data [1.1 2.1 2.0 1.9 0.0 -0.1 -0.05]
      likes (foreach 3 []
              (let [mu (sample (normal 0.0 10.0))
                    sigma (sample (gamma 1.0 1.0))]
                (normal mu sigma)))
      pi (sample (dirichlet [1.0 1.0 1.0]))
      z-prior (discrete pi)]
  (foreach 7 [y data] 
    (let [z (sample z-prior)]
      (observe (get likes z) y)
      z)))
\end{foppl}

\begin{align}
  \sigma_k 
  &\sim 
  \text{Gamma}(1.0, 1.0),
  &&
  {\text{for}~k = 1,2,3},
  \\
  \mu_k 
  &\sim 
  \text{Normal}(0.0, 10.0),
  &&
  {\text{for}~k = 1,2,3},
  \\
  \pi 
  &\sim 
  \text{Dirichlet}(1.0, 1.0, 1.0) ,
  \\
  z_n 
  &\sim 
  \text{Discrete}(\pi),
  &&
  {\text{for}~n = 1,\ldots,7},
  \\
  y_n | z_n=k
  &\sim 
  \text{Normal}(\mu_k, \sigma_k).
\end{align}

Program~\ref{model:gmm} shows a translation of this generative model to the FOPPL. In this model we first sample the 
mean \fop{mu} and standard deviation \fop{sigma} for 3 mixture components. For each observation \fop{y} we then sample a class assignment \fop{z}, after which we observe according to the likelihood of the sampled assignment. The return value from this program is the sequence of latent class assignments, which can be used to ask questions like, ``Are these two datapoints similar?'', etc.




\subsection{Hidden Markov model}

\begin{foppl}[mathescape,label=model:hmm,caption={FOPPL - Hidden Markov model},float=t,floatplacement=t]
(defn hmm-step [t states data trans-dists likes]
  (let [z (sample (get trans-dists 
                       (last states)))]
    (observe (get likes z) 
             (get data t))
    (append states z)))

(let [data [0.9 0.8 0.7 0.0 -0.025 -5.0 -2.0 -0.1 
            0.0 0.13 0.45 6 0.2 0.3 -1 -1]
      trans-dists [(discrete [0.10 0.50 0.40])
                   (discrete [0.20 0.20 0.60])
                   (discrete [0.15 0.15 0.70])]
      likes [(normal -1.0 1.0)
             (normal 1.0 1.0)
             (normal 0.0 1.0)]
      states [(sample (discrete [0.33 0.33 0.34]))]]
  (loop 16 states hmm-step 
    data trans-dists likes))
\end{foppl}

As a second example, let us consider Program~\ref{model:hmm} which denotes
a hidden Markov model (HMM) \citep{rabiner1989tutorial} with known initial state, transition, and observation distributions governing $16$ sequential observations. 

In this program we begin by defining a vector of data points \fop{data}, a vector of transition distributions \fop{trans-dists} and a vector of state likelihoods \fop{likes}. We then loop over the data using a function \fop{hmm-step}, which returns a sequence of states. 

At each loop iteration, the function \fop{hmm-step} does three things. It first samples a new state \fop{z} from the transition distribution associated with the preceding state. 
It then observes data point at time \fop{t} according to the likelihood component of the current state. 
Finally, it appends the state \fop{z} to the sequence \fop{states}. 
The vector of accumulated latent states is the return value of the program and thus the object whose joint posterior distribution is of interest.

\subsection{A Bayesian Neural Network}

\begin{foppl}[mathescape,caption={FOPPL - A Bayesian Neural Network},label=foppl:bnn_edward,float=tbp,floatplacement=thbp]
(let [weight-prior (normal 0 1)
      W_0 (foreach 10 []
            (foreach 1 [] (sample weight-prior)))
      W_1 (foreach 10 []
            (foreach 10 [] (sample weight-prior)))
      W_2 (foreach 1 []
            (foreach 10 [] (sample weight-prior)))

      b_0 (foreach 10 []
            (foreach 1 [] (sample weight-prior)))
      b_1 (foreach 10 []
            (foreach 1 [] (sample weight-prior)))
      b_2 (foreach 1 []
            (foreach 1 [] (sample weight-prior)))

      x   (mat-transpose [[1] [2] [3] [4] [5]])
      y   [[1] [4] [9] [16] [25]]
      h_0 (mat-tanh (mat-add (mat-mul W_0 x)
                             (mat-repmat b_0 1 5)))
      h_1 (mat-tanh (mat-add (mat-mul W_1 h_0)
                             (mat-repmat b_1 1 5)))
      mu  (mat-transpose
             (mat-tanh 
               (mat-add (mat-mul W_2 h_1)
                        (mat-repmat b_2 1 5))))]
   (foreach 5 [y_r y
               mu_r mu]
     (foreach 1 [y_rc y_r
                 mu_rc mu_r]
       (observe (normal mu_rc 1) y_rc)))
   [W_0 b_0 W_1 b_1])
\end{foppl}

%

Traditional neural networks are fixed-dimension computation graphs which means that they too can be expressed in the FOPPL.  In the following we demonstrate this with an example taken from the documentation for Edward~\citep{tran2016edward}, 
a probabilistic programming library based on fixed computation graph. The example shows a Bayesian approach to learning the parameters of a three-layer neural network with input of dimension one, two hidden layers of dimension ten, an independent and identically Gaussian distributed output of dimension one, and \fop{tanh} activations at each layer.  The program inlines five data points and represents the posterior distribution over the parameters of the neural network.  We have assumed, in this code, the existence of matrix primitive functions, e.g. \fop{mat-mul}, whose meaning is clear from context (matrix multiplication), sensible matrix-dimension-sensitive pointwise \fop{mat-add} and \fop{mat-tanh} functionality, vector of vectors matrix storage, etc.  

This example provides an opportunity to reinforce the close relationship between optimization and inference.  The task of estimating neural-network parameters is typically framed as an optimization in which the free parameters of the network are adjusted, usually via gradient descent, so as to minimize a loss function.  This neural-network example can be seen as doing parameter learning too, except using the tools of inference to discover the posterior distribution over model parameters.  In general, all parameter estimation tasks can be framed as inference simply by placing a prior over the parameters of interest as we do here.

It can also be noted that, in this setting, any of the activations of the neural network trivially could  be made stochastic, yielding a stochastic computation graph \citep{schulman2015gradient}, rather than a purely deterministic neural network.  

Finally, the point of this example is not to suggest that the FOPPL is {\em the} language that should be used for denoting neural network learning and inference problems, it is instead to show that the FOPPL is sufficiently expressive to neural networks based on fixed computation graphs.  Even though we have shown only one example of a multilayer perceptron, it is clear that convolutional neural networks, recurrent neural networks of fixed length, and the like, can all be denoted in the FOPPL.

\subsection{Translating BUGS models}


\begin{bugs}[caption={The Pumps example model from BUGS \citep{BUGSpumpsmodel}.},float=t, floatplacement=t, label=bugs:pumps]
# data
list(t = c(94.3, 15.7, 62.9, 126, 5.24,   
           31.4, 1.05, 1.05, 2.1, 10.5),
     y = c(5, 1, 5, 14, 3, 19, 1, 1, 4, 22), 
     N = 10)
# inits
list(a = 1, b = 1)
# model
{
  for (i in 1 : N) {
     theta[i] ~ dgamma(a, b)
     l[i] <- theta[i] * t[i]
     y[i] ~ dpois(l[i])
  }
  a ~ dexp(1)
  b ~ dgamma(0.1, 1.0)
} 
\end{bugs}

\begin{foppl}[mathescape,caption={FOPPL - the Pumps example model from BUGS},float=t, floatplacement=t, label=foppl:pumps]
(defn data []
  [[94.3 15.7 62.9 126 5.24 31.4 1.05 1.05 2.1 10.5]
   [5 1 5 14 3 19 1 1 4 22]
   [10]])

(defn t [i] (get (get (data) 0) i))
(defn y [i] (get (get (data) 1) i))

(defn loop-iter [i _ alpha beta]
  (let [theta (sample (gamma a b))
        l     (* theta (t i))]
    (observe (poisson l) (y i))))

(let [a (sample (exponential 1))
      b (sample (gamma 0.1 1.0))]
  (loop 10 nil loop-iter a b)
  [a b])
\end{foppl}

The FOPPL language as specified is sufficiently expressive to, for instance, compile BUGS programs to the FOPPL. Program~\ref{bugs:pumps} shows one of the examples included with the BUGS system \citep{BUGSpumpsmodel}. This model is a conjugate gamma-Poisson hierarchical model, which is to say that it has the following generative model:
\begin{align}
  a &\sim \text{Exponential}(1),
  \\
  b &\sim \text{Gamma}(0.1, 1),
  \\
  \theta_i &\sim \text{Gamma}(a, b),
  &&
  \text{for~} i = 1,\ldots,10,
  \\
  y_i &\sim \text{Poisson}(\theta_i t_i)
  &&
  \text{for~} i = 1,\ldots,10.
\end{align}


Program~\ref{bugs:pumps} shows this model in the BUGS language. Program~\ref{foppl:pumps} show a translation to the FOPPL that was returned by an automated BUGS-to-FOPPL compiler. Note the similarities between these languages despite the substantial syntactic differences. In particular, both require that the number of loop iterations $N=10$ is fixed and finite. In BUGS the variables whose values are known appear in a separate data block. The symbol $\sim$ is used to define random variables, which can be either latent or observed, depending on whether a value for the random variable is present.
In our FOPPL the distinction between observed and latent random variables is made explicit through the syntactic difference between sample and observe. A second difference is that a BUGS program can in principle be used to compute a marginal on any variable in the program, whereas a FOPPL program specifies a marginal of the full posterior through its return value. As an example, in this particular translation, we treat $\theta_i$ as a nuisance variable, which is not returned by the program, although we could have used the loop construct to accumulate a sequence of $\theta_i$ values. 

These minor differences aside, the BUGS language and the FOPPL essentially define equivalent families of probabilistic programs. An advantage of writing this text using the FOPPL rather than an existing language like BUGS is that FOPPL program are comparatively easy to reason about and manipulate, since there are only 8 expression forms in the language. In the next chapter we will exploit this in order to mathematically define a translation from FOPPL programs to Bayesian networks and factor graphs, keeping in mind that all the basic concepts that we will employ also apply to other probabilistic programming systems, such as BUGS.

\section{A Simple Purely Deterministic Language}
\label{sec:deterministic-expressions}


There is no optimal place to put this section so it appears here, although it is very important for understanding what is written in the remainder of this tutorial.

In subsequent chapters it will become apparent that the FOPPL can be understood in two different ways -- one way as being a language for specifying graphical-model data-structures on which traditional inference algorithms may be run, the other as a language that requires a non-standard interpretation in some implementing language to characterize the denoted posterior distribution.

In the case of graphical-model construction, it will be necessary to have a language for purely deterministic expressions.  This language will be used to express link functions in the graphical model.  More precisely, and contrasting to the usual definition of link function from statistics, the pure deterministic language will encode functions that take values of parent random variables 
and produce distribution objects for children.  
These link functions cannot have random variables inside them; such a variable would be another node in the graphical model instead.

Moreover we can further simplify this link function language by removing user defined functions, effectively requiring their function bodies, if used, to be inlined.  This yields a cumbersome language in which to manually program but an excellent language to target and evaluate because of its simplicity.



We will call expressions in the FOPPL that do not involve user-defined procedure calls and
involve only deterministic computations, e.g.~\fop{(+ (/ 2.0 6.0) 17)} ``0th-order expressions''. Such expressions will play a prominent role when we consider the translation of our probabilistic
programs to graphical models in the next chapter. In order to identify and work with these deterministic 
expressions we define a language with the following extremely simple grammar:
\begin{grammar}[language=Scheme,mathescape,caption={Sub-language for purely deterministic computations},label=lang:fotgt]
   $c ::=$ $\textrm{constant value or primitive operation}$ 
   $v ::=$ $\mathrm{variable}$
   $E ::=$ $c$ | $v$ | (if $E_1$ $E_2$ $E_3$) | ($c$ $E_1 \ldots E_n$) 
\end{grammar}
Note that neither sample nor observe statements appear in the syntax,
and that procedure calls are allowed only for primitive operations, not
for defined procedures. Having these constraints ensures that expressions $E$
cannot depend on any probabilistic choices or conditioning.


%

\vspace{1cm}

The examples provided in this chapter should convince you that many common models and inference problems from statistics and machine learning can be denoted as FOPPL programs.  What remains is to translate FOPPL programs into other mathematical or programming language formalisms whose semantics are well established so that we can define, at least operationally, the semantics of FOPPL programs, and, in so doing, establish in your mind a clear idea about how probabilistic programming languages that are formally equivalent in expressivity to the FOPPL can be implemented.

\chapter{Graph-Based Inference}
\label{ch:graph-based}

	\section{Compilation to a Graphical Model}
	\label{sec:compilation-bounded}


Programs written in the FOPPL specify probabilistic models over finitely many
random variables. In this section, we will make this aspect clear by
presenting the translation of these programs into finite graphical models. 
In the subsequent sections, we will show how this translation can
be exploited to adapt inference algorithms for graphical models to
probabilistic programs.

We specify translation using the following ternary relation $\Downarrow$,
similar to the so called big-step evaluation relation from the 
programming language community.
\begin{align}
        \label{eqn:bigstep} 
        \rho,\phi,e \Downarrow G,E
\end{align}
In this relation, $\rho$ is a mapping from procedure names to their
definitions, $\phi$ is a logical predicate for the flow control context (which will discuss in more detail below), and
$e$ is an expression we intend to compile. This expression is translated to a
graphical model $G$ and an expression $E$ in the deterministic sub-language
described in Section~\ref{sec:deterministic-expressions}. The expression $E$ is deterministic
in the sense that it does not involve sample nor observe. It describes the
return value of the original expression $e$ in terms of random variables in
$G$. Vertices in $G$ represent random variables, and arcs dependencies among
them. For each random variable in $G$, we will define a probability density or
mass in the graph. For observed random variables, we additionally define the
observed value, as well as a logical predicate that indicates whether the
observe expression is on the control flow path, conditioned on the values of
the latent variables.

\subsubsection{Definition of a Graphical Model}

We define a graphical model $G$ as a tuple $(V, A, \m{P}, \m{Y})$ containing (i) a set of
vertices $V$ that represent random variables; (ii) a set of arcs $A \subseteq V
\times V$ (i.e. directed edges) that represent conditional dependencies between
random variables; (iii) a map $\m{P}$ from vertices to deterministic expressions
that specify a probability density or mass function for each random variable;
(iv) a partial map $\m{Y}$ that for each observed random variable contains a deterministic 
expression $E$ for the observed value.

Before presenting a set of translation rules that can be used to compile any
FOPPL program to a graphical model, we will illustrate the intended
translation using a simple example:
\begin{foppl}[mathescape,caption={A simple example FOPPL program.},label={pr:opm}]
  (let [z (sample (bernoulli 0.5))
        mu (if (= z 0) -1.0 1.0)
        d (normal mu 1.0)
        y 0.5]
    (observe d y)
    z)
\end{foppl}
  
\begin{figure}

\centering











\begin{tikzpicture}[x=1.8cm,y=1.2cm]
  \node[obs,
        label=right:\mfop{($p_\mathsf{norm}$\ $y$ (if (=\ $z$ 0) -1.0 1.0) 1.0)}
  ](y){$y$};
  \node[latent, 
        above=of y,
        label=right:\mfop{($p_\mathsf{bern}$\ $z$  0.5)}
  ](z){$z$};
  \edge{z}{y};
\end{tikzpicture}

\caption{The graphical model corresponding to Program~\ref{pr:opm}.}
\label{fig:simple_foppl_example_graphical_model}
\end{figure}
  

This program describes a two-component Gaussian mixture with a single
observation. The program first samples $z$ from a Bernoulli distribution,
based on which it sets a likelihood parameter $\mu$ to $-1.0$ or $1.0$, and
observes a value $y=0.5$ from a normal distribution with mean $\mu$. This program defines a 
joint distribution $p(y=0.5, z)$. The inference problem is then to characterize
the posterior distribution $p(z \,|\, y)$.  Figure~\ref{fig:simple_foppl_example_graphical_model} shows the graphical model and pure deterministic link functions that correspond to 
Program~\ref{pr:opm}.

In the evaluation relation $\rho,\phi,e \Downarrow G,\,E$, the source code of
the program is represented as a single expression $e$. The variable $\rho$ is
an empty map, since there are no procedure definitions. At the top level, the
flow control predicate $\phi$ is $\mfop{true}$. The graphical model
$G=(V, A, \m{P}, \m{Y})$ and the result expression $E$ that this program translates to
are
\begin{align*}
V
=
\{&z, y\},\; 
\\
A
=
\{&(z, y)\},
\\
\m{P}
=\,
[&z \mapsto \mfop{(}p_\mathsf{bern} ~ z ~ \mfop{0.5)},\;
\\
&
y \mapsto \mfop{(}p_\mathsf{norm} ~ y ~ \mfop{(if (= }~z~\mfop{ 0) -1.0 1.0) 1.0)}],\;
\\
\m{Y}
=\,
[&y \mapsto 0.5]
\\
E =\,~ & z
\end{align*}
The vertex set $V$ of the net $G$ contains two variables, whereas the arc set
$A$ contains a single pair $(z,y)$ to mark the conditional dependence
relationship between these two variables. In the map $P$, the probability
mass for $z$ is defined as the target language expression
$\mfop{(}p_\mathsf{bern} ~ z ~ \mfop{0.5)}$. Here $p_\mathsf{bern}$ refers to
a function in the target languages that implements probability mass function
for the Bernoulli distribution. Similarly, the density for $y$ is defined
using $p_\mathsf{norm}$, which implements the probability density function for
the normal distribution. Note that the expression for the program variable
\mfop{mu} has been substituted into the density for $y$. Finally, the map \m{Y} 
contains a single entry that holds the observed value for $y$.

\subsubsection{Assigning Symbols to Variable Nodes}

In the above example we used the symbol $z$ to refer to the
random variable associated with the expression \mfop{(sample (bernoulli 0.5))}
and the symbol $y$ to refer to the observed variable with expression
\mfop{(observe d y)}. In general there will be one node in the network for
each sample and observe expression that is evaluated in a program. In the
above example, there also happens to be a program variable \mfop{z} that holds
the value of the sample expression for node $z$, and a program variable
\mfop{y} that holds the observed value for node $y$, but this is of course
not necessarily always the case.
A particularly common example of this arises in programs that have procedures. Here, the same sample and observe expressions in the procedure body can be evaluated multiple times. Suppose for example that we were to modify our program as follows:
\begin{foppl}[mathescape]%,caption={Programs with sample expressions in procedure calls.},label={pr:opmng}]
(defn norm-gamma
  [m l a b]
  (let [tau (sample (gamma a b))
        sigma (/ 1.0 (sqrt tau))
        mu (sample (normal m (/ sigma (sqrt l)))]
    (normal mu sigma))))

(let [z (sample (bernoulli 0.5))
      d0 (norm-gamma -1.0 0.1 1.0 1.0)
      d1 (norm-gamma 1.0 0.1 1.0 1.0)]
  (observe (if (= z 0) d0 d1) 0.5)
  z)
\end{foppl}
In this version of our program we define two distributions \ang{d0} and
\ang{d1} which are created by sampling a mean \ang{mu} and a precision
\ang{tau} from a normal-gamma prior. We then observe either according to
\ang{d0} or \ang{d1}. Clearly the mapping from program variables to random variables is less obvious here, since each sample expression in the
body of \ang{norm-gamma} is evaluated twice. 

Below, we will define a general set of translation rules that compile a FOPPL
program to a graphical model, in which we assign each vertex in the graphical model a newly generated unique symbol. However, when discussing programs in
this tutorial, we will generally explicitly give names to returns from sample and observe
expressions that correspond to program variables to aid readability.  

Recognize that assigning a label to each vertex is a way of assigning a unique ``address'' 
to each and every random variable in the program.  Such unique addresses are important for the correctness and implementation of generic inference algorithms.  In Chapter~\ref{ch:eval-two}
we develop a more explicit mechanism for addressing in the more difficult situation where not all control flow paths can be completely explored at compile time.

\subsubsection{if-expressions in Graphical Models}

When compiling a program to a graphical model, if-expressions require special
consideration. Before we set out to define translation rules that construct a graphical model for a program, we will first spend some time building intuition about how we would like these translation rules to treat if-expressions. Let us start by considering a simple mixture model, in which only the mean is treated as an unknown variable:
\begin{foppl}[mathescape,caption={A one-point mixture with unknown mean.},label={pr:opm-lazy}]
(let [z (sample (bernoulli 0.5))
      mu (sample (normal (if (= z 0) -1.0 1.0) 1.0))
      d (normal mu 1.0)
      y 0.5]
  (observe d y)
  z)
\end{foppl}
This is of course a really strange way of writing a mixture model. We define a
single likelihood parameter $\mu$, which is either distributed according to
$\text{Normal}(-1,1)$ when $z=0$ and according to $\text{Normal}(1,1)$ when
$z=1$. Typically, we would think of a mixture model as having two components
with parameter $\mu_0$ and $\mu_1$ respectively, where $z$ selects the
component. A more natural way to write the model might be
\begin{foppl}[mathescape,caption={One-point mixture with explicit parameters.},label={pr:opm-eager}]
(let [z (sample (bernoulli 0.5))
      mu0 (sample (normal -1.0 1.0))
      mu1 (sample (normal 1.0 1.0))
      d0 (normal mu0 1.0)
      d1 (normal mu1 1.0)
      y 0.5]
  (observe (if (= z 0) d0 d1) y)
  z)
\end{foppl}
Here we sample parameters $\mu_0$ and $\mu_1$, which then define two component likelihoods \mfop{d0} and \mfop{d1}. The variable $z$ then selects the component likelihood for an observation $y$. 

The second program defines a joint density on four variables $p(y, \mu_1, \mu_0, z)$, whereas the first program defines a density on three variables $p(y, \mu, z)$. However, it seems intuitive that these programs are equivalent in some sense. The equivalence that we would want to achieve here is that both programs define the same marginal posterior on $z$
\begin{align*}
  p(z \,|\, y) = \int p(z,\mu \,|\, y)d\mu  = \int \int p(z, \mu_0, \mu_1 \,|\, y)d\mu_0d\mu_1.
\end{align*}
So is there a difference between these two programs when both return \ang{z}? The second program of course defines additional intermediate variables \mfop{d0} and \mfop{d1}, but these do not change the set of nodes in the corresponding graphical model. The essential difference is that in the first program, the if-expression is placed \emph{inside} the sample expression for \ang{mu}, whereas in the second it sits \emph{outside}. If we wanted to make the first program as similar as possible to the second, then we could write
\begin{foppl}[mathescape,caption={One-point mixture with explicit parameters simplified.},label={pr:opm-eager-simplified}]
(let [z (sample (bernoulli 0.5))
      mu0 (sample (normal -1.0 1.0))
      mu1 (sample (normal 1.0 1.0))
      mu (if (= z 0) mu0 mu1)
      d (normal mu 1.0)
      y 0.5]
  (observe d y)
  z)
\end{foppl}
Because we have moved the if-expression, we now need two sample expressions rather than one, resulting in a network with 4 nodes rather than 3. However, the  distribution on return values remains the same.

This brings us to what turns out to be a fundamental design choice in probabilistic programming systems. Suppose we were to modify the above program to read
\begin{foppl}[mathescape,caption={One-point mixture with samples inside if.},label={pr:opm-if-sample}]
(let [z (sample (bernoulli 0.5))
      mu (if (= z 0) 
           (sample (normal -1.0 1.0)) 
           (sample (normal 1.0 1.0)))
      d (normal mu 1.0)
      y 0.5]
  (observe d y)
  z)
\end{foppl}
Is this program now equivalent to the first program, or to the second? 
The answer to this question depends on how we evaluate if-expressions in our language. 

In almost all mainstream programming languages, if-expressions are evaluated in a lazy manner. In the example above, we would first evaluate the predicate \mfop{(= z 0)}, and then either evaluate the consequent branch, \mfop{(sample (normal -1.0 1.0))}, or the alternative branch, \mfop{(sample (normal 1.0 1.0))}, but never both. The opposite of a lazy evaluation strategy is an eager evaluation strategy. In eager evaluation, an if-expression is evaluated like a normal function call. We first evaluate the predicate and both branches. We then return the value of one of the branches based on the predicate value.

If we evaluate if-expressions lazily, then the program above is more similar to Program~\ref{pr:opm-lazy}, in the sense that the program evaluates two sample expressions. If we evaluate if-expressions eagerly, then the program evaluates three sample expressions and is therefore equivalent to Program~\ref{pr:opm-eager-simplified}. As it turns out, both evaluation strategies offer certain advantages.

Suppose that we use $\mu_0$ and $\mu_1$ to refer to the sample expressions in each branch of Program~\ref{pr:opm-if-sample}. Then the joint  $p(y,\mu_0,\mu_1,z)$ would have a conditional
dependence structure\footnote{It might be tempting to instead define a
distribution $p(y,\mu,z)$ as in the first program, by interpreting the entire
if expression as a single random variable $\mu$. For this particular example
this would work, since both branches sample from a normal distribution.
However, if we were, for example, to modify the $z=1$ branch to sample from a
Gamma distribution instead of a normal, then $\mu \in (-\infty,\infty)$ when
$z=0$ and $\mu \in (0, \infty)$ when $z=1$, which means that the variable
$\mu$ would no longer have a well-defined support.}
\begin{align*}
  p(y, \mu_0, \mu_1, z) 
  =~
  &
  p(y \,|\, \mu_0, \mu_1, z)
  p(\mu_0 | z)
  p(\mu_1 | z)
  p(z)
  .
\end{align*}
Here the likelihood $p(y| \mu_0, \mu_1, z)$ is relatively easy to define,
\begin{align}
  \label{eq:opm-likelihood}
  p(y | \mu_0, \mu_1, z) = p_{\textsf{norm}}(y ; \mu_z, 1).
\end{align}
When translating our source code to a graphical model, the target language expression $\m{P}(y)$ that evaluates this probability would read \ang{($p_{\textsf{norm}}$ $y$ (if (= $z$ 0) $\mu_0$ $\mu_1$) 1)}.

The real question is how to define the probabilities for $\mu_0$ and $\mu_1$. One choice could be to simply set the probability of unevaluated branches to 1. One way to do this in this particular example is to write
\begin{align*}
  p(\mu_0 | z) 
  &= 
  p_{\textsf{norm}}(\mu_0 ; -1, 1)^z
  \\
  p(\mu_1 | z) 
  &= 
  p_{\textsf{norm}}(\mu_1 ; 1, 1)^{1-z}.
\end{align*}
In the target language we could achieve the same effect by using if-expressions defining $P(\mu_0)$ as \ang{(if (= z 0) ($p_{\textsf{norm}}$ $\mu_0$ -1.0 1.0) 1.0)} and defining $\m{P}(\mu_1)$ as \ang{(if (not (= z 0)) ($p_{\textsf{norm}}$ $\mu_1$ 1.0 1.0) 1.0)}.
 
On first inspection this design seems reasonable. Much in the way we would do in a mixture model, we either include $p(\mu_0 | z=0)$ or $p(\mu_1 | z=1)$ in the probability, and assume a probability 1 for unevaluated branches, i.e. $p(\mu_0 | z=1)$
and $p(\mu_1 | z=0)$. 

On closer inspection, however, it is not obvious what support this distribution should have. We might naively suppose that $(y,\mu_0,\mu_1,z) \in \R \times \R \times \R \times \{0,1\}$, but this definition is problematic. To see this, let us try to calculate the marginal likelihood $p(y)$, 
\begin{align*}
  p(y) =~ & p(y,z=0) + p(y,z=1), \\
       =~ & p(z=0) \int d \mu_0 d \mu_1 \: p(y, \mu_0, \mu_1 | z=0) \\
          & + p(z=1) \int d \mu_0 d \mu_1 \; p(y,\mu_0,\mu_1 | z=1), \\
       =~ & 0.5 \int d \mu_1 
            \left(
              \int d \mu_0 \:
              p_{\textsf{norm}}(y ; \mu_0, 1) 
              p_{\textsf{norm}}(\mu_0 ; -1, 1) 
            \right)
            \\
          & + 0.5 \int d \mu_0 
            \left(
              \int d \mu_1 \:
              p_{\textsf{norm}}(y ; \mu_1, 1) 
              p_{\textsf{norm}}(\mu_1 ; 1, 1) 
            \right),
           \\
       =~ & \infty.
\end{align*}
So what is going on here? This integral does not converge because we have not assumed the correct support: We cannot marginalize $\int_{\R} d \mu_0 \: p(\mu_0 | z=0)$ and $\int_{\R} d \mu_1 \: p(\mu_1 | z=1)$ if we assume $p(\mu_0 | z=0) = 1$ and $p(\mu_1 | z=1) = 1$. These uniform densities effectively specify improper priors on unevaluated branches.

In order to make lazy evaluation of if-expressions more well-behaved, we could choose to define the support of the joint as a union over supports for individual branches 
\begin{align}
  \label{eq:if-support}
  (y, \mu_0, \mu_1, z) \in (\R \times \R \times \{ \mfop{nil} \} \times \{0\}) \cup (\R \times \{ \mfop{nil} \}  \times \R \times \{1\})
  .
\end{align}
In other words, we could restrict the support of variables in unevaluated branches to some special value \ang{nil} to signify that the variable does not exist. Of course this can result in rather complicated definitions of the support in probabilistic programs with many levels of nested if-expressions. 

Could eager evaluation of branches yield a more straightforward definition of the probability distribution associated with a program? Let us look at Program~\ref{pr:opm-if-sample} once more. If we use eager evaluation, then this program is equivalent to Program~\ref{pr:opm-eager} which defines a distribution 
\begin{align*}
  p(y,\mu_0,\mu_1,z) 
  = p(y | \mu_0,\mu_1,z) p(z) p(\mu_0) p(\mu_1).
\end{align*}
We can now define $p(\mu_0) = p_\textsf{norm}(\mu_0 ; {-}1, 1)$ and $p(\mu_1) = p_\textsf{norm}(\mu_1 ; 1, 1)$  and assume the same likelihood as in the equation in~\eqref{eq:opm-likelihood}. This defines a joint density that corresponds to what we would normally assume for a mixture model. In this evaluation model, sample expressions in both branches are always incorporated into the joint.

Unfortunately, eager evaluation would lead to counter-intuitive results when observe expressions occur in branches. To see this, Let us consider the following form for our program
\begin{foppl}[mathescape,caption={One-point mixture with observes inside the if expression.},label={pr:opm-if-observe}]
(let [z (sample (bernoulli 0.5))
      mu0 (sample (normal -1.0 1.0))
      mu1 (sample (normal 1.0 1.0))
      y 0.5]
  (if (= z 0)
    (observe (normal mu0 1) y)
    (observe (normal mu1 1) y))
  z)
\end{foppl}
Clearly it is not the case that eager evaluation of both branches is equivalent to lazy evaluation of one of the branches. When performing eager evaluation, we would be observing two variables $y_0$ and $y_1$, both with value 0.5. When performing lazy evaluation, only one of the two branches would be included in the probability density. The lazy interpretation is a lot more natural here. In fact, it seems difficult to imagine a use case where you would want to interpret observe expressions in branches in a eager manner. 

So where does all this thinking about evaluation strategies for if-expressions leave us? Lazy evaluation of if-expressions makes it difficult to characterize the support of the probability distribution defined by a program when branches contain sample expressions. However, at the same time, lazy evaluation is essential in order for branches containing observe expressions to make sense. So have we perhaps made a fundamentally flawed design choice by allowing sample and observe to be used inside if branches?

It turns out that this is not necessarily the case. We just need to understand that observe and sample expressions affect the marginal posterior over program outputs in different ways. Sample expressions that are not on the flow-control path cannot affect the values of any expressions outside their branch. This means they can be safely incorporated into the model as auxiliary variables, since their presence does not change the marginal posterior on the return value. Observed variables, on the other hand, can affect the distribution over return values when eagerly evaluated, even when they are not on the flow-control path.\footnote{The only exception to this rule is observe expressions that are conditionally independent of the program output, which implies that the graphical model associated with the program could be split into two independent networks out of which one could be eliminated without affecting the distribution on return values.}

Based on this intuition, the solution to our problem is straightforward: We can assign probability 1 to observed variables that are not on the same control flow path.
Since observed variables have constant values, the interpretability of their support is not an issue in the way it is with sampled variables. Conversely we assign the same probability to sampled variables, regardless of the branch they occur in. We will describe how to accomplish this in the following sections.

\paragraph{Restricting the language} 
As an aside, an alternative solution could be to restrict the language to forbid the use of observe expressions inside conditionals.
This can often be achieved by simply re-arranging the nesting of observe and if statements, particularly in cases where the same number of observe expressions appear in each branch.
For example, in Program~\ref{pr:opm-if-observe} this could be achieved by refactoring the if block from
\begin{foppl}[mathescape]
  (if (= z 0)
    (observe (normal mu0 1) y)
    (observe (normal mu1 1) y))
\end{foppl}
to the alternative
\begin{foppl}[mathescape]
  (observe 
    (if (= z 0)
      (normal mu0 1)
      (normal mu1 1))
    y)
\end{foppl}
which defines the same conditional density regardless of whether lazy- or eager-evaluated.
Some PPL implementations do, in fact, make such a language restriction --- quite reasonable for cases in which the observed values are a fixed amount of pre-collected data.
However, this can be limiting in other settings, particularly in models which make use of conditioning on quantities other than a static dataset.
For example, many reinforcement learning problems can be thought of as a (possibly stochastic) agent interacting with a (possibly stochastic) environment, where the number of times an agent encounters a reward signal (corresponding to an observe statement) is itself a random variable.
We thus choose not to restrict the FOPPL in such a way, at the expense of some additional complexity in the implementation of observe statements.

\subsubsection{Support-Related Subtleties}

As a last but important bit of understanding to convey before proceeding to the translation rules in the next section it should be noted that the following two programs are allowed by the FOPPL and are not problematic despite potentially appearing to be.
\begin{foppl}[mathescape,caption={Stochastic and potentially infinite discrete support.},label={pr:opm-discrete-changing-support}]
(let [z (sample (poisson 10))
      d (discrete (range 1 z))]
  (sample d))
\end{foppl}

\begin{foppl}[mathescape,caption={Stochastic support and type.},label={pr:opm-changing-support-and-type}]
(let [z (sample (flip 0.5))
      d (if z (normal 1 1) (gamma 1 1)]
  (sample d)).
\end{foppl}
Program~\ref{pr:opm-discrete-changing-support} highlights a subtlety of FOPPL language design and interpretation, that being that the distribution \fop{d} has support that has potentially infinite cardinality.  This is not problematic for the simple reason that samples from \fop{d} cannot be used as a loop bound and therefore cannot possibly induce an unbounded number of random variables.  It does serve as an indication that some care should be taken when reasoning about such programs and writing inference algorithms for the same.  As is further highlighted in Program~\ref{pr:opm-changing-support-and-type}, which adds a seemingly innocuous bit of complexity to the control-flow examples from earlier in this chapter, neither the support nor the distribution type of a random variable need  be the same between two different control flow paths.  The fact that the support might be quite large can yield substantial value-dependent variation in inference algorithm runtimes.  Moreover, inference algorithm implementations must have distribution library support that is robust to the possibility of needing to score values outside of their support.

\subsubsection{Translation rules}

Now that we have developed some intuition for how one might translate a
program to a data structure that represents a graphical model and have been 
introduced to several subtleties that arise in designing ways to do this, we are in a
position to formally define a set of translation rules. We define the
$\Downarrow$ relation for translation using the so called inference-rules
notation from the programming language community. This notation specifies a
recursive algorithm for performing the translation succinctly and
declaratively. The inference-rules notation is
\begin{align}
        \infer{\mathit{bottom}}{\mathit{top}}
\end{align}
It states that if the statement $\mathit{top}$ holds,
so does the statement $\mathit{bottom}$. As a simple example, we could write 
\begin{align}
        \infer{
                \rho, \phi, ({-}\; e) \Downarrow G,({-}\; E)
        }{
                \rho, \phi, e \Downarrow G,E
        }
\end{align}
to state that when $e$ gets translated to $G,E$ under $\rho$ and $\phi$,
then its negation is translated to $G,({-}\;E)$ under the same $\rho$ and $\phi$.

The grammar for the FOPPL in Language~\ref{lang:foppl} describes 8 distinct
expression types: (i) constants, (ii) variable references, (iii) let
expressions, (iv) if expressions, (v) user-defined procedure applications,
(vi) primitive procedure applications, (vi) sample expressions, and finally
(viii) observe expressions. Aside from constants and variable references, each
expression type can have sub-expressions. In the remainder of this
section, we will define a translation rule for f type, under
the assumption that we are already able to translate its sub-expressions,
resulting in a set of rules that can be used to define the translation of
every possible expression in the FOPPL language in a recursive manner.

\paragraph{\bf Constants and Variables} We translate
constants $c$ and variables $v$ in the FOPPL to themselves and the empty graphical model:
\[
        \infer{\rho,\,\phi,\,c \Downarrow G_\mathsf{emp},\, c}{}
        \qquad
        \infer{\rho,\,\phi,\,v \Downarrow G_\mathsf{emp},\, v}{}
\]
where $G_\mathsf{emp}=(\emptyset,\emptyset,[],[])$ 
represents the empty graphical model.

\paragraph{\bf Let} We translate \lsi{(let [$v$ $e_1$] $e_2$)} 
by first translating $e_1$, then substituting the outcome of this translation
for $v$ in $e_2$, and finally translating the result of this substitution:
\[
\infer{ 
        \rho,\,\phi,\, \mfop{(let [$v\ e_1$]$\ $ $e_2$)} 
        \Downarrow 
        (G_1 \oplus G_2),\, E_2
}{ 
        \rho,\,\phi,\,e_1 \Downarrow G_1,\, E_1 
        \qquad 
        \rho,\,\phi,\,e_2[v := E_1] \Downarrow G_2,\, E_2
}
\]
Here $e_2[v := E_1]$ is a result of substituting $E_1$ for $v$ in the expression $e_2$
(while renaming bound variables of $e_2$ if needed). $G_1 \oplus G_2$ is the
combination of two disjoint graphical models:
when $G_1 = (V_1,\,A_1,\,\m{P}_1,\,\m{Y}_1)$ and $G_2 = (V_2,\,A_2,\,\m{P}_2,\,\m{Y}_2)$,
\[
        (G_1 \oplus G_2) = (V_1 \cup V_2,\, A_1 \cup A_2,\,\m{P}_1 \oplus \m{P}_2,\,\m{Y}_1 \oplus \m{Y}_2)
\]
where $\m{P}_1 \oplus \m{P}_2$ and $\m{Y}_1 \oplus \m{Y}_2$ are the concatenation of two finite maps with
disjoint domains. This combination operator assumes that the input graphical models
$G_1$ and $G_2$ use disjoint sets of vertices. This assumption always holds 
because every graphical model created by our translation uses
fresh vertices, which do not appear in other networks previously generated.

We would like to note that this translation rule has not been optimized for computational efficiency. Because $E_2$ is replaced by $v$ in $E_2$, we will evaluate $E_1$ once for each occurrence of $v$. We could avoid these duplicate computations by incorporating deterministic nodes into our graph, but we omit this optimization in favor of readability.

\paragraph{\bf If} Our translation of the if-expression is straightforward.
It translates all the three sub-expressions,
and puts the results from these translations together:
\[ 
\infer{ 
  \rho,\,\phi,\, \mfop{(if $\;e_1\;e_2\;e_3$)} \Downarrow  (G_1 \oplus G_2 \oplus G_3),\, \mlisp{(if $\;E_1\;E_2\;E_3$)}
}{
  \begin{gathered}
    \rho,\,\phi,\, e_1 \Downarrow G_1,\, E_1 
    \qquad
    \rho,\,\mlisp{(and $\;\phi\;E_1$)},\, e_2 \Downarrow G_2,\, E_2 
    \\
    \rho,\,\mlisp{(and $\;\phi\;$ (not $\;E_1$))},\, e_3 \Downarrow G_3,\, E_3
  \end{gathered}
}
\]
As we have discussed, the graphical models $G_1$, $G_2$ and $G_3$ use
disjoint vertices, and so their combination  $G_1 \oplus G_2 \oplus G_3$ is always defined. 

When we translate the sub-expressions for the consequent and
alternative branches, we conjoin the logical predicate $\phi$ with the expression $E_1$ or its negation.  The role of this logical predicate was established before; it serves to include or exclude observe statements that are on or off the current-sample control-flow path.  It will be used in the upcoming translation of 
observe statements.

None of the rules for an expression $e$ so far extends graphical models from $e$'s sub-expressions
with any new vertices. This uninteresting treatment comes from the fact that the programming
constructs involved in these rules perform
deterministic, not probabilistic, computations, and the translation uses graphical models
to express random variables. The next two rules about \lsi{sample} and \lsi{observe} 
show this usage. 

\paragraph{\bf Sample} We translate sample expressions using the following rule: 
\[
\infer{ 
        \rho,\,\phi,\,\mfop{(sample $\;e$)}
        \Downarrow 
        (V \cup \{v\},\, A \cup \{(z,v) \mid z \in Z\},\,\m{P} \oplus [v \mapsto F],\, \m{Y}),\, v
}{
        \begin{array}{ll}
        \rho,\,\phi,\,e \Downarrow (V, A, \m{P}, \m{Y}),\,E 
        &
        \mbox{Choose a fresh variable $v$}
        \\
        Z = \textsc{free-vars}(E) 
        &
        F = \textsc{score}(E,v) \neq \bot
        \end{array}
}
\]
This rule states that we translate \mfop{(sample $\;e$)} in three steps. First, we
translate the argument $e$ to a graphical model $(V, A, \m{P}, \m{Y})$ and a
deterministic expression $E$. Both the argument $e$ and its translation $E$
represent the same distribution, from which $\mfop{(sample e)}$ samples.
Second, we choose a fresh variable $v$, collect all free variables in $E$ that
are used as random variables of the network, and set $Z$ to the set of these
variables. Finally, we convert the expression $E$ that denotes a distribution,
to the probability density or mass function $F$ of the distribution. This
conversion is done by calling $\textsc{score}$, which is defined as follows:
\begin{align*}
        &
        \textsc{score}(\mlisp{(if $\;E_1\;E_2\;E_3$)},\, v)  = 
        \mlisp{(if $\;E_1\;F_2\;F_3$)} 
        \\[-1mm]
        &
        \qquad 
        \mbox{(when $F_i = \textsc{score}(E_i,v)$ for $i \in \{2,3\}$ and it is not $\bot$)}
        \\
        & 
        \textsc{score}(\mlisp{($c\;E_1 \ldots E_n$)},\, v)  = 
        \mlisp{($p_c\; v\ E_1 \ldots E_n$)}
        \\[-1mm]
        &
        \qquad
        \mbox{(when $c$ is a constructor for a distribution and $p_c$ its pdf or pmf)}
        \\
        &
        \textsc{score}(E,\,v) = \bot
        \\[-1mm]
        & \qquad
        \mbox{(when $E$ is not one of the above cases)}
\end{align*}
The $\bot$ (called ``bottom'', indicating terminating failure)  case happens when the argument $e$ in \lsi{(sample $e$)} does
not denote a distribution. Our translation fails in that case.

\paragraph{\bf Observe} Our translation for observe expressions \mfop{(observe $\; e_1 \; e_2$)} is analogous to that of sample expressions, but we additionally need to account for the observed value $e_2$, and the predicate $\phi$:

\[
\infer{ 
  \rho,\, \phi,\, \mfop{(observe $\; e_1 \; e_2$)} 
  \Downarrow 
  (V \cup \{ v \} ,\,
   A \cup B,\,
   P \oplus [v \mapsto F],\, 
   \m{Y} \oplus [v \mapsto E_2]),\, 
  E_2
}{
  \begin{array}{ll}
    \rho,\,\phi,\,e_1 \Downarrow G_1,\,E_1 
    &
    \rho,\,\phi,\,e_2 \Downarrow G_2,\,E_2
    \\
    (V, A, \m{P}, \m{Y}) = G_1 \oplus G_2 \qquad
    &
    \mbox{Choose a fresh variable $v$}
    \\
    F_1 = \textsc{score}(E_1, v) \not= \bot
    &             
    F = \mfop{(if $\; \phi \; F_1 \; 1$)}
    \\
    Z = (\textsc{free-vars}(F) \setminus \{v\})
    &
    \textsc{free-vars}(E_2) = \emptyset
    \\
    B = \{(z,v): z \in Z\}
  \end{array}
}
\]

This rule first translates the sub-expressions $e_1$ and $e_2$. We then construct a network $(V, A, \m{P}, \m{Y})$ by merging the networks of the sub-expressions and pick a new variable $v$ that will represent the observed random variable. As in the case of sample statements, the deterministic expression $E_1$ that is obtained by translating $e_1$ must evaluate to a distribution. We use the \textsc{score} function to construct an expression $F_1$ that represents the probability mass or density of $v$ under this distribution. We then construct a new expression $F = \mfop{(if $\; \phi \; F_1 \; 1$)}$ to ensure that the probability of the observed variable evaluates to 1 if the observe expression occurs in a branch that was not followed. The free variables in this expression are the union of the free variables in $E_1$, the free variables in $\phi$ and the newly chosen variable $v$. We add a set of arcs $B$ to the network, consisting of edges from all free variables in $F$ to $v$, excluding $v$ itself. Finally we add the expression $F$ to $\m{P}$ and store the observed value $E_2$ in $\m{Y}$. 

In order for this notion of an observed random variable to make sense, the expression $E_2$ must be fully deterministic. For this reason we require that $\textsc{free-vars}(E_2) = \emptyset$, which ensures that $E_2$ cannot reference any other random variables in the graphical model. Translation fails when this requirement is not met.  Remember that $\textsc{free-vars}$ refers to all unbound variables in an expression.  Also note an important consequence of $E_2$ being a value, namely, although the return value of an observe may be used in subsequent computation, no graphical model edges will be generated with the observed random variable as a parent.  An alternative rule could return a \fop{null} or \fop{nil} value in place of $E_2$ and, as a result, might  potentially be ``safer'' in the sense of ensuring clarity to the programmer.  Not being able to bind the observed value would mean that there is no way to imagine that an edge could be created where one was not.

\paragraph{\bf Procedure Call} The remaining two cases are those for procedure calls, one for a user-defined procedure $f$ and one for a primitive function $c$. In both cases, we first translate arguments. In the case of primitive functions we then translate the expression for the call by substituting translated arguments into the original expression, and merging the graphs for the arguments 
\[
\infer{
        \rho,\,\phi,\,\mfop{($c \;e_1\ldots e_n$)} \Downarrow G_1 \oplus \ldots \oplus G_n,\,
        {\mfop{($c \;E_1\ldots E_n$)}}
}{
        \rho,\,\phi,\,e_i \Downarrow G_i,\,E_i \mbox{ for all $1 \leq i \leq n$} 
}
\]
For user-defined procedures, we additionally transform the procedure body. We do this by replacing all instances of the variable $v_i$ with the expression for the argument $E_i$ 
\[
\infer{ 
        \rho,\,\phi,\,\mfop{($f \;e_1\ldots e_n$)} \Downarrow G_1 \oplus \ldots \oplus G_n \oplus G,\,E
}{ 
        \begin{array}{ll}
        \rho,\,\phi,\,e_i \Downarrow G_i,\,E_i \mbox{ for all $1 \leq i \leq n$} 
        & 
        \rho(f) = \mfop{(defn$\;f\;[v_1\ldots v_n]\;e$)}
        \\
        \rho,\,\phi,\,e[v_1:=E_1,\ldots v_n:=E_n]
        \Downarrow
        G,\,E
        \end{array}
}
\]

	\section{Evaluating the Density}
	\label{sec:eval-dens}


Before we discuss algorithms for inference in FOPPL programs, first we make explicit how we can use this representation of a probabilistic program to evaluate the probability of a particular setting of the variables in $V$.
The Bayesian network $G = (V, A, \m{P}, \m{Y})$ that we construct by compiling a FOPPL program is a mathematical representation of a directed graphical model. Like any graphical model, $G$ defines a probability density on its variables $V$. 
In a directed graphical model, each node $v \in V$ has a set of parents
\begin{align} 
  \label{eq:parents}
  \pa(v) := \{ u : (u,v) \in A \}
  .
\end{align}
The joint probability of all variables can be expressed as a product over conditional probabilities
\begin{align}
  \label{eq:joint-v}
  p(V) = \prod_{v \in V} p(v \,|\, \pa(v))
  .
\end{align}
In our graph $G$, each term $p(v \,|\, \pa(v))$ is represented as a deterministic expression $\m{P}(v) = \mlisp{(}c~v~E_{1}~\ldots~E_{n}\mlisp{)}$, in which $c$ is either a probability mass function (for discrete variables) or a probability density function (for continuous variables) and $E_{1}, \ldots, E_{n}$ are expressions that evaluate to parameters $\theta_{1}, \ldots, \theta_{n}$ of this mass or density function. 

Implicit in this notation is the fact that each expression has some set of free variables. To evaluate an expression to a value, we must specify values for each of these free variables. In other words, we can think of each of these expressions $E_{i}$ as a mapping from values of free variables to a parameter value. By construction, the set of parents $\pa(v)$ is nothing but the free variables in $\m{P}(v)$ exclusive of $v$
\begin{align}
    \pa(v) = \textsc{free-vars}(\m{P}(v)) \setminus \{v\}.
\end{align}
Thus, the expression $\m{P}(v)$ can be thought of as a function that maps the $v$ and its parents $\pa(v)$ to a probability or probability density. We will therefore from now on treat these two as equivalent,
\begin{align}
    \label{eq:dens-equiv}
    p(v \,|\, \pa(v))
    \equiv 
    \m{P}(v)
    .
\end{align}
We can decompose the joint probability $p(V)$ into a prior and a likelihood term. In our specification of the translation rule for observe, we require that the expression $\m{Y}(v)$ for the observed value may not have free variables. Each expression $\m{Y}(v)$ will therefore simplify to a constant when we perform partial evaluation, a subject we cover extensively in Section~\ref{sec:partial-eval} of this chapter. We will use $Y$ to refer to all the nodes in $V$ that correspond to observed random variables, which is to say $Y = {\rm dom}(\m{Y})$. Similarly, we can use $X$ to refer to all nodes in $V$ that correspond to unobserved random variables, which is to say $X = V \setminus Y$. Since observed nodes $y \in Y$ cannot have any children we can re-express the joint probability in Equation~\eqref{eq:joint-v} as
\begin{align} 
    p(V)
    = 
    p(Y, X) 
    = p(Y \,|\, X) p(X)
    ,
\end{align}
where
\begin{align}
    \label{eq:pp-posterior}
    p(Y \,|\, X)
    &= 
    \prod_{y \in Y}
    p(y \,|\, \pa(y))
    ,
    &
    p(X)
    &=
    \prod_{x \in X}
    p(x \,|\, \pa(x))
    .
\end{align}
In this manner, a probabilistic program defines a joint distribution $p(Y, X)$. 
The goal of probabilistic program {\em inference} is to characterize the posterior distribution 
\begin{align}
    p(X \,|\, Y) 
    &
    = p(X, Y) / p(Y)
    ,
    &
    p(Y) 
    &
    := 
    \int dX \: p(X, Y)
    .
\end{align}

	\subsection{Conditioning with Factors}
	\label{sec:factor-condition}

The interpretation of a probabilistic program as a model that defines a density $p(Y, X)$ is intuitive, but it is not the most general formulation of densities that a probabilistic program can denote.
Not all inference problems for probabilistic programs target a posterior $p(X \,|\, Y)$ that is defined in terms of unobserved and observed random variables. There are inference problems in which there is no notion of observed data, but in which it is still possible to define some notion of loss, reward, or fitness given a choice of $X$. In the probabilistic programs written in the FOPPL, the \ang{sample} statements in a probabilistic program define a prior $p(X)$ on the random variables, whereas the \ang{observe} statements define a likelihood $p(Y \,|\, X)$. We can replace the likelihood $p(Y\,|\,X)$ with a strictly positive potential function $\psi(X)$, which defines some notion of reward or utility for the variables $X$. This defines an unnormalized density
\begin{align}
    \gamma(X) = \psi(X) p(X).
\end{align}
In this more general setting, the goal of inference is to characterize a target density $\pi(X)$, which we define as
\begin{align}
    \label{eq:pp-target}
    \pi(X) 
    &:= 
    \gamma(X)/Z,
    &
    Z
    &:=
    \int 
    \! dX \:
    \gamma(X)
    .
\end{align}
Here $\pi(X)$ is the analogue to the posterior $p(X \,|\, Y)$, the unnormalized density $\gamma(X)$ is the analogue to the joint $p(Y, X)$, and the normalizing constant $Z$ is the analogue to the marginal likelihood $p(Y)$. 

From a language design point of view, we can now ask how the FOPPL would need to be extended in order to support this more general form of conditioning. For a probabilistic program in the FOPPL, the potential function is a product over terms
\begin{align}
    \psi(X) &= \prod_{y \in Y} \psi_y(X_y),
\end{align}
where we define $\psi_y$ and $X_y$ as 
\begin{align}
    \psi_y(X_y)
    &:=
    p(y\!=\!\m{Y}(y) \,|\, \pa(y)) 
    \equiv
    \m{P}(y)[y := \m{Y}(y)]
    \\
    X_y 
    &:= 
    \textsc{free-vars}(\m{P}(y)) \setminus \{y\}
    =
    \pa(y)
    .
\end{align}
Note that $\m{P}(y)[y := \m{Y}(y)]$ is just some expression that evaluates to either a probability mass or a probability density if we specify values for its free variables $X_y$. Since we never integrate over $y$, it does not matter whether $\m{P}(y)$ represents a (normalized) mass or density function. We could therefore in principle replace $\m{P}(y)$ by any other expression with free variables $X_y$ that evaluates to a number $\ge 0$. 

One way to support arbitrary potential functions is to provide a special form \mang{(factor log-p)} that takes an arbitrary log probability \mang{log-p} (which can be both positive or negative) as an argument. We can then define a translation rule that inserts a new node $v$ with probability $\m{P}(v) = \mang{(exp log-p)}$ and observed value \mang{nil} into the graph:
\begin{align*}
\infer{ 
        \rho,\,\phi,\,\mang{(factor $\; e$)} 
        \Downarrow 
        (V,\,A,\,\m{P} \oplus [v \mapsto F],\, \m{Y} \oplus [v \mapsto \mang{nil}]),\, \mang{nil}
}{
    \begin{gathered}
        \rho,\,\phi,\,e \Downarrow (V,A,\m{P},\m{Y}),\,E
        \qquad
        F=\mang{(if $\;\phi\; $(exp $\; E$) 1)}
        \\
        \text{Choose a fresh variable} ~v
    \end{gathered}
}
\end{align*}

In practice, we don't need to provide separate special forms for \ang{factor} and \ang{observe}, since each can be implemented as a special case of the other. One way of doing so is to define \mang{factor} as a procedure 
\begin{lstlisting}[mathescape]
    (defn factor [log-p]
      (observe (factor-dist log-p) nil))
\end{lstlisting}
in which \ang{factor-dist} is a constructor for a ``pseudo'' distribution object with corresponding potential
\begin{align}
    p_{\text{factor-dist}}(y ; \lambda) 
    := 
    \left\{
    \begin{array}{ll}
        \exp \lambda
        &
        y = \mang{nil}
        \\
        0
        &
        y \not= \mang{nil}
    \end{array}
    \right.
\end{align}
We call this a pseudo distribution, because it defines a (unnormalized) potential function, rather than a normalized mass or density. 

Had we defined the FOPPL language using \mang{factor} as the primary conditioning form, then we could have implemented a primitive procedure $\mang{(log-prob dist v)}$ that returns the log probability mass or density for a value \mang{v} under a distribution \mang{dist}. This would then allow us to define \mang{observe} as a procedure
\begin{lstlisting}[mathescape]
    (defn observe [dist v]
      (factor (log-prob dist v))
      y)
\end{lstlisting}

	\subsection{Partial Evaluation}
	\label{sec:partial-eval}

An important and necessary optimization for our compilation procedure is to perform a partial evaluation step.  This step pre-evaluates expressions $E$ in the target language that do not contain any free variables, which means that they take on the same value in every execution of the program. It turns out that partial evaluation of these expressions is necessary to avoid the appearance of ``spurious'' edges between variables that are in fact not connected, in the sense that the value of the parent does not affect the conditional density of the child.

Because our target language is very simple, we only need to consider if-expressions and procedure calls. We can update the compilation rules for these expressions to include a partial evaluation step
\[ 
\infer{ 
  \rho,\,\phi,\, \mfop{(if $\;e_1\;e_2\;e_3$)} \Downarrow  (G_1 \oplus G_2 \oplus G_3),\, \textsc{eval}(\mlisp{(if $\;E_1\;E_2\;E_3$)})
}{
  \begin{gathered}
    \rho,\,\phi,\, e_1 \Downarrow G_1,\, E_1 
    \qquad
    \rho,\,\textsc{eval}(\mlisp{(and $\;\phi\;E_1$)}),\, e_2 \Downarrow G_2,\, E_2 
    \\
    \rho,\,\textsc{eval}(\mlisp{(and $\;\phi\;$ (not $\;E_1$))}),\, e_3 \Downarrow G_3,\, E_3
  \end{gathered}
}
\]

and
\[
\infer{
        \rho,\, \phi,\, \mfop{($c \;e_1\ldots e_n$)} \Downarrow G_1 \oplus \ldots \oplus G_n,\,
        \textsc{eval}(\mfop{($c \;E_1\ldots E_n$)})
}{
        \rho,\,e_i \Downarrow G_i,\,E_i \mbox{ for all $1 \leq i \leq n$} 
}
\]
The partial evaluation operation $\textsc{eval}(e)$ can incorporate any number of rules for simplifying expressions. We will begin with the rules
\begin{align*}
& 
\textsc{eval}(\mfop{(if $\; c_1 \; E_2 \; E_3$)}) = E_2
\\[-1mm]
&
\qquad
\mbox{when $c_1$ is logically true}
\\
& 
\textsc{eval}(\mfop{(if $\; c_1 \; E_2 \; E_3$)}) = E_3
\\[-1mm]
&
\qquad
\mbox{when $c_1$ is logically false}
\\
&\textsc{eval}(\mfop{($c \; c_1 \; \ldots \; c_n$)}) = c'
\\
&
\qquad
\mbox{when calling $c$ with arguments $c_1, \ldots, c_n$ evaluates to $c'$}
\\
&\textsc{eval}(E) = E
\\
&
\qquad
\mbox{in all other cases}
\end{align*}

These rules state that an if statement \mfop{(if $\; E_1 \; E_2 \; E_3$)} can be simplified when $E_1 = c_1$ can be fully evaluated, by simply selecting the expression for the appropriate branch. Primitive procedure calls can be evaluated when all arguments can be fully evaluated. 

In order to accommodate partial evaluation, we additionally modify the definition of the $\textsc{score}$ function. Distributions in the FOPPL are constructed using primitive procedure applications. This means that a distribution with constant arguments such as \fop{(beta 1 1)} will be partially evaluated to a constant $c$. To account for this, we need to extend our definition of the \textsc{score} conversion with one rule
\begin{align*}
        & 
        \textsc{score}(c,\,v) = 
        \mlisp{($p_c\; v$)}
        \\[-1mm]
        &
        \qquad
        \mbox{(when $c$ is a distribution and $p_c$ is its pdf or pmf)}
\end{align*}

To see how partial evaluation also reduce the number of edges in the Bayesian network, let us consider the expression \mfop{(if true $\; v_1 \; v_2$)}. This expression nominally references two random variables $v_1$ and $v_2$. After partial evaluation, this expression simplifies to $v_1$, which eliminates the spurious dependence on $v_2$. 

Another practical advantage of partial evaluation is that it gives us a simple way to identify expressions in a program that are fully deterministic (since such expressions will be partially evaluated to constants). This is useful when translating observe statements \mfop{(observe $\; e_1 \; e_2$)}, in which the expression $e_2$ must be deterministic. In programs that use the \mfop{(loop $\; c \; v \; e \; e_1 \; \ldots \; e_n$)} syntactic sugar, we can now substitute any fully deterministic expression for the number of loop iterations $c$. For example, we could define a loop in which the number of iterations is given by the dataset size.

{\bf Lists, vectors and hash maps.} Eliminating spurious edges in the dependency graph becomes particularly important in programs that make use of data structures. Let us consider the following example, which defines a 3-state Markov chain
\begin{lstlisting}
(let [A [[0.9 0.1]
         [0.1 0.9]]
      x1 (sample (discrete [1. 1.]))
      x2 (sample (discrete (get A x1)))
      x3 (sample (discrete (get A x2)))]
  [x1 x2 x3])
\end{lstlisting}
Compilation to a Bayesian network will yield three variable nodes. If we refer to these nodes as $v_1$, $v_2$ and $v_3$, then there will be arcs from $v_1$ to $v_2$ and from $v_2$ to $v_3$. Suppose we now rewrite this program using the \mfop{loop} syntactic sugar that we introduced in Chapter \ref{ch:foppl} 
\begin{lstlisting}
(defn markov-step 
  [n xs A]
  (let [k (last xs)
        Ak (get A k)]
    (append xs (sample (discrete Ak)))))

(let [A [[0.9 0.1]
         [0.1 0.9]]
      x1 (sample (discrete [1. 1.]))]
  (loop 2 markov-step [x1] A))
\end{lstlisting}
In this version of the program, each call to \mfop{markov-step} accepts a vector of states \mfop{xs} and appends the next state in the Markov chain by calling \mfop{(append xs (sample (discrete Ak)))}. In order to sample the next element, we need the row \mfop{Ak} for the transition matrix that corresponds to the current state \mfop{k}, which is retrieved by calling \mfop{(last xs)} to extract the last element of the vector. 

The program above generates the same sequence of random variables as the previous one, and has the advantage of allowing us to generalize to sequences of arbitrary length by changing the constant \mfop{2} in the loop to a different value. However, under the partial evaluation rules that we have specified so far, we would obtain a different set of edges. 
As in the previous version of the program, this version of the program evaluates three sample statements. For the first statement, \mfop{(sample (discrete [1. 1.]))} there will be no arcs. Translation of the second sample statement \mfop{(sample (discrete Ak))}, which is evaluated in the body of \mfop{markov-step}, results in an arc from $v_1$ to $v_2$, since the expression for \mfop{Ak} expands to
\begin{lstlisting}[mathescape]
    (get [[0.9 0.1] 
          [0.1 0.9]] 
          (last [$v_1$]))
\end{lstlisting}
However, for the third sample statement there will be arcs from both $v_1$ and $v_2$ to $v_3$, since \mfop{Ak} expands to
\begin{lstlisting}[mathescape]
    (get [[0.9 0.1] 
          [0.1 0.9]] 
          (last (append [$v_1$] $v_2$)))
\end{lstlisting}
The extra arc from $v_1$ to $v_3$ is of course not necessary here, since the expression \mfop{(last (append [$v_1$] $\; v_2$))} will always evaluate to $v_2$. What's more, if we run this program to generate more than 3 states, the node $v_n$ for the $n$-th state will have incoming arcs from all preceding variables $v_1, \ldots, v_{n-1}$, whereas the only real arc in the Bayesian network is the one from $v_{n-1}$.

We can eliminate these spurious arcs by implementing an additional set of partial evaluation rules for data structures, 
\begin{align*}
&\textsc{eval}(\mfop{(vector $\; E_1 \; \ldots \; E_n$)}) = \mfop{[$E_1 \; \ldots \; E_n$]},
\\
&\textsc{eval}(\mfop{(hash-map $\; c_1 \; E_1 \; \ldots \; c_n \; E_n$)}) = \mfop{\{$c_1 \; E_1 \; \ldots \; c_n \; E_n$\}}.
\end{align*}
 These rules ensure that expressions which construct data structures are partially evaluated to data structures containing expressions. We can similarly partially evaluate functions that add or replace entries. For example, we can define the following rules for the \mfop{conj} primitive, which appends an element to a data structure,
\begin{align*}
&\textsc{eval}(\mfop{(append [$E_1 \; \ldots \; E_n$] $\; E_{n+1}$)}) = \mfop{[$E_1$\ $\ldots$\ $E_n$\ $E_{n+1}$]},
\\
&\textsc{eval}(\mfop{(put \{$c_1 \; E_1 \; \ldots \; c_n \; E_n$\}\ $c_{k}$\ $E^\prime_{k}$)}) = \mfop{\{$c_1$\ $E_1$\ $\ldots$\ $c_k$\ $E^\prime_k$\ $\ldots$\ $c_{n}$\ $E_{n}$\}}.
\end{align*}
In the Markov chain example, the expression for \mfop{Ak} in the third sample statement then simplifies to 
\begin{lstlisting}[mathescape]
    (get [[0.9 0.1] 
          [0.1 0.9]] 
          (last [$v_1 \; v_2$]))
\end{lstlisting}
Now that partial evaluation constructs data structures containing expressions, we can use partial evaluation of accessor functions to extract the expression corresponding to an entry 
\begin{align*}
&\textsc{eval}(\mfop{(last [$E_1 \; \ldots \; E_n$])}) = E_n,
\\[6pt]
&\textsc{eval}(\mfop{(get [$E_1 \; \ldots \; E_n$] $\; k$)}) = E_k,
\\
&\textsc{eval}(\mfop{(get \{$c_1 \; E_1 \; \ldots \; c_n \; E_n$\}  $\; c_k$)}) = E_k.
\end{align*}
With these rules in place, the expression for \mfop{Ak} simplifies to
\begin{lstlisting}[mathescape]
  (get [[0.9 0.1] 
        [0.1 0.9]] $v_2$)
\end{lstlisting}
This yields the correct dependency structure for the Bayesian network.




	\section{Inference Algorithms}
	\label{sec:graph-inference-overview}


In the next sections, and in the following chapter, we will consider many different algorithms for Bayesian inference and how to implement them for the FOPPL. 
A natural question is whether we really will need or use all these algorithms --- why not simply provide one ``best'' inference implementation as a default?
Unfortunately, this is not easily done.

Although we will present all inference algorithms in a quite general setting, such that they can be applied to any program written in the FOPPL, most were originally developed with a particular class of models in mind; often, to address deficiencies when applying existing algorithms.
For Monte Carlo based inference, choosing which algorithm to use typically depends on trade-offs between sample efficiency --- i.e.\ the number of iterations of the algorithm needed to produce an acceptable approximation to the posterior distribution --- and the computational cost of the update itself.
As an example, the Gibbs sampling scheme that we describe in the following section iteratively samples each individual latent variable conditioned on all others; depending on the dependency structure of the model this can be quite fast to execute (high computational efficiency per sample), but in cases where two latent variables are very highly correlated in the posterior, then this can be slow to converge (low statistical efficiency), since each individual update will be unable to make large changes to the values of either variable.
In contrast, Hamiltonian Monte Carlo can easily handle correlated latent variables, but each individual sampling step requires multiple potentially expensive gradient evaluations.
Meanwhile, the algorithms in the next chapter which are based on sequential Monte Carlo or particle filtering were designed for inference in state space models, and typically are best suited to models in which evidence is presented incrementally --- in our parlance, meaning that the \ang{sample} statements are interleaved with \ang{observe} statements, rather than many \ang{sample} statements followed by a single block of code containing all observations.

Other approximate inference algorithms, including expectation propagation and variational inference, make an altogether different set of trade-offs. 
With scalability to high-dimensional models and large sample sizes in mind, these algorithms forgo asymptotic guarantees of approximating the posterior, instead fitting a parametric approximation.
As these are typically faster than Monte Carlo methods, they can be useful in time-critical settings, or for very large models. 
The potential downside is that the higher-order moments may be inaccurately estimated; in particular this means that these methods may be inappropriate in cases where precise estimates of uncertainty are critical.

In general, there is no explicit formula for deciding what inference algorithm is the best fit for a particular model and application.
This can sometimes mean it is beneficial to try out multiple algorithms and then decide between them based on external diagnostics (e.g.\ convergence checks for Monte Carlo methods).
A key advantage of using a probabilistic programming language is that all the approaches to inference we will implement for the PPL can automatically be applied to any model we may write.

	\section{Gibbs Sampling}
	\label{sec:gibbs}



So far, we have defined a way to translate probabilistic programs into a data structure for finite graphical models.
One important reason for doing so is that many existing inference algorithms are defined explicitly in terms 
of finite graphical models, and can now be applied directly to probabilistic programs written in the FOPPL.  We will consider such algorithms now, starting with a general family of Markov chain Monte Carlo (MCMC) algorithms. 

MCMC algorithms perform Bayesian inference by drawing samples from the posterior distribution $p(X \mid Y)$ by simulating a Markov chain whose transition kernel is defined such that the stationary distribution is the target posterior $p(X \mid Y)$. These samples are then used to characterize the distribution of the return value $r(X)$. Procedurally, MCMC algorithms begin by initializing latent variables to some value $X^{(0)}$ and generate a sequence of samples $X^{(1)}, \dots, X^{(S)}$ in which each $X^{(s)} \sim q(\cdot Y, \mid X^{(s-1)})$ is sampled from a probability density that is known as a transition kernel. Repeatedly sampling from a transition kernel produces a Markov chain, a sequence of random variables in which each $X^{(s)}$ depends only on $X^{(s-1)}$. Intuitively, we can think of this Markov chain as a biased ``random walk'' in which samples $p(X^{(s)} \mid Y) $ that have a higher probability under the posterior occur more frequently. 

The main technical requirement for MCMC is that the transition kernel leaves the posterior $p(X \mid Y)$ invariant, 
\begin{align}
  p(X' \mid Y) = \int  q(X' \mid Y, X) \: p(X \mid Y) \: dX. 
\end{align}
This condition ensures that if we sample $X \sim p(X \mid Y)$ and generate a new sample $X' \sim q(X' \mid X,Y)$ from the transition kernel, then this new sample $X'$ is also a sample from $p(X' \mid Y)$. A sufficient condition for invariance is that the kernel satisfies detailed balance
\begin{align}
    \label{eq:gibbs-detailed-balance}
    q(X' \mid Y, X) \: p(X \mid Y)
    =
    q(X \mid Y, X') \: p(X' \mid Y).
\end{align}
In this introduction, we will not delve deeply into the exact technical requirements needed to ensure that MCMC produces posterior samples; rather, we will simply describe how these algorithms can be applied in the context of inference in graphs produced by FOPPL compilation in the previous sections. For a review of MCMC methods, see e.g.~\citet{neal1993probabilistic}, \citet{gelman2013bayesian}, or \citet{bishop2006pattern}.

Two widely employed strategies for MCMC inference are Metropolis-Hastings (MH) and Gibbs sampling. The MH algorithm provides a general recipe for constructing a kernel that satisfies detailed balance by applying an accept / reject correction to samples from a proposal. Given a proposal distribution $q(X' \mid V)$, the MH algorithm samples a candidate $X'$ from $q(X' \mid V)$ conditioned on the value of the current sample $X$, and then evaluates the acceptance probability
\begin{align}
\label{eq:gibbs-mh-accept}
\alpha(X', X) = 
\min 
\left 
\{ 1,   
\frac{p(Y, X') q(X \mid Y, X')}{p(Y, X)q(X' \mid Y, X)} \right 
\}.
\end{align}
With probability $\alpha(X',X)$, we ``accept'' the transition $X \to X'$ and with probability $1-\alpha(X',X)$ we ``reject'' the transition and retain the current sample $X \to X$. 
\begin{align*}
  \begin{split}
  X' 
  &\sim 
  q(X' \mid Y, X^{(s-1)}), 
  \\
  u 
  &\sim 
  \text{Uniform}(0,1)
  \end{split}
  &
  X^{(s)}
  =
  &
  \begin{cases}
    X' & u \le \alpha(X', X^{(s-1)}), \\
    X^{(s-1)} & u > \alpha(X', X^{(s-1)}).
  \end{cases}
\end{align*}
If we compare Equation~\eqref{eq:gibbs-detailed-balance} to Equation~\eqref{eq:gibbs-mh-accept}, then we see that any proposal that satisfies detailed balance will be accepted with probability 1. For proposals that do not satisfy detailed balance, the MH acceptance ratio applies a correction; it rejects samples that are proposed frequently, but have a low posterior probability, whereas it replicates samples that are proposed infrequently, but have a high posterior probability.

While MH samplers are very general, they are not necessarily efficient; a proposal that strongly violates detailed balance can have a very low acceptance ratio. In many inference problems, the posterior is high-dimensional and tightly peaked. This makes it difficult to update all variables jointly, since a poor choice for any one variable will cause the proposal to be rejected. 

Gibbs sampling algorithms \citep{geman1984} sidestep the difficulties associated with high-dimensional proposals by updating one variable at a time. In order to update all variables in a model, Gibbs sampling performs a ``sweep'' of updates to individual variables $x$ by sampling from the so-called full conditional distribution
\begin{align}
  q(x \mid Y, X) = p(x \mid Y, X \setminus \{x\}).
\end{align} 
These updates satisfy detailed balance by construction,  since 
\begin{align*}
  p(x\!=\!c' \mid Y, X \setminus \{x\}) 
  \: p(X \mid Y)
  &=
  p(X' \mid Y) \:
  p(x\!=\!c \mid Y, X \setminus \{x\}).
\end{align*}
As long as we can provide some mechanism to generate samples from each of the full conditional distributions 
$p(x \mid X \setminus Y, \{x\})$, Gibbs sampling allows us to replace a single high-dimensional problem with a sequence of lower-dimensional problems. 
This approach is particularly effective in certain classes of models where the conditional distributions can be derived analytically. This typically requires making use of conditional independence and conjugacy properties in a model. In the Beta-Bernoulli model in Section~\ref{sec:model-based-reasoning}, for example, we were able to make use of the fact that the Beta prior is conjugate to the Bernoulli likelihood to derive the exact conditional updates. 

In probabilistic programming, we can exploit conjugacy in certain cases, but we also need to accommodate cases where these optimizations are not possible. A general-purpose solution to this problem is to employ a {\em Metropolis-within-Gibbs} approach, which uses MH updates to target each of the conditional distributions $p(x \mid Y, X \setminus \{x\})$ using some proposal $q(x \mid Y, X)$.

\subsection{Evaluating the Metropolis-Hastings Acceptance Ratio}
 
One of the computational bottlenecks Metropolis-within-Gibbs samplers is evaluating the acceptance ratio in Equation~\eqref{eq:gibbs-mh-accept}. This ratio requires that we evaluate $p(Y, X \setminus \{x\}, x\!=\!c')$ and $p(Y, X \setminus \{x\}, x\!=\!c)$. In a model with unobserved variables $X = \{x_1, \ldots, x_N\}$ and observed variables $Y = \{y_1, \ldots, y_M\}$ this requires $O(N+M)$ computation. When we update each variable individually, we have to evaluate the acceptance ratio $N$ times for each Gibbs sweep, which means that a full update to all variables requires $O(N \: (N+M))$ computation. However, it turns out that we can avoid much of this computation if we make use of the fact that changes to a single variable $x$ leave probabilities for many variables in the model unchanged.

From an implementation point of view, we can compute the acceptance probability in Equation~\eqref{eq:gibbs-mh-accept} by evaluating the expressions $\m{P}(v)$ for each $v \in V$, for a graph $(V,A,\m{P},\m{Y})$. As before we decompose $V = Y \cup X$ into the set of observed variables $Y$ and the set of unobserved variables $X$. We will now use the notation $\m{X}$ to refer to the map from variables to their values,
\begin{align}
  \m{X} &= [x_1 \mapsto c_1, \ldots, x_N \mapsto c_N].
\end{align}
We can then express the joint probability as 
\begin{align}
  p(Y = \m{Y}, X = \m{X})
  =
  \prod_{v \in V}
  \textsc{eval}(\m{P}(v)[Y:=\m{V}, X:=\m{X}]).
\end{align}
When we update a single variable $x$ using a kernel $q(x \mid V)$, 
we are proposing a new mapping $\m{X}' = \m{X}[x \mapsto c']$, 
where $c'$ is the candidate value proposed for $x$.
The acceptance probability for changing the value of $x$ from $c$ to $c'$ 
then takes the form
\begin{align*}
\alpha(\m{X}',\m{X}) = 
\min 
\left\{ 
1,   
\frac{p(Y= \m{Y}, X \!=\! \m{X}') \: q(x \!=\! c \mid Y \!=\! \m{Y}, X \!=\! \m{X}')}
     {p(Y= \m{Y}, X \!=\! \m{X})  \: q(x \!=\! c' \mid Y \!=\! \m{Y}, X \!=\! \m{X})} 
\right \}.
\end{align*}
In this importance ratio, any conditional probabilities for variables $v$ that do not directly depend on $x$ will remain unchanged and therefore cancel out. The probability of a variable $v$ depends on $x$ when either $v = x$, or when $x \in \pa(v)$. In other words, we can avoid unneccesary computation by only computing the probabilities for variables $V_x$  that are influenced by $x$
\begin{align}
  \begin{split}
  V_x 
  :=&~ 
  \{ v: x \in \textsc{free-vars}(\m{P}(v)) \},
  =
  \{ x \} \cup \{v: x \in \pa(v)\},
  \end{split}
\end{align}
%
which simplifies the computation of the ratio of joint probabilities to
\begin{align}
\label{eq:gibbs-ratio-joint}
\frac{p(Y= \m{Y}, X = \m{X}')}
     {p(Y= \m{Y}, X = \m{X})}
=
\frac{\prod_{v \in V_x} \textsc{eval}(\m{P}(v)[Y:=\m{V}, X:=\m{X}'])}
     {\prod_{v \in V_x} \textsc{eval}(\m{P}(v)[Y:=\m{V}, X:=\m{X}])}
\end{align}
This means that we can compute the acceptance ratio in $O(|V_x|)$ time rather than $O(|V|)$. In many classic model families in probabilistic machine learning, such as mixture models and state space models, we have a set of \emph{global} variables (e.g.~the cluster means) whose dimensionality does not increase with the size of the data, and a set of \emph{local} variables (e.g.~the cluster assignments) for each data point. In these models we can typically compute the acceptance ratio for local variables in constant time, leading to a computation time for the full Gibbs sweep that is \emph{linear} in the number of data points, rather than quadratic, which greatly increases the scalability of the sampling algorithm.

\subsection{Choosing Proposals}

To implement a Metropolis-within-Gibbs sampler, we need to specify some
form of proposal. We will here represent the proposal using a map $\m{Q}$ from unobserved variables to expressions in the target language
\begin{align}
  \m{Q} := [x_1 \mapsto E_1, \ldots, x_N \mapsto E_N].
\end{align}
For each variable $x_n$, the expression $E_n$ defines a distribution, which can in
principle depend on other unobserved variables $X$. We can then use this
distribution to both generate samples and evaluate the forward and
reverse proposal densities. To
do so, we first evaluate the expression to a distribution
\begin{align}
  d = \textsc{eval}(\m{Q}(x)[Y := \m{Y}, X := \m{X}]).
\end{align}
We then assume that we have an implementation for functions $\textsc{sample}$ and
$\textsc{log-prob}$ which allow us to generate samples and evaluate the
density function for the distribution
\begin{align}
  c' &= \textsc{sample}(d), 
  &
  q(x = c' \mid Y, X) = \textsc{log-prob}(d, c')
  .
\end{align}

\begin{algorithm}[!t]
  \caption{
    \label{alg:graph-gibbs}
    Gibbs Sampling with Metropolis-Hastings Updates}
  \begin{algorithmic}[1]
    \State \textbf{global} $V,X,Y,\m{P},\m{Y}$
    \Comment{A directed graphical model}
    \State \textbf{global} $\m{Q}$
    \Comment{A map of proposal expressions}
    \Function{accept}{$x, \m{X}', \m{X}$}
      \State $d \gets \textsc{eval}(\m{Q}(x)[Y:=\m{Y}, X:=\m{X}])$
      \State $d' \gets \textsc{eval}(\m{Q}(x)[Y:=\m{Y}, X:=\m{X}'])$
      \State $\log \alpha \gets \textsc{log-prob}(d', \m{X}(x)) - \textsc{log-prob}(d, \m{X}'(x))$
      \State $V_{x} \gets \{ v \colon x \in \textsc{free-vars}(\m{P}(v)) \}$
      \For{$v$ \textbf{in} $V_x$}
        \State $\log \alpha \gets \log \alpha + \textsc{eval}(\m{P}(v)[Y:=\m{Y}, X:=\m{X}'])$
        \State $\log \alpha \gets \log \alpha - \textsc{eval}(\m{P}(v)[Y:=\m{Y}, X:=\m{X}])$
      \EndFor
      \State \Return $\alpha$
    \EndFunction
    \Function{gibbs-step}{$\m{X}$}
    \For {$x$ \textbf{in} $X$}
        \State $d \gets \Call{eval}{\m{Q}(x)[Y:=\m{Y}, X:=\m{X}]}$
        \State $\m{X}' \gets \m{X}$
        \State $\m{X}'(x) \gets \Call{sample}{d}$
        \State $\alpha \gets \Call{accept}{x, \m{X'}, \m{X}}$
        \State $u \sim \text{Uniform}(0,1)$
        \If{$u < \alpha$}
           \State $\m{X} \gets \m{X}'$
        \EndIf
      \EndFor
      \State \Return $\m{X}$
    \EndFunction
    \Function{gibbs}{$\m{X}^{(0)}, S$}
    \For {$s$ \textbf{in} $1, \ldots, S$}
      \State $\m{X}^{(s)} \gets \Call{gibbs-step}{\m{X}^{(s-1)}}$
    \EndFor
    \EndFunction
    \State \Return $\m{X}^{(1)}, \ldots \m{X}^{(S)}$
  \end{algorithmic}
\end{algorithm}

Algorithm \ref{alg:graph-gibbs} shows pseudo-code for a Gibbs sampler with
this type of proposal. In this algorithm we have several choices for the type
of proposals that we define in the map $\m{Q}$. A straightforward option is to
use the prior as the proposal distribution. 

Instead of proposing from the prior, we can also consider a broader class of
proposal distributions. For example, a common choice for continuous random
variables is to propose from a Gaussian distribution with small standard
deviation, centered at the current value of $x$; there exist schemes to tune
the standard deviation of this proposal online during sampling
\citep{latuszynski2013adaptive}. In this case, the proposal is symmetric, which is to say that $q(x' \mid x) = q(x \mid x')$, which means that the acceptance probability simplifies to the same form as in Equation~\eqref{eq:gibbs-ratio-joint}.
\begin{align}
  \alpha(\m{X}', \m{X})
  =
  \frac{p(Y= \m{Y}, X = \m{X}')}
       {p(Y= \m{Y}, X = \m{X})}.
\end{align}


\subsection{Block Gibbs sampling}

An extension of Gibbs sampling is to  perform ``block updates'' over groups of variables, rather than cycling through variables  one at a time.
This can be advantageous in cases where two latent variables are highly correlated. As a pathological example, consider the FOPPL program
\begin{foppl}[mathescape]
(let [$x_0$ (sample (normal $0$ $1$))
      $x_1$ (sample (normal $0$ $1$))]
  (observe (normal (+ $x_0$ $x_1$) 0.01) 2.0))
\end{foppl}
In this program, we observe that the sum of two standard normal random variates is very close to $2.0$. If initialized at any particular pair of values $(x_0, x_1)$ for which $x_0 + x_1 \approx 2.0$, a Gibbs sampler which updates 
one random choice at a time will quickly become ``stuck''; conditioned on $x_0$ we can only make very small changes to $x_1$ and vice versa. In this type of setting, a high-dimensional block proposal that updates both variables can improve the efficiency of a sampler on balance, even if it is in general more difficult to generate high-quality proposals for higher-dimensional distributions.

%
Consider instead a proposal which updates a subset of latent variables $S \subseteq X$, according to a proposal $q(S \mid V \setminus S)$.
The ``trivial'' choice of proposal distribution --- proposing values of each random variable $x$ in $S$ by simulating from the prior $p(x \mid \pa(x))$ --- would, for $S = \{ x_0, x_1 \}$ in this example, sample both values from their independent normal priors.
While this is capable of making larger moves 
(unlike the previous one-choice-at-a-time proposal, it would be possible for this proposal to go in a single step from e.g. $(2.0, 0.0)$ to $(0.0, 2.0)$), 
with this na\"ive choice of block proposal overall performance may actually be worse than that with independent proposals; 
now instead of sampling a single value which needs to be ``near'' the previous value to be accepted, we are sampling two values, where the second value $x_1$ needs to be ``near'' the sampled $x_0 - 2.0$, something quite unlikely for negative values of $x_0$.

Constructing block proposals which have high acceptance rates require taking account of the structure of the model itself.
One way of doing this adaptively, analogous to estimating posterior standard deviations to be used as scale parameters in univariate proposals, is to estimate posterior covariance matrices and using these for jointly proposing multiple latent variables \citep{haario2001adaptive}.

\subsection{Implementation Considerations}

As we noted above, it is sometimes possible to analytically derive the complete conditional distribution of a single variable.  Such cases include all random variables whose value is discrete from a finite set, many settings in which all the densities in $V_x$ are in the exponential family, and so forth. Programming languages techniques can be used to identify such opportunities by performing pattern matching analyses of the source code of the link functions in $V_x.$  If, as is the case in the simplest example, $x$ itself is determined to be governed by a discrete distribution then, instead of using Metropolis within Gibbs, one would merely enumerate all possible values of $x$ under the support of its distribution, score each, normalize, then sample from this exact conditional.  

Inference algorithms vary in their performance, sometimes dramatically.  Metropolis Hastings within Gibbs is sometimes efficient but even more often is not, unless utilizing intelligent block proposals (often, ones customized to the particular model).
This has led to a proliferation of inference algorithms and methods, some but not all of which are directly applicable to probabilistic programming.  In the next section, we consider Hamiltonian Monte Carlo, which incorporates gradient information to provide efficient high-dimensional block proposals.   

%
%


    \section{Compilation to a Factor Graph}
    \label{sec:factor-graph}

In Section~\ref{sec:eval-dens}, we showed that a Bayesian network is a representation of a joint probability $p(Y, X)$ of observed random variables $Y$, each of which corresponds to an \ang{observe} expression, and unobserved random variables $X$, each of which corresponds to a \ang{sample} expression. Given this representation, we can now reason about a posterior probability $p(X \,|\, Y)$ of the sampled values, conditioned on the observed values. In Section~\ref{sec:factor-condition}, we showed that we can generalize this representation to an unnormalized density $\gamma(X) = \psi(X) p(X)$ consisting of a directed network that defines a prior probability $p(X)$ and a potential term (or factor) $\psi(X)$. In this section, we will represent a probabilistic program in the FOPPL as a factor graph, which is a fully undirected network. We will use this representation in Section~\ref{sec:expectation-propagation} to define an expectation propagation algorithm.

A factor graph defines an unnormalized density on a set of variables $X$ in terms of a product over an index set $F$
\begin{align}
    \label{eq:gamma-factor}
    \gamma(X)
    &
    :=
    \prod_{f \in F}
    \psi_f(X_f)
    ,
\end{align}
in which each function $\psi_f$, which we refer to as a factor, is itself an unnormalized density over some subset of variables $X_f \subseteq X$. We can think of this model as a bipartite graph with variable nodes $X$, factor nodes $F$ and a set of undirected edges $A \subseteq X \times F$ that indicate which variables are associated with each factor 
\begin{align}
    X_f := \{x: (x,f) \in A \}.
\end{align}
Any directed graphical model $(V,A,\m{P},\m{Y})$ can be interpreted as a factor graph in which there is one factor $f \in F$ for each variable $v \in V$. In other words we could define
\begin{align}
    \gamma(X) &:= \prod_{v \in V} \psi_v(X_v),
\end{align}
where the factors $\psi_v(X_v)$ equivalent to the expressions $\m{P}(v)$ that evaluate the probability density for each variable $v$, which can be either observed or unobserved,
\begin{align}
    \psi_v(X_v) 
    &\equiv
    \begin{cases}
        \m{P}(v)[v := \m{Y}(v)],
        &
        v \in \text{dom}(\m{Y}), \\
        \m{P}(v), 
        &
        v \not\in \text{dom}(\m{Y}).
    \end{cases}
\end{align}

\begin{figure}
\begin{gather*}
    \infer{\rho,c \Downarrow_f c}{}
    \qquad
    \infer{\rho,v \Downarrow_f v}{}
    \qquad
    \infer{\rho,\mang{(let [$v$\ $e_1$]\ $e_2$)} 
           \Downarrow_f 
           \mang{(let [$v$\ $e^\prime_1$]\ $e^\prime_2$)}}
          {e_1 \Downarrow_f e_1' \qquad e_2 \Downarrow_f e_2'}
    \\[8pt]
    \infer{\rho,\mang{(sample\ $e$)} \Downarrow_f \mang{(sample\ $e^\prime$)}}
          {e \Downarrow_f e'}
    \quad
    \infer{\rho,\mang{(observe\ $e_1$\ $e_2$)} 
           \Downarrow_f 
           \mang{(observe\ $e_1^\prime$\ $e^\prime_2$)}}
          {e_1 \Downarrow_f e_1' \qquad e_2 \Downarrow_f e_2'}
    \\[8pt]
    \infer{
      \rho,\mang{($f$\ $e_1$\ $\ldots$\ $e_n$)}
      \Downarrow_f
      \mang{($f^\prime$\ $e^\prime_1$\ $\ldots$\ $e^\prime_n$)}
    }{
        \begin{aligned}
        \rho,e_i
        &\Downarrow_f
        e'_i
        ~~\text{for}~i=0,\ldots,n 
        &
        \rho(f) 
        &= 
        \mang{(defn [$v_1$ $\ldots$ $v_n$]\ $e_0$)}
        \\
        \rho, e_0 
        &\Downarrow_f
        e'_0
        &
        \rho(f') 
        &= 
        \mang{(defn [$v_1$ $\ldots$ $v_n$]\ $e^\prime_0$)}
        \end{aligned}
    }
    \\[8pt]
    \infer{
      \rho,\mang{($op$\ $e_1$\ $\ldots$\ $e_n$)}
      \Downarrow_f
      \mang{(sample (dirac ($op$\ $e^\prime_1$\ $\ldots$\ $e^\prime_n$)))}
    }
    {\rho,e_i
     \Downarrow_f
     e'_i
     ~\text{for}~i=1,\ldots,n 
      \qquad
      op = \mang{if}
      \quad
      \text{or}
      \quad 
      op = c   
    }
\end{gather*}
\caption{\label{fig:factor-transformation}Inference rules for the transformation $\rho, e \Downarrow_f e'$, which replaces if forms and primitive procedure calls with expressions of the form \mang{(sample (dirac\ $e$))}.}
\end{figure}

A factor graph representation of a density is not unique. For any factorization, we can merge two factors $f$ and $g$ into a new factor $h$
\begin{align}
    \psi_{h}(X_{h})
    &:= 
    \psi_f(X_f) \psi_g(X_g),
    &
    X_{h}
    :=
    X_f \cup X_g.
\end{align}
A graph in which we replace the factors $f$ and $g$ with the merged factor $h$ is then an equivalent representation of the density. The implication of this is that there is a choice in the level of granularity at which we wish to represent a factor graph. The representation above has a comparatively low level of granularity.
We will here consider a more fine-grained representation, analogous to the one used in Infer.NET \citep{minkainfer}. In this representation, we will still have one factor for every variable, but we will augment the set of nodes $X$ to contain an entry $x$ for every deterministic expression in a FOPPL program. We will do this by
defining a source code transformation $\rho,e \Downarrow_f e'$ that replaces each deterministic sub-expressions (i.e.~if expressions and primitive procedure calls) with expressions of the form
\[
    \mang{(sample (dirac\ $e^\prime$))}
\] 
Where \ang{(dirac $e^\prime$)} refers to the Dirac delta distribution with density 
\[
    p_{\mathsf{dirac}}(x \,;\, c) = I[x = c]
\]
After this source code transformation, we can use the rules from Section~\ref{sec:compilation-bounded} to compile the transformed program into a directed graphical model $(V,A,\m{P},\m{Y})$. This model will be equivalent to the directed graphical model of the untransformed program, but contains an additional node for each Dirac-distributed deterministic variable.

The inference rules for the source code transformation $\rho,e \Downarrow_f e'$ are much simpler than the rules that we wrote down in Section~\ref{sec:compilation-bounded}. We show these rules in Figure~\ref{fig:factor-transformation}. The first two rules state that constants $c$ and variables $v$ are unaffected. The next rules state that let, sample, and observe forms are transformed by transforming each of the sub-expressions, inserting deterministic variables where needed. User-defined procedure calls are similarly transformed by transforming each of the arguments $e_1, \ldots, e_n$, and transforming the procedure body $e_0$. So far, none of these rules have done anything other than state that we transform an expression by transforming each of its sub-expressions. The two cases where we insert Dirac-distributed variables are if forms and primitive procedure applications. For these expression forms $e$, we transform the sub-expressions to obtain a transformed expression $e'$ and then return the wrapped expression \ang{(sample (dirac $e'$))}.

As noted above, a directed graphical model can always be interpreted as a factor graph that contains single factor for each random variable. To aid discussion in the next section, we will explicitly define the mapping from the directed graph $(V^\text{dg},A^\text{dg},\m{P}^\text{dg},\m{Y}^\text{dg})$ of a transformed program onto a factor graph $(X,F,A,\Psi)$ that defines a density of the form in Equation~\eqref{eq:gamma-factor}.

A factor graph $(X,F,A,\Psi)$ is a bipartite graph in which $X$ is the set of variable nodes, $F$ is the set of factor nodes, $A$ is a set of undirected edges between variables and factors, and $\Psi$ is a map of factors that will be described shortly. The set of variables is identical to the set of unobserved variables (i.e. the set of sample forms) in the corresponding directed graph
\begin{align}
    X := X^\text{dg} = V^\text{dg} \setminus \text{dom}(\m{Y}^\text{dg}).
\end{align}
We have one factor $f \in F$ for every variable $v \in V^\text{dg}$, which includes \emph{both} unobserved variables $x \in X^\text{dg}$, corresponding to sample expressions, and observed variables $y \in Y^\text{dg}$. We write $F \overset{1-1}= V^\text{dg}$ to signify that there is a bijective relation between these two sets and use $v_f \in V$ to refer to the variable node that corresponds to the factor $f$. Conversely we  use $f_v \in F$ to refer to the factor that corresponds to the variable node $v$. We can then define the set of edges $A \overset{1-1}= A^\text{dg}$ as 
\begin{align}
    A := \{(v,f): (v,v_f) \in A^\text{dg}\}.
\end{align}
The map $\Psi$ contains an expression $\Psi(f)$ for each factor, which evaluates the potential function of the factor $\psi_f(X_f)$. We define $\Psi(f)$ in terms of the of the corresponding expression for the conditional density $\m{P}^\text{dg}(v_f)$,
\begin{align}
    \Psi(f) 
    &:=
    \begin{cases}
        \m{P}^\text{dg}(v_f)[v_f := \m{Y}^\text{dg}(v_f)],
        &
        v_f \in \text{dom}(\m{Y}^\text{dg}), \\
        \m{P}^\text{dg}(v_f), 
        &
        v_f \not\in \text{dom}(\m{Y}^\text{dg}).
    \end{cases}
\end{align}
This defines the usual correspondence between $\psi_f(X_f)$ and $\Psi(f)$, where we note that the set of variables $X_f$ associated with each factor $f$ is equal to the set of  variables in $\Psi(f)$,
\begin{align}
    \psi_f(X_f)
    &\equiv
    \Psi(f)
    ,
    &
    X_f 
    &= 
    \textsc{free-vars}(\Psi(f)).
\end{align}
For purposes of readability, we have omitted one technical detail in this discussion. In Section~\ref{sec:partial-eval}, we spent considerable time on techniques for partial evaluation, which proved necessary to avoid graphs that contain spurious edges for between variable that are in fact conditionally independent. In the context of factor graphs, we can similarly eliminate unnecessary factors and variables. Factors that can be eliminated are those in which the expression $\Psi(f)$ either takes the form \ang{($p_\textsf{dirac}$ $v$ $c$)} or \ang{($p_\textsf{dirac}$ $c$ $v$)}. In such cases we remove the factor $f$, the node $v$, and substitute $v := c$ in the expressions of all other potential functions. Similarly, we can eliminate all variables with factors of the form \ang{($p_\textsf{dirac}$ $v_1$ $v_2$)} by substituting $v_1 := v_2$ everywhere. 

To get a sense of how a factor graph differs from a directed graph, let us look at a simple example, inspired by the TrueSkill model \citep{herbrich2007trueskill}. Suppose we consider a match between two players who each have a skill variable $s_1$ and $s_2$. We will assume that the player 1 beats player 2 when $(s_1 - s_2) > \epsilon$, which is to say that the skill of player 1 exceeds the skill of player 2 by some margin $\epsilon$. Now suppose that we define a prior over the skill of each player and observe that player 1 beats player 2. Can we reason about the posterior on the skills $s_1$ and $s_2$? We can translate this problem to the following FOPPL program
\begin{foppl}
(let [s1 (sample (normal 0 1.0))
      s2 (sample (normal 0 1.0))
      delta (- s1 s2)
      epsilon 0.1
      w (> delta epsilon)
      y true]
  (observe (dirac w) y)
  [s1 s2])
\end{foppl}
This program differs from the ones we have considered so far in that we are using a Dirac delta to enforce a \emph{hard} constraint on observations, which means that this program defines an unnormalized density
\begin{align}
  \gamma(s_1, s_2) 
  = 
  \big( 
    p_\textsf{norm}(s_1 ; 0, 1) 
    \:
    p_\textsf{norm}(s_2 ; 0, 1) 
  \big)^{I[(s_1 - s_2) > \epsilon]}
  .
\end{align}
This type of hard constraint poses problems for many inference algorithms for directed graphical models. For example, in HMC this introduces a discontinuity in the density function. However, as we will see in the next section, inference methods based on message passing are much better equipped to deal with this form of condition. 

When we compile the program above to a factor graph we obtain a set of variables $X=(s_1,s_2,\delta,w)$ and the map of potentials 
\begin{align}
  \Psi = 
  \left[
    \begin{aligned}
      f_{s_1} &\mapsto \mang{($p_\textsf{norm}$\ $s_1$ 0.0\ 1.0)}, \\
      f_{s_2} &\mapsto \mang{($p_\textsf{norm}$\ $s_2$ 0.0\ 1.0)}, \\
      f_{\delta} &\mapsto \mang{($p_\textsf{dirac}$\ $\delta$ (-\ $s_1$\ $s_2$))}, \\
      f_{w} &\mapsto \mang{($p_\textsf{dirac}$\ $w$ (>\ $\delta$\ 0.1))}, \\
      f_{y} &\mapsto \mang{($p_\textsf{dirac}$ true\ $w$)}
    \end{aligned}
  \right].
\end{align}
Note here that the variables $s_1$ and $s_2$ would also be present in the directed graph corresponding to the unstransformed program. The deterministic variables $\delta$
and $w$ have been added as a result of the transformation in Figure~\ref{fig:factor-transformation}. Since the factor $f_y$ restricts $w$ to the value \ang{true}, we can eliminate $f_y$ from the set of factors and $w$ from the set of variables. This results in a simplified graph where $X=(s_1, s_2, \delta)$ and the potentials
\begin{align}
  \Psi = 
  \left[
    \begin{aligned}
      f_{s_1} &\mapsto \mang{($p_\textsf{norm}$\ 0.0\ 1.0)}, \\
      f_{s_2} &\mapsto \mang{($p_\textsf{norm}$\ 0.0\ 1.0)}, \\
      f_{\delta} &\mapsto \mang{($p_\textsf{dirac}$\ $\delta$ (-\ $s_1$\ $s_2$))}, \\
      f_{w} &\mapsto \mang{($p_\textsf{dirac}$ true (>\ $\delta$\ 0.1))}
    \end{aligned}
  \right].
\end{align}

In summary, we have now created an undirected graphical model, in which there is  deterministic variable node $x \in X$ for all primitive operations such as \ang{(> $v_1$ $v_2$)} or \ang{(- $v_1$ $v_2$)}. In the next section, we will see how this representation helps us when performing inference.

    \section{Expectation Propagation}
	\label{sec:expectation-propagation}

One of the main advantages in representing a probabilistic program as a factor graph is that we can perform inference with message passing algorithms. As an example of this we will consider expectation propagation (EP), which forms the basis
of runtime of Infer.NET \citep{minkainfer}, a highly influential probabilistic programming system.

EP considers an unnormalized density $\gamma(X)$ that is defined in terms of a factor graph $(X,F,A,\Psi)$. As noted in the last section, a factor graph defines a density as a product over an index set $F$
\begin{align}
    \pi(X)
    &:=
    \gamma(X)/Z^{\pi},
    &
    \gamma(X)
    &
    :=
    \prod_{f \in F}
    \psi_f(X_f)
    .
\end{align}
We approximate $\pi(X)$ with a distribution $q(X)$ that is similarly defined as a product over factors 
\begin{align}
  q(X) := \frac{1}{Z^q} \prod_{f \in F} \phi_f(X_f).
\end{align}
The objective in EP is to make $q(X)$ as similar as possible to $\pi(X)$ by minimizing the Kullback-Leibler divergence
\begin{align}
    \label{eq:ep-kl-div}
    \argmin_q
    \KL{\pi(X)}{q(X)} 
    &= 
    \argmin_q
    \int \pi(X) \log \frac{\pi(X)}{q(X)} dX
    ,
\end{align}
EP algorithms minimize the KL divergence iteratively by updating one factor $\phi_f$ at a time
\begin{itemize}
\item Define a tilted distribution
\begin{align}
    \label{eq:ep-tilted}
    \pi_{f}(X) 
    &:= 
    \gamma_{f}(X) / Z_f
    ,
    &
    \gamma_{f}(X) 
    :=
    \frac{\psi_f(X_f)}
         {\phi_f(X_f)}
    q(X)
    .
\end{align}
\item Update the factor by minimizing the KL divergence
\begin{align}
    \phi_{f} = \argmin_{\phi_{f}} \KL{\pi_f(X)}{q(X)}.
\end{align}
\end{itemize}

In order to ensure that the KL minimization step is tractable, EP methods rely on the properties of exponential family distributions. We will here consider the variant of EP that is implemented in Infer.NET, which assumes a fully-factorized form for each of the factors in $q(X)$
\begin{align}
    \phi_f(X_f)
    &:=
    \prod_{x \in X_f} 
    \phi_{f \to x}(x)
    .
\end{align}
We refer to the potential $\phi_{f \to x}(x)$ as the message from factor $f$ to the variable $x$. We assume that messages have an exponential form 
\begin{align}
    \label{eq:ep-message}
    \phi_{f \to x}(x)
    &= 
    \exp[\lambda_{f \to x}^\top t(x)]
    ,
\end{align}
in which $\lambda_{f \to x}$ is the vector of natural parameters and $t(x)$ is the vector of sufficient statistics of an exponential family distribution. We can then express the marginal $q(x)$ as an exponential family distribution
\begin{align}
\begin{split}
    q(x) 
    &= 
    \frac{1}{Z^q_x} 
    \prod_{f: x \in V_f} 
    \phi_{f \to x}(x),
    \\
    &=
    h(x)
    \exp
    \left(
    \lambda_x t(x) - a(\lambda_x)
    \right),
\end{split}
\end{align}
where $a(\lambda_x)$ is the log normalizer of the exponential family and $\lambda_x$ is the sum over the parameters for individual messages
\begin{align}
    \label{eq:ep:lambda-v}
    \lambda_x = \sum_{f: x \in X_f} \lambda_{f \to x}.
\end{align}
Note that we can express the normalizing constant $Z^q$ as a product over per-variable normalizing constants $Z^q_x$,
\begin{align}
    Z^q
    &:= 
    \prod_{x \in X} 
    Z^q_x,
    &
    Z^q_x 
    &:= 
    \int dx \:
    \prod_{f: x \in X_f} 
    \phi_{f \to x}(x),
\end{align}
where we can compute $Z^q_x$ in terms of $\lambda_x$ using
\begin{align}
    \label{eq:ep-Z-q-x}
    Z^q_x
    = 
    \exp
    \left(
        a(\lambda_x)
    \right)
    =
    \exp
    \left(
        a \left( \textstyle \sum_{f \,:\, x \in X_f} \lambda_{f \to x} \right)
    \right)
    .
\end{align}
Exponential family distributions have many useful properties. One such property is that expected values of the sufficient statistics $t(x)$ can be computed from the gradient of the log normalizer
\begin{align}
    \label{eq:ep-expect-stat}
    \nabla_{\l_x} a(\l_x) = \Ev_{q(x)}[t(x)].
\end{align}
In the context of EP, this property allows us to express KL minimization as a so-called moment-matching condition. To explain what we mean by this, we will expand the KL divergence
\begin{align}
    \KL{\pi_f(X)}{q(X)}
    =
    \log
    \frac{Z^q}{Z_f}
    +
    \Ev_{\pi_f(X)}
    \left[
    \log \frac{\psi_f(X_f)}
              {\phi_f(X_f)}
    \right]
    .
\end{align}
We now want to minimize this KL divergence with respect the parameters $\lambda_{f \to v}$. When we ignore all terms that do not depend on these parameters, we obtain
\begin{align*}
    &
    \nabla_{\lambda_{f \to x}}
    \KL{\pi_f(X)}{q(X)}
    =
    \\
    &
    \qquad
    \nabla_{\lambda_{f \to x}}
    \left(
      \log Z_x^q
      - 
      \Ev_{\pi_f(X)}[\log \phi_{f \to x}(x)]
    \right)
    =
    0
    .
\end{align*}
When we substitute the message $\phi_{f \to x}(x)$ from Equation~\eqref{eq:ep-message}, the normalizing constant $Z^q_x(\l_x)$ from Equation~\eqref{eq:ep-Z-q-x}, and apply  Equation~\eqref{eq:ep-expect-stat}, then we obtain the moment matching condition
\begin{align}
    \nonumber
    \Ev_{q(x)}[t(x)]
    &=
    \nabla_{\lambda_{f \to x}}
    \Ev_{\pi_f(X)}[\log \phi_{f \to x}(x)],
    \\
    \label{eq:ep-moment-match}
    &=
    \nabla_{\lambda_{f \to x}}
    \Ev_{\pi_f(X)}[\lambda^\top_{f \to x} t(x)],
    \\
    \nonumber
    &=\Ev_{\pi_f(X)}[t(x)].
\end{align}
If we assume that the parameters $\lambda^*_x$ satisfy the condition above, then we can use Equation~\eqref{eq:ep:lambda-v} to define the update for the message $\phi_{f \to x}$ \begin{align}
    \label{eq:ep-lambda-proj}
    \lambda_{f \to x}
    \gets 
    \lambda^*_x
    -
    \sum_{f' \not= f \,:\, x \in X_{f'}}
    \lambda_{f' \to x}
    .
\end{align}

\begin{algorithm}[!t]
  \caption{
    \label{alg:ep}
    Fully-factorized Expectation Propagation}
  \begin{algorithmic}[1]
    \Function{proj}{$G, \lambda, f$}
        \State $X,F,A,\Psi \gets G$
        \State $\gamma_f(X) \gets \psi_f(X) q(X) / \phi_f(X) $
        \Comment{Equation~\eqref{eq:ep-tilted}}
        \State $Z_f \gets \int d X \: \gamma_f(X)$
        \Comment{Equation~\eqref{eq:ep-Z-f}}
        \For{$x$ in $X_f$}
            \State $\bar{t} \gets 1/Z_f \int dX \: \gamma_f(X) t(x)$
            \Comment{Equation~\eqref{eq:ep-expect-t}}
            \State $\lambda^*_{x} \gets \Call{moment-match}{\bar{t}\,}$
            \Comment{Equation~\eqref{eq:ep-moment-match}}
            \State $\lambda_{f \to x} \gets \lambda_{x}^* - \sum_{f'\not=f \,:\, x \in X_{f'}} \lambda_{f' \to x}$
            \Comment{Equation~\eqref{eq:ep-lambda-proj}}
        \EndFor
        \State \Return $\lambda, \log Z_f$
    \EndFunction
    \Function{ep}{$G$}
        \State $X,F,A,\Psi \gets G$
        \State $\lambda \gets$ \textsc{initialize-parameters}$(G)$
        \For{$f$ in \textsc{schedule(G)}}
            \State $\lambda,  \log Z_f \gets$ \textsc{proj}$(G,\lambda,f)$
        \EndFor
        \For{$x$ in $X$}
            \State $\log Z^q_x \gets a(\lambda_x)$
            \Comment{Equation~\eqref{eq:ep-Z-q-x}}
        \EndFor
        \State $\log Z^\pi \gets \sum_{f} \log Z_f + \sum_{x} \log Z^q_x$
        \Comment{Equation~\eqref{eq:ep-Z-approx}}
        \State \Return $\lambda, \log Z^\pi$
    \EndFunction
  \end{algorithmic}
\end{algorithm}

In order to implement the moment matching step, we have to solve two integrals. The first computes the normalizing constant $Z_f$. We can express this integral, which is nominally an integral over all variables $X$, as an integral over the variables $X_f$ associated with the factor $f$,
\begin{align}
    \nonumber
    Z_f 
    = 
    \int dX \frac{\psi_f(X_f)}{\phi_f(X_f)} q(X)
    &=
    \int dX_f \: 
    \frac{\psi_f(X_f)}
         {\phi_f(X_f)} 
    \prod_{x \in X_f} 
    \frac{1}{Z^q_x}
    \prod_{f' \,:\, x \in V_{f'}}  \phi_{f' \to x}(x),
    \\
    \label{eq:ep-Z-f}
    &=
    \int dX_f \:
    \psi_f(X_f)  
    \prod_{x \in X_{f}}
    \frac{1}{Z^q_x}
    \phi_{x \to f}(x).
\end{align}
Here, the function $\phi_{x \to f}(x)$ is known as the message from the variable $v$ to the factor $f$, which is defined as 
\begin{align}
    \phi_{x \to f}(x) 
    &:=
    \prod_{x \in X_{f}} \prod_{f'\not=f \,:\, x \in X_{f'}  } \phi_{f' \to x}(x).
\end{align}
These messages can also be used to compute the second set of integrals for the sufficient statistics
\begin{align}
    \label{eq:ep-expect-t}
    \bar{t} 
    =
    \Ev_{\pi_f(V)}[t(v)]
    = 
    \frac{1}{Z_f} 
    \int 
    dV_f \: 
    t(x) \psi_f(X_f) 
    \prod_{x \in X_{f}}
    \frac{1}{Z^q_x}
    \phi_{x \to f}(x).
\end{align}

Algorithm~\ref{alg:ep} summarizes these computations. We begin by initializing parameter values for each of the messages. We then pick factors $f$ to update according to some schedule. For each update we then compute $Z_f$. For each $x \in X_f$ we then compute $\bar{t}$, find the parameters $\lambda^*_x$ that satisfy the moment-matching condition and then use these parameters to update parameters $\lambda_{f \to x}$. Finally, we note that EP obtains an approximation to the normalizing constant $Z^\pi$ for the full unnormalized distribution $\pi(X) = \gamma(X) / Z^\pi$. This approximation can be computed from the normalizing constants of the tilted distributions $Z_f$ and the normalizing constants $Z^q_x$,
\begin{align}
    \label{eq:ep-Z-approx}
    Z^\pi 
    \simeq
    \prod_{f \in F}
    Z_f
    \prod_{x \in X}
    Z^q_x
    .
\end{align}

\subsection{Implementation Considerations}
 
There are a number of important considerations when using EP for probabilistic programming in practice. The type of schedule implemented by the function $\textsc{schedule}(G)$ is perhaps the most important design consideration. In general, EP updates are not guaranteed to converge to a fixed point, and choosing a schedule that is close to optimal is an open problem. In fact, a significant portion of the development effort for Infer.NET \citep{minkainfer} has focused on identifying heuristics for choosing this schedule.

As with HMC, there are also restrictions to the types of programs that are amenable to inference with EP. To perform EP, a FOPPL program needs to satisfy the following requirements 
\begin{itemize}
  \item[1.] We need to be able to associate an exponential family distribution with each variable $x$ in the program. 

  \item[2.] For every factor $f$, we need to be able to compute the integral for $Z_f$ in Equation~\eqref{eq:ep-Z-f}. 

  \item[3.] For every message $\phi_{f \to x}(x)$, we need to be able to compute the sufficient statistics $\bar{t}$ in Equation~\eqref{eq:ep-expect-t}.
\end{itemize}

The first requirement is relatively easy to satisfy. The exponential family includes the Gaussian, Gamma, Discrete, Poisson, and Dirichlet distributions, which covers the cases of real-valued, positive-definite, discrete with finite cardinality and discrete with infinite cardinality.

The second and third requirement impose more substantial restrictions on the program. To get a clearer sense of these requirements, let us return to the example that we looked at in Section~\ref{sec:factor-graph}
\begin{foppl}
(let [s1 (sample (normal 0 1.0))
      s2 (sample (normal 0 1.0))
      delta (- s1 s2)
      epsilon 0.1
      w (> delta epsilon)
      y true]
  (observe (dirac w) y)
  [s1 s2])
\end{foppl}
After elimination of unnecessary factors and variables, this program defines a model with variables $X=(s_1, s_2, \delta)$ and potentials
\begin{align}
  \Psi = 
  \left[
    \begin{aligned}
      f_{1} &\mapsto \mang{($p_\textsf{norm}$\ 0.0\ 1.0)}, \\
      f_{2} &\mapsto \mang{($p_\textsf{norm}$\ 0.0\ 1.0)}, \\
      f_{3} &\mapsto \mang{($p_\textsf{dirac}$\ $\delta$ (-\ $s_1$\ $s_2$))}, \\
      f_{4} &\mapsto \mang{($p_\textsf{dirac}$ true (>\ $\delta$\ 0.1))}
    \end{aligned}
  \right].
\end{align}
In fully-factorized EP, we assume an exponential family form for each of the variables $s_1$, $s_2$ and $d_{12}$. The obvious choice here is to approximate each variable with an unnormalized Gaussian, for which the sufficient statistics are $t(x)=(x^2, x)$. The Gaussian marginals $q(s_1)$, $q(s_2)$ and $q(d_{12})$ will then approximate the the corresponding marginals $\pi(s_1)$, $\pi(s_2)$, and $\pi(d_{12})$ of the target density.

Let us now consider what operations we need to implement to compute the integrals in Equation~\eqref{eq:ep-Z-f} and Equation~\eqref{eq:ep-expect-t}. We will start with the case of the integral for $Z_f$ when updating the factor $f_3$,
\begin{align}
  \begin{split}
    Z_f 
    = 
    \frac{1}{Z^q_{s_1}Z^q_{s_2}Z^q_{\delta}}
    \int \! & d s_1 \, d s_2 \, d \delta ~
    I[\delta = s_1 - s_2]
    \\
    &\quad
    \phi_{s_1 \to f_3}(s_1)
    \phi_{s_2 \to f_3}(s_2)
    \phi_{\delta \to f_3}(\delta).
  \end{split}
\end{align}
We can the substitute $\delta := s_1 - s_2$ to eliminate $\delta$, which yields an integral over $s_1$ and $s_2$
\begin{align*}
    Z_f 
    = 
    \frac{1}{Z^q_{s_1}Z^q_{s_2}Z^q_{\delta}}
    \int \! & d s_1 \, d s_2 ~
    \phi_{s_1 \to f_3}(s_1)
    \phi_{s_2 \to f_3}(s_2)
    \phi_{\delta \to f_3}(s_1 - s_2).
\end{align*}
Each of the messages is an unnormalized Gaussian, so this is an integral over a product of 3 Gaussians, which we can compute in closed form.

Now let us consider the case of the update for factor $f_4$. For this factor the integral for $Z_f$ takes the form
\begin{align}
  \begin{split}
  Z_f 
  &= 
  \frac{1}{Z^q_{\delta}}
  \int_{-\infty}^\infty \! 
  d \delta ~
  ~I[\delta > 0.1]~
  \phi_{\delta \to f_4}(\delta),
  \\
  &=
  \frac{1}{Z^q_{\delta}}
  \int_{0.1}^\infty \! 
  d \delta ~
  \phi_{\delta \to f_4}(\delta)
  .
  \end{split}
\end{align}
This is just an integral over a truncated Gaussian, which is also something that we can approximate numerically. 

We now also see why it is advantageous to introduce a factor for each primitive operation. In the case above, if we were to combine the factors $f_3$ and $f_4$ into a single factor, then we would obtain the integral
\begin{align}
  \begin{split}
    Z_f 
    = 
    \frac{1}{Z^q_{s_1}Z^q_{s_2}}
    \int \! & d s_1 \, d s_2 
    I[s_1 - s_2 > 0.1]
    \\
    &\quad
    \phi_{s_1 \to f_{3+4}}(s_1)
    \phi_{s_2 \to f_{3+4}}(s_2)
    .
  \end{split}
\end{align}
Integrals involving constraints over multiple deterministic operations will be much harder to compute in an automated manner than integrals involving constraints over atomic operations. Representing each deterministic operation as a separate factor avoids this problem.

To provide a full implementation of EP for the FOPPL, we need to be able to solve the integral for $Z_f$ in Equation~\eqref{eq:ep-Z-f} and the integrals for the sufficient statistics in Equation~\eqref{eq:ep-expect-t} for each potential type. This requirement imposes certain constraints on the programs we can write. The cases that we have to consider are stochastic factors (\ang{sample} and \ang{observe} expressions) and deterministic factors (if expressions and primitive procedure calls).

For \ang{sample} and \ang{observe} expressions, potentials have the form $\Psi(f) = \mang{($p$\ $v_0$\ $v_1$\ $\ldots$\ $v_n$)}$ and $\Psi(f) = \mang{($p$\ $c_0$\ $v_1$\ $\ldots$\ $v_n$)}$ respectively. For these potentials, we have to integrate over products of densities, which can in general be done only for a limited number of cases, such as conjugate prior and likelihood pairs. This means that the exponential family that is chosen for the messages needs to be compatible with the densities in \ang{sample} and \ang{observe} expressions.

Deterministic factors take the form \mang{($p_\textsf{dirac}$\ $v_0$\ $E$)} where $E$ is an expression in which all sub-expressions are variable references,
\begin{grammar}
  $E ::=$ (if $v_1$ $v_2$ $v_3$) | ($c$ $v_1 \ldots v_n$) 
\end{grammar}
For if expressions \ang{(if $v_1$ $v_2$ $v_3$)}, it is advantageous to employ constructs known as gates \citep{minka2009gates}, which treat the if block as a mixture over two distributions and propagate messages by computing expected values of over the indicator variable accordingly. 

In the case of primitive procedure calls, we need to provide implementations of the integrals that only depend on the primitive $c$, but also on the type of exponential family that is used for the messages $v_1$ through $v_n$. For example, if we consider the expression \ang{(- $v_1$ $v_2$)}, then our implementation for the integrals will be different when $v_1$ and $v_2$ are both Gaussian, both Gamma-distributed, or when one variable is Gaussian and the other is Gamma-distributed.

\chapter{Inference with Evaluators}
\label{ch:eval-one}

In the previous chapter, our inference algorithms operated on a graph
representation of a probabilistic model, which we created through a
compilation of a program in our first-order probabilistic programming
language. Like any compilation step, the construction of this graph is
performed ahead of time, prior to running inference. We refer to graphs that 
can be constructed at compile time as having static support.

There are many models in which the graph of conditional dependencies is
dynamic, in the sense that it cannot be constructed prior to performing
inference. One way that such graphs arise is when the number of random
variables is itself not known at compile time. For example, in a model that performs object
tracking, we may not know how many objects will appear, or for how long they
will be in the field of view. We will refer to these types of models as having dynamic support.

There are two basic strategies that we can employ to represent models with
dynamic support. One strategy is to introduce an upper bound on the number of
random variables. For example, we can specify a maximum number of objects that
can be tracked at any one time. When employing this type of modeling strategy,
we additionally need to specify which variables are needed at any one time.
For example, if we had random variables corresponding to the position of each
possible object, then we would have to introduce auxiliary variables to
indicate which objects are in view. This process of "switching" random
variables "on" and "off" allows us to approximate what is fundamentally a
dynamic problem with a static one.

The second is strategy is to implement inference methods that dynamically
instantiate random variables. For example, at each time step an inference
algorithm could decide whether there are any new objects have appeared in the
field of view, and then create random variables for the position of these
objects as needed. A particular strategy for dynamic instantiation of
variables is to generate values for variables by simply running a program. We
refer to such strategies as evaluation-based inference methods.

Evaluation-based methods differ from their compilation-based counterparts in
that they do not require a representation of the dependency graph to be known
prior to execution. Rather, the graph is either built up dynamically at run
time, or never explicitly constructed at all. This means that many
evaluation-based strategies can be applied to models that can in principle
instantiate an unbounded number of random variables.

One of the main things we will change in evaluation-based methods is how we
deal with if-expressions. In the previous chapter we realized that
if-expressions require special consideration in probabilistic programs. The
question that we identified was whether lazy or eager evaluation should
be used in if expressions that contain sample and/or observe expressions. We
showed that lazy evaluation is necessary for observe expressions, since
these expressions affect the posterior distribution on the program output.
However, for sample expressions, we have a choice between evaluation
strategies, since we can always treat variables in unused branches as
auxiliary variables. Because lazy evaluation makes it difficult to
characterize the support, we adopted an eager evaluation strategy, in which
both branches of each if expression are evaluated, but a symbolic flow
control predicate determines when observe expressions need to be
incorporated into the likelihood.

In practice, this eager evaluation strategy for if expressions has its
limitations. The language that we introduced in chapter \ref{ch:foppl} was
carefully designed to ensure that programs always evaluate a bounded set of
sample and observe expressions. Because of this, programs that are written in
the FOPPL can be safely eagerly evaluated. It is very easy to create a
language in which this is no longer the case. For example, if we simply allow  function definitions to be recursive, then we can now write programs such as this one

\begin{lstlisting}[mathescape]
(defn sample-geometric [alpha]
  (if (= (sample (bernoulli alpha)) 1)
    1
    (+ 1 (sample-geometric p)))) 

(let [alpha (sample (uniform 0 1)) 
      k (sample-geometric alpha)]
  (observe (poisson  k) 15)
  alpha)
\end{lstlisting}

In this program, the recursive function \ang{sample-geometric} defines the
functional programming equivalent of a while loop. At each iteration, the
function samples from a Bernoulli distribution, returning 1 when the sampled
value is 1 and recursively calling itself when the value is 0. Eager
evaluation of if expressions would result in an infinite recursion for this
program, so the compilation strategy that we developed in the previous chapter
would clearly fail here. This makes sense, since the expression
\ang{(sample (bernoulli p))} can in principle be evaluated an unbounded number
of times, implying that the number of random variables in the graph is
unbounded as well.

Even though we can no longer compile the program above to a static graph, it
turns out that we can still perform inference in order to characterize the
posterior on the program output. To do so, we rely on the fact that we can
always simply run a program (using lazy evaluation for if expressions) to
generate a sample from the prior. In other words, even though we might not be
able to characterize the support of a probabilistic program, we can still
generate a sample that, by construction, is guaranteed to be part of the
support. If we additionally keep track of the probabilities associated with
each of the observe expressions that is evaluated in a program, then we can
implement sampling algorithms that generate proposals using evaluation-based mechanisms.

While many evaluation-based methods in principle apply to models with
unbounded numbers of variables, there are in practice some subtleties that
arise when reasoning about such inference methods. In this chaper, we will
therefore assume that programs are defined using the first order language form
Chapter \ref{ch:foppl}, but that a lazy evaluation strategy is used for if
expressions. Evaluation-based methods for these programs are still easier to
reason about, since we know that there is some finite set of sample and
observe expressions that can be evaluated. In the next chapter, we will
discuss implementation issues that arise when probabilistic programs can have
unbounded numbers of variables.

	\section{Likelihood Weighting}
	\label{sec:eval-likelihood}

One of the simplest evaluation-based methods for inference is likelihood weighting, which is a form of importance sampling in which the proposal is the prior. In order to see how importance sampling methods can be implemented using evaluation-based strategies, we will first discuss what operations need to be performed in importance sampling. We then briefly discuss how we could implement likelihood weighting for a program that has been compiled to a graphical model. We will then move on to discussing how we can implement importance sampling by repeatedly running the program.

\subsection{Background: Importance Sampling}
\label{subsec:importance-sample}
Like any Monte Carlo technique, importance sampling approximates the posterior distribution  $p(X|Y)$  with a set of (weighted) samples. The trick that importance sampling methods rely upon is that we can replace an expectation over $p(X|Y)$, which is generally hard to sample from, with an expectation over a proposal distribution $q(X)$, which is chosen to be easy to sample from
\begin{align*}
  \Ev_{p(X | Y)}[r(X)]
  &=
  \int dX \: 
  p(X|Y) r(X),
  \\
  &=
  \int dX \: 
  q(X) \frac{p(X|Y)}{q(X)} r(X)
  =
  \Ev_{q(X)}
  \left[
    \frac{p(X|Y)}{q(X)} 
    r(X)
  \right].
\end{align*}
The above equality holds as long as $p(X|Y)$ is absolutely continuous with respect to $q(X)$, which informally means that if according to $p(X|Y)$, the random variable $X$ has a non-zero probability
of being in some set $A$, then $q(X)$ assigns a non-zero probability to $X$ being in the same set.
If we draw samples $X^l \sim q(X)$ and define importance weights $w^l := p(X^l | Y) / q(X^l)$ then we can express our Monte Carlo estimate as an average over weighted samples $\{(w^l, X^l)\}_{l=1}^L$,
\begin{align*}
  \Ev_{q(X)}
  \left[
    \frac{p(X|Y)}{q(X)} 
    r(X)
  \right]
  &\simeq
  \frac{1}{L}
  \sum_{l=1}^L
  w^l r(X^l).
\end{align*}
Unfortunately, we cannot calculate the importance ratio $p(X|Y)/q(X)$. This requires evaluating the posterior $p(X|Y)$, which is what we did not know how to do in the first place. However, we are able to evaluate the joint $p(Y,X)$, which allows us to define an unnormalized weight,
\begin{align}
  W^l := \frac{p(Y,X^l)}{q(X^l)} = p(Y) \: w^l.
\end{align}
If we substitute $p(X|Y) = p(Y,X)/p(Y)$ then we can re-express the expectation over $q(X)$ in terms of the unnormalized weights,
\begin{align}
  \Ev_{q(X)}
  \left[
    \frac{p(X|Y)}
         {q(X)} 
    r(X)
  \right]
  &
  =
  \frac{1}{p(Y)}
  \Ev_{q(X)}
  \left[
    \frac{p(Y,X)}
         {q(X)} 
    r(X)
  \right]
  ,
  \\
  \label{eq:is-unnormalized}
  &\simeq
  \frac{1}{p(Y)}
  \frac{1}{L}
  \sum_{l=1}^L
  W^l
  r(X^l)
  ,
\end{align}
This solves one problem, since the unnormalized weights $W^l$ are quantities that we can calculate directly, unlike the normalized weights $w^l$. However, we now  have a new problem: We also don't know how to calculate the normalization constant $p(Y)$. Thankfully, we can derive an approximation to $p(Y)$ using the same  unnormalized weights $W^l$ by considering the special case $r(X)=1$,
\begin{align}
  p(Y) 
  &= 
  \Ev_{q(X)}
  \left[
    \frac{p(Y,X)}
         {q(X)}
    1
  \right]
  \simeq
  \frac{1}{L}
  \sum_{l=1}^L
  W^l
  .
\end{align}
In other words, if we define $\hat{Z} := \frac{1}{L} \sum_{l=1}^L W^l$ as the average of the unnormalized weights, then $\hat{Z}$ is an unbiased estimate of the marginal likelihood $p(Y) = \Ev[\hat{Z}]$. We can now use this estimate to approximate the normalization term in Equation~\eqref{eq:is-unnormalized},
\begin{align}
  \Ev_{q(X)}
  \left[
    \frac{p(X|Y)}
         {q(X)} 
    r(X)
  \right]
  &\simeq
  \frac{1}{p(Y)}
  \frac{1}{L}
  \sum_{l=1}^L
  W^l
  r(X^l),
  \\
  &\simeq
  \frac{1}{\hat{Z}}
  \frac{1}{L}
  \sum_{l=1}^L
  W^l
  r(X^l)
  =
  \sum_{l=1}^L
  \frac{W^l}
       {\sum_{k} W^k}
  r(X^l).
\end{align}

To summarize, as long as we can evaluate the joint $p(Y,X^l)$ for a sample $X^l \sim q(X)$, then we can perform importance sampling using unnormalized weights $W^l$.
As a bonus, we obtain an estimate $\hat{Z} \simeq p(Y)$ of the marginal likelihood as a by-product of this computation, a number which turns out to be of practical importance for many reasons, not least because it allows for Bayesian model comparison \citep{rasmussen2001occam}.

Likelihood weighting is a special case of importance sampling, in which we use the prior as the proposal distribution, i.e. $q(X) = p(X)$. The reason this strategy is known as likelihood weighting is that unnormalized weight evaluates to the likelihood when $X^l \sim p(X)$,
\begin{align}
  W^l 
  = 
  \frac{p(Y,X^l)}{q(X^l)}
  =
  \frac{p(Y|X^l)p(X^l)}{p(X^l)}
  =
  p(Y | X^l)
  .
\end{align}

We have played a little fast and loose with notation here with the aim of greater readability.  Throughout, we have focused on the fact that a FOPPL program represents a marginal projection of the posterior distribution, but in the above we temporarily pretended that a FOPPL program represented the full posterior distribution on $X$.  It is entirely correct and acceptable to reread the above with $r(X)$ being the return value projection of $X$.  The most important fact that we have skipped in this entire work up until now is that this posterior marginal will almost always be used in an outer host program to compute an expectation, say of a test function $f$ applied to the posterior distribution of the return value $r(X)$.  Note that no matter what the test function is, $\Ev_{p(X | Y)}[f(r(X))] \approx \sum_{l=1}^L w_k f(r(X^l))$ meaning that $\{(w^l, r(X^l))\}_{l=1}^L$ is the a weighted sample-based posterior marginal representation that can be used to approximate any expectation.

\subsection{Graph-based Implementation}

Suppose that we compiled our program to a graphical model as described in Section~\ref{sec:compilation-bounded}. We could then implement likelihood weighting using the following steps: 
\begin{itemize}  
  \item[1.] For each $x \in X$: sample from the prior $x^l \sim p(x \,|\, \pa(x))$. 

  \item[2.] For each $y \in Y$: calculate the weights $W_y^l = p(y \,|\, \pa(y))$.

  \item[3.] Return the weighted set of return values $r(X^l)$
  \begin{align*}
      \sum_{l=1}^L \frac{W^l}{\sum_{k=1}^L W^k} \delta_{r(X^l)} ,
      &
      W^l := \prod_{y \in Y} W_y^l
      .
  \end{align*}
\end{itemize}
where $\delta_x$ denotes an atomic mass centered on $x$.

Sampling from the prior for each $x \in X$ is more or less trivial. The only thing we need to make sure of is that we sample all parents $\pa(x)$ before sampling $x$, which is to say that we need to loop over nodes $x \in X$ according to their topological order. As described in Section~\ref{sec:eval-dens}, the terms $W_y^l$ can be calculated by simply evaluating the target language expression $\m{P}(y)[y := \m{Y}(y)]$, substituting the sampled value $x^l$ for each $x \in \pa(y)$.

\subsection{Evaluation-based Implementation}

\begin{algorithm}[!p]
  \caption{
    \label{alg:eval-base} Base cases for evaluation of a FOPPL program.}
  \begin{algorithmic}[1]
  \State \textbf{global} $\rho$
  \Comment{Procedure definitions}
  \Function{eval}{$e$, $\sigma$, $\ell$}
    \Match{$e$}
      \Case{\mang{(sample\ $d$)}}
        \State $\ldots$
        \Comment{Algorithm-specific}
      \EndCase
      \Case{\mang{(observe\ $d$\ $y$)}}
        \State $\ldots$
        \Comment{Algorithm-specific}
      \EndCase
      \Case{$c$}
        \State \Return $c$, $\sigma$
      \EndCase
      \Case{$v$}
        \State \Return $\ell(v)$, $\sigma$
      \EndCase
      \Case{\mang{(let [$v_1$\ $e_1$]\ $e_0$)}}
        \State $c_1, \sigma \gets$ \Call{eval}{$e_1$, $\sigma$, $\ell$}
        \State \Return \Call{eval}{$e_0$, $\sigma$, $\ell[v_1 \mapsto c_1]$}
      \EndCase
      \Case{\mang{(if\ $e_1$\ $e_2$\ $e_3$)}}
        \State $e_1', \sigma \gets$ \Call{eval}{$e_1$, $\sigma$, $\ell$}
        \If{$e_1'$}
          \State \Return \Call{eval}{$e_2$, $\sigma$, $\ell$}
        \Else
          \State \Return \Call{eval}{$e_3$, $\sigma$, $\ell$}
        \EndIf
      \EndCase
      \Case{\mang{($e_0$\ $e_1$\ $\ldots$\ $e_n$)}}
        \For{$i$ \textbf{in} $1, \ldots, n$}
          \State $c_i, \sigma \gets$ \Call{eval}{$e_i$, $\sigma$, $\ell$}
        \EndFor
        \Match{$e_0$}
          \Case{$f$}
            \State $(v_1, \ldots, v_n), e'_0 \gets \rho(f)$
            \State \Return \Call{eval}{$e'_0$, $\sigma$, $\ell[v_1 \mapsto c_1, \ldots, v_n\mapsto c_n]$}
          \EndCase
          \Case{$c$}
            \State \Return $c(c_1, \ldots, c_n)$, $\sigma$
          \EndCase
        \EndMatch
      \EndCase
    \EndMatch
  \EndFunction
  \end{algorithmic}
  \hrule\smallskip
  Constants $c$ are returned as is. Symbols $v$ return a constant from the
  local environment $\ell$. When evaluating the body $e_0$ of a \ang{let} form
  or a procedure $f$, free variables are bound in $\ell$. Evaluation of
  \ang{if} expressions is lazy. The \ang{sample} and \ang{observe} cases are
  algorithm-specific.
\end{algorithm}

So how can we implement this same algorithm using an evaluation-based
strategy? The basic idea in this implementation will be that we can
generate samples by simply running the program. More precisely, we will sample
a value $x \sim d$ whenever we encounter an expression \ang{(sample $d$)}. By
definition, this will generate samples from the prior. We can then calculate
the likelihood as a side-effect of running the program. To do so, we initalize
a state variable $\state$ with a single entry  $\log W=0$, which tracks the log of the unnormalized
importance weight. Each time we encounter an expression 
\ang{(observe $d$ $y$)}, we calculate the log likelihood $\log p_d(y)$ and
update the log weight to $\log W \gets \log W + \log p_d(y)$, ensuring that
$\log W = \log p(Y | X)$ at the end of the execution.

In order to define this method more formally, let us specify what we
mean by ``running'' the program. In Chapter \ref{ch:foppl}, we
defined a program $q$ in the FOPPL as
\begin{lstlisting}[mathescape]
  $q ::=$ $e$ | (defn $f$ [$x_1 \ldots x_n$] $e$) $q$
\end{lstlisting}
In this definition, a program is a single expression $e$, which
evaluates to a return value $r$, which is optionally preceded by one or more
definitions for procedures that may be used in the program. Our language
contained eight expression types
\begin{lstlisting}[mathescape]
  $e ::=$ $c$ | $v$ | (let [$v$ $e_1$] $e_2$) |  (if $e_1$ $e_2$ $e_3$) 
     | ($f$ $e_1$ $\ldots$ $e_n$) | ($c$ $e_1$ $\ldots$ $e_n$) 
     | (sample $e$) | (observe $e_1$ $e_2$) 
\end{lstlisting}
Here we used $c$ to refer to a constant or primitive operation in the
language, $v$ to refer to a program variable, and $f$ to refer to a
user-defined procedure.

In order to ``run'' a FOPPL program, we will define a function that evaluates
an expression $e$ to a value $c$. We can define this function recursively; if
we want to evaluate the expression \ang{(+ (* 2 3) (* 4 5))} then we would
first recursively evaluate the sub-expressions \ang{(* 2 3)} and 
\ang{(* 4 5)}. We then obtain values \ang{6} and \ang{20} that can be used to
perform the function call \ang{(+ 6 20)}. As long as our evaluation function
knows how to recursively evaluate each of the eight expression forms above,
then we can use this function to evaluate any program written in the FOPPL.

Algorithm~\ref{alg:eval-base} shows pseudo-code for a function
$\textsc{eval}(e,\sigma,\ell)$ that implements evaluation of each of the
non-probabilistic expression forms in the FOPPL (that is, all forms except
sample and observe). 
The arguments to this function are an expression $e$, a mapping of inference state variables $\sigma$ and a mapping of local variables $\ell$, which we refer to as the local environment. The map $\sigma$ allows us to store variables needed for inference, which are computed as a side-effect of the computation. The map $\ell$ holds the local variables that are bound in let forms and procedure calls.
 As in Section~\ref{sec:compilation-bounded}, we also assume a mapping $\rho$, which we refer to as the global environment. For each
procedure $f$ the global environment holds a pair $\rho(f)=([v_1,\ldots,v_n],e_0)$ consisting of the argument variables and the body of the procedure.

In the function $\textsc{eval}(e,\sigma,\ell)$, we use the \textbf{match} statement to pattern-match \citep{wiki_patternmatching} the expression $e$ against each of the 6 non-probabilistic expression forms. These forms are then evaluated as follows:
\begin{itemize}
  \item Constant values $c$ are returned as is. 
  \item For program variables $v$, the evaluator returns the value $\ell(v)$
  that is stored in the local environment.
  \item For let forms \ang{(let [$v_1$ $e_1$] $e_0$)}, we first evaluate $e_1$
  to obtain a value $c_1$. We then evaluate the body $e_0$ relative to the
  extended environment $\ell[v_1 \mapsto c_1]$. This ensures that every reference
  to $v_1$ in $e_0$ will evaluate to $c_1$.
  \item For if forms \ang{(if $e_1$ $e_2$ $e_3$)}, we first evaluate the
  predicate $e_1$ to a value $c_1$. If $c_1$ is logically true, then we
  evaluate the expression for the consequent branch $e_2$; otherwise we
  evaluate the alternative branch $e_3$. Since we only evaluate one of the two
  branches, this implements a {\em lazy evaluation strategy} for if expressions.
  \item For procedure calls \ang{($f$ $e_1$ $\ldots$ $e_n$)}, we first
  evaluate each of the arguments to values $c_1, \ldots, c_n$. We then
  retrieve the argument list $[v_1, \ldots, v_n]$ and the procedure body $e_0$
  from the global environment $\rho$. As with let forms, we then evaluate the
  body $e_0$ relative to an extended environment $\ell[v_1 \mapsto c_1, \ldots,
  v_n \mapsto c_n]$.
  \item For primitive calls \ang{($c_0$ $e_1$ $\ldots$ $e_n$)}, we similarly
  evaluate each of the arguments to values $c_1, \ldots, c_n$. We assume that
  the primitive $c_0$ is a function that can be called in the language that
  implements $\textsc{eval}$. The value of the expression is therefore simply
  the value of the function call $c_0(c_1, \ldots, c_n)$.
\end{itemize}

The pseudo-code in Algorithm~\ref{alg:eval-base} is remarkably succinct given
that this function can evaluate any non-probabilistic program in our first
order language. Of course, we are hiding a little bit of complexity. Each of
the cases in matches against a particular expression template.
Implementing these matching operations can require a bit of extra code. That
said, you can write your own LISP interpreter, inclusive of the parser, in
about 100 lines of Python \citep{norviglispy}.

Now that we have formalized how to evaluate non-probabilistic expressions, it
remains to define evaluation for sample and observe forms. As we described at
a high level, these evaluation rules are algorithm-dependent. For likelihood
weighting, we want to draw from the prior when evaluating sample expressions and
update the importance weight when evaluting observe expressions. In
Algorithm~\ref{alg:lw-eval} we show pseudo-code for an implementation of these
operations. We assume a variable $\log W$, that holds the log
importance weight. 

Sample and observe are now implemented as follows:
\begin{itemize}[itemsep=1pt]
    \item For sample forms \ang{(sample $e$)}, we first evaluate the
    distribution argument $e$ to obtain a distribution value $d$. We then call
    $\textsc{sample}(d)$ to generative a sample from this distribution. Here \textsc{sample} is a function in the language that implements the evaluator, which needs to be able to generate samples of each distribution type in the FOPPL (in other words, we can think of \textsc{sample} as a required method for each type of distribution object).

    
    \item For observe forms \ang{(observe $e_1$ $e_2$)} we first evaluate the
    argument $e_1$ to a distribution $d_1$ and the argument $e_2$ to a value
    $c_2$. We then update a variable $\sigma(\log W)$, which is stored in the inference state, by adding $\textsc{log-prob}(d_1, c_2)$,
    which is the log likelihood of $c_2$ under the distribution $d_1$. Finally
    we return $c_2$. The function $\textsc{log-prob}$ similarly needs to be able to compute log probability densities for each distribution type in the FOPPL.
\end{itemize}

Given a program with procedure definitions $\rho$ and body $e$, the likelihood
weighting algorithm repeatedly evaluates the program, starting from an initial state $\sigma \gets [\log W \to 0]$. It returns the value $r^l$ and the
final log weight $\sigma(\log W^l)$ for each execution.

To summarize, we have now defined an evaluated-based inference algorithm that
applies generally to probabilistic programs written in the FOPPL. This
algorithm generates a sequence of weighted samples by simply 
running the program repeatedly. Unlike the algorithms that we defined in the previous
chapter, this algorithm does not require any explicit representation of the
graph of conditional dependencies between variables. In fact, this
implementation of likelihood weighting does not even track how many sample and
observe statements a program evaluates. Instead, it draws from the prior as
needed and accumulates log probabilities when evaluating observe expressions.

\begin{algorithm}[!t]
  \caption{
    \label{alg:lw-eval}
    Evaluation-based likelihood weighting}
  \begin{algorithmic}[1]
  \State \textbf{global} $\rho, e$
  \Comment{Program procedures, body}
  \Function{eval}{$e$, $\sigma$, $\ell$}
    \Match{$e$}
      \Case{$\mang{(sample}~e\mang{)}$}
        \State $d, \sigma \gets$ \Call{eval}{$e$, $\sigma$, $\ell$}
        \State \Return \Call{sample}{$d$}, $\sigma$
      \EndCase
      \Case{$\mang{(observe}~e_1~e_2\mang{)}$}
        \State $d_1, \sigma \gets$ \Call{eval}{$e_1$, $\sigma$, $\ell$}
        \State $c_2, \sigma \gets$ \Call{eval}{$e_2$, $\sigma$, $\ell$}
        \State $\sigma(\log W) \gets \sigma(\log W) +$ \Call{log-prob}{$d_1$, $c_2$}
        \State \Return $c_2$, $\sigma$
      \EndCase
      \State $\ldots$ 
      \Comment{Base cases (as in Algorithm \ref{alg:eval-base})}
    \EndMatch
  \EndFunction

  \Function{likelihood-weighting}{$L$}
    \State $\sigma \gets [\log W \mapsto 0]$
    \For{$l$ in $1, \ldots, L$}
      \Comment{Initialize state}
      \State $r^l, \sigma^l \gets$ \Call{eval}{$e$, $\sigma$, [\,]}
      \Comment{Run program}
      \State $\log W^l \gets \sigma(\log W)$
      \Comment{Store log weight}
    \EndFor
    \State \Return $((r^1, \log W^1), \ldots, (r^L, \log W^L))$
  \EndFunction
  \end{algorithmic}

\end{algorithm}

\subsubsection{Aside 1: Relationship between Evaluation and Inference Rules}

In order to evaluate an expression $e$, we first evaluate its sub-expressions
and then compute the value of the expression from the values of the
sub-expressions. In Section~\ref{sec:compilation-bounded} we implicitly
followed the same pattern when defining inference rules for our translation.
For example, the rule for translation of a primitive call was
\[
\infer{
  \rho,\,\phi,\,\mang{($f \;e_1\ldots e_n$)} 
  \Downarrow G_1 \oplus \ldots \oplus G_n,\,{\mang{($c \;E_1\ldots E_n$)}}
}{
    \rho,\,\phi,\,e_i \Downarrow G_i,\,E_i \mbox{ for all $1 \leq i \leq n$} 
}
\]
This rule states that if we were implementing a function
$\textsc{translate}$ then $\textsc{translate}(\rho, \phi, e)$ should
perform the following steps when $e$ is of the form 
\ang{($f$ $e_1$ $\ldots$ $e_n$)}:
\begin{itemize}
  \item[1.] Recursively call $\textsc{translate}(\rho, \phi, e_i)$ to obtain a pair
  $G_i,E_i$ for each of the sub-expresions $e_1, \ldots, e_n$.
  \item[2.] Merge the graphs $G \gets G_1 \oplus \ldots \oplus G_n$ 
  \item[3.] Construct an expression $E \gets \mang{($c \;E_1\ldots E_n$)}$
  \item[4.] Return the pair $G,E$
\end{itemize}
In other words, inference rules do not only formally specify how our
translation should behave, but also give us a recipe for how to
implement a recursive \textsc{translate} operation for each
expression type.

This similarity is not an accident. In fact, inference rules are commonly used
to specify the big-step semantics of a programming language, which defines the
value of each expression in terms of the values of its sub-expressions. We can
similarly use inference rules to define our evaluation-based likelihood 
weighting method. We show these inference rules in Figure~\ref{fig:lw-bigstep}.

\begin{figure}[!t]
\begin{align*}
\infer{
\rho,\ell,c
\Downarrow
c,0
}
{}
\quad
\infer{
\rho,\ell,v
\Downarrow
c
}
{\ell(v) = c}
\quad
\infer{
  \rho,\ell,\mang{(let [$v_1$\ $e_1$]\ $e_0$)}
  \Downarrow
  c_0, l_0 + l_1
}{
  \begin{array}{c}
    \rho,\ell,e_1
    \Downarrow
    c_1,l_1 
    \quad
    \rho,\ell \oplus [v_1 \mapsto c_1],e_0
    \Downarrow
    c_0, l_0
  \end{array}
}
\end{align*}
\vspace{-20pt}
\begin{align*}
\infer{
\rho,\ell,\mang{(if\ $e_1$\ $e_2$\ $e_3$)}
\Downarrow
c_2,l_1+l_2
}
{
\begin{array}{c}
\rho,\ell,e_1
\Downarrow
\mang{true},l_1
\\
\rho,\ell,e_2
\Downarrow
c_2,l_2
\end{array}
}
\quad
\infer{
\rho,\ell,\mang{(if\ $e_1$\ $e_2$\ $e_3$)}
\Downarrow
c_3,l_1+l_3
}
{
\begin{array}{c}
\rho,\ell,e_1
\Downarrow
\mang{false},l_1
\\
\rho,\ell,e_3
\Downarrow
c_3,l_3
\end{array}
}
\end{align*}
\vspace{-20pt}
\begin{align*}
\infer{
  \rho,\ell,\mang{($f$\ $e_1$\ $\ldots$\ $e_n$)}
  \Downarrow
  c_0, l_0 + l_1 + \ldots + l_n
}{
  \begin{array}{c}
    \rho(f) = [v_1, \ldots, v_n], e_0
    \quad
    \rho,\ell,e_i
    \Downarrow
    c_i,l_i 
    ~\text{for}~i=1,\ldots,n 
    \\
    \rho,\ell \oplus [v_1 \mapsto c_1, \ldots, v_n \mapsto c_n],e_0
    \Downarrow
    c_0,l_0
  \end{array}
}
\end{align*}
\vspace{-20pt}
\begin{align*}
\infer{
  \rho,\ell,\mang{($c$\ $e_1$\ $\ldots$\ $e_n$)}
  \Downarrow
  c_0, l_1 + \ldots + l_n
}{
  \begin{array}{c}
    \rho,\ell,e_i
    \Downarrow
    c_i,l_i 
    ~\text{for}~i=1,\ldots,n 
    \quad
    c(c_1, \ldots,c_n) = c_0
  \end{array}
}
\end{align*}
\vspace{-20pt}
\begin{align*}
\infer{
  \rho,\ell,\mang{(sample\ $e$)}
  \Downarrow
  c, l
}{
  \begin{array}{c}
    \rho,\ell,e
    \Downarrow
    d,l 
    \quad
    c \sim d
  \end{array}
}
\quad
\infer{
  \rho,\ell,\mang{(observe\ $e_1$\ $e_2$)}
  \Downarrow
  c_2, l_0 + l_1 + l_2 
}{
  \begin{array}{c}
    \rho,\ell,e_1
    \Downarrow
    d_1,l_1 
    \quad
    \rho,\ell,e_2
    \Downarrow
    c_2,l_2
    \quad
    \log p_{d_1}(c_2) = l_0
  \end{array}
}
\end{align*}

\caption{
\label{fig:lw-bigstep}
Big-step semantics for likelihood weighting. These rules define an evaluation operation $\rho,\ell,e \Downarrow c,l$, in which $\rho$ and $\ell$  refers to the global and local environment, refers to the local environment, $e$ is an expression, $c$ is the value of the expression and $l$ is its log likelihood.}
\end{figure}

\subsubsection*{Aside 2: Side Effects and Referential Transparency}

The implementation in Algorithm \ref{alg:lw-eval} highlights a fundamental
distinction between sample and observe forms relative to the non-probabilistic
expression types in the FOPPL. If we do not include sample and observe in our 
syntax, then our first order language is not only deterministic, but it is also
pure in a functional sense. In a purely functional language, there are no side
effects. This means that every expression $e$ will always evaluate to the same
value. An implication of this is that any expression in a program can be
replaced with its corresponding value without affecting the behavior of the
rest of the program. We refer to expressions with this property as
referentially transparent, and expressions that lack this property as
referentially opaque.

Once we incorporate sample and observe into our language, our language is no
longer functionally pure, in the sense that not all expressions are
referentially transparent. In our implementation in Algorithm
\ref{alg:lw-eval}, a sample expression does not always evaluate to the same
value and is therefore referentially opaque. By extension, any
expression containing a sample form as a sub-expression is also
opaque. An observe expression \ang{(observe $e_1$ $e_2$)} always evaluates to
the same value as long as $e_2$ is referentially transparent. However observe
expressions have a side effect, which is that they increment the log weight stored in the inference state $\sigma(\log W)$. If we replaced an 
observe form \ang{(observe $e_1$ $e_2$)} with the expression for its observed value 
$e_2$, then the program would still produce the same distribution on return 
values when sampling from the prior, but the log weight $\sigma(\log W)$ would be 0 after every execution.


The distinction between referentially transparent and opaque expressions also
implicitly showed up in our compilation procedure in Section
\ref{sec:compilation-bounded}. Here we translated an opaque program into a set
of target-language expressions for conditional probabilities, which were
referentially transparent. In these target-language expressions, each
sub-expression corresponding to sample or observe was replaced with a free
variable $v$. If a translated expression has no free variables, then
the original untranslated expression is referentially transparent. In
Section~\ref{sec:partial-eval}, we implicitly exploited this property to
replace all target-language expressions without free variables with their
values. We also relied on this property in Section~\ref{sec:compilation-bounded} 
to ensure that observe forms \ang{(observe $e_1$ $e_2$)} always contained a 
referentially transparent expression for the observed value $e_2$.

	\section{Metropolis-Hastings}
	\label{sec:eval-mh}



In the previous section, we used evaluation to generate samples from the program prior while calculating the likelihood associated with these samples as a side-effect of the computation. We can use this same strategy to define Markov-chain Monte Carlo (MCMC) algorithms. We already discussed such an algorithm, namely, Gibbs Sampling in Section \ref{sec:gibbs}. The algorithm implicitly relied on the fact that we were able to represent a probabilistic program as a static graphical model. It explicitly made use of the conditional dependency graph in order to identify the minimal set of variables needed to compute the acceptance ratio. 

Metropolis-Hastings (MH) methods, which we also mentioned in Section~\ref{sec:gibbs} generate a Markov chain of program return values $r(X)^1, \ldots, r(X)^S$  by accepting or rejecting a newly proposed sample according to the following pseudo-algorithm.
\begin{itemize}[leftmargin=1.5em,topsep=4pt,itemsep=2pt,parsep=2pt,partopsep=2pt]
  \item[-] Initialize the current sample $X$. Return $X^1 \gets X$.
  \item[-] For each subsequent sample $s=2,\ldots,S$
  \begin{itemize}[leftmargin=1.5em,topsep=4pt,itemsep=2pt,parsep=2pt,partopsep=2pt]
    \item[-] Generate a proposal $X' \sim q(X' \,|\, X)$ 
    \item[-] Calculate the acceptance ratio
    \begin{align}
      \label{eq:mh-accept}
      \alpha
      =
      \frac{p(Y',X')q(X \,|\, X')}
           {p(Y,X) q(X' \,|\, X)}
    \end{align}
    \item[-] Update the current sample $X \gets X'$ with probability $\text{max}(1, \alpha)$,  otherwise keep $X \gets X$. Return $X^{\text{s}} \gets X$.
  \end{itemize}
\end{itemize}
An evaluation-based implementation of a MH sampler needs to do two things. It needs to be able to run the program to generate a proposal, conditioned on the values $\m{X}$ of sample expressions that were evaluated previously. The second is that we need to be able to compute the acceptance ratio $\alpha$ as a side effect. 


\begin{algorithm}[!t]
  \caption{
    \label{alg:eval-imh}
    Evaluation-based Metropolis-Hastings with independent proposals from the prior.}
  \begin{algorithmic}[1]
  \State \textbf{global} $\rho, e$
  \Function{eval}{$e$, $\sigma$, $\ell$}
      \State $\ldots$ 
      \Comment{As in Algorithm \ref{alg:lw-eval}}
  \EndFunction
  \Function{independent-mh}{$S$}
    \State $\sigma \gets [\log W \mapsto 0]$
    \State $r \gets$ \Call{eval}{$e$, $\sigma$, $[\,]$}
    \State $\log W \gets \log W$
    \For{$s$ \textbf{in} $1, \ldots, S$}
        \State $r', \sigma' \gets$ \Call{eval}{$e$, $\sigma$, $[\,]$}
        \State $\log W' \gets \sigma'(\log W)$
        \State $\alpha \gets W' / W$
        \State $u \sim \Call{uniform-continuous}{0,1}$
        \If{$u < \alpha$}
          \State $r, \log W \gets r', \log W'$
        \EndIf
        \State $r^s \gets r$
      \EndFor
     \State \Return $(r^1, \ldots, r^S)$
  \EndFunction
  \end{algorithmic}
\end{algorithm}


Let us begin by considering a simplified version of this algorithm. Suppose that we defined $q(X' | X) = p(X')$. In other words, at each step we generate a sample $X' \sim p(X)$ from the program prior, which is independent of the previous sample $X$. We already know that we can generate these samples simply by running the program. The acceptance ratio now simplifies to: 
\begin{align}
  \alpha = \frac{p(Y',X') q(X\,|\,X')}
                {p(Y,X) q(X'\,|\,X)}
         = \frac{p(Y'\,|\,X') p(X')  p(X)}
                {p(Y\,|\,X) p(X)  p(X')}
         = \frac{p(Y'\,|\,X')}
                {p(Y\,|\,X)}
\end{align}
In other words, when we propose from the prior, the acceptance ratio is simply the ratio of the likelihoods. Since our likelihood weighting algorithm computes $\sigma(\log W) = \log p(Y \mid X)$ as a side effect, we can re-use the evaluator from Algorithm~\ref{alg:lw-eval} and simply evaluate the acceptance ratio as $W' / W$, where $W' = p(Y' | X')$ is the likelihood of the proposal and $W = p(Y | X)$ is the likelihood associated with the previous sample. Pseudo-code for this implementation is shown in Algorithm~\ref{alg:eval-imh}.

\subsection{Single-Site Proposals}
\label{sec:lmh-single-site}

Algorithm~\ref{alg:eval-imh} is so simple because we have side-stepped the difficult operations in the more general MH algorithm: In order to generate a proposal, we have to run our program in a manner that generates a sample $X' \sim q(X' | X)$ which is conditioned on the values associated with our previous sample. In order to evaluate the acceptance ratio, we have to calculate the probability of the reverse proposal $q(X | X')$. Both these operations are complicated by the fact that $X$ and $X'$ potentially  refer to different subsets of sample expressions in the program. To see what we mean by this, Let us take another look at Example~\ref{pr:opm-if-sample}, which we introduced in Section~\ref{sec:compilation-bounded}
\begin{lstlisting}[mathescape]
(let [z (sample (bernoulli 0.5))
      mu (if (= z 0) 
           (sample (normal -1.0 1.0)) 
           (sample (normal 1.0 1.0)))
      d (normal mu 1.0)
      y 0.5]
  (observe d y)
  z)
\end{lstlisting}
In Section~\ref{sec:compilation-bounded}, we would compile this model to a Bayesian network with three latent variables $X = \{\mu_0,\mu_1,z\}$ and one observed variable $Y = \{y\}$. In this section, we evaluate if-expressions lazily, which means that we will either sample $\mu_1$ (when $z=1$) or $\mu_0$ (when $z=0$), but not both. This introduces a complication: What happens when we update $z=0$ to $z=1$ in the proposal? This now implies that $X$ contains a variable $\mu_0$, which is not defined for $X'$. Conversely, $X'$ needs to instantiate a value for the variable $\mu_1$ which was not defined in $X$. 

In order to define an evaluation-based algorithm for constructing a proposal, 
we will construct a map $\sigma(\m{X})$, such that $\m{X}(x)$ refers to the value of a variable $x$. In order to calculate the acceptance ratio, we will similarly construct a map $\sigma(\log \m{P})$. Section~\ref{sec:compilation-bounded} contained a target-language expression $\log \m{P}(v)$ that evaluates to the density for each variable $v \in X \cup Y$. In our evaluation-based algorithm, we will store the log density 
\begin{align} 
  \sigma(\log \m{P}(x)) 
  = 
  \textsc{log-prob}(d, \m{X}(x)).
\end{align} 
for each sample expression \ang{(sample $d$)}, as well as the log density 
\begin{align}
  \sigma(\log \m{P}(y)) = \textsc{log-prob}(d, c)
\end{align} 
for each observe expression \ang{(observe $d$ $c$)}.

With this notation in place, let us define the most commonly used evaluation-based proposal for probabilistic programming systems: the single-site Metropolis-Hastings update. In this algorithm we change the value for one variable $x_0$, keeping the values of other variables fixed whenever possible. To do so, we sample $x_0$ from the program prior, as well as any variables $x \not\in \dom(\m{X})$. For all other variables, we reuse the values $\m{X}(x)$. This strategy can be summarized in the following pseudo-algorithm:

\begin{itemize}
  \item[-] Pick a variable $x_0 \in \dom(\m{X})$ at random from the current sample.
  \item[-] Construct a proposal $\m{X}',\m{P}'$ by re-running the program:
  \begin{itemize}
    \item[-] For expressions \ang{(sample $d$)} with variable $x$: 
    \begin{itemize}
    \item[-] If $x=x_0$, or $x \not\in \dom(\m{X})$, then sample $\m{X}'(x) \sim d$. \\Otherwise, reuse the value $\m{X}'(x) \gets \m{X}(x)$.
    \item[-] Calculate the probability $\m{P}'(x) \gets \textsc{prob}(d, \m{X}'(x))$.
    \end{itemize}
    \item[-] For expressions \ang{(observe $d$ $c$)} with variable $y$: 
    \begin{itemize}
    \item[-] Calculate the probability $\m{P}'(y) \gets \textsc{prob}(d, c)$
    \end{itemize}
  \end{itemize}
\end{itemize}

\begin{algorithm}[!t]
\caption{
  \label{alg:accept-lmh}
  Acceptance ratio for single-site proposals
}
\begin{algorithmic}[1]
  \Function{accept}{$x_0, \m{X}', \m{X}, \log \m{P}', \log \m{P}$}
  \State $X'^{\text{sampled}} \gets \{x_0\} \cup (\dom(\m{X}') \setminus \dom(\m{X}))$
  \State $X^{\text{sampled}} \gets \{x_0\} \cup (\dom(\m{X}) \setminus \dom(\m{X}'))$
  \State $\log \alpha \gets \log |\dom(\m{X})| - \log |\dom(\m{X}')|$
  \For{$v ~\textbf{in}~ \dom(\log \m{P}') \setminus X'^{\text{sampled}}$}
    \State $\log \alpha  \gets \log \alpha + \log \m{P}'(v)$
  \EndFor
  \For{$v ~\textbf{in}~ \dom(\log \m{P}) \setminus X^{\text{sampled}}$}
      \State $\log \alpha  \gets \log \alpha - \log \m{P}(v)$
  \EndFor
  \State \Return $\alpha$
  \EndFunction
\end{algorithmic}
\end{algorithm}

What is convenient about this proposal strategy is that it becomes comparatively easy to evaluate the acceptance ratio $\alpha$. In order to evaluate this ratio, we will rearrange the terms in Equation~\eqref{eq:mh-accept} into a ratio of probabilities for $X'$ and a ratio of probabilities for $X$:
\begin{align}
\alpha
&=
\frac{p(Y',X') q(X|X')}
     {p(Y,X) q(X'|X)}
\\
&=
\frac{p(Y',X')}
     {q(X'|X, x_0)}
\frac{q(X|X', x_0)}
     {p(Y,X)}
\frac{q(x_0|X')}
     {q(x_0|X)}
.
\end{align}
Here the ratio $q(x_0|X')/q(x_0|X)$ accounts for the relative probability of selecting the initial site. Since $x_0$ is chosen at random, this is 
\begin{align}
\frac{q(x_0|X')}
     {q(x_0|X)}
=
\frac{|X|}
     {|X'|}
.
\end{align}
We can now express the ratio $p(Y',X')/q(X'|X, x_0)$ in terms of the probabilities $\m{P}'$. The joint probability is simply the product
\begin{align}
  p(Y',X') = p(Y'|X') p(X') = \prod_{y \in Y'} \m{P}'(y) \prod_{x \in X'} \m{P}'(x),
\end{align}
where $X'=\dom(\m{X}')$ and $Y'=\dom(\m{P}') \setminus X'$.

To calculate the probability $q(X' | X, x_0)$ we decompose the set of variables $X' = X'^{\text{sampled}} \cup X'^{\text{reused}}$ into the set of sampled variables $X'^{\text{sampled}}$ and the set of reused variables $X'^{\text{reused}}$. Based on the rules above, the set of sampled variables is given by
\begin{align} 
  X'^{\text{sampled}} = \{x_0\} \cup (\dom(\m{X}') \setminus \dom(\m{X})).
\end{align}
Since all variables in $X'^{\text{sampled}}$ were sampled from the program prior, the proposal probability is
\begin{align}
  q(X'|X,x_0) = \prod_{x \in X'^{\text{sampled}}} \m{P}'(x).
\end{align}
Since some of the terms in the prior and the proposal cancel, the ratio $p(Y',X')/q(X'|X, x_0)$ simplifies to
\begin{align}
  \frac{p(Y',X')}
       {q(X'|X,x_0)} 
  & = \prod_{y \in Y'}
      \m{P}'(y)
      \prod_{x \in X'^{\text{reused}}}
      \m{P}'(x)
\end{align}
We can define the ratio $p(Y,X)/q(X | X', x_0)$ for the reverse transition  by noting that this transition would require sampling a set of variables $X^{\text{sampled}}$ from the prior whilst reusing a set of variables $X^{\text{reused}}$
\begin{align}
  \frac{p(Y,X)}
       {q(X|X,x_0)} 
  & = \prod_{y \in \mathcal{Y}}
      \m{P}(y)
      \prod_{x \in X^{\text{reused}}}
      \m{P}(x)
  .
\end{align}
Here the set of reused variable $X^{\text{reused}}$ for the reverse transition is, by definition, identical that of the forward transition $X'^{\text{reused}}$,
\begin{align}
  X'^{\text{reused}}
  =
  (\dom(\m{X}') \cap \dom(\m{X})) \setminus \{x_0\}
  =
  X^{\text{reused}}.
\end{align}
Putting all the terms together, the acceptance ratio becomes:
\begin{align}
\label{eq:lmh-accept}
\alpha = 
\frac{|\dom(\m{X})|}
     {|\dom(\m{X}')|}
\frac{\prod_{y \in \mathcal{Y}} \m{P}'(y) \prod_{x \in X'^{\text{reused}}} \m{P}'(x)}
     {\prod_{y \in \mathcal{Y}} \m{P}(y) \prod_{x \in X^{\text{reused}}} \m{P}(x)}
.
\end{align}

If we look at the terms above, then we see that the acceptance ratio for single-site proposals is a generalization of the acceptance ratio that we obtained for independent proposals. When using independent proposals, we could express the acceptance ratio $\alpha = W' / W$ in terms of the likelihood weights $W' = p(Y', X') / q(X') = p(Y' \,|\, X')$. In the single-site proposal, we treat retained variables $X'^{\text{reused}} = X^{\text{reused}}$ as if they were observed variables. In other words, we could define 
\begin{align}
  W' 
  = 
  \frac{p(Y', X')} 
       {q(X' | X, x_0)}.
\end{align}


\subsubsection{Addressing Transformation}
\label{sec:addressing-transform}

\begin{figure}[!t]
\begin{align*}
\infer{
\rho,c
\Downarrow_\alpha
c
}
{}
\qquad
\infer{
\rho,v
\Downarrow_\alpha
v
}
{}
\qquad
\infer{
  \rho,\mang{(let [$v_1$\ $e_1$]\ $e_0$)}
  \Downarrow_\alpha
  \mang{(let [$v_1$\ $e^\prime_1$]\ $e^\prime_0$)}
}{
  \begin{array}{c}
    \rho,e_1
    \Downarrow_\alpha
    e'_1
    \quad
    \rho,e_0
    \Downarrow_\alpha
    e'_0
  \end{array}
}
\end{align*}
\vspace{-20pt}
\begin{align*}
\infer{
  \rho,\mang{($op$\ $e_1$\ $\ldots$\ $e_n$)}
  \Downarrow_\alpha
  \mang{($op$\ $e^\prime_1$\ $\ldots$\ $e^\prime_n$)}
}{
  \begin{array}{c}
    \rho,e_i
    \Downarrow_\alpha
    e'_i
    ~\text{for}~i=1,\ldots,n 
    \quad
    op = \mang{if} 
    ~~\text{or}~~ 
    op = c
  \end{array}
}
\end{align*}

\vspace{-20pt}

\begin{align*}
\infer{
  \rho,\mang{($f$\ $e_1$\ $\ldots$\ $e_n$)}
  \Downarrow_\alpha
  e_1''
}{
  \begin{array}{c}
    \rho,e_i
    \Downarrow_\alpha
    e'_i
    ~~\text{for}~i=0,\ldots,n 
    \qquad
    \rho(f) = \mang{(defn [$v_1$ $\ldots$ $v_n$]\ $e_0$)}
    \\
    \rho,\mang{(let [$v_n\;e^\prime_n$]\ $e^\prime_0$)}  
    \Downarrow_\alpha
    e''_n
    ~~
    \rho,\mang{(let [$v_{i-1}\;e^\prime_{i-1}$]\ $e^{\prime\prime}_{i}$)}  
    \Downarrow_\alpha
    e''_{i-1}
    ~\text{for}~i=n,\ldots,2
    \end{array}
}
\end{align*}

\vspace{-20pt}

\begin{align*}
\infer{
  \rho,\mang{(sample\ $e$)}
  \Downarrow_\alpha
  \mang{(sample\ $v$\ $e^\prime$)}
}{
  \begin{array}{c}
    \rho,e
    \Downarrow_\alpha
    e'
    \quad
    \text{fresh}~v
  \end{array}
}
~~
\infer{
  \rho,\mang{(observe\ $e_1$\ $e_2$)}
  \Downarrow_\alpha
  \mang{(observe\ $v$\ $e^\prime_1$\ $e^\prime_2$)}
}{
  \begin{array}{c}
    \rho,e_1
    \Downarrow_\alpha
    e'_1
    \quad
    \rho,e_2
    \Downarrow_\alpha
    e'_2
    \quad
    \text{fresh}~v
  \end{array}
}
\end{align*}
\caption{\label{fig:addressing} Addressing transformation for FOPPL programs.}
\end{figure}

In defining the acceptance ratio in Equation~\eqref{eq:lmh-accept}, we have tacitly assumed that we can associate a variable $x$ or $y$ with each sample or observe expression. This is in itself not such a strange assumption, since we did just that in Section~\ref{sec:compilation-bounded}, where we assigned a unique variable $v$ to every sample and observe expression as part of our compilation of a graphical model. In the context of evaluation-based methods, this type of unique identifier for a sample or observe expression is commonly referred to as an address. 

If needed, unique addresses can be constructed dynamically at run time. We will get back to this in Chapter~\ref{ch:eval-two}, Section~\ref{sec:addressing}. For programs in the FOPPL, we can create addresses using a source code transformation that is similar to the one we defined in Section~\ref{sec:compilation-bounded}, albeit a much simpler one. In this transformation we replace all expressions of the form \ang{(sample $e$)} with expressions of the form \ang{(sample $v$ $e$)} in which $v$ is a newly created variable. Similarly, we replace \ang{(observe $e_1$ $e_2$)} with \ang{(observe $v$ $e_1$ $e_2$)}. Figure ~\ref{fig:addressing} defines this translation $\rho,e \Downarrow_\alpha e'$. As in Section~\ref{sec:compilation-bounded}, this translation accepts a map of function definitions $\rho, e$ and returns a transformed expression $e'$ in which addresses have been inserted into all sample and observe expressions. 


\subsubsection{Evaluating Proposals}


\begin{algorithm}[!t]
\caption{
  \label{alg:lmh-eval}
  Evaluator for single-site proposals
}
\begin{algorithmic}[1]
  \State \textbf{global} $\rho$
  \Function{eval}{$e$, $\sigma$, $\ell$}
    \Match{$e$}
      \Case{\mang{(sample\ $v$\ $e$)}}
        \State $d, \sigma \gets$ \Call{eval}{$e$, $\sigma$, $\ell$}
        \If{$v \in \dom(\sigma(\m{C})) \setminus \{\sigma(x_0)\}$}
          \State $c, \gets \sigma(\m{C}(v))$
          \Comment{Retain previous value}
        \Else
          \State $c \gets$ \Call{sample}{$d$}
          \Comment{Sample new value}
        \EndIf
        \State $\sigma(\m{X}(v)) \gets c$
        \Comment{Store value}
        \State $\sigma(\log \m{P}(v)) \gets \Call{log-prob}{d, c}$
        \Comment{Store log density}
        \State \Return $c$, $\sigma$
      \EndCase
      \Case{\mang{(observe\ $v$\ $e_1$\ $e_2$)}}
        \State $d, \sigma \gets$ \Call{eval}{$e_1$, $\sigma$, $\ell$}
        \State $c, \sigma \gets$ \Call{eval}{$e_2$, $\sigma$, $\ell$}
        \State $\sigma(\log \m{P}(v)) \gets \Call{log-prob}{d, c}$
        \Comment{Store log density}
        \State \Return $c$, $\sigma$
      \EndCase
      \State $\ldots$ 
      \Comment{Base cases (as in Algorithm \ref{alg:eval-base})}
    \EndMatch
  \EndFunction
\end{algorithmic}
\end{algorithm}

\begin{algorithm}[!t]
  \caption{
    \label{alg:lmh}
    Single-site Metropolis Hastings}
  \begin{algorithmic}[1]
  \State \textbf{global} $\rho, e$
  \Function{eval}{$e$, $\sigma$, $\ell$}
  \State $\ldots$
  \Comment{As in Algorithm~\ref{alg:lmh-eval}}
  \EndFunction
  \Function{accept}{$x_0,\m{X}',\m{X},\log \m{P}', \log \m{P}$}
  \State $\ldots$
  \Comment{As in Algorithm~\ref{alg:accept-lmh}}
  \EndFunction
  \Function{single-site-mh}{$S$}
    \State $\sigma_0 \gets [x_0 \gets \mang{nil}, \m{C} \mapsto, [\,], \m{X} \mapsto [\,], \log \m{P} \mapsto [\,]]$
    \State $r, \sigma \gets$ \Call{eval}{$e$, $\sigma_0$, $[\,]$}
    \For{$s$ \textbf{in} $1, \ldots, S$}
        \State $v \sim \Call{uniform}{\dom(\sigma(\m{X}))}$
        \State $\sigma' \gets \sigma_0[x_0 \mapsto v, \m{C} \mapsto \sigma(\m{X})]$
        \State $r', \sigma' \gets$ \Call{eval}{$e$, $\sigma'$, $[\,]$}
        \State $u \sim \Call{uniform-continuous}{0,1}$
        \State $\alpha \gets \Call{accept}{x_0, \sigma'(\m{X}), \sigma(\m{X}), \sigma'(\log \m{P}), \sigma(\log \m{P})}$
        \If{$u < \alpha$}
          \State $r,\sigma \gets r', \sigma'$
        \EndIf
        \State $r^s \gets r$
      \EndFor
     \State \Return $(r^1, \ldots, r^S)$
  \EndFunction
  \end{algorithmic}
\end{algorithm}

Now that we have incorporated addresses that uniquely identify each sample and observe expression, we are in a position to formally define the pseudo-algorithm for single-site Metropolis Hastings that we oulined in Section~\ref{sec:lmh-single-site}.

In Algorithm~\ref{alg:lmh-eval}, we define the evaluation rules for sample and observe expressions. We assume that the inference state $\sigma$ holds a value $\sigma(x_0)$, which is the site of the proposal, a map $\sigma(\m{X})$ map $\sigma(\log \m{P})$, which holds the log density for each variable, and finally a ``cache'' $\sigma(\m{C})$ of values that we would like to condition the execution on.

For a sample expression with address $v$, we reuse the value $\m{X}(v) \gets \m{C}(v)$ when possible, unless we are evaluating the proposal site $v=x_0$. In all other cases, we sample $\m{X}(v)$ from the prior. For both sample and observe expressions we calculate the log probability $\log \m{P}(v)$. 

The Metropolis Hastings implementation is shown in Algorithm~\ref{alg:lmh}. This algorithm initializes the state $\sigma$ sample by evaluating the program, storing the values $\sigma(\m{X})$ and log probabilities $\sigma(\log \m{P})$ for the current sample. For each subsequent sample the algorithm then selects the initial site $x_0$ at random from the domain of the current sample $\sigma(\m{X})$. We then rerun the program accordingly to construct a proposal and either accept or reject according to the ratio defined in Algorithm~\ref{alg:accept-lmh}.

	\section{Sequential Monte Carlo}

\label{sec:eval-smc}

One of the limitations of the likelihood weighting algorithm that we
introduced in Section~\ref{sec:eval-likelihood} is that it is essentially a
``guess and check'' algorithm; we \emph{guess} by sampling a proposal $X^l$ from
the program prior and then \emph{check} whether this is in fact a good
proposal by calculating a weight $W^l = p(Y | X^l)$ according to the
probabilities of observe expressions in the program. The great thing about
this algorithm is that it is both simple and general. Unfortunately it is not
necessarily \emph{efficient}. In order to get a high weight sample, we have to
generate reasonable values for all random variables $X$. This means that
likelihood weighting will work well in programs with a small number of sample
expressions, where we can expect to ``get lucky'' for all sample expressions
with reasonable frequency. However, the frequency with which we generate good proposals decreases exponentially with the number of sample expressions in the program.

Sequential Monte Carlo (SMC) methods solve this problem by turning a sampling
problem for a high dimensional distribution into a sequence of sampling
problems for lower dimensional distributions. In their most general form, SMC
methods consider a sequence of unnormalized densities $\gamma_1(X_1), \ldots,
\gamma_N(X_N)$, where each $\gamma_n(X_n)$ has the form that we discussed in
Section~\ref{sec:factor-condition}. Here $\gamma_1(X_1)$ is typically a 
low dimensional distribution, for which it is easy to perform importance 
sampling, whereas $\gamma_N(X_N)$ is a high dimensional distribution, for which
want to generate samples. For each $\gamma_n(X_n)$ in between increases in
dimensionality to interpolate between these two distributions. For a FOPPL program, 
we can define $\gamma_N(X_N) = \gamma(X) = p(Y,X)$ as the joint density
associated with the program. 

Given a set of unnormalized densities $\gamma_n(X_n)$, SMC sequentially
generates weighted samples $\{(X_n^l, W_n^l)\}_{l=1}^L$ by performing 
importance sampling for each of the normalized densities $\pi_n(X_n)
= \gamma_n(X_n)/Z_n$ according to the following rules

\begin{itemize}[leftmargin=1.5em,topsep=4pt,itemsep=2pt,parsep=2pt,partopsep=2pt]
\item[-] Initialize a weighted set $\{(X^l_1, W^l_1)\}_{l=1}^L$ using importance sampling
\begin{align}
    X_1^l 
    &\sim 
    q_1(X_1),
    &
    W_1^l 
    &:=
    \frac{\gamma_1(X_1^l)}
         {q_1(X_1^l)}
    .
\end{align}
\item[-] For each subsequent generation $n=2,\ldots,N$:

\begin{itemize}[leftmargin=1.5em,topsep=4pt,itemsep=2pt,parsep=2pt,partopsep=2pt]
\item[1.] Select a value $X_{n-1}^k$ from the preceding set by sampling an
ancestor index $a_{n-1}^l=k$ with probability proportional to $W^k_{n-1}$
\begin{align}
    \label{eq:smc-resampling}
    a_{n-1}^l 
    &\sim 
    \text{Discrete}
    \left(
        \frac{W^1_{n-1}}
             {\sum_{l} W^l_{n-1}}, 
        \ldots, 
        \frac{W^L_{n-1}}
             {\sum_{l} W^l_{n-1}}, 
    \right),
\end{align}
\item[2.] Generate a proposal conditioned on the selected particle
\begin{align}
    \label{eq:smc-propose}
    X_n^l
    &\sim
    q_n(X_n \,|\, X^{a^l_{n-1}}_{n-1})
    ,
\end{align}
and define the importance weights 
\begin{align}
    \label{eq:smc-weight}
    W_n^l
    &:=
    W^l_{n \setminus n-1}
    \hat{Z}_{n-1}
\end{align}
where $W^l_{n \setminus n-1}$ is the incremental weight
\begin{align}
    \label{eq:smc-weight-incremental}
    W^l_{n \setminus n-1}
    &:=
    \frac{\gamma_n(X^l_n)}
         {\gamma_{n-1}(X^{a^l_{n-1}}_{n-1}) q_n(X^l_n \,|\, X^{a^l_{n-1}}_{n-1}) }
    ,
\end{align}
and $\hat{Z}_{n-1}$ is defined as the average weight
\begin{align}
    \hat{Z}_{n-1}
    =
    \frac{1}{L} 
    \sum_{l=1}^L
    W_{n-1}^l
    .
\end{align}

\end{itemize}
\end{itemize}

The defining operation in this algorithm is in Equation~\eqref{eq:smc-resampling}, which is known as the resampling step. We can think of this operation as performing ``natural selection'' on the sample set; samples $X^k_{n-1}$ with a high weight $W^k_{n-1}$ will be used more often to construct proposals equation in \eqref{eq:smc-propose}, whereas samples with a low weight will with high probability not be used at all. In other words, SMC uses the weight of a sample at generation $n-1$ as a heuristic for the weight that it will have at generation $n$, which is a good strategy whenever weights in subsequent densities are strongly correlated.


\subsection{Defining Intermediate Densities with Breakpoints}

As we discussed in Section~\ref{sec:factor-condition}, a FOPPL program defines
an unnormalized distribution $\gamma(X) = p(Y,X)$. When inference is performed
with SMC we define the final density as $\gamma_N(X_N) = \gamma(X)$. In
order to define intermediate densities $\gamma_n(X_n) = p(Y_n, X_n)$ we
consider a sequence of \emph{truncated} programs that evaluate successively
larger subsets of the sample and observe expressions
\begin{align}
    &
    X_1 \subseteq X_2 \subseteq \ldots \subseteq X_N
    = 
    X,
    \\
    &
    Y_1 \subseteq Y_2 \subseteq \ldots \subseteq Y_N
    = 
    Y
    .
\end{align}
The definition of a \emph{truncated} program that we employ here is programs
that halt at a breakpoint. Breakpoints can be specified explicitly by the
user, constructed using program analysis, or even dynamically defined at run
time. The sequence of breakpoints needs to satisfy the following two
properties in order.
\begin{itemize}
    \item[1.] The breakpoint for generation $n$ must always occur after the
    breakpoint for generation $n-1$.
    \item[2.] Each breakpoint needs to occur at an expression that is
    evaluated in every execution of a program. In particular, this means that
    breakpoints should not be associated with expressions inside branches of
    if expressions.
\end{itemize}
In this section we will assume that we first apply the addressing
transformation from Section~\ref{sec:addressing-transform} to a FOPPL program.
We then assume that the user identifies a sequence of symbols
$y_1,\ldots,y_{N-1}$ for observe expressions that satisfy the two properties
above. An alternative design, which is often used in practice, is to simply
break at every observe and assert that each sample has halted at the same
point at run time.

\subsection{Calculating the Importance Weight}

Now that we have defined a notion of intermediate densities $\gamma_n(X_n)$
for FOPPL programs, we need to specify a mechanism for generating proposals
from a distribution $q_n(X_n | X_{n-1})$. The SMC analogue of likelihood 
weighting is to simply sample from the program prior $p(X_n | X_{n-1})$, 
which is sometimes known as a bootstrapped proposal. For this proposal, we can
express $\gamma_n(X_n)$ in terms of $\gamma_{n-1}(X_{n-1})$ as
\begin{align*}
    \gamma_n(X_n) 
    &= 
    p(Y_n, X_n)
    \\
    &=
    p(Y_n | Y_{n-1}, X_n) p(X_n | X_{n-1}) p(Y_{n-1}, X_{n-1})
    \\
    &=
    p(Y_n | Y_{n-1}, X_n) p(X_n | X_{n-1}) \gamma_{n-1}(X_{n-1})
    .
\end{align*}
If we substitute this expression back into Equation~\eqref{eq:smc-weight-incremental}, then
the incremental weight $W^l_{n \setminus n-1}$ simplifies to
\begin{align}
    \label{eq:smc-weight-incremental-simplified}
    W_{n \setminus n-1}^l
    &=
    \frac{p(Y^l_n \,|\, X^l_{n})}
         {p(Y^{a_{n-1}^l}_{n-1} \,|\, X^{a_{n-1}^l}_{n-1})}
    =
    \prod_{y \in Y^l_{n \setminus n-1} }
    p(y \,|\, X_n^l)
    ,
\end{align}
where $Y^l_{n \setminus n-1} $ is the set difference between the
observed variables at generation $n$ and the observed variables at generation
$n-1$. 
\begin{align*}
  Y^l_{n \setminus n-1} 
  = 
  \text{dom}(\m{Y}^l_n) \setminus \text{dom}(\m{Y}^{a_{n-1}^l}_{n-1})
  .
\end{align*}
In other words, for a bootstrapped proposal, the importance weight at each
generation is defined in terms of the joint probability of observes that 
have been evaluated at breakpoint $n$ but not at $n-1$.

\subsection{Evaluating Proposals}


\begin{algorithm}[!t]
\caption{
  \label{alg:smc-eval}
  Evaluator for bootstrapped sequential Monte Carlo
  }
\begin{algorithmic}[1]
  \State \textbf{global} $\rho,e$
  \Function{eval}{$e$, $\sigma$, $\ell$}
    \Match{$e$}
      \Case{\mang{(sample\ $v$\ $e$)}}
        \State $d, \sigma \gets$ \Call{eval}{$e$, $\sigma$, $\ell$}
        \If{$v \not\in \dom(\sigma(\m{X}))$}
          \State $\sigma(\m{X}(v)) \gets$ \Call{sample}{$d$}
        \EndIf
        \State \Return $\sigma(\m{X}(v))$, $\sigma$
      \EndCase
      \Case{\mang{(observe\ $v$\ $e_1$\ $e_2$)}}
        \State $d, \sigma \gets$ \Call{eval}{$e_1$, $\sigma$, $\ell$}
        \State $c, \sigma \gets$ \Call{eval}{$e_2$, $\sigma$, $\ell$}
        \State $\sigma(\log \Lambda) \gets \sigma(\log \Lambda) + \Call{log-prob}{d, c}$
        \If{$v = \sigma(y_{r})$}
          \State \textbf{error} \Call{resample-breakpoint}{\,}
        \EndIf
        \State \Return $c$, $\sigma$
      \EndCase
      \State $\ldots$ 
      \Comment{Base cases (as in Algorithm \ref{alg:eval-base})}
    \EndMatch
  \EndFunction
  \Function{propose}{$\m{X}$, $y$}
    \State $\sigma \gets [y_{r} \mapsto y, \m{X} \mapsto \m{X}, \log \Lambda \mapsto 0]$
    \Try
        \State $r, \sigma \gets \Call{eval}{e, \sigma, [\,]}$
        \State \Return $r, \sigma(\log \Lambda)$
    \Catch{\Call{resample-breakpoint}{\,}}
        \State \Return $\sigma(\m{X}), \sigma(\log \Lambda)$
    \EndTry
  \EndFunction
\end{algorithmic}
\end{algorithm}

\begin{algorithm}[!t]
\caption{
  \label{alg:smc}
  Sequential Monte Carlo with bootstrapped proposals
  }
\begin{algorithmic}[1]
  \State \textbf{global} $\rho,e$
  \Function{eval}{$e, \sigma, \ell$}
    \State $\ldots$
    \Comment{As in Algorithm~\ref{alg:smc-eval}}
  \EndFunction
  \Function{propose}{$\m{X}$, $y$}
    \State $\ldots$
    \Comment{As in Algorithm~\ref{alg:smc-eval}}
  \EndFunction
  \Function{smc}{$L, y_1, \ldots, y_{N-1}$}
    \State $\log \hat{Z}_0 \gets 0$
    \For {$l$ \textbf{in} $1, \ldots, L$}
        \State $\m{X}_1^l, \log \Lambda_1^l \gets \Call{propose}{[], y_1}$
        \State $\log W^l_{1} \gets \log \Lambda^l_1$
    \EndFor
    \For {$n$ \textbf{in} $2, \ldots, N$}
            \State $\log \hat{Z}_{n-1} \gets \Call{log-mean-exp}{\log W_{n-1}^{1:L}}$
        \For {$l$ \textbf{in} $1, \ldots, L$}
            \State $a_{n-1}^l \sim \Call{discrete}{W^{1:L}_{n-1}/ \sum_l W^l_{n-1}}$
            \If {$n < N$}
                \State $(\m{X}_n^l, \log \Lambda_n^l) \gets \Call{propose}{\m{X}^{a^l_{n-1}}_{n-1}, y_n}$
            \Else
                \State $(r^l, \log \Lambda_N^l) \gets \Call{propose}{\m{X}^{a^l_{N-1}}_{N-1}, \texttt{nil}}$
            \EndIf
            \State $\log W^{l}_n \gets \log \Lambda^{l}_n - \log \Lambda^{a_{n-1}^l}_{n-1} + \log \hat{Z}_{n-1}$
        \EndFor
    \EndFor 
    \State \Return $((r^1, \log W^1_N), \ldots, (r^L, \log W^L_N))$
  \EndFunction
\end{algorithmic}
\end{algorithm}

To implement SMC, we will introduce a function $\textsc{propose}(\m{X}_{n-1},
y_n)$. This function evaluates the program that truncates at the observe
expression with address $y_n$, conditioned on previously sampled values
$\m{X}_{n-1}$, and returns a pair $(\m{X}_{n}, \log \Lambda_{n})$ containing a
map $\m{X}_n$ of values associated with each sample expression and the log
likelihood $\log \Lambda_n = \log p(Y_n | X_n)$. To construct the proposal 
for the final generation we will call 
$\textsc{propose}(\m{X}_{N-1}, \mang{nil}, y_{N-1})$, which returns a pair 
$(r,\log \Lambda)$ in which the return value $r$ replaces the values $\m{X}$.

In Algorithm~\ref{alg:smc-eval} we define this function and its evaluator.
When evaluating sample expressions, we reuse previously sampled values $\m{X}(v)$
for previously sampled variables $v$ and sample from the prior for new
variables $v$. When evaluating observe expressions, we accomulate log
probability into a state variable $\log \Lambda$ as we have done with likelihood
weighting. When we reach the observe expression with a specified symbol 
$y_{r}$, we terminate the program by throwing a special-purpose 
\textsc{resample-breakpoint} error. In the function \textsc{propose}, we initialize 
$\m{X} \gets \m{X}_{n-1}$ and $y \gets y_n$. The evaluator will then reuse all the
previously sampled values $\m{X}_{n-1}$ and run the program until the observe with
address $y_n$, which samples $\m{X}_n | \m{X}_{n-1}$ from the program prior. We then
catch the \textsc{resample-breakpoint} error to return $(\m{X}_n, \log \Lambda_n)$ for a
program that truncates at $y_n$, and return $(r, \log \Lambda)$ when no such error occurs.



\subsection{Algorithm Implementation}

In Algorithm~\ref{alg:smc} we use this proposal mechanism to calculate the 
importance weight at each generation as according to Equation~\eqref{eq:smc-weight-incremental-simplified}
\begin{align}
  \log W_n = \log \Lambda_n - \log \Lambda_{n-1} + \hat{Z}_{n-1}
\end{align}
We calculate $\log \hat{Z}_{n-1}$ at each iteration by evaluating the function
\begin{align} 
  \textsc{log-mean-exp}(\log W_{n-1}^{1:L}) 
  =
  \log \left( 
    \frac{1}{L} \sum_{l=1}^L W_{n-1}^l 
  \right)
  .
\end{align}

\subsection{Computational Complexity}

The proposal generation mechanism in Algorithm~\ref{alg:smc-eval} has a lot in 
common with the mechanism for single-site Metropolis Hastings proposals in 
Algorithm~\ref{alg:lmh-eval}. In both evaluators, we rerun a program conditioned
on previously sampled values $\m{X}$. The advantage of this type of proposal 
strategy is that it is relatively easy to define and understand; a program in 
which all sample expressions evaluate to their previously sampled values is 
fully deterministic, so it is intuitive that we can condition on values of 
random variables in this manner. 

Unfortunately this implementation is not particularly efficient. SMC is most
commonly used in settings where we evaluate one additional observe expression
for each generation, which means that the cardinality of the set of variables 
$|Y^l_{n \setminus n-1}|$ that determines the incremental weight 
in Equation~\eqref{eq:smc-weight-incremental-simplified} is either 1 or $\mathcal{O}(1)$.
Generally this implies that we can also generate proposals and evaluate the 
incremental weight in constant time, which means that a full SMC sweep with $L$ 
samples and $N$ generations requires $\mathcal{O}(LN)$ computation.
For this particular proposal strategy, each proposal step will require
$\mathcal{O}(n)$ time, since we must rerun the program for the first $n$
steps, which means that the full SMC sweep will require $\mathcal{O}(L N^2)$
computation. 

For this reason, the SMC implementation in this section is more a
proof-of-concept implementation than an implementation that one would use in
practice. We will define a more realistic implementation of SMC in
Section~\ref{sec:hoppl-smc}, once we have introduced an execution model based on
continuations, which eliminates the need to rerun the first $n-1$ steps at
each stage of the algorithm.



	\section{Black Box Variational Inference}
	\label{sec:eval-bbvi}

\label{sec:eval-bbvi}

In the sequential Monte Carlo method that we developed in the last section, we
performed resampling at observes in order to obtain high quality importance
sampling proposals. A different strategy for importance sampling is to learn 
a parameterized proposal distribution $q(X ; \lambda)$ in order to maximize
some notion of sample quality. In this section we will learn proposals by
performing variational inference, which optimizes the evidence lower bound
(ELBO) 
\begin{align}
  \begin{split}
  \label{eq:bbvi-elbo}
  \mathcal{L}(\lambda)
  &:=
  \Ev_{q(X;\lambda)}
  \left[
    \log \frac{p(Y,X)}{q(X; \lambda)}
  \right],
  \\
  &\phantom{:}=
  \log p(Y) - \KL{q(X;\lambda)}{p(X|Y)}
  \le
  \log p(Y).
  \end{split}
\end{align}
In this definition, $\KL{q(X;\lambda)}{p(X|Y)}$ is the KL divergence between
the distribution $q(X;\lambda)$ and the posterior $p(X|Y)$, 
\begin{align}
  \KL{q(X;\lambda)}{p(X|Y)}
  :=
  \Ev_{q(X;\lambda)}
  \left[
    \log 
    \frac{q(X;\lambda)}
         {p(X|Y)}
  \right]
  .
\end{align}
The KL divergence is a positive definite measure of dissimilarity between two
distributions; it is 0 when $q(X;\lambda)$ and $p(X|Y)$ are identical and
greater than 0 otherwise, which implies $\mathcal{L}(\lambda) \le \log p(Y)$. We can therefore maximize $\mathcal{L}(\lambda)$ with respect to
$\lambda$ to minimize the KL term, which yields a distribution $q(X ;
\lambda)$ that approximates $p(X | Y)$.
 
In this section we will use variational inference to learn a distribution 
$q(X; \lambda)$ that we will then use as an importance sampling proposal. 
We will assume an approximation $q(X;\lambda)$ in which all variables $x$ are
independent, which in the context of variational inference is known as a mean
field assumption
\begin{align}
  \label{eq:bbvi-mean-field}
  q(X; \lambda)
  = 
  \prod_{x \in X}
  q(x ; \lambda_x)
  .
\end{align}

\subsection{Likelihood-ratio Gradient Estimators}
\label{sec:bbvi-grad-est}

Black-box variational inference (BBVI) \citep[]{wingate2013automated,ranganath2014blackbox} optimizes $\mathcal{L}(\lambda)$ by performing gradient updates using a noisy estimate of the gradient $\hat{\nabla} \mathcal{L}(\lambda)$
\begin{align}
  \label{eq:bbvi-sgd}
  \lambda_t 
  &= 
  \lambda_{t-1} 
  + 
  \eta_t \hat{\nabla}_\lambda \mathcal{L}(\lambda) 
  \big\vert_{\lambda = \lambda_{t-1}},
  &
  \sum_{t=1}^\infty \eta_t
  &= 
  \infty,
  &
  \sum_{t=1}^\infty \eta_t^2 < \infty.
\end{align}
BBVI uses a particular type of estimator for the gradient, which is alternately referred to as a likelihood-ratio estimator or a REINFORCE-style estimator. In general, likelihood-ratio estimators compute a Monte Carlo approximation to an expectation of the form
\begin{align}
  \begin{split}
  \label{eq:grad-expect}
  \nabla_\lambda
  \Ev_{q(X;\lambda)}
  [r(X;\lambda)]
  &=
  \int \!\! dX \:
  \nabla_\lambda q(X; \lambda) r(X;\lambda)
  +
  q(X; \lambda) \nabla_\lambda  r(X;\lambda)
  \\
  &=
  \int \!\! dX \:
  \nabla_\lambda q(X; \lambda) r(X;\lambda)
  +
  \Ev_{q(X ; \lambda)}[\nabla r(X;\lambda)]
  .
  \end{split}
\end{align}
Clearly, this expression is equal to the ELBO in Equation~\eqref{eq:bbvi-elbo}
when we substitute $r(X;\lambda):=\log \big(p(Y,X) / q(X ; \lambda)\big)$. For
this particular choice of $r(X;\lambda)$, the second term in the equation above is 0, 
\begin{align}
  \begin{split}
  \Ev_{q(X ; \lambda)}
  \left[
    \nabla_\lambda 
    \log  
    \frac{p(Y,X)}
         {q(X;\lambda)}
  \right]
  &=
  -
  \Ev_{q(X ; \lambda)}
  \left[
    \nabla_\lambda 
    \log  
    q(X;\lambda)
  \right]
  \\
  &=
  -
  \int \!\! dX \:
  q(X ; \lambda)
  \nabla_\lambda
  \log 
  q(X ; \lambda)
  \\
  &=
  -
  \int \!\! dX \:
  \nabla_\lambda
  q(X ; \lambda)
  =
  -\nabla_\lambda
  1
  =0
  ,
  \end{split}
\end{align}
where the final equalities make use of the fact that, by definition, $\int dX \: q(X ; \lambda) = 1$ since a probability distribution is normalized.

If we additionally substitute $\nabla_\lambda q(X;\lambda) := q(X;\lambda) \nabla_\lambda \log  q(X;\lambda)$ in Equation~\eqref{eq:grad-expect}, then we can express the gradient of the ELBO as

\begin{align}
\nabla_\lambda \mathcal{L}(\lambda)
&=
\Ev_{q(X ; \lambda)}
\left[
  \nabla_\lambda \log q(X;\lambda) 
  \left(
    \log \frac{p(Y,X)}{q(X;\lambda)}
    -
    b
  \right)
\right]
,
\end{align}
where $b$ is arbitrary constant vector, which does not change the expected value since $\Ev_{q(X ; \lambda)}[\nabla_\lambda \log q(X ; \lambda)] = 0$.

The likelihood-ratio estimator for the gradient of the ELBO approximates the expectation with a set of samples $X^l \sim q(X ; \lambda)$. If we define the standard importance weight $W^l = p(Y^l, X^l) / q(X^l ; \lambda)$, the the likelihood-ratio estimator is defined as 
\begin{align}
\label{eq:bbvi-elbo-reinforce}
\hat{\nabla}_\lambda \mathcal{L}(\lambda)
&:=
\frac{1}{L}
\sum_{l=1}^L
  \nabla_\lambda \log q(X^l;\lambda) 
  \left(
    \log W^l
    -
    \hat{b}
  \right)
.
\end{align}
Here we  set $\hat{b}$ to a value that minimizes the variance of the estimator. If we use $(\lambda_{v,1}, \ldots, \lambda_{v,D_v})$ to refer to the  components of the parameter vector $\lambda_v$, then the variance reduction constant $\hat{b}_{v,d}$ for the component $\lambda_{v,d}$ is defined as
\begin{align}
  \label{eq:bbvi-bhat}
  \hat{b}_{v,d}
  &:=
  \frac{\text{covar}(F^{1:L}_{v,d}, G^{1:L}_{v,d})} 
       {\text{var}(G^{1:L}_{v,d})}
  ,
  \\
  \label{eq:bbvi-F}
  F^l_{v,d}
  &:=
  \nabla_{\lambda_{v,d}} \log q(X_v^l;\lambda_v)
  \log W^l
  ,
  \\
  \label{eq:bbvi-G}
  G^l_{v,d}
  &:=
  \nabla_{\lambda_{v,d}} \log q(X_v^l;\lambda_v)
  .
\end{align}


\begin{algorithm}[!t]
\caption{
  \label{alg:bbvi-eval}
  Evaluator for Black Box Variational Inference
}
\begin{algorithmic}[1]
  \State \textbf{global} $\rho$
  \Function{eval}{$e$, $\sigma$, $\ell$}
    \Match{$e$}
      \Case{\mang{(sample\ $v$\ $e$)}}
        \State $d,\sigma \gets$ \Call{eval}{$e$, $\sigma$, $\ell$}
        \If{$v \not\in \text{dom}(\sigma(\m{Q}))$}
            \State $\sigma(\m{Q}(v)) \gets p$
            \Comment{Initialize proposal using prior}
        \EndIf
        \State $c \sim \Call{sample}{\sigma(\m{Q}(v))}$
        \State $\sigma(G(v)) \gets \Call{grad-log-prob}{\sigma(\m{Q}(v)), c}$
        \State $\log W_v \gets \Call{log-prob}{d, c} - \Call{log-prob}{\sigma(\m{Q}(v)),c}$
        \State $\sigma(\log W) \gets \sigma(\log W) + \log W_v$
        \State \Return $c$, $\sigma$
      \EndCase
      \Case{\mang{(observe\ $v$\ $e_1$\ $e_2$)}}
        \State $d, \sigma \gets$ \Call{eval}{$e_1$, $\sigma$, $\ell$}
        \State $c, \sigma \gets$ \Call{eval}{$e_2$, $\sigma$, $\ell$}
        \State $\sigma(\log W) \gets \sigma(\log W) + \Call{log-prob}{d, c}$
        \State \Return $c$, $\sigma$
      \EndCase
      \State $\ldots$ 
      \Comment{Base cases (as in Algorithm \ref{alg:eval-base})}
    \EndMatch
  \EndFunction
\end{algorithmic}
\end{algorithm}

\begin{algorithm}[!t]
\caption{
  \label{alg:bbvi}
  Black Box Variational Inference
}
\begin{algorithmic}[1]
  \State \textbf{global} $\rho,e$
  \Function{eval}{$e, \sigma, \ell$}
    \State $\ldots$
    \Comment{As in Algorithm~\ref{alg:bbvi-eval}}
  \EndFunction
  \Function{optimizer-step}{$\m{Q}$, $\hat{g}$}
    \For{$v$ \textbf{in} $\text{dom}(\hat{g})$}
        \State $\lambda(v) \gets \Call{get-parameters}{\m{Q}(v)}$
        \State $\lambda'(v) \gets \lambda(v) + \ldots$
        \Comment{SGD/Adagrad/Adam update}
        \State $\m{Q}(v) \gets \Call{set-parameters}{\m{Q}(v), \lambda'}$
    \EndFor
    \Return $\m{Q}$
  \EndFunction
  \Function{elbo-gradients}{$G^{1:L}$, $\log W^{1:L}$}
    \For{$v$ \textbf{in} $\text{dom}(G^1) \cup \ldots \cup \text{dom}(G^L)$}
      \For{$l$ \textbf{in} $1, \ldots, L$}
        \If{$v \in \text{dom}(G^l)$}
          \State $F^{l}(v) \gets G^{l}(v) \log W^{1:L}$
        \Else
          \State $F^{l}(v), G^{l}(v) \gets 0, 0$ 
        \EndIf
      \EndFor
      \State $\hat{b} \gets \Call{sum}{\Call{covar}{F^{1:L}(v), G^{1:L}(v)}} / \Call{sum}{\Call{var}{G^{1:L}(v)}}$
      \State $\hat{g}(v) \gets \Call{sum}{F^{1:L}(v) - \hat{b} \: G^{1:L}(v)} / L$
    \EndFor
    \State \Return $\hat{g}$
  \EndFunction
  \Function{bbvi}{$T$, $L$}
    \State $\sigma \gets [\log W \mapsto 0, \m{Q} \mapsto [\,], G \mapsto [\,]]$
    \For{$t$ \textbf{in} $1,\ldots,T$}
        \For{$l$ \textbf{in} $1,\ldots,L$}
            \State $r^{t,l}, \sigma^{t,l} \gets \Call{eval}{e, \sigma, [\,]}$
            \State $G^{t,l},\log W^{t,l} \gets \sigma^{t,l}(G), \sigma^{t,l}(\log W)$
        \EndFor
        \State $\hat{g} \gets \Call{elbo-gradients}{G^{t,1:L}, \log W^{t,1:L}}$
        \State $\sigma(\m{Q}) \gets \Call{optimizer-step}{\sigma(\m{Q}), \hat{g}}$
    \EndFor
    \State \Return $((r^{1,1}, \log W^{1,1}), \ldots, (r^{1,L}, \log W^{1,L}), \ldots, (r^{T,L}, \log W^{T,L}))$
  \EndFunction
\end{algorithmic}
\end{algorithm}

\subsection{Evaluator for Gradient Estimation}

From the equations above, we see that we need to calculate two sets of  quantities in order to estimate the gradient of the ELBO. The first consists of the importance weights $\log W^l$. The second consists of the gradients of the log proposal density for each variable $G^l_{v} = \nabla_{\lambda_{v}} \log q(X^l_v | \lambda_v)$. 

In Algorithm~\ref{alg:bbvi-eval} we define an evaluator that extends the likelihood-ratio evaluator from Algorithm~\ref{alg:lw-eval} in two ways: 
\begin{itemize}
  \item[1.] Instead of sampling proposals from the program prior, we now propose from a distribution $\m{Q}(v)$ for each variable $v$ and 
  update the importance weight $\log W$ accordingly.

  \item[2.] When evaluating a sample expression, we additionally calculate the gradient of the log proposal density $G(v) = \nabla_{\lambda_v} \log q(X_v | \lambda_v)$. For this we assume an implementation of a function $\textsc{grad-log-prob}(d,c)$ for each primitive distribution type supported by the language.
\end{itemize}

Algorithm~\ref{alg:bbvi} defines a BBVI algorithm based on this evaluator. The function $\textsc{elbo-gradients}$ returns a map $\hat{g}$ in which each entry $\hat{g}(v) := \hat{\nabla}_{\lambda_v} \mathcal{L}(\lambda)$ contains the gradient components for the variable $v$ as defined in Equations~\eqref{eq:bbvi-elbo-reinforce}-\eqref{eq:bbvi-G}. The main algorithm \textsc{bbvi} then simply runs the evaluator $L$ times at each iteration and then passes the computed  gradient estimates $\hat{g}$ to a function $\textsc{optimizer-step}$, which can either implement the vanilla stochastic gradient updates defined in Equation~\eqref{eq:bbvi-sgd}, or more commonly updates for an extension of stochastic gradient descent such as Adam \citep{kingma2015adam} or Adagrad \citep{duchi2011adaptive}.
 
\subsection{Computational Complexity and Statistical Efficiency}

From an implementation point of view, BBVI is a relatively simple algorithm. The main reason for this is the mean field approximation for $q(X ; \lambda)$ in Equation~\eqref{eq:bbvi-mean-field}. Because of this approximation, calculating the gradients $\nabla_\lambda \log q(X;\lambda)$ is easy, since we can calculate the gradients $\nabla_{\lambda_v} \log q(X_v ; \lambda_v)$ for each component independently, which only requires that we implement gradients of the log density for each primitive distribution type. 

One of the main limitations of this BBVI implementation is that the gradient estimator tends to be relatively high variance, which means that we will need a relatively large number of samples per gradient step $L$ in order to ensure convergence. Values of $L$ of order $10^2$ or $10^3$ are not uncommon, depending on the complexity of the model. For comparison, methods for variational autoencoders that compute the gradient of a reparameterized objective \citep{kingma2014auto,rezende2014stochastic} can be evaluated with $L=1$ samples for many models. In addition to this, the number of iterations $T$ that is needed to achieve convergence can easily be order $10^3$ to $10^4$. This means that BBVI we may need order $10^6$ or more samples before BBVI starts generating high quality proposals. 
We will revisit this algorithm and discuss alternative gradient estimator implementations in Chapter~\ref{sec:gbli}.

When we compile a program to a graph $(V,A,\m{P}, \m{Y})$ we can perform an additional optimization to reduce the variance. To do so, we replace the term 
$\log W$ in the objective with a vector in which each component $\log W_v$
contains a weight that is restricted to the variables in the Markov blanket,
\begin{align}
  \log W_v 
  = 
  \sum_{w \in \textsc{mb}(v)\}} \frac{p(w | \pa(w))}{q(w | \lambda_w)}
  ,
\end{align}
where the Markov blanket $\textsc{mb}(v)$ of a variable $v$ is 
\begin{align}
  \begin{split}
  \textsc{mb}(v) 
  = 
  \pa(v) 
  &\cup 
  \{w: w \in \pa(v) \} 
  \\
  &\cup \left\{w: \exists u \Big(v \in \pa(u) \wedge  w \in \pa(u) \Big) \right\}.
  \end{split}
\end{align}
This can be interpreted as a form of Rao-Blackwellization \citep{ranganath2014blackbox}, which reduces the variance by ignoring the components of the weight that are not directly associated with the sampled value $X_v$. In a graph-based implementation of BBVI, one can easily construct this Markov blanket, which we rely upon in the implementation of Gibbs sampling~\ref{sec:gibbs}.


\chapter{A Probabilistic Language With Recursion}
\label{ch:hoppl}

%
%

In the three preceding chapters we have introduced a first-order probabilistic
programming language and described graph- and evaluation-based inference
methods. The defining characteristic of the FOPPL is that it is suitably
restricted to ensure that there can only ever be a finite number of random
variables in any model denoted by a program.

In this chapter we relax this restriction by introducing a higher-order
probabilistic programming language (HOPPL) that supports programming language
features, such as higher-order procedures and general recursion.  
HOPPL programs can denote models with an unbounded number of random variables.
This rules out graph-based evaluation strategies immediately, since an
infinite graph cannot be represented on a finite-capacity computer.  However,
it turns out that evaluation-based inference strategies can still be made to
work by considering only a finite number of random variables at any particular
time, and this is what will be discussed in the subsequent chapter.


In the FOPPL, we ensured that programs could be compiled to a finite graph by
placing a number of restrictions on the language:
\begin{itemize}
\item The \fop{defn} forms disallow recursion;
\item Functions are not first class objects, which means that it is not
possible to write higher-order functions that accept functions as arguments;
\item The first argument to the \fop{loop} form, the loop depth, has to be a
constant, as \fop{loop} was syntactic sugar unrolled to nested let expressions
at compile time.
\end{itemize}

Say that we wish to remove this last restriction, and would
like to be able to loop over the range determined by the runtime value of a
program variable. 

This
means that the looping construct cannot be syntactic sugar, but must instead be
a function that takes the loop bound as an argument and repeats the execution
of the loop body up to this dynamically-determined bound.

If we wanted to implement a loop function that supports a dynamic number of
loop iterations, then we could do so as follows
\begin{hoppl}[mathescape]%,caption={FOPL: Linear regression with sugared let},label=linear_regression_simplified]
(defn loop-helper [$i$ $c$ $v$ $f$ $a_1$ $\ldots$ $a_n$]
  (if (= $i$ $c$)
      $v$
      (let [$v'$ ($f$ $i$ $v$ $a_1$ $\ldots$ $a_n$)]
         (loop-helper (+ $i$ 1) $c$ $v'$ $f$ $a_1$ $\ldots$ $a_n$))))
(defn loop [$c$ $v$ $f$ $a_1$ $\ldots$ $a_n$]
  (loop-helper 0 $c$ $v$ $f$ $a_1$ $\ldots$ $a_n$)).
\end{hoppl}
In order to implement this function we have to allow the \fop{defn} form to
make recursive calls, a large departure from the FOPPL restriction. Doing so
gives us the ability to write programs that have loop bounds that are
determined at runtime rather than at compile time, a feature that most
programmers expect to have at their disposal when writing any program.
However, as soon as loop is a function that takes a runtime value as a bound, 
then we could write programs such as
\begin{hoppl}[mathescape]
(defn flip-and-sum [$i$ $v$] 
  (+ $v$ (sample (bernoulli 0.5))))
(let [$c$ (sample (poisson 1))]
  (loop $c$ 0 flip-and-sum)).
\end{hoppl}
This program, which represents the distribution over the sums of the outcomes
of a Poisson distributed number of of fair coin flips, is one of the shortest
programs that illustrates concretely what we mean by a program that denotes an
infinite number of random variables. Although this program is not particularly
useful, we will soon show many practical reasons to write programs like this.
If one were to attempt the loop desugaring approach of the FOPPL here one
would need to desugar this loop for all of the possible constant values $c$
could take.  
As the support of the Poisson distribution is unbounded above, one
would need to desugar a loop indefinitely, leading to
an infinite number of random variables (the Bernoulli draws) in the expanded
expression. 
The corresponding graphical model would have an infinite number
of nodes, which means that it is no longer possible to compile this model to a
graph.

The unboundedness of the number of random variables is the central issue. It
arises naturally when one uses stochastic recursion, a common way of
implementing certain random variables. Consider the example
\begin{hoppl}[mathescape]
(defn geometric-helper [n dist] 
  (if (sample dist)
       n
       (geometric-helper (+ n 1))))
(defn geometric [p]
 (let [dist (flip p)]
    (geometric-helper 0 dist))).
\end{hoppl}
This is a well-known sampler for geometrically distributed random variables. 
Although a primitive for the geometric distribution would definitely be provided by a probabilistic programming language (e.g.\ in the FOPPL), 
the point of this example is to demonstrate that the use of infinitely many random variables arises with the introduction of stochastic recursion.
Notably, here, it could be that this particular computation never terminates, as at each stage of the recursion \hop{(sample dist)} could return \hop{false}, with probability \hop{p}.
Leveraging referential transparency, one could attempt to inline the helper function above as
\filbreak
\begin{hoppl}[mathescape]
(defn geometric [p]
 (let [dist (flip p)]
    (if (sample dist)
      0
      (if (sample dist)
        1
        (if (sample dist)
          2
          $\vdots$
            (if (sample dist)
               $\infty$
               (geometric-helper (+ $\infty$ 1))))))))
\end{hoppl}
but the problem in attempting to do so quickly becomes apparent. Without a
deterministic loop bound, the inlining cannot be terminated, showing that the
denoted model has an infinite number of random variables. No inference
approach which requires eager evaluation of if statements, such as the graph
compilation techniques in the previous chapter, can be applied in general.

While expanding the class of denotable models is important, the primary reason
to introduce the complications of a higher-order modeling language is
that ultimately we would like simply to be able to do probabilistic
programming using any {\em existing} programming language as the modeling
language. If we make this choice, we need to be able to deal with all of the
possible models that could be written in said language and, in general, we
will not be able to syntactically prohibit stochastic loop bounds or
conditioning on data whose size is known only at runtime. Furthermore, in the
following chapter we will show how to do probabilistic programming using not
just an existing language syntax but also an existing compiler and runtime
infrastructure. Then, we may not even have access to the source code of the
model. A probabilistic programming approach that extends an existing language
in this manner will typically target a family of models that are, roughly
speaking, in the same class as models that can be defined using the HOPPL.


\section{Syntax}
\label{sec:hoppl-grammar}

Relative to the first-order language in Chapter~\ref{ch:foppl}, the higher-order
language that we introduce here has two additional features. The first is that
functions can be recursive. The second is that functions are first-class
values in the language, which means that we can define higher-order functions
(i.e.~functions that accept other functions as arguments). The syntax for the
HOPPL is shown in Language~\ref{higher_order_prob_prog_lang}.

\begin{grammar}[mathescape,caption={Higher-order probabilistic programming language (HOPPL)},label=higher_order_prob_prog_lang,float=htp,floatplacement=htbp]
  $v ::=$ $\textrm{variable}$
  $c ::=$ $\textrm{constant value or primitive operation}$
  $f ::=$ $\textrm{procedure}$
  $e ::=$ $c$ | $v$ | $f$ | (if $e$ $e$ $e$) | ($e$ $e_1 \ldots e_n$) | (sample $e$) 
     | (observe $e$ $e$) | (fn [$v_1 \ldots v_n$] $e$) 
  $q ::=$ $e$ | (defn $f$ [$v_1 \ldots v_n$] $e$) $q$.
\end{grammar}

While a procedure had to be declared globally in the FOPPL, functions in the
HOPPL can be created locally using an expression
\hop{(fn $[v_1 \ldots v_n]\ e$)}. Also, the HOPPL lifts the restriction of the FOPPL
that the operators in procedure calls are limited to globally declared
procedures $f$ or primitive operations $c$; as the case
\hop{($e\ e_1 \ldots e_n$)} in the grammar indicates, a general expression $e$ may appear as
an operator in a procedure call in the HOPPL. Finally, the HOPPL drops the
constraint that all procedures are non-recursive. When defining a procedure
$f$ using
\hop{(defn$\ f\ [v_1 \ldots v_n]\ e$)} in the HOPPL, we are no longer forbidden to call $f$ in the body 
$e$ of the procedure.


Again not that the key distinction between FOPPL and HOPPL programs is finite versus unbounded random variable cardinality.  The choice of names and acronyms, FOPPL and HOPPL, for these two specific languages  describes the distinction in language features available in each.  However, please note that there are other mechanisms (i.e.~loops with stochastic guards in imperative languages without first class functions) for introducing unbounded variable cardinality models, so care should be taken when describing a language as being ``a FOPPL'' or ``a HOPPL'' rather than referring specifically to these specific FOPPL and HOPPL languages.  In designing these languages our aim, however, was to establish the difference between finite random variable cardinality languages (FRVCL) and unbounded random variable cardinality languages (URVCL).  These acronyms are terrible.  At points we casually refer to FRVCL as FOPPLs and URVCL as HOPPLs but in so doing what we really mean to communicate is the difference between FRVCLs and URVCLs not that one has first class functions and the other does not.  The PPL community has had a difficult time demarking this crucial difference, using terms such as ``Turing complete'' and ``universal'' to actually describe URVCLs.  

Such higher-order language features are present in Church,
Venture,  Anglican, and WebPPL and are required to reason about languages like Probabilistic-C, Turing, Pyro, and PyProb.  In the following we illustrate the benefits of having these features by short evocative source code examples of some kinds of advanced probabilistic models that can now be expressed.  In the next chapter we describe a class of inference algorithms
suitable for performing inference in the models that are denotable in such an expressive higher-order probabilistic programming language.

\section{Syntactic sugar}
\label{sec:hoppl-sugar}
We will define syntactic sugar that re-establishes some of the convenient syntactic features of the HOPPL.  Note that the syntax of the HOPPL omits the \lsi{let} expression. This is because it can be defined in terms of nested functions as
\begin{hoppl}[mathescape]%,caption={Desugaring HOPPL let expression},label=hoppl:desugared_let]
(let [$x$ $e_1$] $e_2$) = ((fn [$x$] $e_2$) $e_1$).
\end{hoppl}
For instance,
\begin{hoppl}%[mathescape,caption={Example of desugaring let with multiple bindings and body expressions in the HOPPL},label=_ho_desugaring_let_example]

(let [a (+ k 2)
      b (* a 6)]
  (print (+ a b))
  (* a b))
\end{hoppl}
gets first desugared to the following expression
\begin{hoppl}%[mathescape,caption={Example of Desugaring Let With Multiple Bindings and Body Expressions},label=_ho_desugaring_let_example]

(let [a (+ k 2)]
  (let [b (* a 6)]
    (let [c (print (+ a b))]
      (* a b))))
\end{hoppl}
where \hop{c} is a fresh variable.  This can then be desugared to the expression without \hop{let} as follows
\begin{hoppl}%[mathescape,caption={Example of Desugaring Let With Multiple Bindings and Body Expressions},label=ho_desugared_let_example]

((fn [a] 
   ((fn [b]
     ((fn [c] (* a b))
      (print (+ a b))))
    (* a 6)))
 (+ k 2)).
\end{hoppl}

While we already described a HOPPL \mang{loop} implementation in the preceding text,
we have elided the fact that the FOPPL \mang{loop} accepts a variable number of arguments, a language feature
we have not explicitly introduced here.
An exact replica of the FOPPL loop can be implemented as HOPPL sugar, with loop desugaring 
occurring prior to the let desugaring. If we define the helper function
\begin{hoppl}[mathescape]
(defn loop-helper [$i$ $c$ $v$ $g$]
  (if (= $i$ $c$)
      $v$
      (let [$v'$ ($g$ $i$ $v$)]
         (loop-helper (+ $i$ 1) $c$ $v'$ $g$))))
\end{hoppl}
the expression \hop{(loop $c$ $e$ $f$ $e_1$ $\cdots$ $e_n$)} can be desugared to
\begin{hoppl}[mathescape]
(let [bound $c$
      initial-value $e$
      $a_1$ $e_1$
         $\vdots$ 
      $a_n$ $e_n$
      $g$ (fn [$i$ $w$] ($f$ $i$ $w$ $a_1$ $\ldots$ $a_n$))]
  (loop-helper 0 bound initial-value $g$)).
\end{hoppl}
With this loop and let sugar defined, and the implementation of foreach straightforward, any valid FOPPL program is also valid in the HOPPL.

\section{Examples}
\label{sec:hoppl-examples}

In the HOPPL, we will employ a number of design patterns from functional
programming, which allow us to write more conventional code than was necessary
to work around limitations of the FOPPL.
Here we give some examples of higher-order function implementations and usage
in the HOPPL before revisiting models previously discussed in
Chapter~\ref{ch:foppl} and introducing new examples which depend on new
language features.

\subsection*{Examples of higher-order functions}


We will frequently rely on higher-order functions \mang{map} and \mang{reduce}. 
We can write these explicitly as HOPPL functions which take functions as arguments, and do so here by way of introduction
to HOPPL usage before considering generative model code.

\paragraph{Map.} The higher-order function \mang{map} takes two arguments: a function and a sequence.
It then returns a new sequence, constructed by applying the function to every individual element of the sequence.
\begin{hoppl}[mathescape] % ,caption={Definition of \mang{map}}]
(defn map [f values]
  (if (empty? values)
    values
    (prepend (map f (rest values))
             (f (first values)))))
\end{hoppl}
Here \mang{prepend} is a primitive that prepends a value to the beginning of a
sequence. This ``loop'' works by applying \mang{f} to the first element of the
collection \mang{values}, and then recursively calling \mang{map} with the
same function on the rest of the sequence. At the base case, for an empty
input \mang{values}, we return the empty sequence of values. 

\paragraph{Reduce.} The \mang{reduce} operation, also known as ``fold'', takes
a function and a sequence as input, along with an initial state; unlike
\mang{map}, it returns a single value. The fixed-length \mang{loop} construct
we defined as syntactic sugar in the FOPPL can be thought of as a poor-man's
\mang{reduce}. The function passed to \mang{reduce} takes a state and a value,
and computes a new state. We get the output by repeatedly applying the
function to the current state and the first item in the list, recursively
processing the rest of the list.
\begin{hoppl}[mathescape]%,caption={Definition of \mang{reduce}}]
(defn reduce [f x values]
  (if (empty? values)
    x
    (reduce f (f x (first values)) (rest values))))
\end{hoppl}
Whereas \hop{map} is a function that maps a sequence of values onto a sequence
of function outputs, \hop{reduce} is a function that produces a single result.
An example of where you might use \hop{reduce} is when writing a function that
computes the sum of all entries in a sequence:
\begin{hoppl}[mathescape]
(defn sum [items]
  (reduce + 0.0 items))
\end{hoppl}
Note that the output of \hop{reduce} depends on the return type of the
provided function. For example, to return a list with the same entries as the
original list, but reversed, we can use a \hop{reduce} with a function that
builds up a list from back-to-front:
\begin{hoppl}[mathescape]
(defn reverse [values]
  (reduce prepend [] values))
\end{hoppl}

\paragraph{No need to inline data.} A consequence of allowing unbounded
numbers of random variables in the model is that we no longer need to
``inline'' our data. In the FOPPL, each \mang{loop} needed to have an explicit
integer literal representing the total number of iterations in order to
desugar to \mang{let} forms. 
As a result, each program that we wrote had to hard-code the total number of
instances in any dataset. Flexible looping structures mean we can read data
into the HOPPL in a more natural way; assuming libraries for e.g.~file access,
we could read data from disk, and use a recursive function to loop through
entries until reaching the end of the file.

For example, consider the hidden Markov model in the FOPPL given by
Program~\ref{model:hmm}. In that implementation, we hard coded the number of
loop iterations (there, 16) to the length of the data. In the HOPPL, suppose
instead we have a function which can read the data in regardless of its
length.
\begin{hoppl}[mathescape]
(defn read-data []
  (read-data-from-disk "filename.csv")) 

;; Sample next HMM latent state and condition 
(defn hmm-step [trans-dists obs-dists]
  (fn [states data]
    (let [state (sample (get trans-dists 
                             (last states)))]
      (observe (get obs-dists state) data)
      (conj states state))))

(let [trans-dists [(discrete [0.10 0.50 0.40])
                   (discrete [0.20 0.20 0.60])
                   (discrete [0.15 0.15 0.70])]
      obs-dists [(normal -1.0 1.0)
                 (normal 1.0 1.0)
                 (normal 0.0 1.0)]
      state [(sample (discrete [0.33 0.33 0.34]))]]
  ;; Loop through the data, return latent states
  (reduce (hmm-step trans-dists obs-dists) 
          [state] 
          (read-data)))
\end{hoppl}

The \mang{hmm-step} function now takes a vector containing the current states,
and a {\em single} data point, which we \observe. Rather than using an
explicit iteration counter $n$, we can use \mang{reduce} to traverse the data
recursively, building up and returning a vector of visited states.

\subsection*{Open-universe Gaussian Mixtures}

The ability to write loops of unknown or random iterations is not just a handy tool for writing more readable code;
as hinted by the recursive geometric sampler example, it also increases the expressivity of the model class.
Consider the Gaussian mixture model example we implemented in the FOPPL in Program~\ref{model:gmm}: 
there we had two explicit loops,
one over the number of data points, but the other over the number of mixture components, which we had to fix at compile time.
As an alternative, we can re-write the Gaussian mixture to define a distribution over the number of components.
We do this by introducing a prior over the number of mixture components; this prior could be e.g.~a Poisson distribution,
which places non-zero probability on all positive integers.

To implement this, we can define a higher-order function, \mang{repeatedly},
which takes a number $n$ and a function $f$, and constructs a sequence of length
$n$ where each entry is produced by invoking $f$.
\begin{hoppl}
(defn repeatedly [n f]
  (if (<= n 0)
    []
    (append (repeatedly (- n 1) f) (f))))
\end{hoppl}
The \mang{repeatedly} function can stand in for the fixed-length loops that we used to sample mixture components from the prior 
in the FOPPL implementation.
An example implementation is in Program~\ref{model:open-universe-gmm}.
\begin{hoppl}[mathescape,label=model:open-universe-gmm,caption={HOPPL: An open-universe Gaussian mixture model with an unknown number of components}] %,float=tp,floatplacement=tbp]
(defn sample-likelihood []
  (let [sigma (sample (gamma 1.0 1.0))
        mean (sample (normal 0.0 sigma))]
    (normal mean sigma)))

(let [ys [1.1 2.1 2.0 1.9 0.0 -0.1 -0.05]
      K (sample (poisson 3)) ;; random, with mean 3
      ones (repeatedly K (fn [] 1.0))
      z-prior (discrete (sample (dirichlet ones)))
      likes (repeatedly K sample-likelihood)]
  (map (fn [y] 
         (let [z (sample z-prior)]
           (observe (nth likes z) y)
           z))
       ys))
\end{hoppl}

Here we still used a fixed, small data set (the \mang{ys} values, same as before, are inlined) but the model code would not change
if this were replaced by a larger data set.
Models such as this one, where the distribution over the number of mixture components $K$ is unbounded above, are sometimes
known as {\em open-universe} models: given a small amount of data, we may infer there are only a small number of clusters;
however, if we were to add more and more entries to \mang{ys} and re-run inference, we do not discount the possibility that there
are additional clusters (i.e.~a larger value of $K$) than we had previously considered.

Notice that the way we wrote this model interleaves sampling from $z$ with observing values of $y$,
rather than sampling all values $z_1, z_2, z_3, \dots$ up front.
While this does not change the definition of the model (i.e.~does not change the joint distribution over observed and latent variables),
writing the model in a formulation which moves \observe statements as early as possible (or alternatively delays calls to \sample)
yields more efficient SMC inference.

\subsection*{Sampling with constraints}

One common design pattern involves simulating from a distribution, subject to constraints.
Obvious applications include sampling from truncated variants of known distributions, such as a normal distribution with a positivity constraint;
however, such rejection samplers are in fact much more common than this.
In fact, samplers for most standard distributions (e.g.~Gaussian, gamma, Dirichlet) are implemented under the hood as
rejection samplers which propose from some known simpler distribution, and evaluate an acceptance criteria;
they continue looping until the criteria evaluates to true.

In a completely general form, we can write this algorithm as a higher-order function which takes two functions as arguments:
a \mang{proposal} function which simulates a candidate point,
and \mang{is-valid?} which returns true when the value passed satisfies the constraint.
\begin{hoppl}[mathescape]
(defn rejection-sample [proposal is-valid?]
  (let [value (proposal)]
    (if (is-valid? value) 
      value
      (rejection-sample proposal is-valid?))))
\end{hoppl}
This sort of accept-reject algorithm can take an unknown number of iterations, and thus cannot be expressed in the FOPPL.

The \mang{rejection-sample} function can be used to implement samplers for distributions which do not otherwise have samplers, for example when sampling from constrained in simulation-based models in the physical sciences.

\subsection*{Program synthesis}

As a more involved modeling example which cannot be written without exploiting higher-order language features,
we consider writing a generative model for mathematical functions.
The representation of functions we will use here is actually literal code written in the HOPPL: 
that is, our generative model will produce samples of function bodies \mang{(fn [] $\dots$)}.
For purposes of illustration, suppose we restrict to simple arithmetic functions of a single variable, 
which we could generate using the grammar
\begin{lstlisting}[mathescape]
  $\mathit{op}$ $::=$ + | - | * | /
  $\mathit{num}$ $::=$ 0 | 1 | $\dots$ | 9
  e $::=$ $\mathit{num}$ | $x$ | ($\mathit{op}$ e e)
  f $::=$ (fn [$x$] ($\mathit{op}$ e e))
\end{lstlisting}
We can sample from the space of all functions $f(x)$ generated by composition of digits with \hop{+}, \hop{-}, \hop{*}, and \hop{/},
by starting from the initial rule for expanding $f$ and recursively applying rules to fill in values of 
$\mathit{op}$, $\mathit{num}$, and $e$ until only terminals remain.
To do so, we need to assign a probability for sampling each rule at each stage of the expansion.
In the following example, when expanding each $e$ we choose a number with probability $0.4$, the symbol $x$ with probability $0.3$, and a new function application with probability $0.3$; both operations and numbers $0,
\ldots,9$ are chosen uniformly.
\begin{hoppl}[mathescape,caption={generative model for function of a single variable},label=prog:gen-arithmetic]
(defn gen-operation []
  (sample (uniform [+ - * /])))

(defn gen-expr []
  (let [expr-prior (discrete [0.4 0.3 0.3])
        expr-type (sample expr-prior)]
    (case expr-type 
      0 (sample (uniform-discrete 0 10)) 
      1 (quote x)
      2 (let [operation (gen-operation)]
          (list operation 
                (gen-expr)
                (gen-expr))))))

(defn gen-function [] 
  (list (quote fn) [(quote x)] 
    (list (gen-operation) 
          (gen-expr) 
          (gen-expr))))
\end{hoppl}
In this program we make use of two constructs that we have not previously encountered. The first is the \hop{(case $v$ $e_1$ $\ldots$ $e_n$)} form, which is syntactic sugar that allows us to select between more than two branches, depending on the value of the variable $v$. The second is the \hop{list} data type. A call \hop{(list 1 2 3)} returns a list of values \hop{(1 2 3)}. We differentiate a list from a vector by using round parentheses \hop{(...)} rather than squared parentheses \hop{[...]}. 

In this program we see one of the advantages of a language which inherits from
LISP and Scheme: programmatically generating code in the HOPPL is quite
straightforward, requiring only standard operations on a basic \mang{list}
data type. The function \mang{gen-function} in Program~\ref{prog:gen-arithmetic} 
returns a list, not a ``function''. That is, it does not directly produce a HOPPL 
function which we can call, but rather the source code for a function. In defining 
the source code, we used the \hop{quote} function to wrap keywords and symbols
in the source code, e.g. \hop{(quote x)}. This primitive prevents the source code 
from being evaluated, which means that the variable name \hop{x} is included into 
the list, rather than the value of the variable (which does not exist).
Repeated invocation of \mang{(gen-function)} produces samples from the grammar, which can be used as a basic diagnostic:
\begin{hoppl}[mathescape,caption={Unconditioned samples from a generative model for arithmetic expressions, produced by calling \mang{(gen-function)}}]
(fn [x] (- (/ (- (* 7 0) 2) x) x))
(fn [x] (- x 8))
(fn [x] (* 5 8))
(fn [x] (+ 7 6))
(fn [x] (* x x))
(fn [x] (* 2 (+ 0 1)))
(fn [x] (/ 6 x))
(fn [x] (- 0 (+ 0 (+ x 5))))
(fn [x] (- x 6))
(fn [x] (* 3 x))
(fn [x]
  (+ (+ 2 
        (- (/ x x) 
           (- x (/ (- (- 4 x) (* 5 4)) 
                   (* 6 x)))))
     x))
(fn [x] (- x (+ 7 (+ x 4))))
(fn [x] (+ (- (/ (+ x 3) x) x) x))
(fn [x] (- x (* (/ 8 (/ (+ x 5) x)) (- 0 1))))
(fn [x] (/ (/ x 7) 7))
(fn [x] (/ x 2))
(fn [x] (* 8 x))
\end{hoppl}
Most of the generated expressions are fairly short, with many containing only a single function application.
This is because the choice of probabilities in Program~\ref{prog:gen-arithmetic} is biased towards 
avoiding nested function applications; the probability of producing a number or the variable $x$
is $0.7$, a much larger value than the probability $0.3$ of producing a function application.
However, there is still positive probability of sampling an expression of any arbitrarily large size --- there is nothing which
explicitly bounds the number of function applications in the model.
Such a model could not be written in the FOPPL without introducing a hard bound on the recursion depth. 
In the HOPPL we can allow functions to grow long if necessary, while still preferring short results, thanks to the
eager evaluation of \mang{if} statements and the lack of any need to enumerate possible random choices.

Note that some caution is required when defining models which can generate a countably infinite number of
latent random variables: it is possible to write programs which do not necessarily terminate.
In this example, had we assigned a high enough probability to the expansion rule \mang{$e\ \rightarrow\ $(op $\ e$ $\ e$)},
then it is possible that, with positive probability, the program {\em never} terminates.
In contrast, it is not possible to inadvertently write an infinite loop in the FOPPL.

If we wish to fit a function to data, it is not enough to merely generate the source code for the function --- we also
need to actually evaluate it.
This step actually requires invoking either a compiler or an interpreter to parse the symbolic representation of the function
(i.e., as a list containing symbols) and evaluate it to a user-defined function, 
just as if we had included the expression \mang{(fn [$x$] $\dots$)} in our original program definition.
The magic word is \mang{eval}, which we assume to be supplied as a primitive in the HOPPL target language.
We use \mang{eval} to evaluate code that has previously been quoted with \hop{quote}.
Consider the function \hop{(fn [x] (- x 8))}.
Using \hop{quote}, we can define source code (in the form of a list)
that could then be evaluated to produce the function itself,
\begin{hoppl}[mathescape]
;; These two lines are identical:
(eval (quote (fn [x] (- x 8))))
(fn [x] (- x 8))
\end{hoppl}
For our purposes, we will want to evaluate the generated functions at particular inputs to see how well they conform to some specific target data, e.g.
\begin{hoppl}[mathescape]
;; Calling the function at x=10 (outputs: 2)
(let [f (eval (quote (fn [x] (- x 8))))]
  (f 10))
\end{hoppl}

Running a single-site Metropolis-Hastings sampler, using an algorithm similar to that in Section~\ref{sec:eval-mh}
(which we will describe precisely in Section~\ref{sec:hoppl-mh}),
we can draw posterior samples given particular data.
Some example functions are shown in Figure~\ref{fig:function-induction}, conditioning on three input-output pairs.
\begin{figure}[t]
\centering
\includegraphics[width=0.48\textwidth]{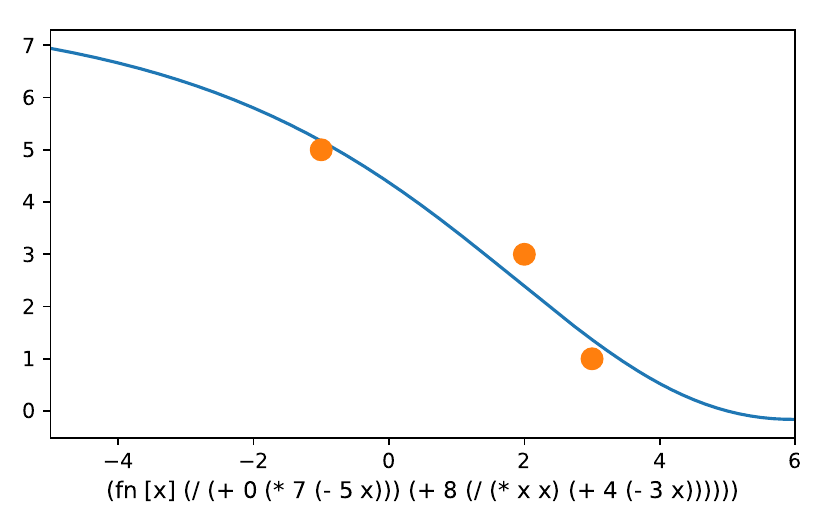}
\includegraphics[width=0.48\textwidth]{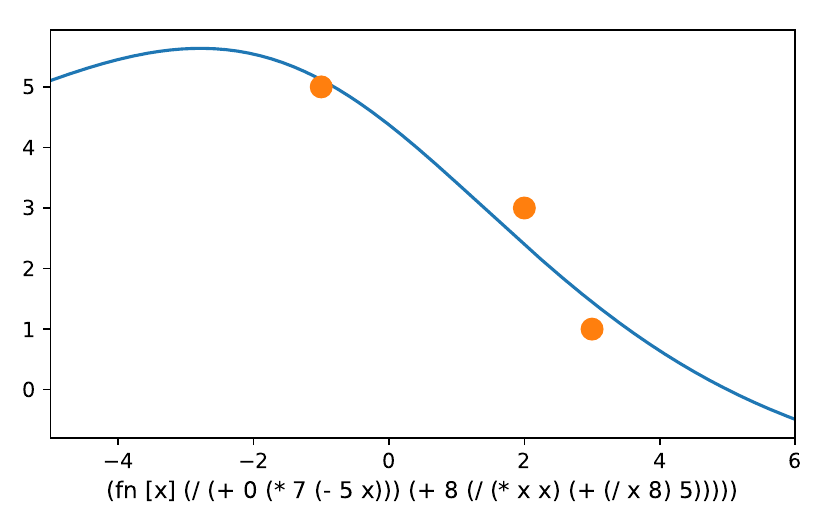}
\includegraphics[width=0.48\textwidth]{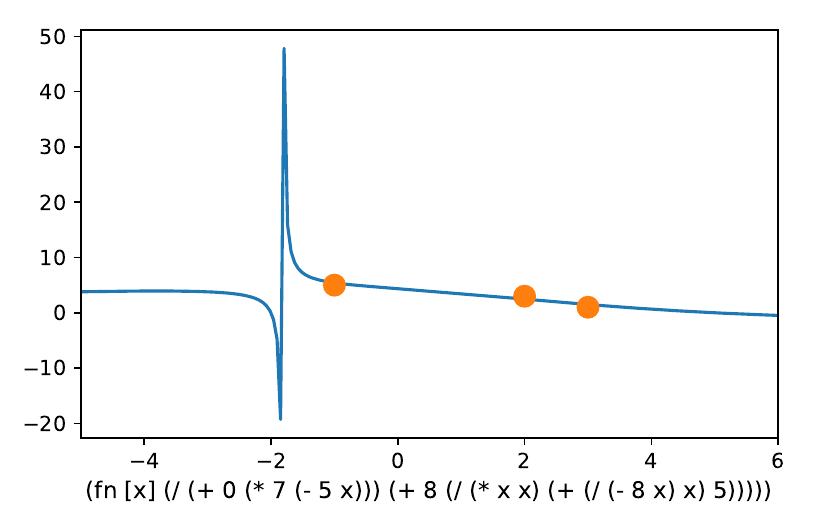}
\includegraphics[width=0.48\textwidth]{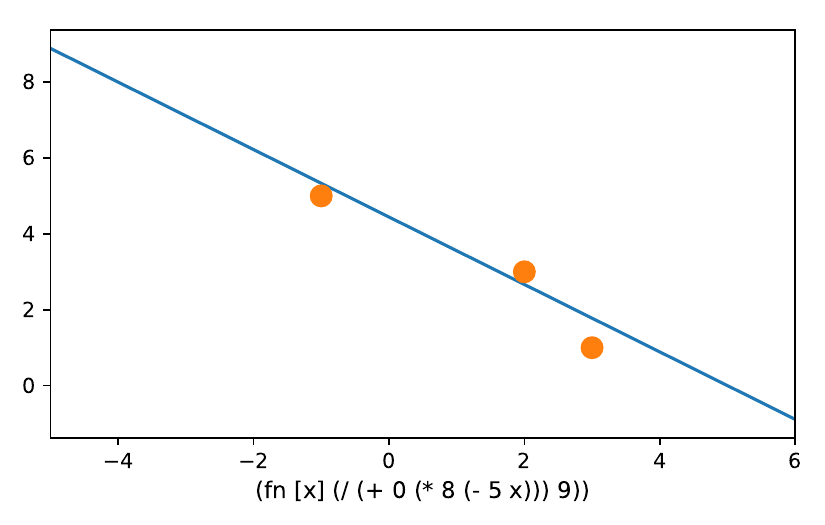}
\caption{Examples of posterior sampled functions, drawn from the same MH chain.}
\label{fig:function-induction}
\end{figure}

\subsection*{Captcha-breaking}

We can also now revisit the Captcha-breaking example we discussed in Chapter 1,
and write a generative model for Captcha images in the HOPPL.
Unlike the FOPPL, the HOPPL is a fully general programming language, and could
be used to write functions such as a Captcha renderer which produces images
similar to those in Figure~\ref{fig:captcha-posterior}.
If we write a \hop{render} function, which takes as input a string of text to encode and
a handful of parameters governing the distortion, and returns the a rendered image, it is straightforward
to then include this function in a probabilistic program that then can be used for inference.
We simply define a distribution (perhaps even uniform) over text and parameters
\begin{hoppl}[mathescape]
;; Define a function to sample a single character
(defn sample-char []
  (sample (uniform ["a" "b" $\ldots$ "z"
                    "A" "b" $\ldots$ "Z"
                    "0" "1" $\ldots$ "9"])))

;; Define a function to generate a Captcha
(defn generate-captcha [text]
  (let [char-rotation (sample (normal 0 1))
        add-distortion? (sample (flip 0.5))
        add-lines? (sample (flip 0.5))
        add-background? (sample (flip 0.4))]
    ;; Render a Captcha image
    (render text char-rotation 
            add-distortion? add-lines? add-background?)))
\end{hoppl}
and then to perform inference on the text 
\begin{hoppl}[mathescape]
(let [image ( ... ) ;; read target Captcha from disk
      num-chars (sample (poisson 4))
      text (repeatedly num-chars sample-char)
      generated (generate-captcha text)]
  ;; score using any image similarity measure
  (factor (image-similarity image generated))
  text)
\end{hoppl}
Here we treated the \hop{render} function as a black box,
and just assumed it could be called by the HOPPL program.
In fact, so long as \hop{render} has purely deterministic behavior and no side-effects it
can actually be written in another language entirely, or even be a black-box precompiled binary;
it is just necessary that it can be invoked in some manner by the HOPPL code
(e.g.\ through a foreign function interface, or some sort of inter-process communication).

\chapter{Inference Across a Messaging Interface}
\label{ch:eval-two}

Programs in the HOPPL can represent an unbounded number of random variables.
In such programs, the compilation strategies that we developed in
Chapter~\ref{ch:graph-based} will not terminate, since the program represents
a graphical model with an infinite number of nodes and vertices. In
Chapter~\ref{ch:eval-one}, we developed inference methods that generate
samples by evaluating a program. In the context of the FOPPL, the defining
difference between graph-based methods and evaluation-based methods lies in
the semantics of if forms, which are evaluated eagerly in graph-based methods
and lazily in evaluation-based methods. In this chapter, we generalize
evaluation-based inference to probabilistic programs in general-purpose
languages such as the HOPPL. A simple yet important insight behind this
strategy is that every terminating execution of an HOPPL program works on only
finitely many random variables, so that program evaluation provides a
systematic way to select a finite subset of random variables used in the
program. 



As in Chapter~\ref{ch:eval-one}, the inference algorithms in this chapter use
program evaluation as one of their core subroutines. However, to more clearly
illustrate how evaluation-based inference can be implemented by extending
existing languages, we abandon the definition of inference algorithms in terms
of evaluators in favor of a more language-agnostic formulation; we define
inference methods as non-standard schedulers of HOPPL programs. The guiding
intuition in this formulation is that the majority of operations in HOPPL
programs are deterministic and referentially transparent, with the exception
of \ang{sample} and \ang{observe}, which are stochastic and have side-effects.
In the evaluators in Chapter~\ref{ch:eval-one}, this is reflected in the fact
that only \ang{sample} and \ang{observe} expressions are algorithm specific;
all other expression forms are always evaluated in the same manner. In other
words, a probabilistic program is a computation that is mostly
inference-agnostic. The abstraction that we will employ in this chapter is
that of a program as a deterministic computation that can be interrupted at
\ang{sample} and \ang{observe} expressions. Here, the program cedes control to
an inference controller, which implements probabilistic and stochastic
operations in an algorithm-specific manner.

Representing a probabilistic program as an interruptible computation can also
improve computational efficiency. If we implement an operation that ``forks''
a computation in order to allow multiple independent evaluations, then we can
avoid unnecessary re-evaluation during inference. In the single-site
Metropolis-Hastings algorithm in Chapter~\ref{ch:eval-one}, we re-evaluate a
program in its entirety for every update, even when this update only changes
the value of a single random variable. In the sequential Monte Carlo
algorithm, the situation was even worse; we needed to re-evaluate the program
at \observe, which lead to an overall runtime that is quadratic in the number
of observations, rather than linear. As we will see, forking the computation
at
\sample and \observe expressions avoids this re-evaluation,
while this forking operation almost comes for free in languages such as the HOPPL,
in which there are no side effects outside of \sample and \observe.

\section{Explicit separation of model and inference code}
\label{sec:hoppl-message}

A primary advantage of using a higher-order probabilistic programming language
is that we can leverage existing compilers for real-world languages, rather
than writing custom evaluators and custom languages. In the interface we
consider here, we assume that a probabilistic program is a deterministic
computation that is interrupted at every \sample and \observe expression. 
Inference is carried out using a ``controller'' process. The controller
needs to be able to start executions of a program, receive the return value
when an execution terminates, and finally control
program execution at each \sample and \observe expression. 

The inference controller interacts with program executions via a messaging
protocol. When a program reaches a \sample or \observe expression, it sends a
message back to the controller and waits a response. This message will
typically include a unique identifier (i.e.~an address) for the random
variable, and a representation of the fully-evaluated arguments to $\sample$
and $\observe$. The controller then performs any operations that are necessary
for inference, and sends back a message to the running program. The message
indicates whether the program should continue execution, fork itself and
execute multiple times, or halt. In the case of \sample forms, the inference
controller must also provide a value for the random variable (when
continuing), or multiple values for the random variable (when forking).

This interface defines an abstraction boundary between program execution and
inference. From the perspective of the inference controller, the deterministic
steps in the execution of a probabilistic program can be treated as a black
box. As long as the program executions implement the messaging interface, 
inference algorithms can  be implemented in a language-agnostic manner. In
fact, it is not even necessary that the inference algorithm and the program
are implemented in the same language, or execute on the same physical machine.
We will make this idea explicit in Section\ \ref{sec:hoppl-client-server}.

\paragraph{Example: likelihood weighting.} To build intuition, we begin by
outlining how a controller could implement likelihood weighting using a
messaging interface (a precise specification will be presented in
Section~\ref{sec:hoppl-likelihood}). In the evaluation-based implementation of
likelihood weighting in Section~\ref{sec:eval-likelihood}, we evaluate \sample
expressions by drawing from the prior, and increment the log importance weight
at every \observe expression. The controller for this inference strategy would
repeat the following operations:
\begin{itemize}
  \item The controller starts a new execution of the HOPPL program,
  and initializes its log weight $\log W = 0.0$;
  \item The controller repeatedly receives messages from the running program,
  and dispatches based on type:
  \begin{itemize}
    \item At a $\mang{(sample $\ d$)}$ form, the controller samples $x$ from
    the distribution $d$ and sends the sampled value $x$ back to the program 
    to continue execution;
    \item At an $\mang{(observe $\ d$ $\ c$)}$ form, the controller increments
    $\log W$ with the log probability of $c$ under $d$, and sends a message
    to continue execution;
    \item If the program has terminated with value $c$, the
    controller stores a weighted sample $(c, \log W)$ and exits the loop.
   \end{itemize}
\end{itemize}

\paragraph{Messaging Interface.}  In the inference algorithm above, a program
pauses at every \sample and
\observe form, where it sends a message to the inference process and awaits 
a response. In likelihood weighting, the response is always to continue
execution. To support algorithms such as sequential Monte Carlo, the program
execution process will additionally need to implement a forking operation, which 
starts multiple independent processes that each resume from the same
point in the execution.

To support these operations, we will define an interface in which an
inference process can send three messages to the execution process:
\begin{itemize}
    \item[1.] 
    $(\mang{"start"}, \id)$: Start a new execution with process id $\id$.
    \item[2.] 
    $(\mang{"continue"}, \id, c)$:
    Continue execution for the process with id $\id$, using $c$ as the argument value.
    \item[3.] 
    $(\mang{"fork"}, \id, \id', c)$:
    Fork the process with id $\id$ into a new process with id $\id'$ and continue execution  with argument $c$.
    \item[4.] $(\mang{"kill"}, \id)$:
    Terminate the process with id $\id$.
\end{itemize}

Conversely, we will assume that the program execution process can send three
types of messages to the inference controller:
\begin{itemize}
    \item[1.] 
    $(\mang{"sample"}, \id, \alpha, d)$:
     The execution with id $\id$ has reached a \sample expression with address $\alpha$ and distribution $d$. 
    \item[2.] 
    $(\mang{"observe"}, \id, \alpha, d, c)$:
    The execution with id $\id$ has reached an \observe expression with address $\alpha$, distribution $d$, and value $c$. 
    \item[3.] 
    $(\mang{"return"}, \id, c)$: 
    The execution with id $\id$ has terminated with return value $c$.
\end{itemize}

\paragraph{Implementations of interruption and forking.} To implement this
interface, program execution needs to support interruption, resuming and
forking. Interruption is relatively straightforward. In the case of the HOPPL,
we will assume two primitives \ang{(send $\mu$)} and \ang{(receive $\id$)}. At
every \sample and \observe, we send a message $\mu$ to the inference process,
and then receive a response with process id $\sigma$. The call to
\ang{receive} then effectively pauses the execution until a response arrives.
We will discuss this implementation in more detail in Section~\ref{sec:hoppl-client-server}.

Support for forking can be implemented in a number of ways. In
Chapter~\ref{ch:eval-one} we wrote evaluators that could be conditioned on a
trace of random values to re-execute a program in a deterministic manner. This
strategy can also be used to implement forking; we could simply re-execute the
program from the start, conditioning on values of
\sample expressions that were already evaluated in the parent execution. As we
noted previously, this implementation is not particularly efficient, since it
requires that we re-execute the program once for every \observe in the
program, resulting a computational cost that is quadratic in the number of
\observe expressions, rather than linear.

An alternative strategy is to implement an evaluator which keeps track of the
current execution state of the machine; that is, it explicitly manages all
memory which the program is able to access, and keeps track of the current
point of execution. To interrupt a running program, we simply store the memory
state. The program can then be forked by making a (deep) copy of the saved
memory back into the interpreter, and resuming execution. The difficulty with
this implementation is that although the asymptotic performance may be better
--- since the computational cost of forking now depends on the size of the
saved memory, not the total length of program execution --- there is a large
fixed overhead cost in running an interpreted rather than compiled language,
with its explicit memory model.

In certain cases, it is possible to leverage support for process control in
the language, or even the operating system itself. An example of this is
probabilistic C \citep{paige2014compilation}, which literally uses the system call \ang{fork} to
implement forking. In the case of Turing \citep{ge2016turing}, the implementing language
(Julia) provides coroutines, which specify computations that may be interrupted and resumed later.
Turing provides a copy-on-write implementation for cloning
coroutines, which is used to support forking of a process in a manner that
avoids eagerly copying the memory state of the process.

As it turns out, forking becomes much more straightforward when we restrict
the modeling language to prohibit mutable state. In a probabilistic variant of
such a language, we have exactly two stateful operations: \sample and
\observe. All other operations are guaranteed to have no side effects.
In languages without mutable state, there is no need to copy the memory
associated with a process during forking, since a variable cannot be modified
in place once it has been defined.

In the HOPPL, we will implement support for interruption and forking of
program executions by way of a transformation to continuation-passing style
(CPS), which is a standard technique for supporting interruption of programs
in purely functional languages. This transformation is used by both Anglican,
where the underlying language Clojure uses data types which are by default
immutable, as well as by WebPPL, where the underlying Javascript language is
restricted to a purely-functional subset. Intuitively, this transformation
makes every procedure call in a program happen as the last step of its caller,
so that the program no longer needs to keep a call stack, which stores
information about each procedure call. Such stackless programs are easy to
stop and resume, because we can avoid saving and restoring their call stacks,
the usual work of any scheduler in an operating system.


In the remainder of this chapter, we will first describe two source code
transformations for the HOPPL. The first transformation is an addressing
transformation, somewhat analogous to the one that we introduced in
Section~\ref{sec:eval-mh}, which ensures that we can associate a unique
address with the messages that need to be sent at each \sample and
\observe expression. The second transformation converts the HOPPL program to
continuation passing style. Unlike the graph compiler in Chapter 3 and the
custom evaluators in Chapter 4, both these code transformations take HOPPL
programs as input and then yield output which are still HOPPL programs ---
they do not change the language. If the HOPPL has an existing efficient
compiler, we can still use that compiler on the addressed and CPS-transformed
output code. Once we have our model code transformed into this format, we show
how we can implement a thin client-server layer and use this to define HOPPL
variants of many of the evaluation-based inference algorithms from Chapter 4;
this time, without needing to write an explicit evaluator.

	\section{Addressing Transformation}
	\label{sec:addressing}

\label{sec:hoppl-addr}

An addressing transformation modifies the source code of the program to a new
program that performs the original computation whilst keeping track of an
\emph{address}: a representation of the current execution point of the
program. This address should uniquely identify any \sample and \observe
expression that can be reached in the course of an execution of a program.
Since HOPPL programs can evaluate an unbounded number of \sample and \observe
expressions, the transformation that we used to introduce addresses in
Section~\ref{sec:eval-mh} is not applicable here, since this transformation
inlines the bodies of all function applications to create an exhaustive list
of \sample and \observe statements, which may not be possible for HOPPL programs.

The most familiar notion of an address is a stack trace, which is encountered
whenever debugging a program that has prematurely terminated: the stack trace
shows not just which line of code (i.e.~lexical position) is currently being
executed, but also the nesting of function calls which brought us to that
point of execution.
In functional programming languages like the HOPPL, a stack trace effectively
provides a unique identifier for the current location in the program
execution. In particular, this allows us to associate a unique address with
each \sample and \observe expresssion at run time, rather than at compile time, 
which we can then use in our implementations of inference methods.

The addressing transformation that we present here follows the design
introduced by \citet{wingate2011lightweight}; all function calls, \sample
statements, and \observe statements are modified to take an additional
argument which provides the current address. We will use the symbol $\alpha$
to refer to the address argument, which must be a fresh variable that does
not occur anywhere else in the program. As in previous chapters, we will
describe the addressing transformation in terms of a $(e, \alpha \Downarrow_{\mathrm{addr}} e')$
relation, which translates a HOPPL expression $e$ and a variable $\alpha$ to a new expression which incorporates addresses. We additionally define
a secondary $\downarrow_{\mathrm{addr}}$ relation that operates on the top-level
HOPPL program $q$. This secondary evaluator serves to define the top-level
outer address; that is, the base of the stack trace.

\paragraph{Variables, procedure names, constants, and if.} Since addresses
track the call stack, evaluation of expressions that do not increase the depth
of the call stack leave the address unaffected. Variables $v$ and procedure
names $f$ are invariant under the addressing transformation:
\begin{align*}
\infer{
v, \alpha
\Downarrow_{\mathrm{addr}}
v
}
{}
\qquad
\infer{
f, \alpha
\Downarrow_{\mathrm{addr}}
f
}
{}
\end{align*}
Evaluation of constants similarly ignores addressing. Ground types
(e.g.~booleans or floating point numbers) are invariant, whereas primitive
procedures are transformed to accept an address argument. Since we are not
able to ``step in'' primitive procedure calls, these calls do not increase the
depth of the call stack. This means that the address argument to primitive
procedure calls can be ignored.
\begin{align*}
\infer{
c, \alpha
\Downarrow_{\mathrm{addr}}
c
}
{
c~\text{is a constant value}
}
\qquad
\infer{
c, \alpha
\Downarrow_{\mathrm{addr}}
\mang{(fn [$\alpha$\ $v_1$\ $\ldots$\ $v_n$] ($c$\ $v_1$\ $\ldots$\ $v_n$))}
}
{
c~\text{is a primitive function with $n$ arguments}
}
\end{align*}
User-defined functions are similarly updated to take an extra address
argument, which may be referenced in the function body:
\begin{align*}
\infer{
  \mang{(fn [$v_1$\ $\ldots$\ $v_n$]\ $e$)}, \alpha
  \Downarrow_{\mathrm{addr}}
  \mang{(fn [$\alpha'$\ $v_1$\ $\ldots$\ $v_n$]\ $e^\prime$)}
}{
  \begin{array}{c}
    e, \alpha
    \Downarrow_{\mathrm{addr}}
    e'
  \end{array}
}
\end{align*}
Here, the translated expression $e'$ may contain a free variable $\alpha$,
which (as noted above) must be a unique symbol that cannot occur anywhere in
the original expression $e$. 

Evaluation of \mang{if} forms also does not modify the address in our
implementation, which means that translation only requires
translation of each of the sub-expression forms.
\begin{align*}
\infer{
  \mang{(if\ $e_1$\ $e_2$\ $e_3$)}, \alpha
  \Downarrow_{\mathrm{addr}}
  \mang{(if\ $e^\prime_1$\ $e^\prime_2$\ $e^\prime_3$)}
}{
  \begin{array}{c}
    e_1, \alpha
    \Downarrow_{\mathrm{addr}}
    e'_1
    \quad
    e_2, \alpha
    \Downarrow_{\mathrm{addr}}
    e'_2
    \quad
    e_3, \alpha
    \Downarrow_{\mathrm{addr}}
    e'_3
  \end{array}
}
\end{align*}
This is not the only choice one could make for this rule, as making an address more complex is
completely fine so long as each random variable remains uniquely identifiable.  If one were to 
desire interpretable addresses one might wish to add to the address, in a manner somewhat similar to the rules that immediately follow, a value that indicates the conditional branch.  This 
could be useful for debugging or other forms of graphical model inspection.


\paragraph{Functions, sample, and observe.} So far, we have simply threaded an
address through the entire program, but this address has not been modified in
any of the expression forms above. We increase the depth of the call stack
at every function call:
\begin{align*}
\infer{
  \mang{($e_0$\ $e_1$\ $\ldots$\ $e_n$)}, \alpha
  \Downarrow_{\mathrm{addr}}
  \mang{($e^\prime_0$\ (push-addr\ $\alpha$\ $c$)\ $e^\prime_1$\ $\ldots$\ $e^\prime_n$)}
}{
  \begin{array}{c}
    e_i, \alpha
    \Downarrow_{\mathrm{addr}}
    e'_i
    ~\text{for}~i=0,\ldots,n 
    \quad
    \text{Choose a fresh value~} c 
  \end{array}
}
\end{align*}
In this rule, we begin by translating the expression $e_0$, which returns a
transformed function $e_0'$ that now accepts an address argument. This address
argument is updated to reflect that we are now nested one level deeper in the
call stack. To do so, we assume a primitive \ang{(push-addr $\alpha$ $c$)}
which creates a new address by combining the current address $\alpha$ with some
unique identifier $c$ which is generated at translation time. The translated
expression will contain a new free variable $\alpha$ since this variable is
unbound in the expression
\ang{(push-addr $\alpha$ $c$)}. We will bind $\alpha$ to a top-level address
using the $\downarrow_\mathrm{addr}$ relation.

If we take the stack trace metaphor literally, then we can
think of  $\alpha$ a list-like data structure, and of \mang{push-addr} as an
operation that appends a new unique identifier $c$ to the end of this list. Alternatively,
\mang{push-addr} could perform some sort of hash on $\alpha$ and $c$ to yield an
address of constant size regardless of recursion depth. The identifier $c$ can
be any, such as an integer counter for the number of function calls in the
program source code, or a random hash. Alternatively, if we want addresses to
be human-readable, then $c$ can be string representation of the expression
\ang{($e_0$ $e_1$ $\ldots$ $e_n$)} or its lexical position in the source
code.

The translation rules \sample and \observe can be thought of as special cases
of the rule for general function application.
\begin{align*}
\infer{
  \mang{(sample\ $e$)}
  \Downarrow_{\mathrm{addr}}
  \mang{(sample\ (push-addr\ $\alpha$\ $c$)\ $e^\prime$)}
}{
  \begin{array}{c}
    e, \alpha
    \Downarrow_{\mathrm{addr}}
    e'
    \quad
    \text{Choose a fresh value~} c 
  \end{array}
}
\end{align*}
\begin{align*}
\infer{
  \mang{(observe\ $e_1$\ $e_2$)}
  \Downarrow_{\mathrm{addr}}
  \mang{(observe\ (push-addr\ $\alpha$\ $c$)\ $e^\prime_1$\ $e^\prime_2$)}
}{
  \begin{array}{c}
    e_1, \alpha
    \Downarrow_{\mathrm{addr}}
    e'_1
    \quad
    e_2, \alpha
    \Downarrow_{\mathrm{addr}}
    e'_2
    \quad
    \text{Choose a fresh value~} c 
  \end{array}
}
\end{align*}
The result of this translation is that each \sample and \observe expression in
a program will now have a unique address associated with it. These addresses 
are constructed dynamically at run time, but are well-defined in the sense that
a \sample or \observe expression will have an address that is fully determined
by its call stack. This means that this address scheme is valid for any HOPPL
program, including programs that can instantiate an unbounded number of
variables.

\paragraph{Top-level addresses and program translation.} Translation of
function calls introduces an unbound variable $\alpha$ into the expression. To
associate a top-level address to a program execution, we define a
relation $\downarrow_\mathrm{addr}$ that translates the program body and wraps it in
a function.
\begin{align*}
\infer{
  e, \alpha
  \downarrow_{\mathrm{addr}}
  \mang{(fn [$\alpha$]\ $e^\prime$)}
}{
  \begin{array}{c}
    \text{Choose a fresh variable $\alpha$}
    \qquad
    e, \alpha
    \Downarrow_{\mathrm{addr}}
    e'
  \end{array}
}
\end{align*}
For programs which include functions that are user-defined at the top level,
this relation also inserts the additional address argument into each of the function definitions.
\begin{align*}
\infer{
  \mang{(defn\ $f$\ [$v_1$\ $\ldots$\ $v_n$]\ $e$)\ $q$}
  \downarrow_\mathrm{addr}
  \mang{(defn\ $f$\ [$\alpha$\ $v_1$\ $\ldots$\ $v_n$]\ $e^\prime$)\ $q^\prime$}
}{
  \begin{array}{c}
    \text{Choose a fresh variable $\alpha$}
    \qquad
    e, \alpha
    \Downarrow_{\mathrm{addr}}
    e'
    \qquad
    q
    \downarrow_\mathrm{addr}
    q'
  \end{array}
}
\end{align*}
These rules translate our program into an address-augmented version which is still in the same language,
up to the definitions of \sample and \observe, which are redefined to take a single additional argument.

	\section{Continuation-Passing-Style Transformation}
	\label{sec:hoppl-cps-semantics}

Now that each function call in the program has been augmented with an address
that tracks the location in the program execution, the next step is to
transform the computation in a manner that allows us to pause and resume,
potentially forking it multiple times if needed. The
continuation-passing-style (CPS) transformation is a standard method from
functional programming that achieves these goals.

A CPS transformation linearizes a computation into a sequence of stepwise
computations. Each step in this computation evaluates an expression in the
program and passes its value to a function, known as a continuation, which
carries out the remainder of the computation. We can think of the continuation
as a ``snapshot'' of an intermediate state in the computation, in the
sense that it represents both the expressions that have been evaluated so far,
and the expressions that need to be evaluated to complete the computation.

In the context of the messaging interface that we define in this chapter, a
CPS transformation is precisely what we need to implement pausing, resuming,
and forking. Once we transform a HOPPL program into CPS form, we gain access
to the continuation at every \sample and \observe expression. This continuation
can be called once to continue the computation, or multiple times to fork the
computation.

There are many ways of translating a program to continuation passing style. We
will here describe a relatively simple version of the transformation; for
better optimized CPS transformations, see \citet{appel-continuation-book}. We
define the $\Downarrow_c$ relation
\[
        e,\,\kappa,\,\state \Downarrow_c e'.
\]
Here $e$ is a HOPPL expression, and $\kappa$ is the continuation. The last
$e'$ is the result of CPS-transforming $e$ under the continuation $\kappa$. As
with other relations, we define the $\Downarrow_c$ relation by considering
each expression form separately and using inference-rules notation. As with
the addressing transformation, we then use this relation to define the CPS
transformation of program $q$, which is specified by another relation
\[
    q,\,\state \downarrow_c q'.
\]
For purposes of generality, we will incorporate an argument $\state$, which is
not normally part of a CPS transformation. 
This variable serves to store mutable state, or any information that needs to
be threaded through the computation. For example, if we wanted to implement
support for function memoization, then $\state$ would hold the memoization
tables. 

In Anglican and WebPPL, $\state$ holds any state that needs to be tracked by
the inference algorithm, and hereby plays a role analogous to that of the
variable $\sigma$ that we thread through our evaluators in
Chapter~\ref{ch:eval-two}. In the messaging interface that we define in this
chapter, all inference state is stored by the controller process. Moreover,
there is no mutable state in the HOPPL. As a result, the only state that we
need to pass to the execution is the process id, which is needed to allow an
execution to communicate its id when messaging the controller process. For
notational simplicity, we therefore use $\state$ to refer to both the CPS
state and the process id in the sections that follow.

\paragraph{Variables, Procedure Names and Constants} 
\[
    \infer{
        v,\,\kappa,\,\state \Downarrow_c \mang{(}\kappa\ \state\ v\mang{)} 
    }{}
    \qquad
    \infer{
        f,\,\kappa,\,\state \Downarrow_c \mang{(}\kappa\ \state\ f\mang{)} 
    }{}
    \qquad
    \infer{
        c,\,\kappa,\,\state \Downarrow_c \mang{(}\kappa\ \state\ \cps{c}\mang{)} 
    }{
        \textsc{cps}(c) = \cps{c}                        
    }
\]
When $e$ is a variable $v$ or a procedure name $f$, the CPS transform simply
calls the continuation on the value of the variable. The same is true for
constant values $c$ of a ground type, such as boolean values, integers and real
numbers. The case that requires special treatment is that of constant primitive
functions $c$, which need to be transformed to accept a continuation and a
state as arguments. We do so using a subroutine $\textsc{cps}(c)$, which
leaves constants of ground type invariant and transforms this primitive
functions into a procedure
\[
    \cps{c}
    =
    \textsc{cps}(c) 
    = 
    \mang{(fn$\ [v_1\ v_2\ \kappa\ \state]\ $($\kappa$\ $\state$ ($c$\ $v_1$\ $v_2$)))}.
\]
The transformed procedure accepts $\kappa$ and $\state$ as additional
arguments. When called, it evaluates the return value \ang{($c$ $v_1$ $v_2$)}
and passes this value to the continuation $\kappa$, together with the
state $\state$. For all the usual operators $c$, such as \ang{+} and \ang{*},
we represent CPS variants with $\cps{c}$, such as $\cps{\mang{+}}$
and $\cps{\mang{*}}$.

\paragraph{If Forms.} Evaluation of if forms involves two steps. First we
evaluate the predicate, and then we either evaluate the consequent or
the alternative branch. When transforming an if form to CPS, we turn this
order ``inside out'', which is to say that we first transform the consequent
and alternative branches, and then use the transformed branches to define a
transformed if expression that evaluates the predicate and selects the correct
branch
\[ 
    \infer{ 
        \mang{(if $\ e_1\ e_2\ e_3$)},\,\kappa,\,\state \Downarrow_c e'
    }{ 
        \begin{array}{ll}
            e_2,\,\kappa,\,\state \Downarrow_c e'_2 
            & 
            e_3,\,\kappa,\,\state \Downarrow_c e'_3 
            \\ 
            \mbox{Choose a fresh variable $v$}\ \  
            & 
            e_1, \mang{(fn [$\state$\ $v$] (if\ $v$\ $e^\prime_2$\ $e^\prime_3$))},\, \state \Downarrow_c e' 
        \end{array} 
    }
\]
The inference rule begins by transforming both branches $e_1$ and $e_2$ under
the continuation $\kappa$. This yields expressions $e_1'$ and $e_2'$ that pass 
the value of each branch to the continuation. Given these expressions, we then
define a new continuation \ang{(fn [$\state$ $v$] (if $v$ $e'_2$ $e'_3$))}
which accepts the value of a predicate and selects the appropriate branch. We
then use this continuation to transform the expression for the predicate
$e_1$.

This inference rule illustrates one of the basic mechanics of the CPS
transformation, which is to create continuations \emph{dynamically} during
evaluation. To see what we mean by this, let us consider the expression
\hop{(if true 1 0)}, which translates to
\begin{hoppl}
    ((fn [$\sigma$ $v$]
      (if $v$ 
       ($\kappa$ $\sigma$ 1)
       ($\kappa$ $\sigma$ 0))) $\sigma$ true)
\end{hoppl}
The CPS transformation accepts a HOPPL program and two variables $\kappa$ and
$\sigma$, and returns a HOPPL program in which $\kappa$ and $\sigma$ are free
variables. When we evaluate this program, we pass the state and the value of
the predicate to a newly created anonymous procedure that calls the
continuation $\kappa$ on the value of the appropriate branch. The important
point 
is that the CPS transformation creates the
\emph{source code} for a procedure, not a procedure itself. In other words,
the top-level continuation is not created until we evaluate the transformed
program. This property will prove essential when we define the CPS
tranformation for procedure calls.

\paragraph{Procedure Definition} To tranform an anonymous procedure, we need
to transform the procedure to accept continuation and state arguments, and
transform the procedure body to pass the return value to the continuation. We
do so using the following rule
\[ 
    \infer{ 
        \mang{(fn [$v_1\;\dots\;v_n$] $\ e$)},\,\kappa,\,\state
        \Downarrow_c
        \mang{($\kappa\ \state$\ (fn [$v_1\;\dots\;v_n\ $} \mang{ $\kappa^\prime\ \state$]  }\mang{\ $e^\prime$))} 
    }{ 
        \mbox{Choose a fresh variable $\kappa'$}
        && 
        e,\, \kappa',\,\state \Downarrow_c e'
    }
\]
We introduce a new continuation variable $\kappa'$, and transform
the procedure body $e$ recursively under this new $\kappa'$. Then, we use
the transformed body $e'$ to define a new procedure, which is passed to the
original continuation $\kappa$. Note that the original continuation expects a
procedure, not the return value of a procedure. For instance,
\[
    \mang{(fn $\ $[]$\ 1$)},\,\kappa,\,\state
    \Downarrow_c
    \mang{($\kappa\ \state\ $(fn $\ $[}\kappa'\ \state\mang{]$\ $(}\kappa'\ \state\ 1\mang{)))}
\]
The continuation parameter $\kappa'$ takes the result of the original
procedure $1$ while the current continuation $\kappa$ takes the
CPS-transformed version of the procedure itself.

\paragraph{Procedure Call} To evaluate a procedure call, we normally evaluate
each of the arguments, bind the argument values to argument variables and then
evaluate the body of the procedure. When performing the CPS transformation we
once again reverse this order
\[ 
    \infer{ 
        \mang{($e_0\ e_1 \ldots e_n$)},\, \kappa,\,\state \Downarrow_c e'_0 
    }{ 
        \begin{array}{l} 
            \mbox{Choose fresh variables $v_0,\ldots,v_n$}
            \\ 
            e_n,\, \mang{(fn [$\state\ v_n$] ($v_0\ v_1\ldots v_n\ \kappa\ \state$))},\, \state \Downarrow_c e'_n 
            \\ 
            e_i,\, \mang{(fn [$\state\ v_{i}$]}\ e'_{i+1}\mang{)},\, \state \Downarrow_c e'_i 
            \ \ \mbox{for $i = (n-1),\ldots,0$}
        \end{array} 
    }
\]
We begin by constructing a continuation 
\ang{(fn [$\state$ $v_n$] ($v_0$ $v_1$ $\ldots$ $v_n$ $\kappa$ $\state$))}
that calls a transformed procedure $v_0$ with continuation $\kappa$ and state
$\sigma$. Note that this continuation is ``incomplete'', in the sense that $v_0,
\ldots, v_{n-1}$ are unbound variables that are not passed to the
continuation. In order to bind these variables, we transform the expression,
and put the result $e_n'$ inside another expression that creates the
continuation for variable $v_{n-1}$. We continue this transformation-then-nesting
recursively until we have defined source code that creates a continuation 
\ang{(fn [$\sigma$ $v_0$] $\ldots$)}, which accepts the transformed procedure
as an argument. It is here where the ability to create continuations
dynamically, which we highlighted in our earlier discussion of if expressions, becomes
essential.

To better understand what is going on, let us consider the HOPPL expression
\ang{(+ 1 2)}. Based on the rules we defined above, we know that 1 and 2 are
invariant and that the primitive function \ang{+} will be transformed to a
procedure $\cps{\mang{+}}$ that accepts a continuation and a state as
additional arguments. The CPS tranform of \ang{(+ 1 2)} is
\begin{hoppl}
    ((fn [$\sigma$ $v_0$]
       ((fn [$\sigma$ $v_1$]
          ((fn [$\sigma$ $v_2$]
             ($v_0$ $v_1$ $v_2$ $\kappa$ $\sigma$)
           ) $\sigma$ 2)
        ) $\sigma$ 1)
     ) $\sigma$ $\cps{\text{+}}$)
\end{hoppl}
This expression may on first inspection not be the easiest to read. It is
equivalent to the following nested \ang{let} expressions, which are much
easier understand (you can check this by desugaring)
\begin{hoppl}
    (let [$\sigma$ $\sigma$
          $v_0$ $\cps{\text{+}}$]
      (let [$\sigma$ $\sigma$
            $v_1$ 1]
        (let [$\sigma$ $\sigma$
              $v_2$ 2]
          ($v_0$ $v_1$ $v_2$ $\kappa$ $\sigma$))))
\end{hoppl}
In order to highlight where continuations are defined, we can equivalently
rewrite the expression by assigning each anonymous procedure to a variable
name
\begin{hoppl}
    (let [$\kappa_0$ (fn [$\sigma$ $v_0$]
              (let [$\kappa_1$ (fn [$\sigma$ $v_1$]
                        (let [$\kappa_2$ (fn [$\sigma$ $v_2$]
                                  ($v_0$ $v_1$ $v_2$ $\kappa$ $\sigma$))]
                          ($\kappa_2$ $\sigma$ 2)))]
                ($\kappa_1$ $\sigma$ 1)))
      ($\kappa_0$ $\sigma$ $\cps{\text{+}}$))
\end{hoppl}
In this form of the expression we see clearly that we define 3 continuations
at runtime in a nested manner. The outer continuation $\kappa_0$ accepts
$\sigma$ and $\cps{\mang{+}}$. This continuation $\kappa_0$ in turn
defines a continuation $\kappa_1$, which accepts $\sigma$ and the first
argument. The continuation $\kappa_1$ defines a third continuation $\kappa_2$,
which accepts the $\sigma$ and the second argument, and calls the
CPS-transformed function.

This example illustrates how continuations record both the remainder of the
computation and variables that have been defined thus far. $\kappa_2$
references $v_0$ and $v_1$, which are in scope because $\kappa_2$ is defined
inside a call to $\kappa_1$ (where $v_1$ is defined), which is in turn defined
in a call to $\kappa_0$ (where $v_0$ is defined). In functional programming
terms, we say that the continuation $\kappa_2$ \emph{closes} over the
variables $v_0$ and $v_1$. If we want to interrupt the computation, then we
can return a tuple \ang{[$\kappa_2$ $\sigma$ $v_2$]}, rather than evaluating
the continuation call \ang{($\kappa_2$ $\sigma$ $v_2$)}. The
continuation $\kappa_2$ then effectively contains a snapshot of the variables
$v_0$ and $v_1$.



\paragraph{Observe and Sample}
\[ 
\begin{array}{c}
    \infer{ 
	\mang{(observe\ $e_\mathrm{addr}$\ $e_1$\ $e_2$)},\, \kappa,\,\state \Downarrow_c e'_\mathrm{addr}
    }{ 
        \begin{array}{l}
	    \mbox{Choose fresh variables $v_\mathrm{addr},v_1,v_2$} 
            \\ 
            e_2,\, 
		\mang{(fn [$\state\ v_2$] (}\cps{\mang{observe}}\mang{\ $v_\mathrm{addr}$\ $v_1$\ $v_2$\ $\kappa$\ $\state$))},\, \state
            \Downarrow_c 
            e'_2
            \\
            e_1,\, \mang{(fn [$\state\ v_1$]\ $e_2^\prime$}\mang{)},\, \state \Downarrow e'_1
	    \\
	    e_\mathrm{addr},\, \mang{(fn [$\state\ v_\mathrm{addr}$]\ $e_1^\prime$}\mang{)},\, \state \Downarrow e'_\mathrm{addr}
        \end{array} 
    }
    \\
    \\
    \infer{ 
	\mang{(sample\ $e_\mathrm{addr}$\ $e$)},\, \kappa,\,\state \Downarrow_c e'_\mathrm{addr} 
    }{ 
        \begin{array}{l} 
	    \mbox{Choose a fresh variable $v_\mathrm{addr},v$} 
            \\ 
	    e,\, \mang{(fn [$\state\ v$] (}\cps{\mang{sample}}\mang{\ $v_\mathrm{addr}$\ $v$\ $\kappa$\ $\state$))},\, \state \Downarrow_c e' 
	    \\
	    e_\mathrm{addr},\, \mang{(fn [$\state\ v_\mathrm{addr}$]\ $e^\prime$}\mang{)},\, \state \Downarrow e'_\mathrm{addr}
        \end{array}
    }
\end{array}
\]
These two rules are unique for the CPS transform of probabilistic programming
languages. They replace \observe and \sample operators with their CPS
equivalents $\cps{\observe}$ and $\cps{\sample}$, which take two additional
parameters $\kappa$ for the current continuation and $\state$ for the current
state. In this translation we assume that an addressing
tranformation has already been applied to add an address $e_\mathrm{addr}$ as an argument to sample and observe.

Implementing $\cps{\observe}$ and $\cps{\sample}$ corresponds to writing an
inference algorithm for probabilistic programs. When a program execution hits
one of $\cps{\observe}$ and $\cps{\sample}$ expressions, it suspends the
execution, and returns its control to an inference algorithm with information
about address $\alpha$, parameters, current continuation $\kappa$ and current
state $\state$. In the next section we will discuss how we can implement these
operations.

\paragraph{Program translation}

The CPS tranformation of expression defined so far enables the
translation of programs. It is shown in the following inference rules 
in terms of the relation $q \downarrow_c q'$, which means that
the CPS transformation of the program $q$ is $q'$:
\[
\infer{
  \mang{(fn\ [$\alpha$]\ $e$)} \downarrow_c \mang{(fn\ [$\alpha$\ $\state'$]\ $e^\prime$)}
}{ 
  \mbox{Choose fresh variables $v,\state,\state'$}
  &&
  e,\,\mang{(fn\ [$\state$\ $v$]\ (return\ $v$\ $\state$))},\,\state'
  \Downarrow_c 
  e'
}
\]
\[
\infer{
  \mang{(defn\ $f$\ [$v_1$\ $\ldots$\ $v_n$]\ $e$)\ $q$}
  \downarrow_c
  \mang{(defn\ $f$\ [$v_1$\ $\ldots$\ $v_n$\ $\kappa$\ $\state$]\ $e^\prime$)\ $q^\prime$}
}{
  \mbox{Choose fresh variables $\kappa,\state$}
  &&
  e,\,\kappa,\,\state
  \Downarrow_c
  e'
  &&
  q
  \downarrow_c
  q'
}
\]
The main difference between the CPS transformation of programs and that of
expressions is the use of the default continuation in the first rule,
which returns its inputs $v,\state$ by calling the \ang{return} function.

	\section{Message Interface Implementation}
	\label{sec:hoppl-client-server}

Now that we have inserted addresses into our programs, and transformed them into CPS,
we are ready to perform inference. To do so, we will implement the messaging interface that we outlined in Section~\ref{sec:hoppl-message}. In this interface, an inference controller process starts copies of the probabilistic program, which are interrupted at every \sample and \observe expression. Upon interruption, each program execution sends a message to the controller, which then carries out any inference operations that need to be performed. These operations can include sampling a new value, reusing a stored value, computing the log probabilies, and resampling program executions. After carrying out these operations, the controller sensds messages to the program executions to continue or fork the computation.



As we noted previously, the messaging interface creates an abstraction
boundary between the controller process and the execution process. As long as
each process can send and receive messages, it need not have knowledge of the
internals of the other process. This means that the two processes can be
implemented in different languages if need be, and can even be executed on
different machines. 

In order to clearly highlight this separation between model execution and
inference, we will implement our messaging protocol using a client-server
architecture. The server carries out program executions, and the client is the
inference process, which sends requests to the server to start, continue, and
fork processes. We assume the existence of an interface that supports
centrally-coordinated asynchronous message passing in the form of request and
response. Common networking packages such as ZeroMQ \citep{powellzeromq} 
provide abstractions for these patterns. We will also assume a mechanism for
defining and serializing messages, e.g.~protobuf \citep{protobuf}. At an
operational level, the most important requirement in this interface is that we
are able to serialize and deserialize distribution objects, which effectively
means that the inference process and the modeling language must implement the
same set of distribution primitives.

\paragraph{Messages in the Inference Controller.} In the language that
implements the inference controller (i.e.~the client), we assume the existence
of a $\textsc{send}$ method with 4 argument signatures, which we previously
introduced in Section~\ref{sec:hoppl-message}
\begin{itemize}
    \item[1.] $\textsc{send}(\mang{"start"}, \id)$: Start a new process with id \id.
    \item[2.] $\textsc{send}(\mang{"continue"}, \id, c)$: Continue process \id{} with argument $c$.
    \item[3.] $\textsc{send}(\mang{"fork"}, \id, \id', c)$: Fork process \id{} into a new process with id $\id'$ and continue execution with argument $c$.
    \item[4.] $\textsc{send}(\mang{"kill"}, \id)$: Halt execution for process \id{}.
\end{itemize}
In addition, we assume a method $\textsc{receive}$, which listens for responses from the execution server. 

\paragraph{Messages in the Execution Server.} The execution server, which runs CPS-transformed HOPPL programs, can itself entirely be implemented in the HOPPL. The execution server must be able to receive requests from the inference controller and return responses. We will assume that primitive functions \ang{receive} and \ang{send} exist for this purpose. The 3 repsonses that we defined in Section~\ref{sec:hoppl-message} were
\begin{itemize}
    \item[1.] \ang{(send "sample" $\id$ $\alpha$ $d$)}: The process $\id$ has arrived at a \sample expression with address $\alpha$ and distribution $d$.
    \item[2.] \ang{(send "observe" $\id$ $\alpha$ $d$ $c$)}: The process $\id$ has arrived at an \observe expression with address $\alpha$, distribution $d$, and value $c$.
    \item[3.] \ang{(send "return" $\id$ $c$)}: Process $\id$ has terminated with value $c$.
\end{itemize}

To implement this messaging architecture, we need to change the behavior of \sample and \observe. Remember that in the CPS transformation, we make use of CPS analogues $\cps{\sample}$ and $\cps{\observe}$. To interrupt the computation, we will provide an implementation that returns a tuple, rather than calling the continuation. Similarly, we will also implement \hop{return} to yield a tuple containing the state (i.e. the process id) and the return value
\begin{hoppl}
  (defn $\cps{\text{sample}}$ [$\id$ $\alpha$ $d$  $\kappa$] 
    ["sample" $\id$ $\alpha$ $d$ $\kappa$])

  (defn $\cps{\text{observe}}$ [$\id$ $\alpha$ $d$ $c$ $\kappa$]
    ["observe" $\id$ $\alpha$ $d$ $c$ $\kappa$])

  (defn return [$\id$ $c$]
    ["return" $\id$ $c$])
\end{hoppl}

Now we will assume that execution server reads in some program source from a file, parses the source, applies the address transformation and the cps transformation, and then evaluates the source code to create the program 
\begin{hoppl}
  (def program 
    (eval (cps-transform 
            (address-transform 
              (parse "program.hoppl")))))
\end{hoppl}

Now that this program is defined, we will implement a request handler that accepts a process table and an incoming message. 
\begin{hoppl}
  (defn handler [processes message]
    (let [request-type (first message)]
      (case request-type
        "start" (let [[$\id$] (rest message)
                      output (program default-addr $\id$)]
                  (respond processes output))
        "continue" (let [[$\id$ $c$] (rest message)
                         $\kappa$ (get processes $\id$)
                         output ($\kappa$ $\id$ $c$)]
                     (respond processes output))
        "fork" (let [[$\id$ $\id'$ $c$] (rest message)
                     $\kappa$ (get processes $\id$)
		     output ($\kappa$ $\id'$ $c$)]
		 (respond (put processes $\id'$ $\kappa$) output))
	"kill" (let [[$\id$] (rest message)] 
		 (remove processes $\id$)))))
\end{hoppl}
To process a message, the handler dispatches on the request type. For \hop{"start"},  it starts a new process by calling the compiled \ang{program}. For \hop{"kill"}, it simply deletes the continuation from the process table. For \hop{"continue"} and \hop{"fork"}, it retrieves one of continuations from the process table and continues executions. For each request type the program/continuation will return a tuple that is the output from a call to \cps{\sample}, \cps{\observe}, or \hop{return}. The handler then calls a second function 
\begin{hoppl}
  (defn respond [processes output]
    (let [response-type (first output)]
      (case response-type
        "sample" (let [[$\id$ $\alpha$ $d$ $\kappa$] (rest output)]
                    (send "sample" $\id$ $\alpha$ $d$)
                    (put processes $\id$ $\kappa$))
        "observe" (let [[$\id$ $\alpha$ $d$ $c$ $\kappa$] (rest output)]
                    (send "observe" $\id$ $\alpha$ $d$ $c$)
                    (put processes $\id$ $\kappa$))
        "return" (let [[$\id$ $c$] (rest output)]
                   (send "return" $\sigma$ $c$)
                   (remove processes $\id$)))))
\end{hoppl}
This function dispatches on the response type, sends the appropriate message, and returns a process table that is updated with the continuation if needed. Now that we have all the machinery in place, we can define the execution loop as a recursive function
\begin{hoppl}
  (defn execution-loop [processes]
    (let [processes (handler processes (receive))]
      (execution-loop processes)))
\end{hoppl}
\newcommand{\algorithmicbreak}{\textbf{break}}
\newcommand{\Break}{\State \algorithmicbreak}

\newcommand{\algorithmicerror}{\textbf{error}}
\newcommand{\Error}{\State \algorithmicerror}


	\section{Likelihood Weighting}
    \label{sec:hoppl-likelihood}

Setting up the capability to run, interrupt, resume, and fork HOPPL programs,
required a fair amount of work. However, the payoff is that we have now implemented an interface which we can use to easily write many different inference algorithms. We illustrate this benefit with a series of inference
algorithms, starting with likelihood weighting.

Algorithm~\ref{alg:hoppl-likelihood} shows an explicit definition of the
inference controller for likelihood weighting that we described high-level at
the beginning of this chapter. In this implementation, we launch $L$
executions of the program. For each execution, we define a unique process id
$\id$ using a function \textsc{newid}, and intialize the log weight to $\log
W_\id \gets 0$. We then repeatedly listen for responses. At
\hop{"sample"} interrupts, we draw a sample from the prior and continue execution. At \hop{"observe"} interrupts, we update the log weight of the process with id $\sigma$ and continue execution with the observed value. When an execution completes with a \hop{"return"} interrupt, we output the return value $c$ and the accumulated log weight $\log W_\sigma$ by calling a procedure \textsc{output}, which depending on our needs could either save to disk, print to standard output, or store the sample in some database. 

Note that this controller process is fully asynchronous. This means that if we were to implement the function \hop{execution-loop} to be multi-threaded, then we can trivially exploit the embarrassingly parallel nature of the likelihood weighting algorithm to speed up execution.



\begin{algorithm}[!t]
\caption{
\label{alg:hoppl-likelihood} Inference controller for Likelihood Weighting}
\begin{algorithmic}[1]
\Repeat
	\For {$\ell = 1, \ldots, L$}
		\Comment{Start $L$ copies of the program}
		\State $\id \gets \textsc{newid}()$
    	\State $\log W_\id \gets 0$
    	\State $\textsc{send}(\mhop{"start"},\, \id)$
	\EndFor
	\State $l \gets 0$ 
	\While {$l < L$} 
		\State $\mu \gets$ \textsc{receive}$()$
		\Switch{$\mu$}
			\Case{$(\mhop{"sample"},\, \id,\, \alpha,\, d)$}
				\State $x \gets$ \textsc{sample}$(d)$
				\State \textsc{send}$(\mhop{"continue"},\, \id,\, x)$
			\EndCase
			\Case{$(\mhop{"observe"},\, \id,\, \alpha,\, d,\, c)$}
				\State $\log W_\id \gets \log W_\id + \textsc{log-prob}(d,\, c)$
				\State \textsc{send}$(\mhop{"continue"},\, \id,\, c)$
			\EndCase
			\Case{$(\mhop{"return"},\, \id,\, c)$}
				\State $l \gets l + 1$
				\State \textsc{output}$(c,\, \log W_\id)$
			\EndCase
		\EndSwitch
	\EndWhile
\Until{forever}
\end{algorithmic}
\end{algorithm}

	\section{Metropolis-Hastings}
	\label{sec:hoppl-mh}

We next implement a single-site Metropolis-Hastings algorithm using this interface.
The full algorithm, given in Algorithm~\ref{alg:hoppl-lmh}, has an overall structure which
closely follows that of the evaluation-based algorithm for the first-order language given
in Section~\ref{sec:eval-mh}.
\begin{algorithm}[thp]
\caption{
\label{alg:hoppl-lmh} Inference Controller for Metropolis-Hastings}
\begin{algorithmic}[1]

\State $\ell \gets 0$
\Comment{Iteration counter}
\State $r, \m{X}, \log \m{P} \gets \mang{nil}, [], []$
\Comment{Current trace}
\State $\m{X}', \log \m{P}'$
\Comment{Proposal trace}
\Function{accept}{$\beta, \m{X}', \m{X}, \log \m{P}', \log \m{P}$}
\State \dots \Comment{as in Algorithm~\ref{alg:accept-lmh}}
\EndFunction
\Repeat
	\State $\ell \gets \ell + 1$
	\State $\beta \sim \textsc{uniform}(\dom(\m{X}))$
	\Comment{Choose a single address to modify}
	\State \textsc{send}(\mhop{"start"}, \textsc{newid}())
	\Repeat
		\State $\mu \gets$ \textsc{receive}$()$
		\Switch{$\mu$}
			\Case{$(\mhop{"sample"},\, \id,\, \alpha,\, d)$}
				\If{$\alpha \in \dom(\m{X}) \setminus \{ \beta \}$}
					\State $\m{X}'(\alpha) \gets \m{X}(\alpha)$
				\Else
	        				\State $\m{X}'(\alpha) \gets$ \textsc{sample}$(d)$
				\EndIf
        				\State $\log \m{P}'(\alpha) \gets$ \textsc{log-prob}$(d, \m{X}'(\alpha))$
        				\State \textsc{send}$(\mang{"continue"}, \id, \m{X}'(\alpha))$
			\EndCase
			\Case{$(\mhop{"observe"},\, \id,\, \alpha,\, d,\, c)$}
        				\State $\log \m{P}'(\alpha) \gets$ \textsc{log-prob}$(d,c)$
        				\State \textsc{send}$(\mang{"continue"}, \id, c)$
			\EndCase
			\Case{$(\mhop{"return"},\, \id,\, c)$}
				\If{$\ell = 1$}
					\State $u \gets 1$
					\Comment{Always accept first iteration}
				\Else
	    				\State $u \sim \textsc{uniform-continuous}(0,1)$
				\EndIf
	    			\If{$u < \textsc{accept}(\beta, \m{X}', \m{X}, \log \m{P}', \log \m{P})$}
		    			\State $r, \m{X}, \log \m{P} \gets c, \m{X}', \log \m{P}'$
		    		\EndIf
				\State \textsc{output}$(r, 0.0)$
				\Comment{MH samples are unweighted}
				\Break
			\EndCase
		\EndSwitch
	\Until{forever}
\Until{forever}
\end{algorithmic}
\end{algorithm}

The primary difference between this algorithm and that of
Section~\ref{sec:eval-mh} is due to the dynamic addressing. In the FOPPL, each
function is guaranteed to be called a finite number of times. This means that
we can unroll the entire computation, inlining functions, and literally
annotate every \sample and \observe that can ever be evaluated with a unique
identifier. In the HOPPL,  programs can have an unbounded number of addresses
that can be encountered, which necessitates the addressing transformation that
we defined in Section~\ref{sec:hoppl-addr}.

As in the evaluator-based implementation in Section~\ref{sec:eval-mh}, the
inference controller maintains a trace $\m{X}$ for the current sample and a
trace $\m{X}'$ for the current proposal, which track the values for \sample
form that is evaluated. We also maintain maps $\log
\m{P}$ and $\log \m{P}'$ that hold the log probability for each \sample and
\observe form. The acceptance ratio is calculated in exactly the same way as
in Algorithm~\ref{alg:accept-lmh}.

As with the implementation in Chapter~\ref{ch:eval-one}, an inefficiency
in this algorithm is that we need to re-run the entire program
when proposing a change to a single random choice. The graph-based MH sampler
from Section~\ref{sec:gibbs}, in contrast, was able to avoid re-evaluation of
expressions that do not reference the updated random variable. Recent work has
explored a number of ways to avoid this overhead. In a CPS-based
implementation, we store the continuation function at each address $\alpha$.
When proposing an update to variable $\alpha$, we know that none of the steps
in the computation that precede $\alpha$ can change. This means we can skip
re-execution of this part of the program by calling the continuation at
$\alpha$. The implementation in Anglican makes use of this optimization
\citep{tolpin2016design}. A second optimization is callsite caching
\citep{ritchie2016c3}, which memoizes return values of functions in a manner
that accounts for both the argument values and the environment that
a function closes over, allowing re-execution in the proposal to be skipped
when the arguments and environment are identical.

%
%

	\section{Sequential Monte Carlo}
	\label{sec:hoppl-smc}

\begin{algorithm}[tbp]
\caption{
\label{alg:hoppl-smc} Inference Controller for SMC}
\begin{algorithmic}[1]
\Repeat
	\State $\log \hat Z \gets 0.0$
	\For{$l$ \textbf{in} $1,\ldots,L$}
		\Comment{Start $L$ copies of the program}
		\State \textsc{send}(\mang{"start"},  \textsc{newid}())
    \EndFor    
	\State $l \gets 0$
	\Comment{Particle counter}
	\While {$l < L$}
		\State $\mu \gets$ \textsc{receive}$()$
		\Switch{$\mu$}
			\Case{$(\mhop{"sample"},\, \id,\, \alpha,\, d)$}
				\State $x \gets$ \textsc{sample}$(d)$
				\State \textsc{send}$(\mang{"continue"}, \id, x)$
			\EndCase
			\Case{$(\mhop{"observe"},\, \id,\, \alpha,\, d,\, c)$}
				\State $l \gets l+1$
				\State $\id_l, \log W_l \gets \sigma, \textsc{log-prob}(d,c)$
				\If{$l=1$}
					\State $\alpha_\text{cur} \gets \alpha$ 
					\Comment{Set address for current \observe}
				\EndIf
				\If{$l>1$}
					\State $\textbf{assert}~\alpha_\text{cur} = \alpha$
					\Comment{Ensure same address}
				\EndIf
				\If{$l=L$}
					\State $o_1, \ldots, o_L \gets$ \textsc{resample}$(W_1, \ldots, W_L)$
					\State $\log \hat{Z} \gets \log \hat{Z} + \log \frac{1}{L} \sum_{l=1}^L W_l$
					\For{$l' = 1, \ldots, L$}

						\For{$i = 1, \ldots, o_l$}
							\State \textsc{send}(\mhop{"fork"}, $\id_{l'}$, \textsc{newid}(), $c$)
						\EndFor
						\State \textsc{send}(\mhop{"kill"}, $\id_{l'}$)
					\EndFor
					\State $l \gets 0$
					\Comment{Reset particle counter}
				\EndIf
			\EndCase
			\Case{$(\mhop{"return"},\, \id,\, c)$}
				\State $l \gets l + 1$
				\State \textsc{output}$(c, \log \hat{Z})$
			\EndCase
		\EndSwitch
	\EndWhile
\Until{forever}
\end{algorithmic}
\end{algorithm}

While the previous two algorithms were very similar to those presented for the
FOPPL, running SMC in the HOPPL context is slightly more complex, though doing
so opens up significant opportunities for scaling and efficiency of inference.
We will need to take advantage of the $\mang{"fork"}$ message, and due to the
(potentially) asynchronous nature in which the HOPPL code is executed, we will
need to be careful in tracking execution ids of particular running copies of
the model program.

An inference controller for SMC is shown in Algorithm~\ref{alg:hoppl-smc}. As
in the implementation of likelihood weighting, we start $L$ executions in
parallel, and then listen for responses. When an execution reaches a sample
interrupt, we simply sample from the prior and continue execution. When one 
of the executions reaches an observe, we will need to perform a resampling step, 
but we cannot do so until all executions have arrived at the same observe.
For this reason we store the address of the current observe in a variable
$\alpha_\text{cur}$, and use a particle counter $l$ to track how many of
executions have arrived at the current observe. For each execution, we store
the process id $\sigma_l$ and the incremental log weight $\log W_l$ at the
observe. Note that, since the order in which messages are received from the
running programs is nondeterministic, the individual indices $1,\dots,L$ for
different particles do not hold any particular meaning from one observe to
the next. 

An important consideration in this algorithm, which also applies to the
implementation in Section~\ref{sec:eval-smc}, is that resampling in SMC must
happen at some sequence of interrupts that are reached in every
execution of a program. In Section~\ref{sec:eval-smc} we performed resampling
at a user-specified sequence of breakpoint addresses $y_1, \ldots, y_N$. Here,
we simply assume that the HOPPL program will always evaluate the same sequence
of observes in the same order, and throw an error when this is not the case. A
limitation of this strategy is that it cannot handle \observe forms that
appear conditionally; e.g.~\observe forms that appear inside branches of
\hop{if} forms. If we needed to support SMC inference for such programs, then
we could carry resampling at a subset of \observe forms which are guaranteed
to appear in the same order in every execution of the program. This could be
handled by manually augmenting the \observe form (and the
\mang{"observe"} message) to annotate which observes should be treated as
triggering a resample. Alternatively, one could implement an addressing scheme
in which addresses are ordered, which is to say that we can define a
comparison $\alpha < \alpha'$ that indicates whether an interrupt at address
$\alpha$ precedes an interrupt at address at $\alpha'$ during evaluation. 
When addresses are ordered, we can implement a variety of resampling policies
that generalize from SMC \citep{whiteley2016role}, such as policies that resample the
subset of executions at an address $\alpha$ once all remaining executions have
reached interrupts with addresses $\alpha' > \alpha$.

This SMC algorithm can additionally be used as a building-block for particle MCMC
algorithms \citep{andrieu2010particle}, which uses a single SMC sweep of $L$ 
particles as a block proposal in a Metropolis-Hastings algorithm.
Particle MCMC algorithms for HOPPL languages are discussed in detail in
\cite{wood2014anglican} and \citet{paige2014compilation}.

\chapter{Inference in Differentiable Models}
\label{ch:hmc}

In this book, we have so far discussed two ways of representing programs as densities. In Chapter~\ref{ch:graph-based}, we compiled probabilistic programs in the FOPPL to directed or undirected graphical models. In this representation, the probabilistic program denotes a density over a set of unobserved and observed variables that are known at compile time. In Chapter~\ref{ch:eval-two}, we considered how to perform evaluation-based inference in programs in the HOPPL by defining a messaging interface between a program execution process and an inference process. In this representation, the probabilistic program is opaque to the inference algorithm, we do not know what observed and unobserved variables the probabilistic program will instantiate, or in what order requests will arrive.

As we have seen, both strategies have advantages and disadvantages. Compiling a program to a graph makes it possible to reason about dependencies between variables, which can simplify the computation of acceptance ratios in MCMC algorithms (Section~\ref{sec:gibbs}). Reasoning about dependencies is also required when implementing message passing algorithms (Section~\ref{sec:expectation-propagation}). At the same time, language restrictions are needed to ensure that we can compile a program to a graph. Functions must be first order, and there can be no stochastic control flow. This means that we cannot express models that dynamically instantiate variables, e.g.~to determine the number of mixture components in a dataset at inference time. More generally, even when defining programs that always instantiate the same variables, using a first-order language can be somewhat inconvenient. We have to inline our data into the program, and loops must be designed in a manner that ensures that the number of iterations will be known at compile time.

By contrast, writing programs in a language like the HOPPL is often much more convenient. Almost any existing language can be turned into a probabilistic programming language by defining \hop{sample} and \hop{observe} primitives that perform callbacks to an inference backend. Using an existing language can be a big advantage, since it allows probabilistic programs to make use of deterministic functions from existing libraries, for example to perform inference in a simulation-based model. Moreover, we can even implement the inference backend in a different language from the probabilistic program. As we will see in Section~\ref{sec:pyprob-proposals}, this makes it possible to design variational inference methods that ``invert'' stochastic simulators in a low-level language like C, by learning neural proposals in a high-level language like Python. However, we have seen that implementing inference methods that communicate across this messaging interface requires careful thought, because such methods must now be designed to accommodate use cases in which the set of random variables varies from execution to execution.

In this chapter, we will consider a language design that is a middle ground between these two implementation strategies. This design is used by the Stan probabilistic programming system \citep{stan_software_2014}, which is arguably the most successful and widely used system that has been developed to date, as well as in a number of other systems, including PyMC3 \citep{salvatier2016probabilistic}, Rainier \citep{kirsch2018rainier}, and Infergo \citep{tolpin2019deployable}. In Stan, the modeling language is based on C/C++ and Matlab, and this language is higher-order. However, the language is also statically typed, and all random variables must be declared at compile time. As a result, probabilistic programs define dynamic computations with static support. This is to say that we know exactly which random variables a Stan program will instantiate at compile time, but that the density function associated with the program will be computed dynamically at run time. This strikes a middle ground between providing a convenient language for model implementations, whilst avoiding the implementational complexity associated with probabilistic programs that do not instantiate the same set of variables in each execution.

Stan combines this language design with algorithms for learning and inference that are based on gradient computations. In this Chapter, we will focus on Hamiltonian Monte Carlo (HMC; \citet{neal2011}). Other algorithms include automatic differentiation variational inference (ADVI; \citet{kucukelbir2017automatic}), a method that that makes use of reparameterized gradient estimates, which we will discuss in Section~\ref{sec:gbli}, and penalized maximum likelihood estimation with L-BFGS. In all of these algorithms, the main computational requirement is that we need to be able to calculate the gradient of the log density $\nabla_X \log p(Y, X)$ of a program. To compute the gradient of a dynamic density computation, Stan makes use of automatic differentiation (AD; see \citet{baydin2017automatic} for a recent review).

The gradient-based inference methods in Stan impose several additional requirements on programs. The main requirement is that all variables in the program must be continuous, since we cannot differentiate the density with respect to discrete variables. A secondary, more subtle requirement is that we have to be careful when writing programs with control flow, since this can introduce discontinuities in the corresponding program density. These discontinuities do not necessarily invalidate inference (as long as they have a zero measure), but they can significantly reduce inference efficiency. The language design in Stan is ``hands off'' in this respect -- it is up to the user to consider whether discontinuities can give rise to problems -- but it is also possible to design languages that explicitly account for discontinuties to facilitate optimized inference implementations \citep{pmlr-v89-zhou19b}.

In this Chapter, our goal will be to discuss the style of language design and inference implementation in Stan, and more generally in systems with similar functionality like PyMC3 \citep{salvatier2016probabilistic}, Rainier \citep{kirsch2018rainier}, and Infergo \citep{tolpin2019deployable}. As in other chapters, our description will omit certain technical details, but will be sufficiently low-level to get a sense of implementation. For a much more comprehensive discussion of the Stan probabilistic programming system specifically, we refer to the excellent Stan User's Guide and Reference Manual \citep{hoffman13}, a book in its own right.
We will here discuss inference in Stan-like languages in a manner that is notationally coherent with respect to the rest of this book. To do so, we will begin by describing how we can modify the HOPPL to ensure that programs define a density with respect to a statically determinable set of variables (Section~\ref{sec:hmc-density}). This results in a program that accepts values for both observed and unobserved programs as inputs, and in which evaluation of the program serves to compute the density as a side effect. We will then cover the basics of inference with Hamiltonian Monte Carlo methods (Section~\ref{sec:hmc-inference}) and discuss how gradient computations serve to construct high-quality proposals. In Section~\ref{sec:hmc-ad}, we discuss how reverse-mode automatic differentiation works, and how it can be implemented in the variant of the HOPPL from Section~\ref{sec:hmc-density}. We conclude with a discussion of implementation considerations in Section~\ref{sec:hmc-implementation}.

	\section{Higher-order Probabilistic Programs with Static Support}
	\label{sec:hmc-density}

Programs in Stan and similar frameworks differ in a subtle way from the programs that we have seen in this book so far. In our discussion of graph compilation techniques for FOPPL programs (Chapter~\ref{ch:graph-based}), we saw that we can compile first-order probabilistic programs to a corresponding directed graphical model, or a factor graph. This graph is a programmatic specification of a density function; it allows us to compute the density $p(Y, X)$ if we specify values for all unobserved variables $\X = [x_1 \mapsto c_1, \dots, x_n \mapsto c_n]$. In this interpretation of FOPPL programs, we evaluated if-expressions eagerly in order to ensure that a program defines a density over all variables on all flow-control paths.

In Chapter~\ref{ch:eval-one}, we discussed an equivalent interpretation of FOPPL programs as a specification of a model that we can run, which made it possible to design inference methods by implementing evaluators in which \hop{sample} and \hop{observe} expressions mutate inference state. Here if-expressions are evaluated lazily rather than eagerly, which meant that the trace $\X$ for an evaluation may not always reference the same set of random variables $X = \text{dom}(\X)$.
This same evaluation-based view of inference formed the basis for the messaging interface that we developed in Chapter~\ref{ch:eval-two}, where we replaced the evaluator for FOPPL programs with an execution process for HOPPL programs.

\subsection{Static Addressing in the HOPPL}

Stan programs can perform dynamic computations that arise from if-expressions, loops, and recursion, but cannot “create” random variables in a dynamic manner. To illustrate this idea, we will introduce a variant of the HOPPL that is syntactically very different from the Stan language, but that is, loosely speaking, equivalent in terms of programming functionality. To do so, we will make a very localized change to the HOPPL
\begin{grammar}[mathescape,caption={A statically-addressed HOPPL. This language differs from the HOPPL (Language~\ref{higher_order_prob_prog_lang}) in that it requires a static address $v$, in the form of a fresh variable name, for each sample expression.},label=shoppl]
  $v ::=$ $\textrm{variable}$
  $c ::=$ $\textrm{constant value or primitive operation}$
  $e ::=$ $c$ | $v$ | $f$ | (if $e$ $e$ $e$) | ($e$ $e_1 \ldots e_n$) | (sample $v$ $e$)
      | (observe $e_1$ $e_2$) | (fn [$v_1 \ldots v_n$] $e$)
  $q ::=$ $e$ | (defn $f$ [$v_1 \ldots v_n$] $e$) $q$.
\end{grammar}
This grammar is nearly identical to that of the HOPPL (Language~\ref{higher_order_prob_prog_lang}), but modifies the way unobserved random variables are declared. Whereas in the HOPPL, this was previously done using the form \hop{(sample $e$)}, we here require the user to provide a variable name $v$
\[
  \mhop{(sample\ $v$\ $e$)}
\]
We have previously introduced similar expression forms as part of addressing transformations for the FOPPL (Section~\ref{sec:eval-mh}) and the HOPPL (Section~\ref{sec:addressing}). In the FOPPL, the language design guaranteed that a program can evaluate a finite set of unique sample expressions, which can be enumerated and assigned unique addresses at compile time. For the HOPPL, we devised a way to construct addresses dynamically at run time, which also associated a unique address with each sample expression that could be evaluated in a program.

The design here differs from these addressing transformations in that the user must specify an address in the form of a variable name $v$. We deliberately do not allow the user to construct addresses dynamically at run time. This would be the case if we defined the form \hop{(sample $e_1$ $e_2$)}, in which any computation $e_1$ could be used to compute an address. However since we here define the form \hop{(sample $v$ $e$)}, the address must be inlined as a variable name into the source code of the program. This means that, by construction, a program can only instantiate a finite set of unique unobserved variables. This set of variables is statically determinable, since it is just the set free variables in the program, i.e.~any variables that are not bound when they are first referenced.

\begin{hoppl}[caption={Bayesian linear regression in the statically-addressed HOPPL}, label={prog:regr-hoppl}]
(let [data (read-csv "data.csv")
      y-values (map first data)
      x-values (map second data)
      a (sample a (normal 0 1))
      b (sample b (normal 0 1))]
  (map (fn [y x]
         (let [fx (+ (* a x) b)]
           (observe (normal fx 0.1) y)))
       y-values x-values)
  [a b])
\end{hoppl}
This style of programming defines a middle ground between writing programs in a FOPPL and writing programs in a HOPPL. Programs in the statically-addressed HOPPL do not need to inline data and can make use of higher-order functions like \hop{map} to loop over data points. At the same time, explicit addressing ensures that the variable identifiers $X$ (here \hop{a} and \hop{b}) are inlined into the program, rather than computed dynamically at run time. This simplifies many aspects of the corresponding inference implementation, since we no longer need to account for cases where the set of variables may vary from execution to execution.


\subsection{Computing the Unnormalized Density as a Side Effect}
\label{sec:hmc-density-translation}

Like all probabilistic programs in this book, programs in the statically-addressed HOPPL defines a target density $\pi(X) = \gamma(X) / Z$ in terms of an unnormalized density $\gamma(X)$. As we discussed in Section~\ref{sec:eval-dens}, $\gamma(X) = p(Y \mid X) \: p(X)$ can be a joint density over observed and unobserved variables, or a density $\gamma(X) = \psi(X) \: p(X)$ in which the likelihood is replaced by an arbitrary factor. However, the static addressing that we have introduced here does come with some caveats. 

To understand these caveats, we will begin by clarifying what density a program with static addressing denotes. In reference to terminology that we will use in our discussion of Hamiltonian Monte Carlo methods in the next section, we will define the \emph{potential energy} function of a program as its negative logarithm of the unnormalized density
\begin{align}
    U(X)
    =
    - \log \gamma(X)
    .
\end{align}
To compute this energy, we will define a translation of a statically-addressed HOPPL expression $e$ into a function that accepts a the free variables $X$ as inputs and computes both the return value \hop{v} of a program and its potential energy \hop{U}\footnote{To streamline our discussion, we will only consider expressions $e$, and not programs $q$ that can also contain function definitions, but this does not fundamentally change the set of programs that can be expressed in the language.} as a side effect. As we have discussed previously in this book, sample, observe, and factor expressions typically have side effects in the inference computation. In this translated function, we will use side effects to compute the potential energy \hop{U}.

We have previously either tracked side effects by threading an inference state $\sigma$ through the evaluator (Chapter~\ref{ch:eval-one}), or by defining a stateful inference process that communicates with the execution process (Chapter~\ref{ch:eval-two}).
In this Section, we will be a little more informal. We will assume that we will translate the expression $e$ of the original program into a non-probabilistic language that supports mutable state. For this purpose, we will assume the existence of three additional expression forms
\begin{align*}
  &\mhop{(mutable\ $c$)} &
  &\mhop{(set\!\ $v$\ $e$)} &
  &\mhop{(immutable\ $v$)}
\end{align*}

\begin{hoppl}[mathescape,caption={The embedding of an expression $e$ into a function that accepts a set of inputs $X$ that correspond to the free variables in the expression $e$. The function computes the return value \hop{v} as well as the potential energy \hop{U}.},float=t, floatplacement=t, label={prog:sahoppl-energy}]
(fn [$x_1$ $\dots$ $x_n$]
   (let [U (mutable 0.0)
         sample (fn [$x$ $d$]
                  (let [logpx (log-prob $d$ $x$)]
                    (set! U (- U logpx))
                    $x$))
         observe (fn [$d$ $y$]
                   (let [logpy (log-prob $d$ $y$)]
                     (set! U (- U logpy))
                     $y$))
         factor (fn [$c$]
                  (set! U (- U $c$)))]
     (let [v $e$]
       [v (immutable U)])))
\end{hoppl}
With these language constructs in place, we define the translation of an expression $e$ in Program~\ref{prog:sahoppl-energy}. This program defines a mutable variable \hop{U}, along with normal functions \hop{sample}, \hop{observe}, and \hop{factor} that have the following side effects:
\begin{itemize}
  \item For each call \hop{(sample $x$ $d$)}, the variable $x$
   will either a free variable in $X=(x_1, \dots, x_n)$, or a bound variable. In both cases, we decrease the potential by $\textsc{log-prob}(d, x)$ and return $x$.\footnote{This contruction implies that \mhop{(let [x\ $c$] (sample\ $x$\ $d$))} is equivalent to \mhop{(observe\ $d$\ $c$)} in terms of the potential energy that a program computes.}

  \item For each call \hop{(observe $d$ $y$)}, we decrease the potential energy by $\textsc{log-prob}(d, y)$ and return $y$.

  \item For each call \hop{(factor $c$)} we decrease the potential energy by $c$ and return nothing.
\end{itemize}
After defining these functions, we evaluate the program expression $e$ to compute the return value \hop{v}, which will now compute \hop{U} as a side effect. We then return \hop{v} and the dereferenced value of \hop{U}.

\subsection{Densities with Control Flow and Non-Unique Variables}

The translation in Program~\ref{prog:sahoppl-energy} transforms a program in the statically-addressed HOPPL into a completely deterministic function that accepts the free variables $X$ as inputs and computes the return value \hop{v} and the potential energy \hop{U}, i.e.~the negative unnormalized log density. In the next section, we will discuss how to use this density to perform inference with Hamiltonian Monte Carlo methods. However, before we do so, we will briefly reflect on some of the more counterintuitive aspects of this statically-addressed language definition and of the densities that programs in this language denote. 

One aspect of the language definition that requires special attention is how we deal with sample expressions inside control flow. As a particularly pathological example, let us consider the program
\begin{hoppl}[caption={A program that always returns 0.}, label={prog:false}]
(if false
  (sample x (normal 0 1))
  0)
\end{hoppl}
In the FOPPL, we would either evaluate this if-expression eagerly or lazily. In an eager evaluation model, the return value of the program is 0, and the density defines a Gaussian prior over a single random variable \hop{x}. In a lazy evaluation model, this program would return 0 and an empty trace $\X = []$, which implies that its density is 1. As we discussed in Section~\ref{sec:compilation-bounded}, both interpretations are equivalent in terms of the posterior marginal on return values that the program denotes.

In the statically-addressed HOPPL, the interpretation of this program is a bit more problematic. If we provide the translated function with a trace $\X = [\mhop{x} \mapsto c_x]$, the program will return a value \hop{v} that is 0, and an energy \hop{U} that is also 0, since the value for the variable \hop{x} will never be referenced. This means that this program defines an improper constant density $\gamma(X)=1$.

This behavior is not hypothetical; the corresponding Stan program would have the same semantics. The thing to remember in this programming model is that ``sample'' does not really mean ``define a new random variable that is distributed according to the specified density''. Instead, it really means ``increment the log density'', or equivalently ``decrease the potential energy''.

This brings us to a second aspect of this language design, which is whether static addresses should be unique. As an example, let us consider the following modification of the regression example
\begin{hoppl}[caption={A program with non-unique addresses}, label={prog:non-unique}]
(let [data (read-csv "data.csv")
      y-values (map first data)
      x-values (map second data)]
  (map (fn [y x]
         (let [a (sample a (normal 0 1))
               b (sample b (normal 0 1))
               fx (+ (* a x) b)]
           (observe (normal fx 0.1) y)))
       y-values x-values)
  [a b])
\end{hoppl}
In this somewhat nonsensical example, we repeatedly ``sample'' the variables \hop{a} and \hop{b} when looping over data points. It is ambiguous whether we should consider this example a valid program. We could argue that programs should evaluate exactly one sample expression for each unique address. If this is the case, then we should probably also disallow the pathological example in Program~\ref{prog:false} earlier, since that example never evaluates the sample expression for the variable \hop{x} on the unreachable branch of the conditional.

In Stan, a program analogous to Program~\ref{prog:non-unique} would once again be valid, and it has the semantics that we would expect based on our computation for the potential energy that we defined in Program~\ref{prog:sahoppl-energy}. When we supply the resulting function with a trace $\X = [\mhop{a} \mapsto c_a, \mhop{b} \mapsto c_b]$, repeated evaluation will return the same stored values $\X(\mhop{a})$ and $\X(\mhop{b})$ at each iteration, but will also repeatedly update the potential energy \hop{U} as a side effect. This means that, when provided with $N$ data points will define an unnormalized density
\begin{align}
    \gamma(a, b) = p(a)^N p(b)^N \prod_{n=1}^N p(y_n \mid a, b; x_n).
\end{align}
Again, we need to interpret the sample form not so much as the instantiation of a new random variable, but rather as a command that adds a term to the log density, or equivalently subtracts a term from the potential energy.

Despite these caveats, this programming model is relatively easy to use. Probabilistic programs in Stan-like languages have a well-defined prior whenever the dynamic density computation evaluates exactly one sample expression for each random variable, which will be true for the vast majority of programs that a user will write in practice.

	\section{Hamiltonian Monte Carlo}
	\label{sec:hmc-inference}

In Section~\ref{sec:hmc-density} we have described how to compute the potential energy function associated with a statically-addressed HOPPL program, which then defines a target density $\pi(X)$ of the form
\begin{align}
    \label{eq:hmc-target}
    \pi(X) = \frac{1}{Z} \exp \left\{ - U(X) \right\}.
\end{align}
In Program~\ref{prog:sahoppl-energy}, we defined a concrete computation that embeds a program expression $e$ into a function that computes the energy $U(X)$ as a side effect. Given this computation, it becomes possible to implement any number of inference methods. We could perform importance sampling by defining a proposal density $q(X)$, which could in principle be any program without observe expressions that denotes a density over the same set of variables as the target density. We could also implement MH methods by designing one or more transition kernels $q(X' \mid X)$. In practice however, this style of programming is often combined with methods that rely on gradient computations, and Hamiltonian Monte Carlo methods are particularly widely used in this context.

\paragraph{Hamiltonian Dynamics} HMC is an MCMC algorithm that makes use of gradients to construct an efficient transition kernel, particularly for high-dimensional densities. HMC is applicable to any target density $\pi(X)$ of the general form defined in Equation~\eqref{eq:hmc-target}, in which each variable $x \in X$ must be continuous. To construct proposals, HMC methods make use of an analogy to classical mechanics, which we will discuss at a high level before we define more formally how proposals are constructed. In this analogy, we interpret $X$ as the position of a ``particle'' (technically a point mass) and the function $U(X)$ as an energy ``landscape''. In this landscape, ``wells'' are regions of low energy $U(X)$, which will correspond to peaks in the target density $\pi(X)$. An MCMC sampler defines a biased random walk through this landscape, in which regions of lower energy (i.e.~higher density) are visited more frequently.

To construct such a biased random walk, HMC simulates the trajectory of a particle in this energy landscape. For this purpose, it introduces a momentum variable $R$ and a corresponding kinetic energy $K(R)= \frac{1}{2} R^\top M^{-1} R$, in which the matrix $M$ is known as the ``mass matrix'' of the particle. In classical mechanics (as in all of physics), energy must be conserved. This means that kinetic energy $K(R)$ can be converted into potential energy $U(X)$ and vice versa, but that the sum of these two quantities, which is is known as the Hamiltonian
\begin{align}
   \label{eq:hmc-dHdt}
    \frac{d}{dt}H
  & = (\nabla_R H)^\top \frac{dR}{dt} + (\nabla_X H)^\top \frac{dX}{dt}
  = 0.
\end{align}
The solution to this equation is to define the equations of motion that define the so-called Hamiltonian Dynamics\footnote{We can verify this solution by subtituting these identities into Equation~\ref{eq:hmc-dHdt}, which yields $dH/dt = - (\nabla_R H)^\top (\nabla_X H) + (\nabla_X H)^\top \nabla_R H = 0$}
\begin{align}
  \label{eq:hmc-eq-motion}
  \begin{split}
    \frac{dX}{dt} &= \nabla_R H(X,R) = M^{-1} R,  \\
    \frac{dR}{dt} &= -\nabla_X H(X,R) = - \nabla_X U(X).
  \end{split}
\end{align}

\paragraph{MCMC with Hamiltonian Dynamics} HMC methods approximate the equations of motion in Hamiltonian Dynamics to define an MCMC transition operator. To do so, HMC methods use an \emph{auxiliary variable} construction. Rather than sampling from the density $\pi(X)$, HMC generates samples from a \emph{extended} target density on $X$ and $R$
\begin{align}
  \begin{split}
  \tilde{\pi}(X, R)
  &=
  \frac{1}{\tilde{Z}}
  \exp \big\{ -H(X, R)\big\},
  \\
  &=
  \frac{1}{\tilde{Z}}
  \exp
  \left\{
    - U(X)
    - \frac{1}{2} R^\top M^{-1} R
  \right\}.
  \end{split}
\end{align}
In the extended target density, $R$ is known as an \emph{auxiliary} variable. The density on the extended space is defined in such a manner that the marginal of the extended density $\tilde{\pi}(X) = \pi(X)$ is equal to the original target density. To see that this property indeed holds in the case of the construction above, we note that we can factorize $\pi(X, R)$ into a product of unnormalized densities
\begin{align}
  \tilde{\pi}(X, R)
  &=
  \frac{1}{\tilde{Z}}
  \: \gamma(X) \: \gamma(R),
  &
  \gamma(X)
  &=
  \exp \big\{ -U(X) \big\}.
\end{align}
In other words, we see that $X$ and $R$ are uncorrelated in the density $\pi(X, R)$ and that the unnormalized marginal $\gamma(X)$ corresponds precisely to the unnormalized density of the original model. This means that if we sample $X, R \sim \tilde{\pi}(X, R)$ and simply discard $R$, then $X$ is a sample from the marginal $\tilde{\pi}(X) = \pi(X)$. In short, the introduction of the auxiliary variables $R$ does not change the distribution over $X$ in the sampler.

It is not immediately obvious why introducing the auxiliary variable $R$ would make it easier to generate samples, rather than more difficult, since this increases the dimensionality of the sampling problem by a factor 2. As it turns out, we can use the auxiliary variables $R$ to propose large moves for $X$ whilst at the same time ensuring that these moves have a high acceptance probability. 

To see why this is the case, let us consider the MH acceptance ratio for a proposal $X', R' \sim q(X',R' \mid X, R)$,
\begin{align}
  \alpha
  =
  \min
  \left\{
  1,
  \frac{\tilde{\gamma}(X', R') \: q(X, R \mid X', R')}
       {\tilde{\gamma}(X, R) \: q(X', R' \mid X, R)}
  \right\}
  .
\end{align}
For any reversible proposal, i.e.~a proposal for which
\begin{align}
   q(X',R' \mid X, R) = q(X,R \mid X',R'),
\end{align}
the acceptance ratio simplifies to
\begin{align}
  \label{eq:hmc-accept}
  \alpha
  =
  \min
  \left\{
  1,
  \frac{\exp \big\{H(X', R')\}}
       {\exp \big\{H(X, R)\}}
  \right\}
  .
\end{align}
The essential idea in HMC is that if we can design a proposal that is reversible and that also preserves the Hamiltonian, then this proposal can be accepted with probability 1. HMC uses this intuition to define an MCMC sampler in which moves preserve the Hamiltonian:
\begin{itemize}
\item Initialize $X^0$.
\item For sampler iteration $s=1,\dots,S$ perform two updates
\begin{itemize}
  \item[1.] Perform a Gibbs update to sample the initial momentum $R_0 \sim \tilde{\pi}(R \mid X^{s-1})$. The density $\tilde{\pi}(R \mid X) = \tilde{\pi}(R)$, is just a multivariate normal, since
  \[
    \gamma(R)
    = \exp \left\{
        - \frac{1}{2} R^\top M^{-1} R
      \right\}
    \propto \text{Normal}(R ; 0, M).
  \]
  Here $M$ can be tuned to optimize the acceptance ratio of the HMC method (see Section~\ref{sec:hmc-implementation}). 

  \item[2.] Perform an MH update using Hamiltonian Dynamics. Define $X_0=X^{s-1}$ and use numerical integration of Hamilton's Equations~\eqref{eq:hmc-eq-motion} to compute a discretized trajectory
  \[
      \big((X_0,R_0), \dots, (X_T, R_T)\big)
  \]
  Define $X'=X_T$ and $R'=-R_T$ to compute the acceptance ratio $\alpha$ (Equation~\ref{eq:hmc-accept}). With probability $\alpha$, accept and define $X^s=X'$ and $R^s=R'$. With probability $1-\alpha$, reject and define $X^s=X^{s-1}$ and $R^s=R_0$
\end{itemize}
\end{itemize}
The proposal mechanism in step 2 is reversible by virtue of the fact that we define $R'=-R_T$. Since Hamilton's equations are reversible, this means that integration starting from $X_T$ and $-R_T$ will return us to $X_0$ and $R_0$. In other words, as long as numerical integration preserves the Hamiltonian, the proposal will be accepted with probability 1. When the Hamiltonian is not preserved, the MH acceptance step serves to correct for integration errors.

We will describe how to perform the numerical integration in Section~\ref{sec:hmc-implementation}. However before we do so, we will discuss how to compute the derivatives in Hamilton's Equations~\eqref{eq:hmc-eq-motion}, which will require an implementation of automatic differentiation.

	\section{Automatic Differentiation}
	\label{sec:hmc-ad}


The essential operation that we need to implement to perform numerical integration of Hamilton's Equations~\eqref{eq:hmc-eq-motion} is computation of the gradient of the potential function
\begin{align}
  \nabla U(X) = -\nabla \log \gamma(X).
\end{align}

To compute the partial derivatives of the potential function, we will use reverse-mode automatic differentiation (\citet{griewank2008evaluating}; see \citet{baydin2015automatic} for a recent introduction). Reverse-mode AD is a technique for computing the gradient of a scalar output of a program with respect to all scalar inputs. It can be implemented in most languages to define differentiable programs. Besides forming the basis for inference methods in Stan, it also forms the basis for loss minimization techniques based on stochastic gradient descent in modern deep learning systems and related deep probabilistic programming systems, which we will discuss in Chapter~\ref{ch:deep-probprog}.

\begin{figure}[!t]
\begin{center}
  \includegraphics[width=0.9\textwidth]{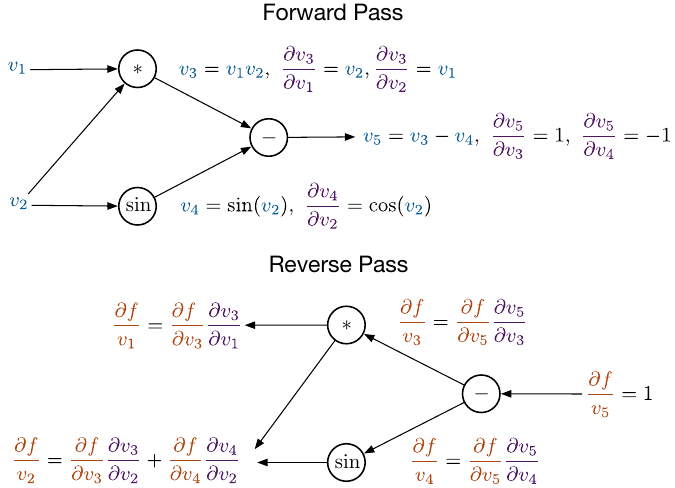}
\end{center}
\caption{
\label{fig:ad-overview}
Reverse-mode AD for the function $f(v_1, v_2) = v_1 v_2 - \sin(v_2)$. Evaluation of the gradient is performed in two steps. The forward pass constructs a computation graph in which nodes correspond to intermediate values $v_3$, $v_4$, and the output $v_5$. For each node, we compute the partial derivatives of the value with respect to its inputs. The reverse pass computes the derivatives of the ouput $f(v_1, v_2) = v_5$ with respect to $v_1$ and $v_2$ by walking the constructed computation graph in the reverse direction to propogate derivatives from each node to its inputs.}
\end{figure}

To illustrate how reverse-mode AD works, we will begin by considering a few simple examples. To begin, let us consider the function
\begin{hoppl}
  (fn [$v_1$ $v_2$]
   (- (* $v_1$ $v_2$) (sin $v_2$)))
\end{hoppl}
We illustrate how to perform reverse-mode AD for this function in Figure~\ref{fig:ad-overview}. Reverse-mode AD is a computation in two stages. During the \emph{forward} pass we construct a \emph{computation graph}. In this computation graph, each node coresponds to a primitive procedure call in the function body, and edges denote inputs to this computation. For each node, we compute the output value of the procedure call, as well as the partial derivative of the output with respect to each of the inputs. During the \emph{backward} pass, we walk the computation graph in the reverse direction starting at the output value. At each node, we apply the chain rule of differentiation to propagate derivatives to parent nodes, which in the deep learning community is often referred to as \emph{backpropagation}.

\begin{hoppl}[label={prog:regr-embedded}, caption={Embedding of the Bayesian linear regression model from Program~\ref{prog:regr-hoppl} using the translation in Program~\ref{prog:sahoppl-energy}. The inputs are the free variables \hop{a} and \hop{b}.  To implement HMC, we need to compute the gradient $\nabla_X U(X)$ with respect to the input values of \hop{a} and \hop{b}.}, float={!t}, floatplacement={t}]
(fn [a b]
   (let [U (mutable 0.0)
         sample (fn [x d]
                  (let [logpx (log-prob d x)]
                    (set! U (- U logpx))
                    x))
         observe (fn [d y]
                   (let [logpy (log-prob d y)]
                     (set! U (- U logpy))
                     y))
         factor (fn [c]
                  (set! U (- U c)))]
     (let [v (let [data (read-csv "data.csv")
                   y-values (map first data)
                   x-values (map second data)
                   a (sample a (normal 0 1))
                   b (sample b (normal 0 1))]
               (map (fn [point]
                      (let [y (first point)
                            x (second point)
                            fx (+ (* a x) b)]
                        (observe (normal fx 0.1) y)))
                    data)
               [a b])]
       [v (immutable U)])))
\end{hoppl}

The example above is easy to understand because the function body is a single expression, in which each sub-expression corresponds to a primitive function call. In practice, we will apply automatic differentiation to programs that can be considerably more complicated. As an example, let us consider the embedding of the Bayesian linear regression example (Program~\ref{prog:regr-hoppl}), which we show in Program~\ref{prog:regr-embedded}. This example makes use of data structures (i.e.~vectors and hash maps), control flow (i.e.~the use of \hop{map} to loop over data), disk access (i.e.~the function \hop{read-csv} that reads in the data), and operations that mutate the variable \hop{U} to compute the potential energy.

However, despite this additional apparent complexity, automatic differentiation for this program can be performed using the same forward and reverse sweep. From the perspective of an AD implementation, the output \hop{U} is still a scalar variable that is computed by way of a series of primitive function calls. To make this connection concrete, suppose we read in a dataset $\big((y_1, x_1), \dots, (y_N, x_N)\big)$. The computation for the potential will update \hop{U} when evaluating the sample expressions for \hop{a} and \hop{b} and then evaluate $N$ observe expressions for each of the data points. If we write out the sequence of operations that mutate \hop{U}, then we see that the full computation for the potential has the form
\begin{hoppl}
  (- ... (- (- (- 0.0
                 (log-prob (normal 0 1) $c_a$))
               (log-prob (normal 0 1) $c_b$))
            (log-prob (normal (+ (* $c_a$ $x_1$) $c_b$) 0.1))
         ...
     (log-prob (normal (+ (* $c_a$ $x_N$) $c_b$) 0.1)
\end{hoppl}
Intuitively, any scalar value that is computed by a program must be  returned by some primitive function in the language, even if this primitive is called inside a let expression, if expression, or a user-defined procedure body. Moreover, any scalar inputs to this primitive call must either themselves be returned by some primitive, or must be root nodes in the computation, which are either constants or free variables.

Based on this intuition, we see that we can essentially ``ignore'' all language constructs other than primitive procedure calls when performing automatic differentiation. To implement the computations that are needed for the forward pass, all we need to do is evaluate the program as normal, whilst ensuring that primitive functions construct a computation graph as a side effect. This means that primitive functions need to perform the following additional operations:
\begin{itemize}
  \item Each time we call a primitive during evaluation, we need to create a new node in the computation graph for each scalar output.
  \item For each new node, we need to compute and store the derivatives of the output with respect to each scalar input.
  \item For each new node, we need to additionally create an edge from the nodes for each of scalar input to each scalar output.
\end{itemize}
If we now define each input variable as a root node in the computation graph at the start of evaluation, then normal program evaluation will build up a graph of the form in Figure~\ref{fig:ad-overview}. In other words, as long as we can design a library of AD-compatible primitives that implement these operations as a side effect, then we can turn any program into a differentiable program by simply rebinding the standard primitives in the language to their AD-compatible counterparts.

At the level of implementation, construction of the computation graph is often done using mutable state. In deep learning frameworks, which are commonly embedded into high-level languages like Python, the AD backend is often implemented in a lower-level language like C, which makes extensive use of mutable state. Here we will sketch out a functionally pure implementation, in which primitives return a boxed value $\ad{c}$ that contains the computation graph for the returned value $c$, rather than mutating a single global computation graph for all values.

\begin{algorithm}[!t]
\caption{
  \label{alg:hmc-lift-ad}
  Primitive function lifting for reverse-mode AD.
}
\begin{algorithmic}[1]
\Function{unbox}{$\ad{v}$}
  \State $c,\_,\_,\_ \gets \ad{v}$
  \State \Return $c$
\EndFunction
\Function{lift-ad}{$f$, $\nabla f$, $n$}
   \Function{$\tilde{f}$}{$\ad{c}_1, \ldots, \ad{c}_n$}
      \State $c_1, \ldots, c_n \gets \Call{unbox}{\ad{c}_1}, \ldots, \Call{unbox}{\ad{c}_n}$
      \State $c \gets f(c_1, \ldots, c_n)$
      \State $v \gets \textsc{fresh-variable}()$
      \State $\dot{c}_1, \ldots, \dot{c}_n \gets \nabla f(c_1, \ldots, c_n)$
      \State \Return $\big(c,v,(\ad{c}_1, \ldots, \ad{c}_n),(\dot{c}_1, \ldots, \dot{c}_n)\big)$
   \EndFunction
   \State \Return $\tilde{f}$
\EndFunction
\end{algorithmic}
\end{algorithm}

Algorithm~\ref{alg:hmc-lift-ad} shows pseudo-code for an operation that constructs an AD-compatible primitive $\tilde{f}$ from a primitive $f$. We will for ease of discussion assume that $f$ returns a single scalar value $c$, and that all inputs are also scalar values. The function \textsc{lift-ad} accepts the primitive $f$, along with second primitive function $\nabla f$ that computes the derivatives of the output with respect to each of the inputs, and the number of arguments $n$ to the primitive and its gradient. It returns a data structure $\ad{c}$ that has the following form
\begin{align}
  \ad{c}
  =
  \big(
     c,
     v,
     (\ad{c}_1, \ldots, \ad{c}_n),
     (\dot{c}_1, \ldots, \dot{c}_n)
   \big).
\end{align}
Here $c$ is the output of $f$, $v$ is a unique variable name, $(\ad{c}_1, \dots, \ad{c}_n)$ are boxed data structures for each of the inputs, and $(\dot{c}_1, \dots, \dot{c}_n)$ is the output of the gradient function $\nabla f$, which is to say that each $\dot{c}_i$ is the derivative of the output with respect to input $i$,
\begin{align}
  \dot{c}_i
  =
  \left.
  \frac{\partial f(v_1, \ldots, v_n)}
       {\partial v_i}
  \right\vert_{v_1=c_1,\ldots,v_n=c_n}
  .
\end{align}

The recursive definition of $\ad{c}$ ensures that each of the input values $\ad{c}_i$, which are also boxed data structures, which can itself contain input values and derivatives. This means that we can walk the computation graph in the reverse direction by recursively unpacking a boxed values $\ad{c}$. There are two base cases in this recursion, which correspond to either constant literals $c$ in the program body $e$, or input variables $x_j$ of the program, which we define boxed values with no inputs or gradients
\begin{align}
  \ad{c} &= \big(c,v,(),()\big),
  &
  \tilde{x}_j &= \big(c_j,x_j,(),()\big).
\end{align}
 Walking the graph allows us to implement reverse-mode AD as follows:
\begin{itemize}
\item Forward pass: Map inputs $x_j$ onto boxed values $\ad{x}_j$. Evaluate the program using lifted primitives $\tilde{f}$ to compute an output $\ad{c}$.
\item Initialize the gradient $\m{G}=[x_1 \mapsto 0, \dots, x_m \mapsto 0]$ and define the derivative of the output with respect to the current node as $g=1$.
\item Walk backwards along computation graph in depth-first order. Evaluate nodes using the following recursion:
\begin{itemize}
  \item[1.] If the variable $v$ of the node is an input variable $x_j$, update the gradient component $\m{G}(x_j) \gets \m{G}(x_j) + g$ using the derivative along the current path.
  \item[2.] When $v$ is not an input node, recursively evaluate each parent node $\ad{c}_i$ with derivative $g \cdot \dot{c}_i$. This covers constant literals, which have no parents, as a special case.
\end{itemize}
\end{itemize}

\begin{algorithm}[!t]
\caption{
  \label{alg:backward}
  The backward recursion in reverse-mode AD.
}
\begin{algorithmic}[1]
\Function{backward}{$\ad{c}_0$, $g_0$, $\m{G}$}
\State $c_0,v_0,(\ad{c}_1, \dots, \ad{c}_n),(\dot{c}_1,\dots,\dot{c}_n) \gets \tilde{c}_0$
\If{$v_0 \in \text{dom}(\m{G})$}
   \State $\m{G}(v_0) \gets \m{G}(v_0) + g_0$
\Else
   \For{$i$ \textbf{in} $1, \ldots, n$}
        \State $g_i \gets g_0 \cdot \dot{c}_i$
        \State $\m{G} \gets \textsc{backward}(\tilde{c}_i, g_i, \m{G})$
   \EndFor
\EndIf
\State \Return $\m{G}$
\EndFunction
\end{algorithmic}
\end{algorithm}

If we apply this recursion to the computation graph in Figure~\ref{fig:ad-overview}, then we see that doing so indeed computes the correct derivatives; it sums derivatives along each unique path from the output back to the input, where the pathwise derivative is the product of the partial derivatives along the path. Algorithm~\ref{alg:backward} shows pseudo-code for an implementation of this backward recursion.

Now that we have described how to perform the forward pass via evaluation and the corresponding backword recursion, let us put the pieces together to implement the gradient computation for $U$.

Reverse-mode AD can be implemented in this manner for most programming language. The main requirement is that we have to define AD-compatible versions $\tilde{f}$ of each primitive function $f$ that we would like to use in the program, which in turn requires that we implement the corresponding function $\nabla f$ that defines the gradient or Jacobian of the primitive. As always, there are some caveats to reverse-mode AD. In particular, implementing reverse-mode AD does not in itself ensure that a program will be differentiable for all possible inputs, since control flow can give rise to discontinuities. We will discuss the implications of this in the context of HMC methods in the next section.



	\section{Implementation Considerations}
	\label{sec:hmc-implementation}

\begin{algorithm}[!t]
  \caption{\label{alg:grad-U} Gradient of the potential of a program.}
  \begin{algorithmic}
    \State \textbf{global} $\tilde{\gamma}$
    \Comment{A program translation with lifted primitives}
    \Function{backward}{$\ad{c}_0$, $g_0$, $\m{G}$}
      \State $\dots$
      \Comment{As in Algorithm~\ref{alg:backward}}
    \EndFunction
    \Function{$\nabla U$}{$\m{X}$}
      \State $[x_1 \mapsto c_1, \dots, x_m \mapsto c_m] \gets \m{X}$
      \For{$j = 1, \dots, m$}
        \State $\ad{x}_j \gets \big(c_j, x_j, (), ()\big)$ 
        \Comment{Create input nodes}
      \EndFor
      \State $\_,\tilde{U} \gets \Call{eval}{\mhop{($\ad{\gamma}$\ $\ad{x}_1$\ $\dots$\ $\ad{x}_m$)}}$
      \Comment{Forward pass}
      \State $\m{G} \gets [x_1 \mapsto 0, \dots, x_m \mapsto 0]$
      \Comment{Initialize gradient}
      \State \Return $\Call{backward}{\tilde{U}, 1, \m{G}}$
      \Comment{Backward pass}
    \EndFunction
  \end{algorithmic}
\end{algorithm}

Now that we have covered how translate a program into a computation for the potential function $U(X)$, and have discussed how to implement
reverse-mode automatic differentiation, we have what we need to design an HMC implementation for programs in the statically addressed HOPPL. We begin by discussing how to compute the gradient of the potential using reverse-mode AD, then discuss how to integrate Hamilton's equations to construct the proposal, and then define the resulting HMC sampler implementation.

\paragraph{Computing the Gradient of the Potential Function} Algorithm~\ref{alg:grad-U} describes how to compute the gradient $\nabla_X U(X)$ using reverse-mode automatic differentiation. Our goal is to compute the potential energy $\ad{U}$ in a manner that is compatible with reverse-mode AD. For this purpose we assume that we have a program body expression $\ad{e}$, which has been defined relative to an environment with AD-compatible primitives, and that this expression has been embedded using Program~\ref{prog:sahoppl-energy} into a function that computes the return value $\ad{v}$ and the potential energy $\ad{U}$.

To perform the forward pass in reverse-mode AD, we need to evaluate the lifted translation \ad{\gamma} relative to a set of inputs. For this purpose we will assume that the intergrator implemetation provides a map $\m{X}=[x_1 \mapsto c_1, \dots, x_m \mapsto c_m]$ that explicity associates variable names with their values. We then define a function $\nabla U$ that evaluates $\nabla U(X)$ at $\m{X}$. To indicate this, we will slightly overload notation to write
\begin{align}
  \nabla U (\m{X})
  \equiv
  \nabla U(X) \big\vert_{X=\m{X}}
  =
  \nabla U(X) \big\vert_{x_1=c_1, \dots, x_m=c_m}.
\end{align}
To evaluate the translated program relative to inputs $\m{X}$, we assume that the inference algorithm has access to an evaluator function $\textsc{eval}$ analogous to the ones that we defined in Chapter~\ref{ch:eval-one}. We use this evaluator to call the program \hop{($\ad{\gamma}$ $\ad{x}_1$ $\dots$ $\ad{x}_m$)} with arguments that have been boxed into data structures $\ad{x}_j$ that define input nodes. This returns the boxed potential energy $\ad{U}$, which we then pass to the backward recursion from Algorithm~\ref{alg:backward} to perform the backward pass and compute the gradient.

\begin{algorithm}[!t]
\caption{\label{alg:leapfrog} Leapfrog integration for Hamilton's Equations~\eqref{eq:hmc-eq-motion}.}
\begin{algorithmic}
  \Function{$\nabla U$}{$\m{X}$}
    \State $\dots$
    \Comment{As in Algorithm~\ref{alg:grad-U}}
  \EndFunction
  \Function{leapfrog}{$\m{X}_0$, $\m{R}_0$, $T$, $\epsilon$}
    \State $\m{R}_{1/2} \gets \m{R}_0 - \frac{1}{2} \epsilon \nabla U(\m{X}_0)$
    \For {$t$ \textbf{in} $1, \ldots, T-1$}
      \State $\m{X}_t \gets \m{X}_{t-1} + \epsilon \m{R}_{t-1/2}$
      \State $\m{R}_{t+1/2} \gets \m{R}_{t-1/2} - \epsilon \nabla U(\m{X}_t)$
    \EndFor
    \State $\m{X}_T \gets \m{X}_{T-1} + \epsilon \m{R}_{T-1/2}$
    \State $\m{R}_{T} \gets \m{R}_{T-1/2} - \frac{1}{2} \epsilon \nabla U(\m{X}_T)$
    \State \Return $\m{X}_T, \m{R}_T$
  \EndFunction
\end{algorithmic}
\end{algorithm}

\paragraph{Leapfrog Integration} 
There are a number of ways to generate particle trajectories in a manner that ensures that proposals are reversible, conserves the Hamiltonian, and is volume-preserving, which is to say that the volume of a region is preserved when mapping all points in the region to new points using an integrator. Most HMC implementations in probabilistic programming systems implement variants of dynamic HMC that derive from the NUTS algorithm \citep{hoffman2014no}, which expands the simulated trajectory until a dynamic stopping condition is reached. In this section, we discuss a simpler variant of HMC that makes use of a standard St\"ormer-Velet ``leapfrog'' integrator.

Algorithm~\ref{alg:leapfrog} shows an implementation of a leapfrog integration scheme for Hamilton's equations. Like any integrator, this scheme discretizes the trajectory for the position into $T$ time points that are spaced at an interval $\epsilon$. The term ``leapfrog'' references the fact that this scheme computes the corresponding discretization for the momentum at time points that are shifted by $\epsilon/2$ relative to those at which we compute the position. 

\begin{algorithm}[!t]
  \caption{
    \label{alg:hmc}
    Hamiltonian Monte Carlo}
  \begin{algorithmic}[1]
    \State \textbf{global} $\gamma$ 
    \Comment{Unlifted program translation}
    \Function{leapfrog}{$\m{X}_0$, $\m{R}_0$, $T$, $\epsilon$}
      \State $\dots$
      \Comment{As in Algorithm~\ref{alg:leapfrog}}
    \EndFunction
    \Function{$H$}{$\m{X}$, $\m{R}$, $M$}
      \State $[x_1 \mapsto c_1, \dots, x_m \mapsto c_m] \gets \m{X}$
      \State $\_, U \gets \Call{eval}{\mhop{($\gamma$\ $c_1$\ $\dots$\ $c_m$)}}$
      \State $K \gets \Call{matmul}{\m{R}, \Call{matmul}{M^{-1}, \m{R}}}$
      \State \Return $U + K$
    \EndFunction
    \Function{hmc}{$\m{X}^{0}$, $S$, $T$, $\epsilon$, $M$}
      \For {$s$ \textbf{in} $1, \ldots, S$}
        \State $\m{R}^{s-1} \sim \Normal(0, M)$
        \State $\m{X}', \m{R}' \gets \Call{leapfrog}{\m{X}^{s-1}, \m{R}^{s-1}, T, \epsilon}$
        \State $u \sim \text{Uniform}(0,1)$
        \If{$u < \exp \big(- H(\m{X}', \m{R}', M) + H(\m{X}^{s-1}, \m{R}^{s-1}, M\big))$}
            \State $\m{X}^{s} \gets \m{X}'$
        \Else
            \State $\m{X}^{s} \gets \m{X}$
        \EndIf
      \EndFor
      \Return $\m{X}^{1}, \ldots, \m{X}^{S}$
    \EndFunction
  \end{algorithmic}
\end{algorithm}

\paragraph{HMC implementation} Algorithm~\ref{alg:hmc} shows an HMC implementation that follows the steps that we described at a higher level Section~\ref{sec:hmc-inference}. For each iteration of the sampler, we perform a Gibbs update for the moment variable $\R$ and propose a move using the leapfrog integrator from Algorithm~\ref{alg:leapfrog}, which we then accept or reject according to the Hamiltonian of the sample and the proposal.

The standard HMC algorithm has 3 hyperparameters –– $T$, $\epsilon$, and $M$ –– and performance can be sensitive to hyperparameters. $T$ needs to be chosen in a manner that ensures a low correlation between samples whilst avoiding unnecessary computation. There is also a trade-off between the step size $\epsilon$ and the numerical stability of the integration scheme, which affects the acceptance rate. Moreover, this step size should also appropriately account for the choice of mass matrix $M$. 

In the HMC implementation in Stan, the $M$ is set by computing an estimate of the posterior covariance during a warmup phase, with
\begin{align}
  M^{-1}_{ij} \simeq \Ev_{\pi(X)}[x_i x_j] - \Ev_{\pi(X)}[x_i]\:\Ev_{\pi(X)}[x_j].
\end{align}
A dynamic HMC algorithm is then used to eliminate the dependence on the number of time steps $T$ by employing a dynamic stopping criterion that minimizes the degree of correlation with respect to the preceding sample. Finally, the step size $\epsilon$ can be adapted to turn the acceptance ratio to a target value.

\paragraph{Reparameterization} One of the implementation aspects for HMC algorithms that we have not discussed up to this point is how to deal with variables that must satisfy constraints, such as samples from a Gamma distribution, which must be positive, or samples from a Dirichlet distribution, which must lie on a simplex. Dealing with such variables requires reparameterizations that define a bijection between the constrained space and an uncontrained space $\mathbb{R}^d$ for the variable. This form of reparameterization is closely related to the one that is commonly used when estimating gradients in variational autoencoders, which is a topic that we will cover in detail in Section~\ref{sec:gbli}. For a discussion of reparameterization in the context of HMC, we refer to the Stan User Guide \citep{hoffman13}.

\paragraph{Programs with Discontinuities} An implementation consideration for HMC that is particularly relevant to probabilistic programming is that not all programs in the statically-addressed HOPPL define densities $\gamma(X)$ that are differentiable at all points in the space. The same is true for program in Stan and most other systems that provide HMC implementations based on reverse-mode automatic differentiation. While Stan enforces the requirement that a program defines a density over a set of continuous variables that is known at compile time, it does not enforce the requirement that the density is differentiable. For example, the following program would be entirely valid when expressed in Stan:
\begin{center}
\begin{minipage}{0.9\textwidth}
\begin{foppl}[mathescape,label=model:gmm:hmc]
(let 
  [x (sample x (normal 0.0 1.0))
   y 0.5]
  (if (> x 0.0)
    (observe (normal 1.0 0.1) y)
    (observe (normal -1.0 0.1) y)))
\end{foppl}
\end{minipage}
\end{center}
This program corresponds to an unnormalized density
\begin{align*}
  \gamma(x) = \text{Norm}(0.5 ; 1, 0)^{I[x > 0]} \text{Norm}(0.5 ; -1, 0)^{I[x \le 0]} \text{Norm}(x ; 0, 1),
\end{align*}
for which the derivative is clearly undefined at $x=0$, since $\gamma(x)$ is discontinuous at this point.

HMC is not necessarily invalid when a program contains discontinuities, but it can be inefficient. As long as the integrator remains reversible, the MCMC transition operator in HMC in principle remains ergodic, which is to say that it converges to the correct target density, allows the sampler to transition between any two points in the sample space in a finite number of steps, and does not get stuck in deterministic cycles. However, discontinuities can dramatically reduce the efficiency of HMC methods. Since numerical integration may not preserve the Hamiltonian, HMC can have a low acceptance ratio, which in turn can lead to poor mixing efficiency. 

Extending HMC to improve sampling efficiency in models that are not differentiable everywhere is a topic of active research. Examples of such approaches include reflective HMC~\citep{AfsharD15} and Discontinuous HMC \citep{nishimura2020discontinuous}. These extensions require the set of non-differentiable points to have measure zero. Intuitively, this means that there is a vanishing probability of evaluating a point where the derivative is undefined. 

It is possible to explicitly restrict probabilistic programming language to ensure that discontinuties have measure zero. \cite{ZhouGKRYW19} do so by introducing a variant of the FOPPL,
\begin{foppl}
  $e$ $::=$ $c$ $\mid$ $v$ $\mid$ (let [$v$ $e_1$] $e_2$) $\mid$ (if ($<$ $e_1$ 0) $e_2$ $e_3$)
       $\mid$ ($c$ $e_1$ $\dots$ $e_n$) $\mid$ (sample $e_1$) $\mid$ (observe $e_2$ $c$)
\end{foppl}
The key difference in this language, relative to the one that we introduced in Chapter~\ref{ch:foppl}, is that the predicate of an if expression is defined in terms of a comparison \fop{($<$ $e_1$ 0)}, where $e_1$ is a real-valued expression rather than a boolean-valued expression. If we now additionally restrict the language to ensure that all distribution primitives are continuous and that all primitive functions are analytic (i.e.~have a converging Taylor series), then the only remaining source of discontinuities in the density arises at the boundaries of predicates, i.e.~points where \fop{($=$ $e_1$ 0)}, which by construction will have measure 0. Moreover, we can track when an integrator crosses the boundary of a branch by defining indicator expressions for each predicate at compile time, analogous to the ones we used in Section~\ref{sec:compilation-bounded} to track whether observe expressions should be included in the density. This makes it possible to define language semantics in which if-expressions are evaluated eagerly, as in Chapter~\ref{ch:eval-one}, and implement extensions of HMC that appropriately account for discontinuities in the density.

In practice, the conditions needed to ensure that discontinuities in a program have measure zero are relatively mild. \citet{mak2021densities} formalized these conditions and demonstrated that programs in a simply-typed stochastic lambda calculus are differentiable almost everywhere, provided a suitable set of analytic primitive functions is used. This result implies that it is in principle also possible to apply HMC to certain classes of programs in a HOPPL. As an example, let us revisit the the program that samples from a geometric distribution, which we originally introduced in Chapter~\ref{ch:eval-one}. We will eliminate discrete variables from this program  reparameterize this program to replace samples from the bernoulli with samples from a uniform distribution, 
\begin{hoppl}
(defn sample-geometric [alpha]
  (let [u (sample (uniform 0.0 1.0))]
    (if (< (- u alpha) 0.0) 
      1.0
      (+ 1.0 (sample-geometric p)))))

(let [alpha (sample (uniform 0.0 1.0))
      k (sample-geometric alpha)]
  (observe (poisson k) 15) 
  alpha)
\end{hoppl}
This program is an example of a so-called tree-representable program \citep{mak2021nonparametric}. While the number of variables that can be instantiated is unbounded, it is possible to characterize the density of the program in terms of a tree in which leaf nodes represent distinct control-flow paths for which the density is piecewise continuous. This makes it possible to define nonparametric HMC methods \citep{mak2021nonparametric} in which the integrator mixes over different levels of recursion.

In short, while we have in this chapter considered a Stan-like variant of the HOPPL with static addressing, in which the set of unique random variables is known at compile time, it is in fact possible to apply gradient-based methods like HMC to more general classes of programs, as long as we ensure that all primitive functions are analytic and all expressions are real-valued. 

While it is possible to apply HMC methods and other gradient-based methods to programs with discontinuities, this of course does not also mean that that these methods are always a suitable choice for inference in programs that are fundamentally discrete. The program above is a good case in point; All samples from uniform distribution have density 1.0, which means the energy of the program is a piecewise-constant function that depends only on the level of recursion 
\begin{hoppl}
  (- (log-prob (poisson k) 15))
\end{hoppl}
While this energy is differentiable almost everywhere, it is also the case that its gradient is 0 wherever it is defined. For this particular program, the gradients therefore carry no information that can guide inference.

	\section{Other Methods for Differentiable Models}
	\label{sec:hmc-other}

In this chapter, we have seen an implementation strategy for probabilistic programming that combines a general higher-order language with differentiable programming. At the level of language design, this style of probabilistic programming defines a modeling language in which the set of latent variables is statically determinable (and typically also statically typed). At the level of the inference backend, this style of probabilistic programming relies on automatic differentiation to implement Hamiltonian Monte Carlo. This design has proven itself to be something of a ``sweet spot''; it allows us to combine a reasonably flexible language for model specification with an algorithm that often performs well in practice, particularly when we are trying to model moderately high-dimensional distributions. 

HMC is in not the only method makes use of differentiable programming to improve inference efficiency. Automatic differentiation is an integral part of deep probabilistic programming systems, which make use of neural networks to parameters variational distributions and generative models. In particular, reverse-mode AD makes it possible to implement reparameterized gradients, which are used in automatic differentiation variational inference (ADVI; \citet{kucukelbir2017automatic}) as well as in variational autoencoders \citep{kingma2014auto,rezende2014stochastic}, which we will discuss in detail in Section~\ref{sec:gbli}.

\chapter{Deep Probabilistic Programming}
\label{ch:deep-probprog}

Up to this point, this book has focused mainly on the design space of probabilistic programming languages and corresponding inference algorithms. In doing so, we have implicitly made two assumptions. The first is that we can design a program that models the data in a sufficiently accurate manner. The second is that we will condition this program on a single set of observations.  In particular, we have considered use cases in which we will be doing inference once per program, and in which we would like to produce the highest quality inference results possible in this one-time-only inference scenario.  

These implicit assumptions are appropriate in a variety of circumstances, notably in traditional Bayesian statistical analyses of small data or in high fidelity simulation-based studies of particular phenomena. However, many use cases for inference in ML and AI pose a different set of challenges.
One challenge is what we will describe as non-programmability. For many data modalities that are commonly considered in ML and AI, including images and natural language, it is near-impossible to fully specify a probabilistic program that defines a sufficiently realistic distribution over data. 
A second challenge is scalability. Models in ML and AI are routinely trained on very large datasets; sometimes labeled, sometimes unlabelled, and sometimes partially labeled. Most inference methods that we have considered so far do not scale to such large datasets without additional modifications. 

In this chapter, we will discuss how these challenges can be addressed by combining inference methods from probabilistic programming with differentiable programming techniques from deep learning research. Probabilistic and differentiable programming are in many ways complementary and can be combined in mutually beneficial manner. 
Whereas probabilistic programming languages provide abstractions for defining random variables, deep learning frameworks provide a domain-specific language (DSL) for differentiable programming, which is typically embedded as a library in an existing language such as Python. The fundamental design choices for this DSL are very much analogous to the ones that we have seen in this book. When the DSL is a first-order language analogous to the FOPPL, as is used in TensorFlow \citep{abadi2015tensorflow}, the model can be compiled to a static computation graph. When the DSL is a higher-order language analogous to the HOPPL, as is used in PyTorch \citep{paszke2017automatic}, we obtain greater flexibility, but the computation graph must be constructed dynamically at run time. 

We have already seen how systems like Stan \citep{carpenter2015stan} and PyMC3 \citep{salvatier2016probabilistic} make use of differentiable programming to implement efficient inference methods like Hamiltonian Monte Carlo. In the next sections, we will discuss how differentiable programming can be used to train probabilistic programs on very large datasets using stochastic gradient descent. 
This forms the basis for learning and inference in deep probabilistic programming systems, such as Edward \citep{tran2016edward}, Pyro \citep{bingham2018pyro}, Probabilistic Torch \citep{siddharth2017learning}, and PyProb \citep{le2016inference}. 

This chapter is organized as follows. In Section~\ref{sec:deep-generative-programs}, we begin with a high-level discussion of probabilistic programs that use networks to parameterize conditional distributions. Incorporating neural networks into probabilistic programs makes it possible to design flexible families of deep generative models, which can represent a range of possible relationships between latent variables in a program and observed data. This is particularly useful when modeling data modalities such as images and text, for which it is difficult to define a realistic likelihood from first principles. In Section~\ref{sec:deep-inference-programs}, we will discuss mechanisms for implementing a corresponding inference program, also known as a guide, that serves to approximate the posterior of the generative program. This makes it possible to amortize the inference computation, by learning a data-dependent variational approximation that facilitates fast inference on new data at test time.

Programs that denote deep generative models or inference models will typically have thousands or even millions of parameters, which need to be estimated from training data. In Section~\ref{sec:gbli}, we review methods that use stochastic gradient descent to learn a generative model by maximizing the marginal likelihood of the data, as well as methods for stochastic variational inference that minimize a divergence between the inference model and the posterior. 

We then turn to implementation strategies for amortized inference in deep probabilistic programming systems. In Section~\ref{sec:pyprob-proposals} we discuss the design employed in PyProb, which implements amortized inference across a messaging interface that is similar to the one that we discussed in Chapter~\ref{ch:eval-two}. This makes it possible to use a differentiable guide in the backend to amortize inference in probabilistic programs. 

A second implementation strategy for amortized inference is to define the inference model in the same language as the generative model. In Section~\ref{sec:webppl-proposals} we discuss the design that is employed in WebPPL, which integrates the inference model into the generative model by associating proposals with unobserved random variables. Finally, in Section~\ref{sec:probtorch-proposals} we discuss a more general design in which the generative model and inference model are implemented as distinct probabilistic programs. This design is imployed in Pyro, Edward2, and Probabilistic Torch. 


	\section{Programs as Deep Generative Models}
	\label{sec:deep-generative-programs}

Programs that define deep generative models are in most respects just like any other program in this introduction; they contain \hop{sample} expressions to denote a prior distribution over unknown variables and \hop{observe} expressions to condition on data. The only distinction between deep generative programs and other programs is that they can use neural networks as primitive functions. There is in some sense nothing special about doing this. Neural networks are pure functions; their outputs are fully determined by their inputs and evaluation has no side effects. This means that they can be incorporated into a program like any other functionally pure primitive. 

As a simple example, let us consider the task of modeling a distribution of images that belong to $K$ different classes, 
\begin{hoppl}
(defn p [y $\eta$ $\theta$]
  (let [z (sample (multinomial 1 $\theta^z$)) 
        v (sample (normal ($\eta^v_\mu$ z $\theta^v$) ($\eta^v_\sigma$ z $\theta^v$)))]
    (observe (normal ($\eta^y_\mu$ v $\theta^y$) ($\eta^y_\sigma$ v $\theta^y$)) y)
    [z v]))
\end{hoppl}
This program defines a deep mixture model with optimizable parameters $\theta = \{\theta^z, \theta^v, \theta^y\}$. We first sample the mixture component \hop{z} from a multinomial distribution. We then define an image encoding \hop{v} using an embedding layer $\eta^v$ (i.e.~a lookup function) that returns a mean \hop{($\eta^v_\mu$ z $\theta^v$)} and standard deviation \hop{($\eta^v_\sigma$ z $\theta^v$)}. Finally, we use a neural network $\eta^y$, which is often referred to as a generator or decoder, to transform the intermediate variable $v$ into a distribution over images, also in the form of a normal distribution with diagonal covariance, which we use to define the likelihood for an image \hop{y}.

Deep probabilistic programming systems provide modeling languages that can be used to implement programs such as this one. In this modeling language, all computations typically operate on a special data type, the tensor, using primitive functions that support automatic differentiation, such as the ones we introduced in our discussion of Hamiltonian Monte Carlo. Primitive functions, including distribution constructors (see, e.g., \citet{dillon2017tensorflow}), typically also support vectorization. This is to say that operations accept tensor-valued arguments as inputs and apply the function to each element in the tensor to construct a tensor of outputs. When constructing a \hop{normal} distribution, for example, we can replace a scalar mean and standard deviation with a tensor-valued mean and standard deviation to construct a tensor of random variables.



From a practical point of view, the main implication of using neural networks to parameterize a program is that we will have to estimate the parameters $\theta$. Whereas traditional probabilistic programs typically have a small number of hyperparameters, which can be specified by hand based on our knowledge of the problem domain, deep probabilistic programs can have thousands or even millions of parameters, which we will have to estimate from data.

We can estimate parameters by learning them, or by inferring them. Inferring parameters will in general be challenging. Given training data $\{Y_1, \dots, Y_N\}$, we could in principle define a prior $p(\theta)$ on the network weights and use inference methods to approximate the posterior marginal $p(\theta \mid Y_1, \dots, Y_N)$. This is known as Bayesian deep learning \cite{}. It is an active area of research that poses conceptual challenges, such as how we should choose the prior $p(\theta)$. It also poses computational challenges, since we need to approximate a high-dimensional posterior in which local optima correspond to distinct modes. 

In this chapter, we will for this reason focus on simpler and more commonly used methods that learn the parameters $\theta$ by maximizing the marginal likelihood of the training data $p(Y_1, \dots, Y_N ; \theta)$,
and particularly focusing on models which can be expressed by a joint distribution of the form
\begin{align*}
p(Y_1, \dots, Y_N, X_1, \dots, X_N ; \theta)
&= 
\prod_{n=1}^N p(Y_n, X_n ; \theta).
\end{align*}
This factorization is a standard formulation for many commonly-used deep generative models (across $N$ data points), and additionally covers the case of repeated inference in the same model (across $N$ different sets of training data) which we will discuss further in Section~\ref{sec:amortized-inference}.
As it turns out, approximating the gradient of the marginal likelihood itself requires inference, since this gradient can be expressed as an expectation with respect to the posterior (see Section~\ref{sec:gbli})
\begin{align*}
  \nabla_\theta \log p(Y_1, \dots, Y_N ; \theta)
  =
  \sum_{n=1}^N
  \Ev_{p(X_n \mid Y_n ; \theta)}
  \big[
  \nabla_\theta
  \log p(Y_n, X_n ; \theta)
  \big]
  .
\end{align*}
The implication of this is that we have to solve $N$ inference problems for every gradient step. This is completely intractable, since $N$ is typically in the order of thousands or even millions of training data points. Our only hope is therefore to use stochastic gradient descent, in which we replace the gradient of the marginal likelihood with an unbiased estimate. To compute this estimate, we follow the normal practice of selecting a mini-batch of examples $Y_n$ at random from the training data.

However, even solving a mini-batches of inference problems is computationally challenging, particularly when we consider that we will have to perform many gradient steps to estimate the parameters $\theta$. A solution to this problem is to ``amortize'' the inference computation by learning a variational distribution $q(X_n \mid Y_n ; \phi)$ that ``inverts'' the generative model by approximating the posterior $p(X_n \mid Y_n ; \theta)$.

	\section{Programs as Inference Models}
	\label{sec:deep-inference-programs}

Learning an amortized variational distribution $q(X_n \mid Y_n ; \phi)$ differs from the black-box variational inference (BBVI) methods that we previously discussed in Section~\ref{sec:eval-bbvi}. In BBVI, we considered a single inference problem, in which we optimized the parameters $\lambda$ of a variational distribution $q(X ; \lambda)$ to approximate the posterior $p(X \mid Y)$. When performing amortized inference, we are interested in solving a collection of $N$ inference problems. In principle, we could do this by defining $N$ variational distributions $q(X_n ; \lambda_n)$, but this means that we would have to optimize $N$ sets of variations parameters $\lambda_n$. Often, a much more practical approach is to train a neural network $\lambda(Y_n, \phi)$ to predict the variational parameters for each item $Y_n$ in the training dataset
\begin{align}
  q\big(X_n \mid Y_n ; \phi \big)
  =
  q\big(X_n ; \lambda(Y_n, \phi)\big).
\end{align}
This approach to inference is known by many names. We will in this book typically use the term amortized inference \citep{gershman2014amortized}, in reference to the fact that learning $\phi$ is an expensive computation that is performed at training time in order to ``amortize'' the cost of inference at test time. It has also been referred to as inference compilation \citep{le2016inference}, since training the neural network can be thought of as ``compiling'' a probabilistic program to its posterior, albeit to an approximation of this posterior that does not strictly preserve the semantic meaning of the original program. 

In the probabilistic programming literature, amortized inference is used when training deep probabilistic programs, but also more generally to accelerate inference in simulation-based models. In the machine learning literature, it has a long history of us in Helmholtz machines and variational autoencoders \citep{dayan1995helmholtz,kingma2014auto,rezende2014stochastic}. In the context of these models, this style of inference is also known as ``autoencoding'' variational inference.

The neural network $\lambda(Y, \phi)$ is often referred to as an encoder network or an inference network. It defines a probabilistic model $q(X \mid Y ; \phi)$ that we will refer to as an inference model, or equivalently as a guide. Further on in this chapter, we will discuss strategies for specifying inference models in a programmatic manner. One strategy is to define the inference network $\lambda(Y, \phi)$ in the language that implements the inference backend. This means that the backend must support automatic differentiation, but that probabilistic programs can be written in a language that need not be differentiable. This is particularly useful when applying amortized inference to simulation-based models, which will not always have been designed with differentiability in mind. We will discuss this approach, which is used in the PyProb probabilistic programming system \citep{le2016inference}, in Section~\ref{sec:pyprob-proposals}.

A second approach is to use a standalone program to define the inference model. For the program \hop{p} from the previous section, we could define a guide program in the same language as \hop{p}
\begin{hoppl}
(defn q [y $\lambda$ $\phi$]
  (let [z (sample (multinomial 1 ($\lambda^z$ y $\phi^z$)))
        v (sample (normal ($\lambda^v_\mu$ y z $\phi^v$) ($\lambda^v_\sigma$ y z $\phi^v$)))]
    [z v]))
\end{hoppl}
This program \hop{q} is a deep probabilistic program, just like \hop{p}. However, unlike \hop{p}, the program \hop{q} does not make use of \hop{observe} expressions, which means that we do not need an inference algorithm to generate samples; we can simply run the program. Doing so generates the cluster assignment \hop{z} from a multinomial distribution, whose parameters are computed from \hop{y} using a neural classifier $\lambda^z$. We subsequently use a second network $\lambda^y$ to predict the image encoding \hop{v} from \hop{y} and \hop{z}.

Using a program \hop{q} to define the inference model gives rise to some technical issues. In particular, while we have used consistent variable names \hop{z} and \hop{y} to make it clear to the reader which variables in \hop{p} correspond to which variables in \hop{q}, there is no general way to unambiguously associate random variables in \hop{q} with random variables in \hop{p} without some form of additional annotation by the user.

One strategy for making this correspondence unambiguous is to interleave the program \hop{p} and the program \hop{q} into a single program that denotes both the generative and the inference model.
To do so, we can replace every instance of an expression \hop{(sample $d_p$)} with a new expression form \hop{(propose $d_p$ $d_q$)}, which associates the variational distribution $d_q$ as a proposal with the prior distribution $d_p$ in the generative model. In the case of the programs \hop{p} and \hop{q}, this interleaved program would have the form
\begin{hoppl}
(defn p-with-q [y $\eta$ $\theta$ $\lambda$ $\phi$]
  (let [z (propose (multinomial 1 $\theta^z$)
                   (multinomial 1 ($\lambda^z$ y $\phi^z$)))
        v (propose (normal ($\eta^v_\mu$ z $\theta^v$) ($\eta^v_\sigma$ z $\theta^v$)))
                   (normal ($\lambda^v_\mu$ y z $\phi^v$) ($\lambda^v_\sigma$ y z $\phi^v$)))]
  (observe (normal ($\eta^y_\mu$ v $\theta^y$) ($\eta^y_\sigma$ v $\theta^y$)) y)
  [z v]))
\end{hoppl}
We will discuss this strategy, which is used to support amortized inference in WebPPL \citep{ritchie2016deep}, in Section~\ref{sec:webppl-proposals}.

A second strategy for making correspondences between variables explicit is to annotate every random variable in \hop{p} and \hop{q} with a unique address $\alpha$, for example by using expressions of the form \hop{(sample $\alpha$ $d$)} and \hop{(observe $\alpha$ $d$ $c$)}. In this setup, variables in \hop{p} and \hop{q} correspond to the same variable if (and only if) they have the same unique address $\alpha$. This strategy is used to implement variational distributions and proposals in a number of (deep) probabilistic programming systems, including Edward \citep{tran2016edward}, Pyro \citep{bingham2018pyro}, Probabilistic Torch \citep{siddharth2017learning}, and Gen \citep{cusumano-towner2019gen}.

A subtlety that arises when using arbitrary programs as inference models is that, even with annotated variables, we will have to check whether the inference model \hop{q} instantiates the same variables as \hop{p}. In the original Edward implementation, which is based on Tensorflow, this problem is solved by compiling both programs to  graphical models, which makes it possible to check correspondences at compile time. However, as we have seen in Chapter~\ref{ch:eval-one}, compiling a program to a graphical model requires a first-order programming language. As it turns out, it is not  necessary for \hop{p} and \hop{q} to instantiate the \emph{same} random variables, as long as we know \emph{which} variables are common to both programs. In Section~\ref{sec:probtorch-proposals}, we will discuss how to handle the general case of programs in higher-order languages, as is used in Pyro, Probabilistic Torch, Edward2 (which supports Tensorflow's eager mode), and Gen.



	\section{Stochastic-Gradient Methods for Learning and Inference}
	\label{sec:gbli}


Before we continue our discussion of design and implementation strategies for deep probabilistic programming systems, we will need to cover some mathematical background. 
In this section, we will discuss general techniques for estimating model parameters by maximizing the marginal likelihood of training data. We will also discuss techniques for learning parameters of amortized inference artifacts by minimizing a divergence between the posterior and the inference model. These techniques form the basis for the deep probabilistic programming methods that we will discuss in the next sections, but are also used generally in machine learning research to train deep generative models. 

We will begin by showing how to perform maximum likelihood estimation of model parameters using stochastic gradient descent. As we will establish, maximum likelihood parameter estimation requires samples from the program posterior.  This connects to the preceding chapters in the book as any of the probabilistic program inference methods described so far can be used to do this. 

We then discuss model parameter learning in the context of variational inference.  We show how model parameter learning is also evidence lower bound maximization in the variational inference framework and explain how variational approximate posteriors can be used for efficient importance-sampling-based model learning. 

The variational framework also presents opportunities to simultaneously learn model parameters and amortized inference neural network artifacts.  We discuss how this works, showing that amortized inference networks predict approximate variational posterior parameters directly from observations.   We relate amortized variational inference to both wake-sleep and variational autoencoders.

This discussion will give us the context we need to discuss the design questions that arise when constructing inference networks for probabilistic programs, and more generally when implementing deep probabilistic programming systems on top of languages endowed with first class automatic differentiation and optimization primitives. We will return to these questions in Section~\ref{sec:deep-design}.




\subsection{Maximizing the Marginal Likelihood}
\label{sec:max-likelihood}

Learning model parameters from data makes sense when a sufficiently large amount of data is available. This is very common in modern machine learning applications, where we often have thousands or even millions of instances.  Model learning is {\em required} when all or part of the generative model program is ``non-programmable'' or otherwise only partially specified.  When either or both of these is the case, we can set up model parameter learning in terms of maximizing what is known as the ``marginal likelihood'' or ``evidence''
\begin{align}
    \theta^* = \argmax_\theta \: \log p(Y_1, \dots, Y_N \mid \theta).
\end{align}
Here $p(Y_1, \dots, Y_N \mid \theta)$ is the evidence of  $Y = \{Y_1, \dots, Y_N\}$ under the generative model parameterized by $\theta$.  

Note that we are making a conscious choice here to {\em maximize} w.r.t.~$\theta$ rather than to impose a prior $p(\theta)$ and estimate the full posterior distribution $p(\theta \mid Y)$, the approach to model learning we have taken so far in this book.  The argument for doing this is simple.  When there is a lot of data, the likelihood $P(Y \mid \theta)$ numerically dominates the prior $p(\theta)$ so effectively that the prior can be ignored.\footnote{The  technical justification for using point estimates is the Bernstein von Mises theorem (\citet[Chapter 9, Section 12]{young2005essentials}, \citet[Chapter 10, Section 2]{van2000asymptotic}).  Intuitively, it states that the Bayesian posterior in reasonable parametric models asymptotically becomes, in the limit of infinite data, a point mass on the maximum likelihood parameter estimate.  This is a strong justification for using point estimates when you have a lot of data. However, this justification does come with caveats; Bernstein von Mises says nothing about model misspecification (i.e.~cases where there is a fundamental mismatch between the parametric model family and the data distribution) and it also breaks down in infinite-dimensional parameter space models (i.e.~general HOPPL programs)}




Maximum likelihood estimation for model parameter learning connects to posterior inference in a natural manner. The reason for this is that the marginal likelihood is an integral with respect to all latent variables $p(Y;\theta) = \int dX \: p(Y,X;\theta)$, which is of course intractable. A direct connection to posterior inference emerges when we consider maximizing the evidence via gradient methods. We can express the gradient of the log marginal likelihood as a posterior expectation,\footnote{
It may not be obvious why Equation~\eqref{eq:grad-log-marginal} holds. To see why this is the case, we can work backwards by expanding the terms on the right hand of the identity
\begin{align}
    \nonumber
    &
    \mathbb{E}_{p(X \mid Y; \theta)}
    \left[
        \nabla_\theta \log p(Y, X; \theta)
    \right]\\
    &\qquad
    =~
    \mathbb{E}_{p(X \mid Y; \theta)}
    \left[
        \nabla_\theta \log p(Y; \theta) + \nabla_\theta \log p(X \,|\, Y; \theta)
    \right],
    \\
    \nonumber
    &\qquad
    =~
    \nabla_\theta \log p(Y; \theta)
    +
    \underbrace
    {\mathbb{E}_{p(X \mid Y; \theta)}
    \big[
       \nabla_\theta \log p(X \,|\, Y; \theta)
    \big]}_{=\nabla_\theta \int dX \: p(X \mid Y ; \theta) =0}.
\end{align}
The second term is 0 by the same reasoning that we previously applied to likelihood-ratio estimators in our discussion of black-box variational inference.
}
\begin{align}
    \label{eq:grad-log-marginal}
    \nabla_\theta \log p(Y; \theta)
    &=
    \mathbb{E}_{p(X \mid Y; \theta)}
    \left[
        \nabla_\theta \log p(Y, X; \theta)
    \right]
    .
\end{align}
In other words, as long as we can approximate expectations with respect to $p(X \mid Y; \theta)$, which has been the focus of this book thus far, we can also approximate the gradient of the marginal likelihood.   And maximun likelihood parameter estimation can be done by gradient ascent of the evidence using the gradient of the evidence with respect to $\theta$.

The identity in Equation~\eqref{eq:grad-log-marginal} suggests a straightforward, if usually inefficient, strategy for performing maximum likelihood estimation in probabilistic programs. Since the gradient can be expressed as an expectation with respect to the posterior, we can perform stochastic gradient descent by computing a Monte Carlo estimate of the gradient. To generate posterior samples of $X$ for this estimator, we can use any of the methods that we have discussed in this book so far. This leads to the following procedure for maximum likelihood estimation:
\begin{itemize}
    \item[1.] Initialize $\theta_0$. Define step sizes $\{\eta_1, \dots, \eta_T\}$ such that
    \begin{align*}
        \textstyle
        \sum_t \eta_t &= \infty,
        & 
        \sum_t \eta_t^2 &< \infty.
    \end{align*}
    \item[2.] Loop over iterations $t=1,\dots,T$:
    \begin{itemize}
    \item[a.] Perform probabilistic program inference to generate samples
    \begin{align*}
        X^l &\sim p(X \,|\, Y ; \theta_{t-1}),
        &
        l = 1, \dots, L.
    \end{align*}

    \item[b.] Update $\theta$ using a Monte Carlo estimate of the gradient
    \begin{align*}
        \theta_t 
        &= 
        \theta_{t-1} 
        + 
        \eta_t 
        \left( 
          \frac{1}{L} 
          \sum_{l=1}^L 
          \nabla_\theta 
          \log p(Y, X^l ; \theta_{t-1})
        \right).
    \end{align*}
    \end{itemize}
\end{itemize}    
This procedure for maximum likelihood estimation is very general.
The only computational requirement is that we should be able to evaluate the gradient of the log density $\log p(Y,X; \theta)$ with respect to the parameters~$\theta$. This in turn implies that we need to be able to compute the gradient of the densities of all individual random variables in the program
\begin{equation*}
    \nabla_\theta \: \log p(Y,X ; \theta)
    =
    \sum_{y \in Y}
    \nabla_\theta \log p(y \,|\, \textsc{pa}(y); \theta)
    +
    \sum_{x \in X}
    \nabla_\theta \log p(x \,|\,\textsc{pa}(x); \theta)
\end{equation*}
Each term in this density corresponds to a distribution object $d$, which is constructed using a primitive function that must be provided as part of the language implementation. This means that we can implement support for maximum likelihood estimation in any system where the distributions library supports evaluation of the gradient of the log density with respect to the parameters. This is exactly the same requirement as the one that we introduced in the context of black-box variational inference, where we assumed an implementation of the function $\textsc{grad-log-prob}(d,c)$ for any distribution $d$ and value $c$. 

Note that the requirements for maximum likelihood estimation are much milder than the requirements for Hamiltonian Monte Carlo, where we needed to compute the gradient $\nabla_X \log p(Y, X ; \theta)$. As we discussed in 
Section~\ref{sec:hmc-ad},
this computation requires a full system for automatic differentiation, which means that we need to be able to compute derivatives with respect to inputs for \emph{every} primitive function in the language that returns real numbers. Moreover, computing $\nabla_X \log p(Y,X ; \theta)$ directly is only possible when all variables $x \in X$ are continuous.

By comparison, in the maximum likelihood procedure above only requires that we can compute derivatives with respect to the \emph{parameters} of each primitive distribution, rather than the \emph{values} of random variables. This means that we can use discrete random variables in the model, as long as the parameters that we wish to learn are continuous.

While automatic differentiation is therefore not strictly a requirement for maximum likelihood estimation in probabilistic programs, there are use cases for automatic differentiation in models that employ reparameterization. We will discuss these approaches in the context of variational inference soon.

\subsubsection{Maximum Likelihood Estimation using Importance Sampling}
\label{sec:importance-weighted-model-learning}

From a practical point of view, the main computational requirement for maximum likelihood estimation is that we need a very cheap way to approximate the posterior. We will need to generate $L$ samples for each of the $T$ gradient steps, which will require thousands (or even millions) of samples in aggregate. This clearly means that we should not start inference from scratch at every gradient step. Rather, we would like to use algorithms in which inference results from the previous step can be used as a starting point for the next step.



A comparatively straightforward way to achieve this is to learn a distribution $q(X ; \lambda)$ that approximates the posterior, for example by using the black-box variational inference procedure form Section~\ref{sec:max-likelihood}. We can use this distribution to approximate the gradient in equation~\eqref{eq:grad-log-marginal} by using it as a proposal in an importance sampler. 

In general, importance sampling expresses an expectation with respect to a \emph{target density}, for which it is difficult to generate samples, in terms of a \emph{proposal density}, for which it should be easy to generate samples. In previous chapters, we have seen examples of importance sampling in which the program prior is used as the proposal, including likelihood weighting and sequential Monte Carlo.  We can also use importance sampling to estimate the gradient of the marginal likelihood in Equation~\eqref{eq:grad-log-marginal}. To do so, we begin by rewriting the expectation with respect to $p(X \mid Y ; \theta)$ as an expectation with respect to $q(X ; \lambda)$,
\begin{align}
\nabla_\theta \log p(Y ; \theta)
&=
\Ev_{p(X|Y; \theta)}
\left[
  \nabla_\theta \: \log p(Y,X ; \theta)
\right],
\\
&=
\Ev_{q(X ; \lambda)}
\left[
  \frac{p(X|Y; \theta)}
       {q(X ; \lambda)}
  \nabla_\theta \: \log p(Y,X ; \theta)
\right].
\end{align} 
As we have previously discussed in the context of likelihood weighting (Section~\ref{subsec:importance-sample}), we cannot directly evaluate the importance ratio $p(X|Y;\theta)/q(X;\lambda)$, since the posterior density is not tractable. The standard solution to this problem is to use samples $X^l \sim q(X ; \lambda)$ to compute a so-called self-normalized estimator
\begin{align}
\label{eq:iw-grad-log-py}
\begin{split}
\nabla_\theta \log p(Y ; \theta)
&=
\frac{1}{P(Y ; \theta)}
\Ev_{q(X ; \lambda)}\!
\left[
  \frac{p(Y, X ; \theta)}
       {q(X ; \lambda)}
  \, \nabla_\theta \, \log p(Y,X ; \theta)
\right],
\\
&\simeq
\frac{1}{\hat{Z}(\theta)}
\frac{1}{L}
\sum_{l=1}^L
w^l \:
\nabla_\theta \: \log p(Y,X^l ; \theta).
\end{split}
\end{align}
In this estimator, the normalizing constant $\hat{Z}(\theta) \simeq p(Y;\theta)$ and the unnormalized weights $w^l$ are defined as
\begin{align}
  \hat{Z}(\theta) &= \frac{1}{L} \sum_{l=1}^L w^l,
  &
  w^l &= \frac{p(Y, X^l ; \theta)}{q(X^l ; \lambda)}.
\end{align}
As we also discussed in Section~\ref{subsec:importance-sample}, self-normalized importance sampling is biased. The reason for this is that the function $f(w)=1/w$ is convex. It therefore follows from Jensen's inequality that
\begin{align}
    \mathbb{E}
    \left[
      \frac{1}{\hat{Z}(\theta)}
    \right] 
    \ge 
    \frac{1}
         {\mathbb{E}[\hat{Z}(\theta)]} 
    =
    \frac{1}{p(Y;\theta)}
    .
\end{align}
The estimate of the normalizer is therefore biased, which implies that the self-normalized estimator is also biased. However, the estimator is consistent, which is to say that the bias converges to 0 as we increase the number of samples $L$. Moreover, for sufficiently large $L$, the bias will be small relative to the variance of the estimator.

The question now arises how we can combine learning of the model parameters $\theta$ with learning of the proposal parameters $\lambda$. In our discussion of black-box variational inference, we have seen that we can also use stochastic gradient descent to learn variational parameters $\lambda$. This means that we can jointly learn these parameters in a single loop over $T$ gradient descent steps. At each step $t$, we can use samples from  $q(X ; \lambda^{t-1})$ to perform a gradient update on both the model parameters and the variational parameters. However, there is a subtle but potentially important concern here; in order for this strategy to work, the variational parameters $\lambda^t$ have to track a ``moving target'', since there will be small changes to the model parameters $\theta^t$ at every time step. We will discuss the implication of this in the next section.



\subsection{Combining Learning and Variational Inference}
\label{sec:advanced-variational-inference}

%


Variational inference methods transform an inference problem into an optimization problem. To do so, they define a parametric family of distributions $q(X; \lambda)$ and optimize $\lambda$ to find a ``best fit'' approximation to the posterior. In Section~\ref{sec:eval-bbvi}, we discussed black-box variational inference, which minimizes the \emph{exclusive} KL divergence between the variational distribution and the posterior,
\begin{align}
\label{eq:ch7-kl-qp}
  \KL{q(X;\lambda)}{p(X|Y ; \theta)}
  =
  \Ev_{q(X;\lambda)}
  \left[
    \log \frac{q(X;\lambda)}{p(X|Y ; \theta)}
  \right].
\end{align}
As we previously discussed, optimizing this KL divergence directly is difficult to the point of being impossible, as the integrand includes the posterior $p(X \,|\, Y ; \theta)$, which is the quantity that we are hoping to approximate in the first place. To get around this problem, we defined an evidence lower bound (ELBO)
\begin{align}
  \label{eq:ch7-elbo}
  \mathcal{L}(Y; \theta, \lambda)
  &:=
  \Ev_{q(X;\lambda)}
  \left[
    \log \frac{p(Y,X ; \theta)}{q(X; \lambda)}
  \right],
\end{align}
This objective takes advantage of the fact that we can decompose the log joint into the log marginal likelihood and the log posterior
\begin{align}
  \log p(Y, X ; \theta) = \log p(Y ; \theta) + \log p(X \,|\, Y ; \theta).
\end{align}
This leads us to a powerful trick; while each of the terms $\log p(Y ; \theta)$ and $\log p(X \,|\, Y; \theta)$ is difficult to approximate, it is very easy to compute their sum, since we can always compute the log joint $\log p(Y, X ; \theta)$ of a probabilistic program. This decomposition shows that \emph{maximizing} the ELBO with respect to $\lambda$ is equivalent to \emph{minimizing} the KL divergence,  
\begin{align}
 \mathcal{L}(Y ; \theta, \lambda) = \log p(Y ; \theta)  - \KL{q(X;\lambda)}{p(X|Y ; \theta)}.
\end{align}

While the ELBO is typically introduced as an objective for learning the variational parameters $\lambda$, we can also use it as an objective for learning the model parameters $\theta$.
To see why this is so, consider the hypothetical case of an ``infinite capacity'' variational family $q^*(X ; \lambda)$, for which it is possible to exactly match the posterior. In this family, there exists some value $\lambda$ such that the KL divergence is zero
\begin{equation}
  \min_\lambda\: \KL{q^*(X;\lambda)}{p(X|Y ; \theta)} = 0.
\end{equation}
If we use this infinitely flexible family in a variational bound, then $\max_\lambda \mathcal{L}^*(Y; \theta, \lambda) = \log p(Y ; \theta)$ and maximizing the ELBO with respect to both $\theta$ and $\lambda$ is equivalent to maximum likelihood estimation,
\begin{align}
  \max_\theta \: \max_\lambda \: \mathcal{L}^*(Y; \theta, \lambda) = \max_\theta \: \log p(Y ; \theta).
\end{align}
Note that the inner maximization resolves the ``moving target'' problem, at the cost of needing to solve an optimization problem for $\lambda$ at every step of the outer optimization problem for $\theta$. If we could somehow solve this problem for an infinite-capacity variational family, then we would have a perfect proposal $q^*(X ; \lambda) = p(X \mid Y, \theta)$. For such a perfect proposal, the importance weight is equal to the marginal likelihood,
\begin{align}
    w &= 
    \frac{p(Y,X;\theta)}{q^*(X ; \lambda)}
    = p(Y ; \theta).
\end{align}
This means that the self-normalized estimator from Equation~\eqref{eq:iw-grad-log-py} simplifies to a normal Monte Carlo estimator, all weights are equal, and that the resulting estimate of the gradient is unbiased.

In practice, we will not have an infinite capacity variational family, and we will typically not fully solve the inner optimization problem for $\lambda$ at every gradient step for $\theta$. This  means that there will be a difference between maximizing the ELBO and maximizing the marginal likelihood. This difference manifests itself as an extra term in the gradient
\begin{align}
    \nabla_\theta \: \mathcal{L}
    =
    \nabla_\theta \: \log p(Y ; \theta)
    -
    \nabla_\theta \:
    \KL{q(X;\lambda)}{p(X \mid Y ; \theta)} 
    .
\end{align}
In this gradient, the second term prevents gradient updates to $\theta$ from making changes to the model that strongly increase the KL relative to the variational approximation. This is sometimes argued to be beneficial, in the sense that it acts as a form of regularization that prevents overfitting in the generative model \citep{shu2018amortized}, or in the sense that it stabilizes the optimizer \citep{schulman2015trust}. However, it can also lead to approximation errors in the learned generative model. This is particularly true when we use a fully-factorized variational family, as we did in black-box variational inference
\begin{align}
  q(X ; \lambda) = \prod_{x \in X} q(x ; \lambda_x).
\end{align}
In this family, all variables are uncorrelated by construction. This means that there is a limit to how well we can approximate the posterior. Intuitively, the best we can do is to choose each individual factor to match the corresponding marginal of the posterior $q(x ; \lambda_x) = p(x \,|\, Y; \theta)$. This bounds the the KL divergence to
\begin{align}
  &\KL{q(X;\lambda)}{p(X|Y ; \theta)}
  =
  \Ev_{q(X; \lambda)}
  \left[
    \log 
    \frac{q(X ; \lambda)}
         {p(X \,|\, Y; \theta)}         
  \right],
  \\
  \nonumber
  &\qquad=
  \Ev_{q(X; \lambda)}
  \left[
    \log 
    \frac{\prod_x q(x ; \lambda_x)}
        {p(X \,|\, Y; \theta)}
    +
    \log 
    \frac{\prod_x p(x \,|\, Y; \theta)}
         {\prod_x p(x \,|\, Y; \theta)}
  \right],
  \\
  \nonumber
  &\qquad=
  \Ev_{q(X; \lambda)}
  \left[
    \log 
    \frac{\prod_x p(x \,|\, Y; \theta)}
         {p(X \,|\, Y; \theta)}         
  \right]
  +
  \sum_x 
  \underbrace{\KL{q(x;\lambda_x)}{p(x|Y ; \theta)}}_{\ge 0},
  \\
  \nonumber
  &\qquad\ge
  \Ev_{q(X; \lambda)}
  \left[
    \log 
    \frac{\prod_x p(x \,|\, Y; \theta)}
         {p(X \,|\, Y; \theta)}         
  \right].
\end{align}
In other words, unless we happen to have a posterior $p(X \,|\, Y; \theta)$ in which all variables are uncorrelated, there will be an ``approximation gap'' between the best approximation and the true posterior. This will have knock-on effects in terms of model learning because the the gradient will be biased in a way that is difficult to characterize.  
Optimizing the ELBO will balance maximizing $\log p(Y;\theta)$ against minimizing $\KL{q(X;\lambda)}{p(X|Y;\theta)}$. This can be seen as a bias towards learned $p(X|Y; \theta)$ that are ``compatible'' with performing variational inference in using the variational family $q(X;\lambda)$.


%
%
%

\subsection{Approximating Gradients of Variational Objectives} 

So far in this book, we have looked at one particular method for variational inference in Section~\ref{sec:eval-bbvi}. This method, black-box variational inference (BBVI) minimizes the exclusive KL divergence by approximating the gradient of the ELBO using a likelihood-ratio estimator. In this section, we will discuss two alternative strategies for stochastic variational inference. The first are methods that minimize the \emph{inclusive} KL divergence (rather than the \emph{exclusive} KL divergence) by approximating its gradient using self-normalized importance sampling. The second are reparameterized gradient estimators, which were popularized as an alternative to likelihood-ratio estimators in the context of variational autoencoders \citep{kingma2014auto,rezende2014stochastic}.


\subsubsection{Self-normalized Gradients of the Inclusive KL Divergence}
\label{sec:inclusive-kl}

Variational inference is often taken as synonymous with maximizing the ELBO, which equates to minimizing the exclusive KL divergence
\begin{align}
    \max_{\lambda} \mathcal{L}(Y ; \theta, \lambda)
    =
    \min_{\lambda}
    \KL{q(X ; \lambda)}{p(X \mid Y ; \theta)}
    .
\end{align}
However, this is certainly not the only conceivable variational objective; we can in principle minimize any number of other divergences between the posterior and the variational distribution. The exclusive KL divergence is a member of a family of divergences known as $f$-divergences
\begin{align}
    D_f \big(
      p(X \mid Y ; \theta)
      \:\big\vert
      \big\vert\:
      q(X ; \lambda)
   \big)
   = 
   \Ev_{q(X ; \lambda)}
   \left[
     f \left(
       \frac{p(X \mid Y ; \theta)}
            {q(X ; \lambda)}
     \right)
   \right]
   .
\end{align}
This family of divergences is parameterized by a function $f(\omega)$. We recover the exclusive KL when we use $f(\omega) = - \log \omega$. For other choices of $f(\omega)$ we obtain the Jensen-Shanon divergence, the $\alpha$-divergence, the $\chi^2$-divergences, and the Hellinger distance. 

The majority of these divergences are currently not widely used as objectives in practice. Unfortunately, changing the divergence in variational inference is not as trivial as, say, changing the loss function in a typical deep learning application. One reason for this is that $f(\omega)$ depends on the importance ratio $\omega = p(X \mid Y ; \theta) / q(X ; \lambda)$, which is intractable. This means that different approximation strategies are needed to compute the gradient of each divergence; we cannot simply change the divergence without deriving new gradient estimators.

However, there is one other divergence that is commonly used when performing variational inference: the inclusive KL divergence
\begin{align}
\label{eq:ch7-kl-pq}
  \KL{p(X|Y ; \theta)}{q(X;\lambda)}
  :=
  \Ev_{p(X|Y ; \theta)}
  \left[
    \log \frac{p(X|Y ; \theta)}{q(X;\lambda)}
  \right].
\end{align}
The inclusive KL divergence is an $f$-divergence with $f(\omega)=\omega \log \omega$. It differs from the exclusive KL divergence in that the arguments to the divergence are flipped, which means that the expectation is taken with respect to the posterior $p(X | Y; \theta)$, rather than the approximating distribution $q(X ; \lambda)$ as in Equation~\eqref{eq:ch7-kl-pq}. While this difference may seem innocuous at first glance, it fundamentally affects how we compute gradient estimates, as well as the properties of the learned $q(X;\lambda)$.

\paragraph{Mode-seeking vs Mode-covering Approximations}
Let us first take a moment to understand how the inclusive KL divergence differs from the exclusive KL divergence, and indeed why we use the terms ``inclusive'' and ``exclusive'' to refer to these divergences. 
The KL divergence is asymmetric with respect to its arguments, which gives rise to an asymmetric notion of similarity between distributions. This asymmetry is most noticeable in regions of the sample space where the posterior $p(X | Y; \theta)$ approaches 0. 
Minimizing the exclusive divergence (Eq.~\eqref{eq:ch7-kl-qp}) will heavily penalize any approximations $q(X;\lambda)$
that assign nonzero probability mass to areas which have zero probability under $p(X|Y ; \theta)$.
Conversely, minimizing Eq.~\eqref{eq:ch7-kl-pq} will heavily penalize approximations in which $q(X; \lambda)$ ascribes near-zero probability to areas of the posterior with positive probability.

In Figure~\ref{fig:which-kl} we illustrate how this difference  manifests itself when learning a variational distribution. When we fit a unimodal variational distribution to a multimodal posterior, we see that minimizing the exclusive KL results in a variational distribution yields a good fit to one of the modes whilst ``excluding'' the second mode. This is sometimes known as ``mode seeking'' behavior \citep{cappe2008adaptive,cornuet2012adaptive}.  By contrast, minimizing the inclusive KL divergence results in a broader, more ``inclusive'' approximation that to avoid missing areas of high probability. This is also known as ``mode covering'' behavior. 

Differences in the resulting approximation often become particularly pronounced when the variational distribution is fully factorized. This means that there will be an approximation gap, since the variational family is not sufficiently expressive to capture correlations between latent variables. In Figure~\ref{fig:which-kl}, illustrate the difference between the exclusive and the inclusive approximation of a posterior with two strongly-correlated variables. We see that minimizing the exclusive divergence underapproximates the posterior variance, whereas the minimizing the inclusive divergence overapproximates the posterior variance.



\begin{figure*}[t]
\centering
\includegraphics[width=0.45\textwidth]{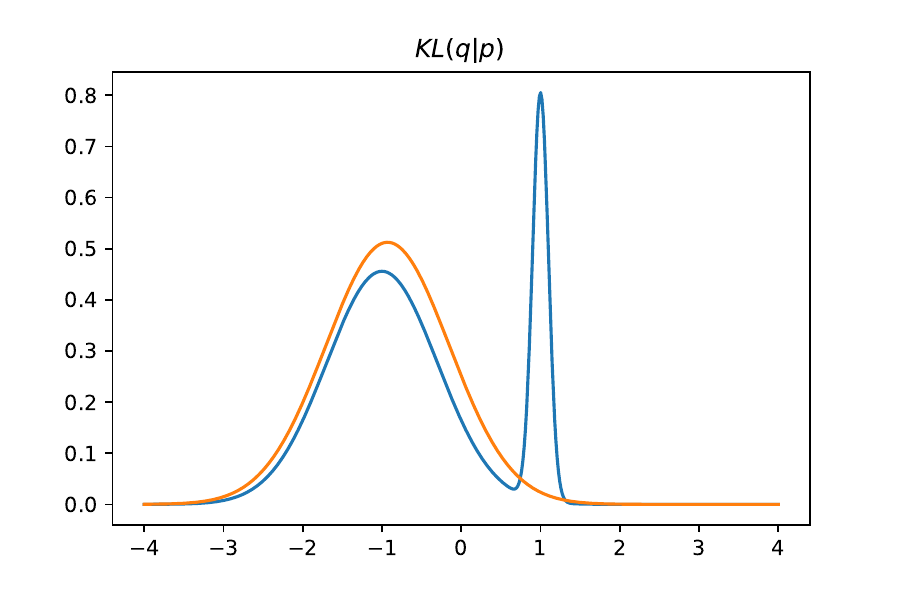}
\includegraphics[width=0.45\textwidth]{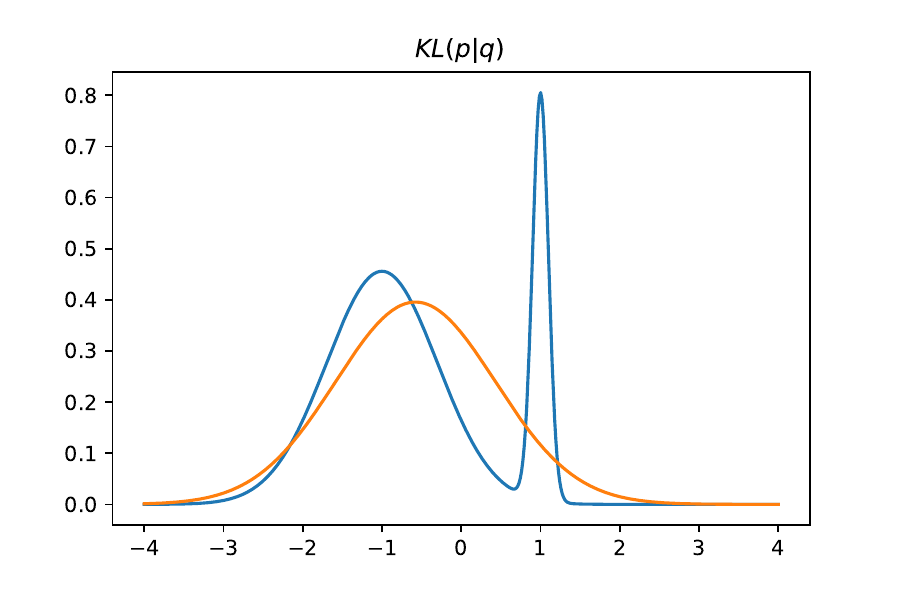}
\includegraphics[width=0.45\textwidth]{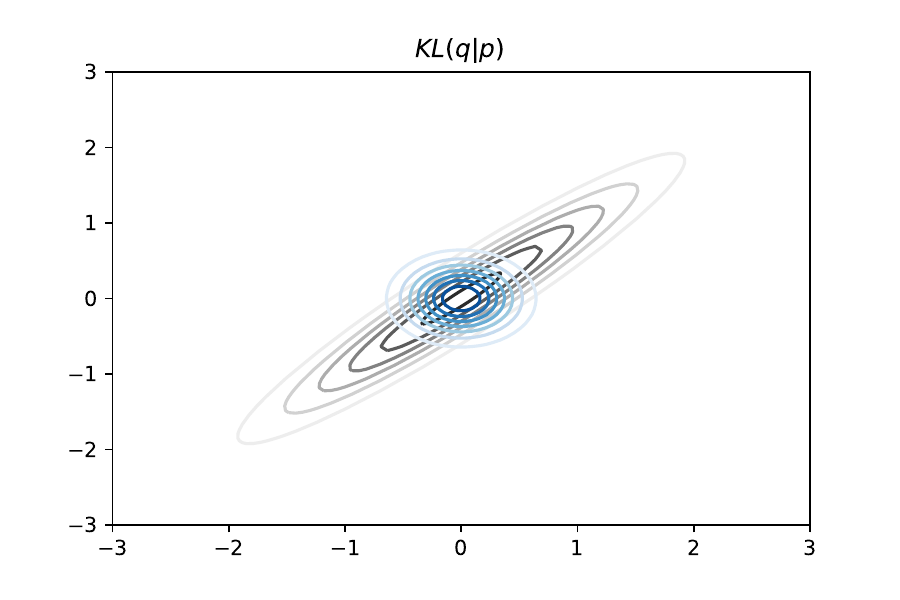}
\includegraphics[width=0.45\textwidth]{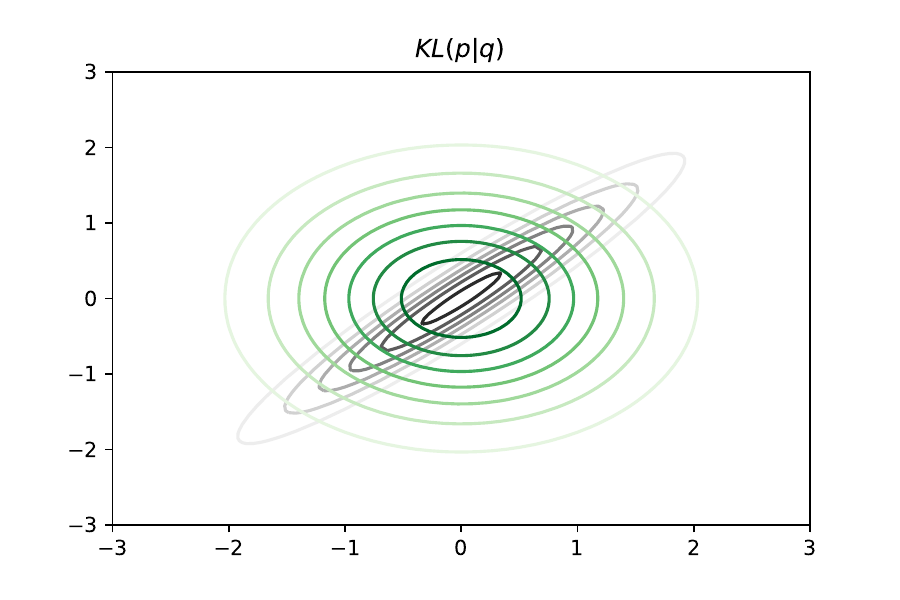}
\caption{Different KL divergences have different minima, when $q(X; \lambda)$ is insufficiently flexible to fully characterize the posterior $p(X | Y)$.}
\label{fig:which-kl}
\end{figure*}

\paragraph{Self-Normalized Gradient Estimation} Using the inclusive KL divergence simplifies gradient computations in one sense and complicates them  in another. What simplifies the computation is the fact that we are computing the gradient of an expectation with respect to the posterior $p(X \mid Y ; \theta)$, which does not depend on $\lambda$. This means that we can move the gradient operator directly into the expectation
\begin{align}
  \label{eq:grad-kl-pq}
  \nabla_\lambda \KL{p(X|Y ; \theta)}{q(X;\lambda)}
  &=
  \Ev_{p(X|Y ; \theta)}
  \left[
    \nabla_\lambda \log \frac{p(X|Y ; \theta)}{q(X;\lambda)}
  \right].
\end{align}
By contrast, the ELBO and the exclusive KL (and indeed all other $f$-divergences) are expectations with respect to $q(X ; \lambda)$, which depends on $\lambda$. Computing the gradient of the gradient of the expectation therefore requires a likelihood-ratio estimator, as is used in BBVI, or a reparameterized estimator, as we will discuss below. Computing the gradient of the inclusive KL is much simpler in this respect, since no likelihood-ratio estimators or reparameterization are needed.

What complicates gradient approximation is that we have to generate samples from the posterior $p(X | Y; \theta)$ rather than samples from $q(X; \lambda)$. This is precisely the same computational problem that we encountered when approximating the gradient of the marginal likelihood in equation~\eqref{eq:grad-log-marginal}; we need some form of inference method to approximate the gradient. This means that we can once again use the variational distribution as a proposal and define a self-normalized estimator analogous to the one in equation~\eqref{eq:iw-grad-log-py}
\begin{align}
  \nabla_\lambda \KL{p(X|Y ; \theta)}{q(X;\lambda)}
  \simeq
  \frac{1}{\hat{Z}(\theta)}
  \frac{1}{L}
  \sum_{l=1}^L
  w^l \:
  \nabla_\lambda \: \log q(X^l ; \lambda). \label{eq:is-grad-inc-kl}
\end{align}


A general theme that emerges here is that variational inference and importance sampling can be mutually beneficial. We can use learned proposals to generate importance-weighted samples from the posterior $p(X | Y ; \theta)$. The resulting self-normalized approximations to the posterior can in turn be used to learn a better proposal $q(X;\lambda)$. 

In this context, minizing the inclusive KL divergence has advantages over minimizing the exclusive KL divergence. Importance sampling can ``correct'' for an overapproximation of the posterior variance by assigning low weights to samples in regions with low posterior density. Conversely, it is more difficult to correct for an underapproximation of the posterior variance, since this requires assigning high weights to outlier samples in regions with high posterior density, but a low proposal density, which means that we will generally need a much larger number of proposals to ``cover'' the posterior.


We also see that we can use importance sampling to combine model learning and variational inference in a straightforward manner. Given a set of proposals $X^l \sim q(X^l ; \lambda)$ the estimator in Equation \eqref{eq:iw-grad-log-py} approximates the gradient of the marginal likelihood and the corresponding estimator from Equation~\eqref{eq:is-grad-inc-kl} approximates the gradient of the inclusive KL divergence. This means that we can perform stochastic gradient descent on both objectives using the same set of samples.

At the same time, this inference strategy does have limitations. Even when learning the proposal using variational inference, maximum likelihood estimation using importance sampling requires a reasonably large number sample budget $L$ to overcome the bias in the self-normalized estimator. Sample budgets in the range of 10 to 1000 are common when learning deep generative models, which is similar to the sample budget that we might employ in BBVI.
\subsubsection{Reparameterized Gradients of the Exclusive KL Divergence} 

We have now discussed two strategies for learning the variational distribution $q(X; \lambda)$.  Both strategies use a Monte Carlo estimator to approximate the gradient of a divergence, and both have advantages and disadvantages. In BBVI, which we discussed in Section~\ref{sec:eval-bbvi}, we minimize the exclusive divergence by maximizing the ELBO. We showed that we can express the gradient of the ELBO as
\begin{align}
\nabla_{\lambda} \mathcal{L}(\lambda)=\mathbb{E}_{q(X ; \lambda)}\left[\nabla_{\lambda} \log q(X ; \lambda)\left(\log \frac{p(Y, X)}{q(X ; \lambda)}-b\right)\right]
\end{align}
We can approximate this expectation using samples $X^l \sim q(X ; \lambda)$, which is known as a likelihood-ratio estimator. This estimator is easy to compute, but can have a high variance. This means that each gradient step requires a moderate to large number of samples.

The second strategy, which we just discussed in the preceding section, is to minimize the inclusive divergence, which is an expectation with respect to the posterior $p(X | Y ; \theta)$. This avoids the need for a likelihood-ratio estimator, since the posterior does not depend on the variational parameters $\lambda$. However, we now have to perform importance sampling to approximate  the posterior, which often requires a computational budget that is similar to that of likelihood-ratio estimators.

In this section, we will discuss a third strategy, which often makes it possible to approximate the gradient of an exclusive divergence with a comparatively small number of samples. This strategy relies on \emph{reparameterization} to express the expectation with respect to $q(X; \lambda)$ in terms of an expectation with respect to a parameter-free distribution, which in turn simplifies the gradient computation. This type of derivative, which is also known as a pathwise derivative (see \cite{Mohamed2019Monte} for a discussion), gained popularity in machine learning in the context of variational autoencoders \citep{kingma2014auto,rezende2014stochastic}, but can approximate gradients for any computation graph that combines deterministic and stochastic nodes \citep{schulman2015gradient}.

To illustrate the idea of reparameterization, we begin by considering a simple case. Suppose that we have a variational distribution $q(x ; \lambda)$ on a single variable $x$ that has the form of a Gaussian with parameters $\lambda = \{\mu, \sigma\}$. We can sample from this distribution by applying a deterministic transformation $g(\epsilon; \lambda)$ to a standard normal
\begin{align}
  x &= g(\epsilon; \lambda) =  \mu + \sigma \: \epsilon,
  &
  \epsilon &\sim \text{Normal}(0, 1). \label{eq:std-norm-reparam}
\end{align}
This construction ensures that $x$ is a Gaussian variable with mean $\mu$ and standard deviation $\sigma$. In other words, computing $x$ from a randomly sampled $\epsilon$ is equivalent to sampling $x$ from $q(x ; \lambda)$. This means that we can rewrite any expectation of a function $r(x)$ with respect to $q(x ; \lambda)$ in terms of an expectation with respect to a distribution $p(\epsilon)$ that has no learnable parameters
\begin{align}
  \mathbb{E}_{q(x ; \lambda)}
  \left[
  r(x)
  \right]
  =
  \mathbb{E}_{p(\epsilon)}
  \left[
  r(g(\epsilon; \lambda))
  \right]
  . \label{eq:example-reparam} 
\end{align}
This ``reparamerization trick'' is very useful when we want to compute the gradient of this expectation with respect to $\lambda$. Because $p(\epsilon)$ has no dependence on $\lambda$, we can express the gradient as
\begin{align}
  \nabla_\lambda
  \mathbb{E}_{q(x ; \lambda)}
  \big[
  r(x)
  \big]
  =
  \mathbb{E}_{p(\epsilon)}
  \left[
  \left.
  \frac{\partial r(x)}
       {\partial x}
  \right\vert_{x=g(\epsilon; \lambda)}
  \nabla_\lambda g(\epsilon; \lambda)
  \right]
  .
\end{align}
In the equation above, we first compute the derivative $\partial r(x) / \partial x$ at the sampled value $x=g(\epsilon; \lambda)$ and then simply apply the chain rule of differentiation and compute the gradient of the sampled value with respect to the parameters. The result is that we can rewrite the gradient of the expectation as an expectation of a gradient. We can now easily approximate this gradient from samples $\epsilon^l \sim p(\epsilon)$
\begin{align}
  \nabla_\lambda
  \mathbb{E}_{q(x ; \lambda)}
  \big[
  r(x)
  \big]
  \simeq
  \frac{1}{L}
  \sum_{l=1}^L
  \left(
  \left.
  \frac{\partial r(x)}
       {\partial x}
  \right\vert_{x=g(\epsilon^l; \lambda)}
  \nabla_\lambda g(\epsilon^l; \lambda)
  \right)
  .
\end{align}
Reparameterized gradients can be used to optimize any objective that can be expressed as an expectation of this form. This includes the ELBO, which is an expectation with respect to $q(X; \lambda)$ of the function
\begin{align}
  r(Y, X, \theta, \lambda) = \log \frac{p(Y,X ; \theta)}{q(X ; \lambda)}.
\end{align}
This means that we can express the ELBO as an expectation with respect to a parameter-free distribution $p(\epsilon)$
\begin{align*}
  \mathcal{L}(Y; \theta, \lambda)
  =
  \Ev_{q(X;\lambda)}
  \left[
    \log \frac{p(Y,X; \theta)}{q(X; \lambda)}
  \right]
  =
  \Ev_{p(\epsilon)}
  \left[
    \log \frac{p(Y,g(\epsilon; \lambda); \theta)}{q(g(\epsilon; \lambda); \lambda)}
  \right].
\end{align*}
Relative to likelihood-ratio and importance-weighted estimators, reparameterized estimators tend to have a lower variance, and therefore tend to be more sample-efficient. In fact, in many cases it is possible to obtain useful gradient estimates using a single sample from $p(\epsilon)$. 

The concept of reparameterization is not unique to Gaussian distributions. The main requirement for reparameterization is that we have to be able to define a differentiable transformation $x = g(\epsilon; \lambda)$. For many continuous univariate distributions, we can transform a variable $\epsilon$ that is uniform on the interval $[0,1]$ using the inverse of the cumulative distribution function
\begin{align}
  g(\epsilon; \lambda) &= F^{-1}_q(\epsilon; \lambda), 
  &
  F_q(x; \lambda)
  =
  \int_{x' \le x}
  dx' \:
  q(x' ; \lambda)
  .
\end{align}
Reparameterization is also possible for certain distributions where inverse transform sampling is not stable, including the gamma, beta, and Dirichlet distributions \citep{jankowiak2018pathwise}.

One of the main limitations of reparameterization is that it is not compatible with discrete random variables. While it is often possible to define a transformation $x = g(\epsilon; \lambda)$ for discrete variables, this tranformation will not be differentiable. A second requirement is that the function $r(Y, X, \theta, \lambda)$ must not only be differentiable with respect to $\theta$ and $\lambda$, but must also be differentiable with respect to $X$.

This second requirement reveals a significant implementation challenge that arises when using reparameterized estimators for variational inference in probabilistic programming: unlike the BBVI likelihood-ratio estimator, this pathwise gradient requires a differentiable model density, since by the multivariable chain rule, the gradient of the ELBO will contain the term
\begin{align*}
  &
  \nabla_\lambda 
  \mathbb{E}_{q(X; \lambda)}
  \big[
     \log p(Y, X ; \theta) 
  \big]
  = \\
  &\qquad
  \mathbb{E}_{p(\epsilon)}
  \left[
    \:
    \sum_{x \in X}
    \left.
    \frac{\partial \log p(Y, X; \theta)}
         {\partial x}
    \right\vert_{x = g_x(\epsilon; \lambda)}
    \nabla_\lambda \: g_x(\epsilon; \lambda)
  \right].
\end{align*}
This implies that reparameterization can only be used in situations where the model density $p(Y,X ; \theta)$ is itself differentiable with respect to the latent variables $X$. As it happens, we have actually already dealt with this requirement in the context of Hamiltonian Monte Carlo methods in Section~\ref{sec:hmc-inference}.
Here the requirements are exactly the same: we need gradients of the model density with respect to $X$, which requires automatic differentiation. While this adds additional restrictions and implementation challenges relative to the BBVI algorithm (which required only gradients of the proposal $q(X; \lambda)$), this additional computational complexity is offset by the sample efficiency of these estimators.

In later sections of this chapter we will show how to use gradient-based learning with both reparameterizable and non-reparameterizable random variables.  Dealing with models that have this mixture of random variable times, and in the context of HOPPL-like model specification languages, a potentially variable number of them, is what we are building towards.  It is a complex business so we are building towards it slowly.


%
%
%
%
\subsection{Amortized Inference}
\label{sec:amortized-inference}

The last subject that we will discuss before returning to implementation aspects of deep probabilistic programming is amortized inference. 
The variational methods that we have reviewed up to this point are designed to perform inference for a given model exactly {\em once}, on a single fixed dataset. 
As we discussed at the start of this chapter, this is not a practical approach in most modern ML and AI applications, where we will typically want to train a model on a large dataset that comprises thousands or even millions of examples.
Morever, we are often interested in applying the learned model to perform inference on previously unseen data at test time, and we would like this inference to be fast.  
This type of rapid, repeated inference has been described as {\em amortized} inference \citep{gershman2014amortized}, in reference to the fact that we pay an up-front computational cost at training time to get an artifact that is useful for rapid inference at test time.




When we are in the situation where we need to do rapid amortized inference, we cannot wait for either one of the asymptotic algorithms from earlier chapters to converge for each unique input $Y_n$ in a dataset, nor can we wait to learn unique variational distribution $q(X_n ; \lambda_n)$ by optimizing $\lambda_n$ to convergence. What we want instead is some kind neural network $\lambda(Y_n, \phi)$, parameterized by $\phi$, that takes data $Y_n$ as input and returns a corresponding set of variational parameters. This network, which we will come to interchangeably refer to as an ``inference network'' or ``encoder,'' then defines a data-dependent variational distribution
\begin{align}
  q(X_n \mid Y_n ; \phi)
  =
  q(X_n \,;\, \lambda(Y_n, \phi)). \label{eq:data-dependent-q}
\end{align}
From a mathematical point of view, all derivations of gradient estimators in this chapter remain perfectly valid if we replace directly optimizable parameters $\lambda_n$, which are unique to each input, with the output of a network $\lambda(Y_n ; \phi)$, whose weights $\phi$ are shared among inputs. In particular, we can  learn $\phi$ by minimizing the exclusive KL using likelihood-ratio or reparameterized estimators, or we can learn $\phi$ by minimizing the inclusive KL using a self-normalized estimator. The only thing that changes is that will now compute gradients with respect to the network weights $\phi$ to indirectly optimize $\lambda(Y_n ; \phi)$ for each input, rather than with respect to directly optimizable parameters $\lambda_n$. Morover, we can use the learned amortized proposal to learn model parameters $\theta$ in exactly the same way as we have described so far. 

The idea of learning amortized variational distributions has a long history in ML and AI research in the context of variational autoencoders and wake-sleep methods for Helmholtz machines. In the remainder of this section, we will place the amortized methods in this book in context by explaining how they are used 
when training these models.


\subsubsection{Variational Autoencoders} 

Variational autoencoders (VAEs) combine learning and amortized inference to train a deep generative model and an inference model. In their original formulation, these models were unstructured, which is to say that they that they defined a single vector-valued latent variable $X$ and an a single vector-valued observed variable $Y$ in the form of a flattened image \citep{kingma2014auto,rezende2014stochastic}. The generative model in an unstructured VAE combines a Gaussian prior $p(X)$ with a neural likelihood $p(Y \mid X ; \theta)$ that is defined in terms of a network $\eta(X, \theta)$. In the inference model, a network $\lambda(Y, \phi)$ defines a Gaussian variational distribrution $q(X \mid Y ; \phi)$,
\begin{align}
    p(Y, X ; \theta) &= p(Y ; \eta(X, \theta)) \: p(X), \\
    q(X \mid Y ; \phi) &= q(X ; \lambda(Y, \phi)).
\end{align}
In this context, the network $\lambda(Y ; \phi)$ is known as an ``encoder'', in reference to the fact that the latent vector $X$, also known as the ``code'', is a compressed representation of the original data. Conversely, the network $\eta(X ; \theta)$ is known as the ``decoder'', in reference to the fact that it attempts to  reconstructs the original input $Y$ from the latent code $X$.

Since their introduction, VAEs have been generalized to structured deep generative models, in which there may be multiple latent variables and observations. At the start of this chapter, we consdired a simple example of such a structured VAE, in which both the generative model \hop{p} and the inference model \hop{q} were defined as programs in the HOPPL,
\begin{hoppl}
(defn p [y $\eta$ $\theta$]
  (let [z (sample (multinomial 1 $\theta^z$)) 
        v (sample (normal ($\eta^v_\mu$ z $\theta^v$) ($\eta^v_\sigma$ z $\theta^v$)))]
    (observe (normal ($\eta^y_\mu$ v $\theta^y$) ($\eta^y_\sigma$ v $\theta^y$)) y)
    [z v]))

(defn q [y $\lambda$ $\phi$]
  (let [z (sample (multinomial 1 ($\lambda^z$ y $\phi^z$))) 
        v (sample (normal ($\lambda^v_\mu$ y z $\phi^v$) ($\lambda^v_\sigma$ y z $\phi^v$)))]
    [z v]))
\end{hoppl}
More generally, variants of structured VAEs have been applied to a wide range of unsupervised tasks, including activity recognition \citep{johnson2016structured}, object recognition \citep{eslami2016attend}, next frame prediction in video \citep{denton2017unsupervised}, neural topic models \citep{miao2016neural,srivastava2017autoencoding,esmaeili2018structuredb}, and neuroimaging analysis \citep{sennesh2020neural}. 

Both structured and unstructured VAEs are trained by maximizing an ELBO with respect to both $\theta$ and $\phi$. Given training data $\{Y_1, \dots, Y_N\}$, we can define the generative model and the inference model as a product over independent and identically distributed instances 
\begin{align}
    p(Y_1, \dots, Y_N, X_1, \dots, X_N ; \theta)
    &= \prod_{n=1}^N p(Y_n, X_n ; \theta),\\
    q(X_1, \dots, X_N \mid Y_1, \dots, Y_N ; \phi)
    &= \prod_{n=1}^N q(X_n \mid Y_n ; \theta).
\end{align}
This means that we can express the ELBO on the training data as a whole as a sum of ELBO terms for individual examples,
\begin{align}
    \begin{split}
    \mathcal{L}(Y_1, \dots, Y_N ; \theta, \phi)
    &=
    \sum_{n=1}^N
    \mathcal{L}(Y_n ; \theta, \phi),\\
    &=
    \sum_{n=1}^N
    \mathbb{E}_{q(X_n \mid Y_n ; \phi)}
    \left[
        \log 
        \frac{p(Y_n, X_n ; \theta)}
             {q(X_n \mid Y_n ; \phi)}
    \right],\\
    &\le
    \log p(Y_1, \dots, Y_N ; \theta).
    \end{split}
\end{align}

To train VAEs, we optimize this objective with respect to $\theta$ and with respect to $\phi$. However, as we discussed in Section~\ref{sec:max-likelihood}, computing a sum over $N$ data points at each gradient step is not computationally feasible. 
Because of this, we typically approximate the gradient using a ``mini-batch'' of $B$ samples selected uniformly at random from the training data without replacement. 
\begin{align}
  \nabla_{\theta,\phi} \:
  \mathcal{L}(Y_1, \dots, Y_N ; \theta, \phi)
  &\simeq
  \frac{N}{B}
  \sum_{b=1}^B
  \
  \nabla_{\theta,\phi} \:
  \mathcal{L}(Y^b ; \theta, \phi)
  .
\end{align}
Doing this allows us to better approximate the gradient while still not performing a full pass over the training data at each gradient step.

Unstructured variational autoencoders use only reparameterizable variational distributions $q(X_n \,;\, \lambda(Y_n, \phi))$.  This means that the inference network produces the parameters $\lambda$ of the reparameterizable distribution in manner that is analogous to Equation~\eqref{eq:std-norm-reparam}, which means that we can substitute $X^b = g(\epsilon, \lambda(Y^b, \phi))$ to compute a reparameterized gradient of the per-sample ELBO
\begin{align*}
  \nabla_{\theta,\phi} \:
  &\mathcal{L}(Y^b \,;\, \theta, \phi) \\
  &=
  \mathbb{E}_{p(\epsilon)}
  \left[
    \nabla_{\theta,\phi} \:
    \log
    \frac{p(Y^b ; \eta(g(\epsilon, \lambda(Y^b, \phi)),\theta)) p(g(\epsilon, \lambda(Y^b, \phi)))}
         {q(g(\epsilon, \lambda(Y^b, \phi)); \lambda(Y^b, \phi))}
  \right]
  .
\end{align*}


In the context of probabilistic programming, variational autoencoders present an opportunity to design models that inherit the flexibility and scalability of neural networks, but can incorporate domain knowledge and inductive biases when needed. To do so, we can use probabilistic programs to define structured deep generative models $p(Y, X ; \theta)$ that incorporate neural networks as learnable deterministic functions.   
This is arguably the single most important aspect of deep probabilistic programming: the ability to partially specify the generative model; encoding the parts of the model you do know, and letting model learning and generic deep neural network function approximation take care of the rest.

However, applying stochastic varational inference to structured VAE in the form of probabilistic programs does give rise to some subtleties. In particular, programs may include variables that are not reparameterizable, such as the multinomial variable in the programs \hop{p} and \hop{q}. A substantial amount of work in ML and AI on VAEs has focused on eliminating discrete variables, for example by replacing them with continuous relaxations \citep{maddison2017concrete,jang2017categorical}. However such relaxations are not always practical in the context of probabilistic programming, because they can force evaluation of all branching paths, which is not possible to do in general in models that employ ``stochastic control flow'' \citep{le2019revisiting}.

One alternative to employing continuous relaxations is to combine reparameterization and likelihood-ratio estimation in models with mixed variables \citep{schulman2015gradient}. We will revisit how to implement such gradient estimators in Section~\ref{sec:webppl-proposals}, where we discuss variational inference with WebPPL-style proposals \citep{ritchie2016deep}. A second alternative, which we previously discussed in Section~\ref{sec:inclusive-kl} is to minimize the inclusive KL divergence, rather than the exclusive divergence. This idea has a long history in the context of wake-sleep methods for Helmholtz machines and their generalizations to reweighted wake-sleep methods.


\subsubsection{Helmholtz Machines} 
\label{sec:helmholtz-machines}

Variational auto-encoders are a modern, reparameterized variational inference-based take on an excellent old architectural idea inspired by cognitive and neuroscience, the Helmholtz machine  \citep{dayan1995helmholtz}.  The basic structural components of a Helmholtz machine are the same as that which we described for VAEs; a stochastic neural network generative model of observations of world and an amortized inference neural network that encodes the current observation into a high-dimensional latent vector computationally efficiently.  The key difference between Helmholtz machines and VAEs is the objective used to train the inference network.  VAEs, as described, jointly optimize a single objective, the ELBO.  Helmholtz machines use two different objectives; one for the inference network called ``sleep'' and another for the generative model called ``wake.''  Recent work on the the ``thermodynamic variational objective'' provides a unifying view of these objectives  \citep{masrani2019thermodynamic}.  The overall training algorithm for Helmholtz machine is called the ``wake-sleep'' algorithm. 

\paragraph{Wake-Sleep}
The original wake-sleep algorithm of \citet{dayan1995helmholtz} makes use of a combination of real data and data that is simulated from the learned generative model. The ``wake phase'' trains the generative model on samples from the data distribution $p^\text{data}(Y)$.  The sleep phase trains the inference network using samples from the generative model.  
Wake-sleep maximizes the expected lower bound with respect to $\theta$ using a single sample variant of the estimator in Equation~\eqref{eq:iw-grad-log-py}. Wake-sleep then trains the proposal distribution using the inclusive KL (rather than the exclusive KL).  



To be specific about the ``sleep phase'' of wake-sleep, it trains the inference network to match the learned generative model posterior $p(X | Y; \theta)$ averaged over samples $p(Y ; \theta)$ from the generative model. 
\begin{align*}
  &-\nabla_\phi \:
  \mathbb{E}_{p(Y \,;\, \theta)}
  \left[
    \text{KL}
    \big(
      p(X \mid Y ; \theta) 
      \:\big\|\: 
      q(X \mid Y ; \phi) 
    \big)
  \right] \\
  &\qquad\qquad
  =\mathbb{E}_{p(Y \,;\, \theta) \: p(X \mid Y \,;\, \theta)}
  \left[
     -\nabla_\phi 
     \log \frac{p(X \mid Y ; \theta)}{q(X \mid Y ; \phi)}
  \right] \\
  &\qquad\qquad
  =  
  \mathbb{E}_{p(Y, X \,;\, \theta)}
  \left[
    \nabla_\phi 
    \log q(X \mid Y; \phi)
  \right]
  .
\end{align*}
By averaging over $Y$ from the model,  it is easy to estimate this gradient, since we can always generate samples $Y^b, X^b \sim p(Y, X ; \theta)$ from the joint.
\begin{align*}
  \nabla_\phi \;
  \mathbb{E}_{p(Y ; \theta)} \!
  \left[
    \text{KL}
    \big(
      p(X \,|\, Y ; \theta) 
      \,\big\|\, 
      q(X \,|\, Y ; \phi) 
    \big)
  \right] 
  \simeq
  \frac{1}{B}
  \sum_{b=1}^B
  \nabla_\phi \; \log q(X^b \,|\, Y^b ; \phi) . 
\end{align*}

\paragraph{Reweighted Wake-Sleep}
While wake-sleep is a perfectly valid learning algorithm, the learning signal for the inference network can be quite poor at the beginning of learning because the model might not be good enough at first to generate samples near the true data distribution.  The effect of this can be sufficiently severe to negatively impact model learning as well because bad approximate posterior samples for learning the model parameters can lead to biased gradients, etc.  The \emph{reweighted} wake-sleep algorithm by \citet{bornschein2014reweighted} addresses this problem by modifying the wake-sleep algorithm to use importance weighting throughout, but particularly in the wake-phase updates to $\theta$ and $\phi$. To do this, we interpret the variational distribution as a proposal and generate weighted samples
\begin{align}
  w^{b,l}
  &=
  \frac{p(Y^b, X^{b,l} ; \theta)}
       {q(X^{b,l} \,|\, Y^b ; \phi)},
  &
  X^{b,l} &\sim q(X \,|\, Y^b ; \phi),
  &
  Y^b \sim p^\text{data}(Y).
\end{align}
We then use mini-batches of these samples to approximate the expected gradient of the marginal likelihood in a manner analogous to Equation~\eqref{eq:iw-grad-log-py}
\begin{align*}
  \mathbb{E}_{p^\text{data}(Y)} \!
  \left[
    \nabla_\theta \; \log p(Y ; \theta)
  \right] 
  \simeq
  \frac{1}{B}
  \sum_{b=1}^B
  \sum_{l=1}^L
  \frac{w^{b,l}}
       {\sum_{l'=1}^L w^{b,l'}}  
  \nabla_\theta \log p(Y^b, X^{b,l} ; \theta).
\end{align*}
Similarly, reweighted wake sleep minimizes the expected inclusive KL divergence in a manner analogous to Equation~\eqref{eq:is-grad-inc-kl},
\begin{align*}
  &
  \mathbb{E}_{p^\text{data}(Y)} \!
  \left[
    -\nabla_\phi \:
    \text{KL}
    \big(
      p(X \,|\, Y ; \theta) 
      \,\big\|\, 
      q(X \,|\, Y ; \phi) 
    \big)
  \right] \simeq \\
  &\qquad\qquad
  \frac{1}{B}
  \sum_{b=1}^B
  \sum_{l=1}^L
  \frac{w^{b,l}}
       {\sum_{l'=1}^L w^{b,l'}}  
  \nabla_\phi \log q(X^{b,l} \mid Y^b ; \phi).
\end{align*}
The wake-phase update for $\theta$ differs from the update in the original wake-sleep algorithm in that we use importance sampling to approximate the gradient of the marginal likelihood, rather than computing the gradient of the lower bound, which, as described before, reduces the bias of the gradient estimate. The wake-phase update for $\phi$ differs from the sleep-phase update for $\phi$ in that we perform inference conditioned samples $Y^b \sim p^\text{data}(Y)$ from the data distribution, rather than samples $Y^b \sim p(Y ; \theta)$ from the model distribution.  This helps overall learning by ensuring that at least the data $Y$ is always from its true distribution.

The reweighted wake-sleep algorithm of \citet{bornschein2014reweighted} combines wake-phase updates to $\theta$ and $\phi$ with wake-sleep sleep-phase updates to $\phi$. In practice, it turns out that sleep-phase updates do not always aid convergence, and can in fact can be detrimental when simulated data are not representative of the training data \citep{le2019revisiting}. For this reason, it is common practice to omit the sleep-phase updates, resulting in an algorithm would be better described as a reweighted ``wake-wake'' rather than reweighted ``wake-sleep,'' since the updates to both $\theta$ and $\phi$ are directly rooted in the training data.

The reweighted ``wake-wake'' procedure combines the self-normalized gradient estimate for the marginal likelihood that we discussed in Section~\ref{sec:importance-weighted-model-learning} with the self-normalized gradient estimate for the inclusive KL divergence that we discussed in Section~\ref{sec:inclusive-kl}. The only difference is that this procedure computes using an amortized proposal $q(X_n \mid Y_n ; \phi)$ rather than using non-amortized proposals $q(X_n ; \lambda_n)$. This implies that, like all methods that minimize the inclusive KL divergence, wake-sleep methods have the advantage of not requiring reparameterization to estimate the parameters $\phi$. 
As a result, wake-sleep methods are directly and simply applicable to models with discrete variables in the generative model, or probabilistic programs with stochastic control flow.  





%
%
%
%




	\section{Design of Deep Probabilistic Programming Systems}
	\label{sec:deep-design}

Having covered methods for gradient-based learning and inference, 
we now return to implementation aspects of deep probabilistic programming systems. As we discussed at the start of this chapter, these systems use neural networks to learn amortized proposals for fast inferenc, or to parameterize partially-specified deep generative models that can be learned from data. As always, our aim is to provide enough detail that readers can both implement a deep probabilistic programming system and be proficient users of existing and future systems. 

In recent years, a large number of deep probabilistic programming systems have been developed. These include Edward \citep{tran2016edward}, Probabilistic Torch \citep{siddharth2017learning}, Pyro \citep{bingham2018pyro}, and PyProb \citep{le2016inference}. From a language-design point of view, these systems combine two sets of foundational ideas. The first are implementation strategies for probabilistic programming systems, which have been the focus of the preceding chapters in this book. The second are techniques for differentiable programming, which build on a long history of research on automatic differentiation (see \citet{baydin2017automatic} for a review), which form the basis of deep learning frameworks like  PyTorch \citep{paszke2017automatic} and TensorFlow \citep{abadi2015tensorflow}, as well as libraries for differentiable numerical computing like JAX \citep{jax2018github}.

There are a number of design questions that arise when we want to combine a generative model with an inference model. Do both models need to be programmed, or can the structure of one be derived from the other? Should a deep probabilistic programming language be a FOPPL or should it be a HOPPL? If the the latter, then how do we align variables in the generative program to variables in the inference program? And lastly, we have discussed different algorithms for parameter learning. How do we actually implement these?  And how do we deal in a general way with programs that have complex control flow and mixture of continuous and discrete random variables?

In the following sections, we will discuss concrete approaches to implementing deep probabilistic programming systems, which each formulate different answers to these questions. In Section~\ref{sec:pyprob-proposals}, we will begin by focusing on amortized inference as a means of accelerating test time inference in probabilistic programs, which is exemplified by the use case of inference in complex stochastic simulators. We will consider a setting in which the amortized inference artifact exists in the form of a completely separate amortized inference ``controller'', operating across the messaging interface that we defined in Chapter~\ref{ch:eval-two}. We then discuss how to implement amortized inference based on the self-normalized estimators that we discussed in Section~\ref{sec:importance-weighted-model-learning} and Section~\ref{sec:inclusive-kl}. 
This design, which is employed in PyProb, is very general in that it does not impose any new requirements on the modeling language. In particular, this style of amortized inference can be applied to probabilistic programs that are not themselves differentiable.

In practice, it is often beneficial, and even necessary, to design the inference network in a manner that accounts for the characteristics of the observed data in a particular probabilistic program. For this reason, it makes sense to design a probabilistic programming language that allows the user to specify both the generative and the inference model, rather than implementing the encoder network in the inference backend. In Section~\ref{sec:webppl-proposals}, we discuss how we can interleave a generative program and an inference program by explicitly associating a proposal with each unobserved random variable. We then discuss how to implement amortized methods that maximize ELBO in a manner that accounts both for reparameterized and non-reparameterized variables in the proposal. This design, which is employed in WebPPL, requires a differentiable language, but makes it possible to use a single program to implement both the generative model and the inference model.

One of the limitations of both PyProb-style and WebPPL-style implementations is that proposals have to generated in the order in which they are instantiated in the generative model. This design is not necessarily optimal, in the sense that it may introduce a mismatch between the conditional dependencies in the inference model and the conditional independencies in the true posterior. To overcome this limitation, we discuss a more general design in which the generative model and inference model are implemented as distinct probabilistic program. We discuss this design, which is employed in Edward, Pyro, and Probabilistic Torch, in Section~\ref{sec:probtorch-proposals}. 



	\section{Implementing Proposals in the Inference Backend}
	\label{sec:pyprob-proposals}


Amortized inference can be seen as a form of approximate ``compilation''.  A probabilistic program denotes a joint distribution $p(Y, X ; \theta)$ in which unobserved random variables $X$ are defined using \hop{sample} expressions and observed random variables $Y$ are defined using \hop{observe} expressions. This joint distribution, which is computable,  uniquely defines a posterior distribution $p(X \mid Y ; \theta)$. Compilation refers to making a meaning-preserving transformation from an expression in a source language to an equivalent expression in a target language. In certain cases, we can compile a program that denotes a joint distribution to a program that denotes the posterior. To do so, we would translate the program to an expression in a target language, potentially the same language as the source, that produces samples directly from  $p(X \mid Y; \theta)$, which is to say that it contains only \hop{sample} statements, no \hop{observe} statements. The probabilistic programming community has developed disintegration methods that can perform this type of exact compilation in special cases \citep{narayanan2020symbolic}, but in general this is hard to do. Amortized methods associate an inference model with a probabilistic program. This inference model defines a tractable distribution $q(X \mid Y ; \phi)$, which is a program in some target language that contains only \hop{sample} expressions. If there exists a set of parameters $\phi$ such that $p(X \mid Y ; \theta) = q(X \mid Y ; \phi)$, then solving the optimization problem for $\phi$ is a form of exact compilation. For this reason, amortized inference is also known as ``inference compilation''.

In this section, we will discuss how close we can get to this ideal of automated exact compilation by way of amortized inference. We will primarily focus on use cases in which we already have a fixed probabilistic program, i.e.~a sufficiently realistic stochastic simulator with known structure and parameters. This means that we can simulate pairs $(Y^l,X^l)$ from the model by replacing all \hop{observe} expressions with \hop{sample} expression. This data can then be used to train the inference network using simple sleep-phase wake-sleep updates. When we additionally have real data available, we can combine these updates with reweighted wake-phase updates as we discussed in Section~\ref{sec:importance-weighted-model-learning} and Section~\ref{sec:inclusive-kl}.



To use amortized inference as a strategy for approximate compilation, we need to make it applicable to programs written in general-purpose languages such as the HOPPL. The design that we will discuss in this section is to implement an approximate compilation backend that interacts with the probabilistic program by way of the messaging interface that we discussed in Chapter~\ref{ch:eval-two}. As before, the program execution process will send messages to the inference backend that indicate the program has either reached a \hop{sample} expression, and \hop{observe} expression, or has completed execution. The backend then generates neural proposals and computes gradient estimates of the variational objectives.

One of the fundamental properties of the messaging interface from Chapter~\ref{ch:eval-two} is that it introduces an abstraction boundary between a deterministic program execution process, and an inference process. The execution process performs all deterministic computations that need to be performed in the probabilistic program. The inference process generates proposals, computes importance weights, and tracks any other form of state that must be computed as a side-effect of program execution. The only state that needs to be tracked in the execution process is a unique process id.

Because of this boundary, the execution process and the inference process need not be implemented in the same language. This is not just a theoretical possibility; it forms the basis for systems such as PyProb \citep{le2016inference}, which provides functionality for learning amortized proposals in Python, which can be applied to models that are written in a variety of languages. This affords the opportunity to perform inference in existing stochastic simulation code in science and engineering, which may be written in languages such as C or C++, and train proposals in languages such as Python, which provide bindings to deep learning frameworks.

The message-based interface for amortized can be more lightweight than the general interface that we considered in Chapter~\ref{ch:eval-two}. In the original interface, we relied on CPS transformations in order to support the \hop{"fork"} directive. However stochastic variational inference can be implemented without forking processes. To support inference amortized inference, we only need to fulfill two requirements:
\begin{itemize}
  \item[1.] Replace all calls to random number generators in the original code with calls that dispatch a request $(\mhop{"sample"}, \id, \alpha, d)$ to the inference backend. This request contains an process id $\id$, an address $\alpha$, and a distribution $d$.
  \item[2.] Add code that conditions the simulation on observed data by dispatching requests $(\mhop{"observe"}, \id, \alpha, d, c)$. This request contains an observed value $c$ in addition to the values that are also present in \hop{"sample"} requests.
\end{itemize}
This interface requires that we associate a unique address $\alpha$ with each request. This can be done in a lightweight manner by defining a (non-unique) base address $\alpha_0$ based on the lexical position of an expression in the model source code, and implementing a function $\alpha = \textsc{new-addr}(\alpha_0, \id)$ in the inference backend that constructs a unique run-time address by counting the number of requests with base address $\alpha_0$ in the process with id $\id$.


In the remainder of this Section, we will discuss how to implement neural proposals and the inference process that calculates the variational objective. Since neural proposals are defined in the backend, this backend needs to support automatic differentiation, but the modeling language need to be differentiable. This greatly simplifies application of amortized inference to stochastic simulators in science and engineering, which are often not implemented in differentiable languages.



\subsection{Implementing Recurrent Proposals}
\label{sec:recurrent-proposals}

One of the constraints of generating proposals across a messaging interface is that we do not control the order in which requests for samples arrive. This means that we have to generate proposals in whichever order they arise in the execution process of the  generative model. If this model includes control flow, it may not even be the case that this order is always the same.
The inference backend will therefore need to define a neural proposal that can construct a distribution $d_q$ whenever a new \hop{"sample"} request arrives. Ideally this proposal should not only be informed by input data $Y$, but also by the proposed values for any preceding requests.

One strategy for designing this type of neural proposal is to use networks that are recurrent, such as long short-term memory (LSTM) networks \citep{le2016inference}. A recurrent network is a differentiable function that operates on some sequence of inputs. At each step of the computation, the network accepts a vector of inputs a hidden state. It then returns an output, along with an updated hidden state. This state can then be passed to the network along with the subsequent inputs to continue the recurrent computation.

PyProb-style neural proposals make use of a recurrent neural network $\lambda(Y, \alpha, d, x, h, \phi)$. As inputs to the network, we provide the input data $Y$, the address $\alpha$ and prior $d$ for the current sample request, the value $x$ for the preceding sample $x$ and hidden state $h$, and the network parameters $\phi$. As its outputs, the network returns variational parameters $\lambda_q$ and an updated hidden state $\lambda_h$. This defines a recurrent proposal
\begin{align*}
  \begin{split}
    x_n
    &\sim
    q
    \big(
      x ; \lambda_q(Y, \alpha_n, d_n,
                    x_{n-1}, h_{n-1}, \phi)
    \big),
    \\
    h_n
    &=
    \lambda_h(Y, \alpha_n, d_n,
                    x_{n-1}, h_{n-1}, \phi).
  \end{split}
\end{align*}
To define a base case, we typically assume an initial sample $x_0 = \phi_x$ and hidden state $h_0 = \phi_h$ that are tunable parameters.

This recurrent architecture is semi-general, in the sense that it is applicable to any probabilistic program. In practice however, this design is really a meta-architecture in the sense that specific neural components need to be tailored to the probabilistic program of interest. A recurrent network such as an LSTM accepts a vector-valued input. This means that we need to design encoder networks for the address $\alpha$, the distribution type $d$, the input data $Y$, and the preceding sample $x$. These networks map inputs to continuous vectors, which can then be concatenated to construct the input to the LSTM.

The component of the model that will in general require the most engineering is the encoder for the input data. The type of network that is appropriate will depend on the input data modality (text, images, video, point clouds), and in certain cases we may wish to combine neural encoding with certain forms of feature engineering, As an example, when performing clustering, we might pre-process input data into a histogram, and use this histogram as an input to a convolutional encoder.

Encoding addresses involves some subtlety. We would like the neural proposal to be able to deal with cases in which the set of addresses that arises in an execution is not fixed, for example because the program instantiates a varying number of clusters. In order for the neural network to generate good proposals at each address, this address implicitly needs to convey information about the role of each variable in the model. The original work on inference compilation by \citet{le2016inference} achieves this by defining an address
\begin{align}
  \alpha = \textsc{push-addr}(\alpha_0, A(\alpha_0))
\end{align}
Here the function $\textsc{push-addr}$ is analogous to the one that we used in our addressing transformation in Chapter~\ref{ch:eval-two}; it combines two identifiers into a (vector-valued) address. The two identifiers here are a prefix $\alpha_0$, which uniquely identifies each sample expression in the source code of the program, and a suffix $A(\alpha_0)$ that tracks the number of evaluations of this particular sample expression in the execution.

\subsection{Computing Gradients of the Variational Objective}
\label{sec:grad-variational}

When implementing amortized variational inference, we have the option of either maximizing a variational lower bound with a likelihood-ratio estimator, as in Section~\ref{sec:eval-bbvi}, or to using reweighted estimators to minimize the inclusive KL divergence with respect to $\phi$, as in Section~\ref{sec:inclusive-kl}. 
In either setting, performing stochastic gradient descent involves a computation of the following form:
\begin{itemize}
\item[-] Initialize the weights $\phi$ for the inference model. Optionally, initialize parameters $\theta$ for the generative model.
\item[-] Learn $\phi$ (and optionally $\theta$) using stochastic gradient descent:
\begin{itemize}
    \item[-] Sample a mini-batch $\{Y^1, \dots, Y^B\}$ from the training data.
    \item[-] For each item $Y^b$ in the mini-batch, generate a set of proposals $\{\mathcal{X}^{1,b}, \dots, \mathcal{X}^{L,b}\}$ and compute weights $\{w^{1,b}, \dots, w^{L,b}\}$.
    \item[-] Average over the weights to compute a Monte Carlo estimate of a loss $\hat{L}_q(\phi)$ and update $\phi$ using the gradient $\nabla_\phi \: \hat{L}_q(\phi)$.
    \item[-] When additionally learning $\theta$, compute a loss $\hat{L}_p(\theta)$ and update $\theta$ using the gradient $\nabla_\theta \: \hat{L}_p(\theta)$.
\end{itemize}
\end{itemize}
In this computation, we combine all samples into a scalar loss $\hat{L}_q(\phi)$. The reason for this is efficiency. When generating $B \cdot L$ proposals for each gradient step, we could in principle generate samples one by one, and perform reverse-mode differentiation for each sample to compute its contribution to the gradient. However, a much more computationally efficient approach is to combine samples into a single scalar objective, and accumulate gradients using a single backward computation.

Aggregating samples into a single scalar objective $\hat{L}_q(\phi)$ involves some subtlety. As we have previously seen, approximating the gradient of an objective using a Monte Carlo estimate is not always the same thing as taking a gradient of a Monte Carlo estiamte of the objective. This means we need to ensure that reverse-mode differentiation of the scalar objective is equivalent to computing the correct Monte Carlo estimate of the gradient.
We will in this section begin by considering the case of computing an objective $\hat{L}_q(\phi)$ for the inclusive KL, and return to the case of exclusive KL in Section~\ref{sec:webppl-proposals}, where we will also cover the use of reparameterized variables. When computing the gradient of the inclusive KL, we need to approximate the expectation
\begin{align}
    \begin{split}
    &
    -\nabla_\phi \:
    \mathbb{E}_{p^\text{data}(Y)}
    \left[
    \text{KL}
    \big(
      p(X \mid Y ; \theta)
      \,\big|\big|\,
      q(X \mid Y ; \phi)
    \big)
    \right]
    =
    \\
    &
    \hspace{10em}
    \mathbb{E}_{p^\text{data}(Y) p(X \mid Y ; \theta)}
    \left[
        \nabla_\phi \log q(X \mid Y ; \phi)
    \right].
    \end{split}
\end{align}
To compute the self-normalized gradient estimate that we derived in Section~\ref{sec:inclusive-kl}, we might naively define an analogous self-normalized loss
\begin{align}
\hat{L}_q(\phi)
=
-
\frac{1}{B}
\sum_{l,b}
\frac{w^{l,b}}
     {\sum_{l'} w^{l',b}}
\log q(X^{l,b} \mid Y^b ; \phi)
.
\end{align}
However, this loss has a math bug. Since the importance weights $w^{l,b}$ themselves depend on $\phi$, the gradient of this loss will comprise two sets of terms
\begin{align*}
\nabla_\phi \:
\hat{L}_q(\phi)
=&
-
\frac{1}{B}
\sum_{l,b}
\frac{w^{l,b}}
     {\sum_{l'} w^{l',b}}
\:
\nabla_\phi \:
\log q(X^{l,b} \mid Y^b \,; \phi)
\\
&
\qquad
-
\frac{1}{B}
\sum_{l,b}
\log q(X^{l,b} \mid Y^b,  ; \theta)
\:
\nabla_\phi
\left(
\frac{w^{l,b}}
     {\sum_{l'} w^{l',b}}
\right)
.
\end{align*}
The first set of terms correspond to the self-normalized gradient of the inclusive KL from Section~\ref{sec:inclusive-kl}. However, we have now accidentally introduced a second set of terms that correspond to gradients of the self-normalized weight.

A solution to this problem is to make careful use of the mechanics of automatic differentiation. As we discussed in Section~\ref{sec:hmc-ad}, automatic differentiation begins which with a forward computation, where all real-valued variables are replaced with ``boxed'' data structures that hold gradient information in addition to the computed values. During the reverse-mode computation, we then propagate gradients backward from variables to their parents.

If we would like to exclude terms from the reverse-mode computation, then we can do so by either zeroing out gradient information in boxed data structures, or by simply ``unboxing'' terms for which we would like to ``block'' gradients. In the automatic differentiation computation that we described in Section~\ref{sec:hmc-ad}, we can use the function $c = \textsc{unbox}(\tilde{c})$ that extracts the literal constant value $c$ from a boxed value $\tilde{c}$. 

In short, to exclude the terms corresponding to the gradients of the self-normalized weights, we can simply unbox the importance weights
\begin{align}
\label{eq:Lq-rww}
\hat{L}_q(\phi)
=
-
\frac{1}{B}
\sum_{l,b}
\frac{\textsc{unbox}(w^{b,l})}
     {\sum_{l'} \textsc{unbox}(w^{l',b})}
\log q(X^{l,b} \mid Y^b \,; \phi)
.
\end{align}
This will ensure that $\nabla_\phi w^{l,b} = 0$ and hereby remove the second set of terms from the gradient computation.

In cases where we are additionally interested in learning model parameters, we can implement an anologous loss to compute the self-normalized gradient of the marginal likelihood from Section~\ref{sec:importance-weighted-model-learning},
\begin{align}
\label{eq:Lp-rww}
\hat{L}_p(\theta)
=
-
\frac{1}{B}
\sum_{l,b}
\frac{\textsc{unbox}(w^{l,b})}
     {\sum_{l'} \textsc{unbox}(w^{l',b})}
\:
\log p(Y^{l,b}, X^{l,b} ; \theta)
.
\end{align}
Notice that both losses contain an extra minus sign, to acount for the fact that we will minimize $\hat{L}_q(\phi)$ and $\hat{L}_p(\theta)$.




\begin{algorithm}[!t]
\caption{Reweighted Loss with PyProb-style Neural Proposals
\label{alg:gradient-reweighted-pyprob}}
\begin{algorithmic}[1]
\Function{reweighted-loss}{$Y$, $\theta$, $\lambda_q$, $\lambda_v$, $\phi$, $h_0$, $L$}
    \For{$l = 1, \dots, L$}
    \State $\id \gets \textsc{newid}()$
    \State \textsc{send}(\hop{"start"}, $\id$, ($y$, $\theta$))
    \State $x_\sigma, h_\id \gets \phi_x, \phi_h$
    \Comment{Initial value $x$ and state $h$}
    \State $A_\sigma \gets []$
    \Comment{Address prefix counter}
    \State $\log w_{\id}, \log p_{\id}, \log q_{\id} \gets 0, 0, 0$
    \EndFor
    \State $l \gets 0$
    \While{$l < L$}
    \Switch{$\textsc{receive}()$}
        \Case{$(\mhop{"sample"},\, \id,\, \alpha_0,\, d_p)$}
            \State $A_\sigma \gets \textsc{inc-count}(A_\sigma, \alpha_0)$
            \State $\alpha \gets \textsc{push-addr}(\alpha_0, A_\sigma(\alpha))$
            \State $d_q, h_\sigma \gets \lambda(Y, \alpha, d_p, x_\id, h_\id, \phi_q)$
            \State $x_\id \gets$ \textsc{sample}$(d_q)$
            \State $\log p_{\id} \gets \log p_{\id} + \textsc{log-prob}(d_p, c)$
            \State $\log q_{\id} \gets \log q_{\id} + \textsc{log-prob}(d_q, c)$
            \State $\log w_\id \gets \log w_\id + \textsc{log-prob}(d_p, c) - \textsc{log-prob}(d_q, c)$
            \State \textsc{send}$(\mhop{"continue"},\, \id,\, x_\id)$
        \EndCase
        \Case{$(\mhop{"observe"},\, \id,\, \alpha_0,\, d,\, c)$}
            \State $\log p_{\id} \gets \log p_{\id} + \textsc{log-prob}(d, c)$
            \State $\log w_{\id} \gets \log w_{\id} + \textsc{log-prob}(d, c)$
            \State \textsc{send}$(\mhop{"continue"},\, \id,\, c)$
        \EndCase
        \Case{$(\mhop{"return"},\, \id,\, c)$}
            \State $l \gets l + 1$
            \State $\mathcal{L}_{\id} \gets \mathcal{L}_{\id} + \log w$
        \EndCase
    \EndSwitch
    \EndWhile
    \State $w \gets \textsc{unbox}(\textsc{exp}(\textsc{softmax}(\log w)))$
    \State $\hat{L}_p \gets \textsc{sum}(-w \cdot \log p)$
    \State $\hat{L}_q \gets \textsc{sum}(-w \cdot \log q)$
    \State \textbf{return} $\hat{L}_p$, $\hat{L}_q$
\EndFunction
\end{algorithmic}
\end{algorithm}

\subsection{Implementing Reweighted ``Wake-Wake''}
\label{sec:pyprob-wake-wake}

Algorithm~\ref{alg:gradient-reweighted-pyprob} illustrates how we can use recurrent neural proposals to compute the objectives $\hat{L}_p(\theta)$ and $\hat{L}_q(\phi)$. We will assume a setting where the outer optimization process samples data $Y$ and a provides the current values for parameters $\theta$ of the generative model and parameters $\phi$ for the proposal. To generate weighted samples, the inference controller starts $L$ executions. It then processes incoming requests as follows:
\begin{itemize}
  \item $(\mhop{"sample"}, \id, \alpha_0, d)$: Increment the count $A_\id(\alpha_0)$ for the prefix and compute a unique address $\alpha = \textsc{push-addr}(\alpha_0, A_\id(\alpha))$. Use the proposal network $\lambda$ to obtain a distribution $d_q$ and an updated hidden state $h_\id$. Sample $x_\id \sim d_q$. Update $\log w_\id$, the joint of the generative model $\log p_\id$ and the joint of the proposal $\log q_\id$. Continue execution with the proposed value $x_\id$.
  \item $(\mhop{"observe"}, \id, \alpha_0, d, c)$: Increment $\log w_\id$ and $\log p_\id$ with the log probability of $c$, and continue execution with this value.
\end{itemize}
Each process $\sigma$ computes a weight $w_\sigma$ along with the joint probability for the generative model $\log p_\sigma$ and the joint probability for the inference model $\log q_\sigma$. We use these quanties to compute the objecitves $\hat{L}_p(\theta)$ and $\hat{L}_q(\phi)$ according to Equation~\ref{eq:Lp-rww} and Equation~\ref{eq:Lq-rww} by averaging over executions $\sigma$.


	\section{Integrating Proposals into Probabilistic Programs}
	\label{sec:webppl-proposals}

Amortized inference with PyProb-style recurrent proposals comes very close to providing a general, automated, approach to inference compilation. The recurrent neural network design in Section~\ref{sec:recurrent-proposals} is generally applicable to probabilistic programs in a HOPPL and hereby defines a single ``compilation target'' for amortized inference. Moreover, we can learn the proposal using model-agnostic stochastic variational methods that minimize the inclusive KL divergence.

At the same time, we have seen that this approach is also not completely agnostic to the program for which we are amortizing inference. While it is possible to design a semi-general recurrent network for inference in programs, this network will typically have subcomponents that need to be adapted to the modality of the input data in a particular program. This means that PyProb-style amortized inference does not entirely separate modeling and inference in the same way that we have seen for other methods in this book; the designer of the probabilistic program will have to write code in the probabilistic programming language to implement the generative model, and some code in the language of the inference backend to implement network subcomponents. 

In this section, we return to an alternative design that we outlined in the introduction of this chapter, which is to use code in the modeling language to implement both the generative model and the inference model. In Section~\ref{sec:deep-generative-programs} and Section~\ref{sec:deep-inference-programs} we introduced a program \hop{p} and program \hop{q} that define a generative model and an inference model, which incorporate neural networks $\eta$ and $\lambda$ respectively
\begin{hoppl}
(defn p [y $\eta$ $\theta$]
  (let [z (sample (multinomial 1 $\theta^z$))
        v (sample (normal ($\eta^v_\mu$ z $\theta^v$) ($\eta^v_\sigma$ z $\theta^v$)))]
    (observe (normal ($\eta^y_\mu$ v $\theta^y$) ($\eta^y_\sigma$ v $\theta^y$)) y)
    [z v]))
\end{hoppl}

\begin{hoppl}
(defn q [y $\lambda$ $\phi$]
  (let [z (sample (multinomial 1 ($\lambda^z$ y $\phi^z$)))
        v (sample (normal ($\lambda^v_\mu$ y z $\phi^v$) ($\lambda^v_\sigma$ y z $\phi^v$)))]
    [z v]))
\end{hoppl}
In Section~\ref{sec:probtorch-proposals} we will return to the general setup in which \hop{p} and \hop{q} are arbitrary programs in a HOPPL, which need not instantiate the same set of random variables. In this section, we begin with a simpler, and in some ways more intuitive setup, which incorporate proposals directly into the generative model. We briefly introduced this design in Section~\ref{sec:deep-inference-programs}, where we defined a program that integrates \hop{q} into \hop{p},
\begin{hoppl}
(defn p-with-q [y $\eta$ $\theta$ $\lambda$ $\phi$]
  (let [z (propose (multinomial 1 $\theta^z$)
                   (multinomial 1 ($\lambda^z$ y $\phi^z$)))
        v (propose (normal ($\eta^v_\mu$ z $\theta^v$) ($\eta^v_\sigma$ z $\theta^v$)))
                   (normal ($\lambda^v_\mu$ y z $\phi^v$) ($\lambda^v_\sigma$ y z $\phi^v$)))]
  (observe (normal ($\eta^y_\mu$ v $\theta^y$) ($\eta^y_\sigma$ v $\theta^y$)) y)
  [z v]))
\end{hoppl}
This basic syntactic design is used in WebPPL \citep{ritchie2016deep}. To incorporate proposals into the program, we have introduced a construct \hop{(propose $d_p$ $d_q$)}. This expression form, whose implementation we will discuss below, associates a proposal density $d_q$ with an unobserved random variable that has density $d_p$ in the generative model. This allows us to simply replace any \hop{sample} form with a corresponding \hop{propose} form. The result is a program in which both the generative program \hop{p} and the inference program \hop{q} are defined together and are executed simultaneously.

It is not the case that we have to replace all \hop{sample} expressions with \hop{propose} expressions in a program. As an example, we can associate a proposal with variable \hop{v} whilst sampling the variable \hop{z} from the prior,
\begin{hoppl}
(defn p-with-partial-q [y $\eta$ $\theta$ $\lambda$ $\phi$]
  (let [z (sample (multinomial 1 $\theta^z$))
        v (propose (normal ($\eta^v_\mu$ z $\theta^v$) ($\eta^v_\sigma$ z $\theta^v$)))
                   (normal ($\lambda^v_\mu$ y $\phi^v$) ($\lambda^v_\sigma$ y $\phi^v$)))]
  (observe (normal ($\eta^y_\mu$ v $\theta^y$) ($\eta^y_\sigma$ v $\theta^y$)) y)
  [z v]))
\end{hoppl}

In this version of the program, the neural proposal for $v$ is parameterized by a network
\hop{($\lambda^v$ y $\phi^v$)}, which only takes \hop{y} as its input. By contrast, in the original program \hop{q} the network \hop{($\lambda^v$ y z $\phi^v$)} takes both \hop{y} and \hop{z} as its inputs. This highlights that we have some degree of leeway in choosing the conditional dependencies in the inference model, and that these dependencies do not necessarily have to mirror those in the generative model.

When we incorporate proposals into a probabilistic program, we need  to make sure that this transformation preserves the semantic meaning of the program. In other words, if we replace a \hop{sample} expression with a \hop{propose} expression, the resulting program should still define the same joint density $p(Y, X ; \theta)$ and posterior $p(X \mid Y ; \theta)$; the choice of proposal should not affect the density that a program denotes.

To ensure that this is indeed the case, we need to define a semantic meaning for the propose form such that the program density is invariant to the choice of proposal. One concrete implementation of \hop{propose} is to define a function
\begin{hoppl}
(defn propose [$d_p$ $d_q$]
  (let [$c$ (sample $d_q$)]
    (factor (- (log-prob $d_p$ $c$)
               (log-prob $d_q$ $c$)))))
\end{hoppl}
This function uses \hop{(sample $d_q$)} to define a random variable that is distributed according to a proposal $d_q$, which is part of the amortized inference program. As the sample is generated from the proposal rather than the generative model, \hop{propose} then uses \hop{factor} to correct the density with the log  density ratio between the prior and the proposal.

Now let us consider whether this definition impacts the meaning of the program \hop{p-with-partial-q}. When evaluating the program we will encounter the following special forms:
\begin{itemize}
  \item A \hop{sample} form that defines a prior $p(z)$.
  \item A call to \hop{propose}. This call contains a \hop{sample} form that defines a density $q(v \mid y ; \phi)$ and a \hop{factor} form (see Section~\ref{sec:factor-condition}) that incorporates an importance ratio $\frac{p(v \mid z)}{q(v \mid y ; \phi)}$ into the program density.
  \item An \hop{observe} form defines a likelihood $p(y \mid v)$.
\end{itemize}
If we multiply all four terms together, we see that the program defines the unnormalized density
\begin{align}
    \gamma(z,v ; y)
    &=
    p(z)
    \:
    q(v \mid y ; \phi)
    \:
   \frac{p(v \mid z)}{q(v \mid y ; \phi)}
    \:
    p(y \mid v)
    \\
    &=
    p(z)  \: p(v \mid z) \: p(y \mid v)
    =
    p(z, v, y).
\end{align}
In other words, by incorporating a factor that computes the log-ratio between the prior and the proposal, we ensure that the density of a program is unaffected when we replace an expression \hop{(sample $d_p$)} with an expression \hop{(propose $d_p$ $d_q$)}.

While inclusion of a proposal leaves the density of a program invariant, it will change the way inference algorithms behave. As an example, in likelihood weighting we generate a proposals using \hop{sample} expressions, and compute importance weights using \hop{observe} and \hop{factor} expressions. In the program \hop{p-with-partial-q}, this will yield the importance weight
\begin{align}
    w(y, v, x ; \phi)
    &=
    p(y \mid v)
    \frac{p(v \mid z)}
         {q(v \mid y ; \phi)}.
\end{align}
This means that we can implement importance sampling with user-specified proposals by simply replacing \hop{sample} with \hop{propose} and running a standard ``likelihood weighting'' algorithm, which now computes a weight that correctly accounts for the proposal, rather than simply computing the weight according to the likelihood.

Given the above implementation of \hop{propose}, it is straightforward to compute self-normalized gradient estimates of the marginal likelihood (Section~\ref{sec:importance-weighted-model-learning}) and the inclusive KL divergence (Section~\ref{sec:inclusive-kl}), which combine to define the reweighted ``wake-wake'' algorithm that is analogous to the one that we described for PyProb-style recurrent proposals in Section~\ref{sec:pyprob-wake-wake}. The corresponding implementation will essentially be the same as the one in Algorithm~\ref{alg:gradient-reweighted-pyprob}, with only distinction being that we can omit lines 5-6 and lines 12-14, since we do not need to manage the state of the recurrent inference network in the backend, or construct addresses for each proposal.

\subsection{Computing the Gradient of the Lower Bound}
\label{sec:grad-elbo-ritchie}
When we integrate WebPPL-style proposals into the probabilistic program, the reweighted ``wake-wake'' style inference from Section~\ref{sec:pyprob-wake-wake} is not the only method that we have at our disposal. Since we are defining the neural proposals in the modeling language, this language will need to provide support for automatic differentiation. This means that we can also maximize the ELBO using reparameterized gradients, as we would in a VAE. This can be more computationally efficient, since reparameterized gradients can in certain models be approximated with a smaller number of samples.

The main point to consider when computing the gradient of a lower bound for a general probabilistic program, is that this program may contain a combination of continuous variables and discrete variables. This is in fact the case in the program \hop{p} and the program \hop{p-with-q} that incorporates the corresponding inference model. For continuous distributions, we will typically want to employ reparameterized variables whenever possible, since reparameterization typically reduces the variance of the resulting gradient estimator. However, for discrete variables, our only option is to compute likelihood-ratio estimators.


As it turns out, there is no problem with combining reparameterized and non-reparameterized variables in a proposal; we can define a single objective that computes the correct gradient estimate for each variable in the model. For a general discussion on this point, we refer to the work by \citet{schulman2015gradient}, who formalize general-purpose gradient estimators for loss functions that are defined using stochastic computation graphs.

To explain the high-level idea behind the general gradient computation, we will make use of a notational convention that
was introduced by \citet{ritchie2016deep} to bring reparameterized and non-reparameterized variables under a commmon denominator. Suppose that we assume a general case in which we sample from a distribution $x \sim q(x ; \phi)$ by transforming a sample from a base distribution
\begin{align}
  \label{eq:webppl-reparam}
  x &= g(\tilde{x}, \phi),
  &
  \tilde{x} \sim \tilde{q}(\tilde{x} ; \phi).
\end{align}
In this general definition, both the base distribution and the transformation depend on the parameters $\phi$. We can now consider reparameterized and non-reparameterized distributions as special cases
\begin{itemize}
    \item Reparameterized proposals combine a parameter-free base density $\tilde{q}(\tilde{x})$ with a parametric transformation $x = g(\tilde{x} , \phi)$.
    \item Non-reparameterized proposals combine a parametric base density $\tilde{q}(\tilde{x}, \phi)$ with parameter-free transform, which is just the identity function $x = g(\tilde{x}) = \tilde{x}$.
\end{itemize}
In other words, we assume a notational convention in which there is always a dependence on $\phi$ in both the base distribution and the transformation, but we will in practice use distributions for which either $\nabla_\phi \: \tilde{q}(\tilde{x} ; \phi) = 0$ or $\nabla_\phi \: g(\tilde{x}, \phi) = 0$.

We can use this notation to define a general lower bound that encompasses both reparameterized and non-reparameterized terms. Suppose that we use $\tilde{q}(\tilde{X} \mid Y ; \phi)$ to denote the base proposal for all variables in a probabilistic program with density $p(Y, X)$, and use $X = g(\tilde{X}, Y, \phi)$ to refer to the transformation of all variables. We can then define a variational lower bound of the form
\begin{align}
\mathcal{L}(\phi)
=
\mathbb{E}_{\tilde{q}(\tilde{X} \mid Y ; \phi)}
\left[
\log w(Y, \tilde{X}, \phi)
\right]
,
\end{align}
where the importance weight is defined as
\begin{align}
    w(Y, \tilde{X}, \phi)
    =
    \frac{p \big( Y, g(\tilde{X}, Y, \phi) \big)}
         {q \big( g(\tilde{X}, Y, \phi) \mid Y ; \phi \big)}
    .
\end{align}
When we compute the gradient of this bound, we obtain the terms
\begin{align}
    \label{eq:grad-elbo-mixed}
    \begin{split}
    \nabla_\phi \mathcal{L}(\phi)
    &=
    \mathbb{E}_{\tilde{q}(\tilde{X} \mid Y ; \phi)}
    \left[
        \nabla_\phi \log w(Y, \tilde{X}, \phi)
    \right]
    \\
    &\qquad
    +
    \mathbb{E}_{\tilde{q}(\tilde{X} \mid Y ; \phi)}
    \left[
        \big(\nabla_\phi
        \log \tilde{q}(\tilde{X} \mid Y ; \phi)\big)
        \:
        \log w(Y, \tilde{X}, \phi)
    \right]
    \end{split}
    .
\end{align}
Here the first term computes reparameterized gradient estimates, and will be 0 for any non-reparameterized variables. The second term computes a likelihood-ratio estimator for non-reparameterized variables, but will be 0 for non-reparameterized variables.

As a concrete example, let us revisit the program \hop{p-with-q}, which combines a proposal $q(z \mid y ; \phi)$ for the discrete variable $z$, which cannot be reparameterized, with a reparameterizable proposal $q(v \mid z,y ; \phi)$ for the continuous image embedding $v$,
\begin{hoppl}
(defn p-with-q [y $\eta$ $\theta$ $\lambda$ $\phi$]
  (let [z (propose (multinomial 1 $\theta^z$)
                   (multinomial 1 ($\lambda^z$ y $\phi^z$)))
        v (propose (normal ($\eta^v_\mu$ z $\theta^v$) ($\eta^v_\sigma$ z $\theta^v$)))
                   (normal ($\lambda^v_\mu$ y z $\phi^v$) ($\lambda^v_\sigma$ y z $\phi^v$)))]
  (observe (normal ($\eta^y_\mu$ v $\theta^y$) ($\eta^y_\sigma$ v $\theta^y$)) y)
  [z v]))
\end{hoppl}
The gradient of the lower bound for this program is
\begin{align*}
    \nabla_\phi \: \mathcal{L}(\phi)
    &=
    \mathbb{E}_{\tilde{q}(\tilde{z}, \tilde{v} \mid y ; \phi)}
    \left[
        \nabla_\phi \log w(y, \tilde{z}, \tilde{v}, \phi)
    \right]
    \\
    &\qquad
    +
    \mathbb{E}_{\tilde{q}(\tilde{z}, \tilde{v} \mid y ; \phi)}
    \left[
        \big(
          \nabla_\phi
          \log \tilde{q}(\tilde{z}, \tilde{v} \mid y ; \phi)
        \big)
        \log w(y, \tilde{z}, \tilde{v}, \phi)
    \right]
    .
\end{align*}
The first term is easy to compute using automatic differentiation, which will automatically expand all terms in the derivative of the weight
\begin{align*}
    \nabla_\phi
    \log w(y, \tilde{z}, \tilde{v}, \phi)
    &=
    \left.
    \frac{\partial}{\partial v}
    \log
    \frac{p(y,z,v)}
         {q(z,v \mid y \,; \phi)}
    \nabla_\phi \:
    g_v(\tilde{v}, y, z, \phi)
    \right\vert_{v = g_v(\tilde{v}, y, z, \phi), z = \tilde{z}}
    \\[4pt]
    &\qquad
    -
    \nabla_\phi
    \log q(z,v \,; \phi)
    \Big\vert_{v = g_v(\tilde{v}, y, z, \phi), z = \tilde{z}}
    .
\end{align*}
In the second term, we can make use of the fact that gradient of the base density will be 0 for the reparameterized variable $v$,
\begin{align}
\nabla_\phi \log \tilde{q}(\tilde{z}, \tilde{v} \mid y ; \phi)
=
\nabla_\phi
\log \tilde{q}(\tilde{z} \mid y ; \phi)
.
\end{align}
We can approximate this term in the same way that we have in BBVI, albeit that we now only compute the terms for non-reparameterized variables, rather than all variables in the model.




\begin{algorithm}[!t]
\caption{
\label{alg:gradient-elbo-loss-webppl} Lower Bound with WebPPL-style proposals}
\begin{algorithmic}[1]
\Function{elbo-loss}{$Y$, $\theta$, $\phi$, $L$}
    \For{$l = 1, \dots, L$}
    \State $\id \gets \textsc{newid}()$
    \State \textsc{send}(\hop{"start"}, $\id$, ($Y$, $\theta$, $\phi$))
    \State $\log w_{\id}, \log \tilde{q}_{\id} \gets 0$
    \EndFor
    \State $l \gets 0$
    \While{$l < L$}
    \Switch{$\textsc{receive}()$}
        \Case{$(\mhop{"sample"},\, \id,\, \alpha,\, d)$}
            \State $x \gets$ \textsc{sample}$(d)$
            \State \textsc{send}$(\mhop{"continue"},\, \id,\, x)$
        \EndCase
        \Case{$(\mhop{"propose"},\, \id,\, \alpha,\, d_p, d_q)$}
            \State $\tilde{d}_q \gets \textsc{base-dist}(d_q)$
            \State $\tilde{x} \gets \textsc{sample}(\tilde{d}_q)$
            \State $x \gets \textsc{transform}(d_q, \tilde{x})$
            \State $\log \tilde{q}_\id \gets \log \tilde{q}_\id + \textsc{log-prob}(\tilde{d}_q, \tilde{x})$
            \State $\log w_{\id} \gets \log w_{\id} \!+\! \textsc{log-prob}(d_p, x) \!-\! \textsc{log-prob}(d_q, x)$
            \State \textsc{send}$(\mhop{"continue"},\, \id,\, x)$
        \EndCase
        \Case{$(\mhop{"observe"},\, \id,\, \alpha,\, d,\, c)$}
            \State $\log w_{\id} \gets \log w_{\id} + \textsc{log-prob}(d, c)$
            \State \textsc{send}$(\mhop{"continue"},\, \id,\, c)$
        \EndCase
        \Case{$(\mhop{"return"},\, \id,\, c)$}
            \State $l \gets l + 1$
        \EndCase
    \EndSwitch
    \EndWhile
    \State $\hat{L}_p \gets \textsc{mean}(-\log \tilde{q} \cdot \textsc{unbox}(\log w) - \log w)$
    \State $\hat{L}_q \gets \hat{L}_p$
    \State \textbf{return} $\hat{L}_p, \hat{L}_q$
\EndFunction
\end{algorithmic}
\end{algorithm}

\subsection{Implementing Autoencoding Variational Inference}
\label{sec:aevi-ritchie}

Algorithm~\ref{alg:gradient-elbo-loss-webppl} describes an inference computation that computes a variational objective. We will once again assume that an outer optimization process samples data $Y$ and aggregate samples into a single objective,
\begin{align*}
    \hat{L}_p(Y, \theta)
    =
    \hat{L}_q(Y, \phi)
    =
    -\hat{\mathcal{L}}(Y, \theta, \phi)
    .
\end{align*}
Our goal is now to compute an estimate $\hat{\mathcal{L}}(Y,\theta,\phi)$ such that $\nabla_\phi \hat{\mathcal{L}}$ is an unbiased estimate of the gradient in Equation~\ref{eq:grad-elbo-mixed}.

To compute the objective, the inference computation starts $L$ executions of the program, each with a unique process id $\id$, and intializes to variables that will be tracked for each execution. The first is the accumulated log importance weight $\log w_\id$, the second is the accumulated log probability $\log \tilde{q}_\id$ under the base distribution of each proposal. To track these two variables, we implement the following operations for each message:
\begin{itemize}
  \item $(\mhop{"sample"}, \id, \alpha, d)$: Sample $x \sim d$ from the program prior and continue execution with the value $x$.
  \item $(\mhop{"propose"}, \id, \alpha, d_p, d_q)$: Sample $x \sim d_q$ from the proposal using the generalized transformation in Equation~\ref{eq:webppl-reparam}. To do so, sample $\tilde{x} \sim \tilde{d}_q$ from the corresponding base distribution and transform the sample $x = \textsc{transform}(d_q, \tilde{x})$.
  Increment $\log q_\id$ by the probability under the base distribution $\textsc{log-prob}(\tilde{d}_q, \tilde{x})$.
  Finally, increment $\log w_\id$ with the log ratio $\textsc{log-prob}(d_p, \tilde{x}) - \textsc{log-prob}(d_q, \tilde{x})$. Continue execution with the sampled value $x$.
  \item $(\mhop{"observe"}, \id, \alpha, d, c)$. Increment $\log w_\sigma$ with the log probability $\textsc{log-prob}(d, c)$. Continue execution with the value $c$.
 \end{itemize}
Once all $L$ processes have terminated, we define the losses by simply averaging over the results from each execution
\begin{align}
  \label{eq:elbo-loss-ritchie}
  \hat{L}_p
  =
  \hat{L}_q
  =
  - \hat{\mathcal{L}}
  =
  -\frac{1}{L} \sum_{\id}
  \log \tilde{q}_\id \cdot \textsc{unbox}(\log w_\id)
  +
  \log w_\id
  .
\end{align}
The implicit requirement in this computation is that both the execution process and the inference process support automatic differentation, at least in cases where we wish to use reparameterized variables. For such variables, we have to be able to compute the gradient of the sampled value with respect to $\phi$, which is to say that we need to call $\textsc{grad}(\textsc{transform}(d_q, \tilde{x}))$. This not only means that the inference backend must support automatic differentiation, but also that the modeling language must support automatic differentiation, since we need to be able to propagate derivatives through the proposal network, which is specified by the user as part of the model.

Because of the reliance on differentiability, this particular implementation of stochastic variational inference is not the best fit for settings where the model and the inference backend are not implemented in the same language. Doing so would require a messaging interface in which we are not only able to serialize distribution objects $d_p$ and $d_q$, also the computation graph in the boxed data structures for the parameters of these distributions. Conversely, when both the execution process and the inference process make use of the same language runtime, then values can be passed across the messaging interface without serialization, which greatly simplifies gradient computations.

	\section{Using Standalone Programs as Proposals}
	\label{sec:probtorch-proposals}

The two strategies for amortization that we have discussed so far have defined proposals that generate samples in the same order as in the generative model. This was true for WebPPL-style proposals, which employ a \hop{(propose $d_p$ $d_q$)} form that requires both a prior $d_p$ and a proposal $d_q$ as its arguments. It was also true for PyProb-style proposals, in which a recurrent network constructs a proposal $d_q$ each time an expression \hop{(sample $\alpha_0$ $d_p$)} is evaluated in the model. In both cases, the proposal $q(x \mid Y, \pa_p(x), \phi)$ for a particular variable $x$ can depend on the data $Y$, as well as on upstream variables in the generative model $\pa_p(x)$, but not on any  downstream variables. 

A limitation of PyProb-style proposals is that not all preceding variables may in fact be relevant to the current proposal. This means that we are imposing a greater burden on the neural network, which must now learn which preceding variables are informative in the context of the current proposal. When using WebPPL-style proposals, we can choose which preceding variables to condition on, but we cannot condition on variables are instantiated later in the execution.

As a concrete example, let us once again look at the deep generative model \hop{p} that we have considered throughout this chapter
\begin{hoppl}
(defn p [y $\eta$ $\theta$]
  (let [z (sample (multinomial 1 $\theta^z$))
        v (sample (normal ($\eta^v_\mu$ z $\theta^v$) ($\eta^v_\sigma$ z $\theta^v$)))]
    (observe (normal ($\eta^y_\mu$ v $\theta^y$) ($\eta^y_\sigma$ v $\theta^y$)) y)
    [z v]))
\end{hoppl}
In our previous discussion, we employed a proposal with the same structure as the prior, which is to say that we first predict $z$ from $y$, and then predict $v$ from both $z$ and $y$,
\begin{align}
 q(z, v \mid y ; \phi) =
 q(z \mid y ; \phi) \:
 q(v \mid y, z ; \phi)
 .
\end{align}
However, we could also design a proposal that begins by predicting $v$ from $y$, and then predicts $z$ from $y$ and $v$,
\begin{align}
 q(z, v \mid y ; \phi) =
 q(v \mid y ; \phi) \:
 q(z \mid y, v ; \phi)
 .
\end{align}
To do so, we could specify a proposal program
\begin{hoppl}
(defn q [$y$ $\lambda_v$ $\lambda_z$ $\phi$]
  (let [v (sample (normal ($\lambda_{v, \mu}$ $y$ $\phi$) ($\lambda_{v, \sigma}$ $y$ $\phi$)))
        z (sample (multinomial 1 ($\lambda_z$ $y$ $v$ $\phi$)))]
    [z v]))
\end{hoppl}
In principle, it is possible to reason about what dependency structures in the proposal yield ``faithful inverses'', in the sense that the proposal can encode the correct conditional dependencies and independencies in the posterior \citep{webb2017faithful}. In practice, the best ordering of variables is often an empirical question. Both proposal designs above are faithful inverses, and both have been used in semi-supervised classification tasks \citep{kingma2014semisupervised,joy2020capturing}. For this reason, it makes sense to consider systems in which the user can specify inference models in the form of arbitrary programs, which makes it possible to determine empirically what dependency structures in the proposal best suit a particular model.

A number of deep probabilistic programming systems have converged on this design. This includes Edward \citep{tran2016edward}, which in its original form was based on a first-order language that is analogous to the FOPPL. It also includes Pyro \citep{bingham2019pyro} and Probabilistic Torch \citep{siddharth2017learning}. These systems are both implemented as domain-specific languages that are embedded in Python, resulting in higher-order languages that are analogous to the HOPPL. The same is true for Edward2 \citep{tran2018simple}, which supports evaluation of dynamic computation graphs using Tensorflow's eager mode, and Gen \citep{cusumano-towner2019gen}, which extends the Julia language and allows functions in TensorFlow to be used as primitives. All these systems support the use of programs as proposals.

In this section, we focus on the implementation questions that arise when using programs as proposals. We begin by considering how we should denote \cmt{hao}{correspondences between .. and ..?} correspondences variables in an inference program with variables in a generative program, and how we should deal with cases in which there are missing variables that are not instantiated by the proposal, or auxiliary variables, which are instantiated by the proposal but are not used in the generative model.

We then turn to a lower-level discussion of the messaging interface that that is implemented in systems like Pyro and Edward2. These systems implement inference using algebraic effects library (see \citet{pretnar2015introduction} for an introduction), which manipulates inference state by way of composable handler functions known as ``messengers'' (Pyro) or ``tracers'' (Edward2). To explain how inference can be implemented using these handlers, we describe a basic, simplified implementation in the HOPPL, which we use to perform importance sampling. We conclude by showing how we can put all the pieces together to compute variational objectives whose gradient can be computed using reverse-mode automatic-differentiation.


\subsection{Importance Sampling with Missing and Auxiliary Variables}
\label{sec:aux-vars}

In a first-order language, associating a custom proposal with a program is comparatively straightforward. In this setting, both the generative model and the inference model define a static graph, and we can explicitly associate nodes in the generative model with nodes in the inference model. In higher-order languages, this is in general not possible, since the sets of random variables that will be instantiated by the generative model and the inference model are not statically determinable. This means that we will in general not be able to ascertain whether a generative program and an inference program will instantiate the same set of variables.

To understand how this affects the underlying implementation, let us begin by considering what quantities we will need to compute. In a higher-order language, the proposal program will generate a trace $\mathcal{X} = [\alpha_1 \mapsto c_1, \dots, \alpha_n \mapsto c_n]$. As we have seen in Section~\ref{sec:gbli} and earlier in this chapter, objectives for stochastic variational inference can be computed in terms of the joint probability of the generative model $p(\Y, \X ; \theta)$, the joint probability of the proposal $q(\X \mid \Y ; \phi)$, and the importance weight,
\begin{align}
  w &= \frac{p(\Y, \X ; \theta)}
           {q(\X \mid \Y ; \phi)}
  ,
  &
  \X &\sim q(\X \mid \Y ; \phi).
\end{align}
If we can compute a valid importance weight $w$ for arbitrary programs \hop{p} and \hop{q}, then computing a variational objective should in principle be straightforward.

The first thing that we have to do to compute an importance weight is explicitly denote, at the level of the language design, what variables in a program \hop{q} correspond to a program \hop{p}. For this purpose, probabilistic programming systems that support programs as proposals normally rely on explicit addressing by the user. To support this type of explicit addressing, we will use the syntax
\begin{align*}
    & \mhop{(sample\ $\alpha$\ $d$)} &
    & \mhop{(observe\ $\alpha$\ $d$\ $c$)}
\end{align*}
This defines a simple and unambiguous notion of correpondence between the generative model and the inference model: two variables in distinct programs are the same when their addresses are identical.



The second problem that arises when using programs as proposals is that there is no guarantee that the proposal haves the same support as the posterior of the generative model. When sampling a trace $\X$ from a proposal in a higher-order language, there may not exist an execution of the generative program that could produce exactly the same trace $\X$.

To check whether a trace $\X$ is valid, and to compute the joint probability $p(\Y, \X ; \theta)$, we have to rerun the program \hop{p}, reusing the stored valued $\X(\alpha)$ whenever we encouter an expression \hop{(sample $\alpha$ $d$)}. When doing so, there are two edge cases that can arise:
\begin{itemize}
\item[1.] \emph{Missing addresses:} We could encounter an address $\alpha$ in the program \hop{p} for which no value $\mathcal{X}(\alpha)$ exists in the proposed trace.

\item[2.] \emph{Auxiliary addresses:} There could be addresses $\alpha$ in the proposed trace $\mathcal{X}$ that are not enountered during  evaluation of \hop{p}.
\end{itemize}
A conservative strategy for computing the importance ratio is to assign  probability $0$ in the event of either edge case. This strategy is correct, in the sense that we cannot sample the trace $\mathcal{X}$ from \hop{p} if the trace is either missing addresses or contains values at auxiliary addresses.






A more permissive strategy is to define an importance sampler on an extended space. In our discussion of WebPPL-style proposals, we already saw that we can simply propose from the prior in cases where no proposed value is available. This means that we can simply ignore these variables when computing  the importance weight, since the prior probability and the proposal probability cancel.

As it turns out, there is a similar argument by which we can ignore any auxiliary terms in the proposal. To show why this is the case, let us condider a target \hop{p} and proposal \hop{q} that cover both edge cases
\begin{hoppl}
(defn p [y $\eta$ $\theta$]
  (let [z (sample (multinomial 1 $\theta^z$))
        v (sample (normal ($\eta^v_\mu$ z $\theta^v$) ($\eta^v_\sigma$ z $\theta^v$)))]
    (observe (normal ($\eta^y_\mu$ v $\theta^y$) ($\eta^y_\sigma$ v $\theta^y$)) y)
    [z v]))

(defn q [$y$ $\lambda$ $\phi$ $\sigma^y$]
  (let [u (sample "u" (normal $y$ $\sigma^y$))
        v (sample "v" (normal ($\lambda^{v}_{\mu}$ u $\phi$) ($\lambda^{v}_{\sigma}$ u $\phi$)))]
    v))
\end{hoppl}
In this example, the generative model the proposal defines a so-called denoising encoder \citep{vincent2010stacked,im2017denoising}. Rather than encoding the input $y$ directly to predict the a feature-vector $v$, we first transform $y$ by adding independent noise of magnitude $\sigma_y$ to each dimension, resulting in a ``corrupted'' variable $u \sim \text{Normal}(y, \sigma_y I)$. We then encode $u$ into a feature-vector $v$. This strategy amounts to a form of regularization \citep{shu2018amortized}; the vector $v$ can only encode properties that remain discernible after the transformation.

We would like to define an importance sampler that produces valid weights for the target program \hop{p} by proposing from \hop{q}. The difficulty here is that the two programs densities have a different support
\begin{align*}
    \gamma(v,z \,; \theta)
    &=
    p(y \mid v \,; \theta) \: p(v \mid z \,; \theta) \: p(z \,; \theta),
    \\
    q(u,v \mid y \,; \phi)
    &=
    q(v \mid u \,; \phi) \: q(u \mid y)
    .
\end{align*}
To compute the a valid importance weight, we need to compute a ratio of densities that have the same support. A standard strategy for doing so is to \emph{extend} both densities; we will extend the target to include the variable $u$ and extend the proposal to include a variable $v$.

As it turns out, we can define densities on an extended space in an implicit manner. To do so, we simply extend the target using the conditional density from the proposal $q(u \mid y)$. Similarly, we extend the proposal using the conditional density from the prior $p(z ; \theta_z)$,
\begin{align*}
    \tilde{\gamma}(u,v,z \,; \theta)
    &=
    p(y \mid v \,; \theta) \: p(v \mid z \,; \theta) \: p(z \,; \theta) \: q(u \mid y),
    \\
    \tilde{q}(u,v,z \mid y \,; \theta,\phi)
    &=
    q(v \mid u \,; \phi) \: q(u \mid y) \: p(z \,; \theta_z)
    .
\end{align*}
We can now use these two densities to compute the importance weight
\begin{align*}
    \tilde{w}
    =
    \frac{\tilde{\gamma}(u,v,z \,; \theta)}
         {\tilde{q}(u,v,z \mid y \,; \theta,\phi)}
    =
    p(y \mid v \,; \theta)
    \:
    \frac{p(v \mid z \,; \theta)}
         {q(v \mid u \,; \phi)}
    .
\end{align*}
In this weight, the terms $q(u \mid y)$ and $p(z \,; \theta)$ cancel, since they are present in both the numerator and denominator. In general, this weight will contain the probabilities for all \hop{observe} expressions in the program \hop{p}, along with the ratio of probabilities for all \hop{sample} expressions with addresses that are present in both \hop{p} and \hop{q}.

We now note that the marginal $\tilde{\gamma}(u,v ; \theta,\phi)$ of the extended density is equal to the original target density
\begin{align*}
    \tilde{\gamma}(v,z ; \theta, \phi)
    &=
    \int du \: \tilde{\gamma}(u,v,z ; \theta, \phi),
    \\
    &=
    \gamma(v, z \,; \theta) \: \int du \: q(u \mid y)
    =
     \gamma(v, z \,; \theta).
\end{align*}
This means that we can interpret $u$ as an auxiliary variable; sampling $u,v,z \sim \tilde{\pi}(u,v,z \,; \theta)$ and discarding $u$ is equivalent to sampling from the marginal on the extended space $x,v \sim \tilde{\pi}(z,v \,; \theta)$. This, in turn, is equivalent to sampling $x,v \sim \pi(z,v \,; \theta)$ from the original target. This means that the weight $w = \tilde{w}$ is not only a valid weight for the density on the extended space, but also a valid weight for the original density.

Based on this intuiton, we can define the following strategy for importance sampling using arbitrary programs \hop{p} and \hop{q}:
\begin{itemize}
\item[1.] Evaluate the program \hop{q}. For each expression \hop{(sample $\alpha$ $d_q$)}, sample $x \sim d_q$ from the proposal. Also store the values
\begin{align*}
    \mathcal{X}_q(\alpha) &= x,
   &
   \log \mathcal{P}_q(\alpha) &= \textsc{log-prob}(d_q, x).
\end{align*}

\item[2.] Evaluate the program \hop{p}. For each expression \hop{(sample $\alpha$ $d_p$)}, reuse the $\mathcal{X}_q(\alpha)$ if available and sample $x \sim d_p$ from the prior if not. As with the proposal, store
\begin{align*}
  \mathcal{X}_q(\alpha) &= x,
  &
  \log \mathcal{P}_p(\alpha)
  &=
  \textsc{log-prob}(d_q, x).
\end{align*}
For each expression \hop{(observe $\alpha$ $d_p$ $c$)}, additionally store
\begin{align*}
  \mathcal{Y}_p(\alpha) &= c,
  &
  \log \mathcal{P}_p(\alpha) &= \textsc{log-prob}(d_p, c).
\end{align*}
\item[3.] Compute the importance weight based on all \hop{observe} expressions in \hop{p} and all \hop{sample} expressions that are common to \hop{p} and \hop{q}
\begin{align}
    \label{eq:gradient-importance-weight-aux}
    w =
    \prod_{y \in \text{dom}(\mathcal{Y}_p)} \mathcal{P}_p(y)
    \prod_{x \in \text{dom}(\mathcal{X}_p) \cap \text{dom}(\mathcal{X}_q)}
    \frac{\mathcal{P}_p(x)}{\mathcal{P}_q(x)}.
\end{align}
\end{itemize}
This computation is very similar to the one that we used when we were computing the Metropolis-Hastings acceptance ratio in Section~\ref{sec:eval-mh}. In both computations, we reuse variables from a proposal trace when possible, and generate from the prior when not. Moreover, both the importance weight and the acceptance ratio depend only on the ratio of probabilities for reused variables; missing and superfluous variables can be ignored in both cases. The main difference is that the previously sampled trace is now a trace that has been generated from a different program \hop{q}, rather than the preceding sample in the MCMC algorithm.

We will discuss a full implementation of the importance sampling algorithm in Section~\ref{sec:effect-is}. However, before we do so, we will take a closer look at the messaging interface implementation that we will use to describethis algorithm.

\subsection{Implementing Messaging with Algebraic Effect Handlers}

Deep probabilistic programming systems such as Pyro and Edward2 implement a messaging interface that is very similar to the one that we described in Chapter~\ref{ch:eval-two}. However, the underlying design of this interface differs from the implementation based on continuation-passing-style (CPS) transformations that we have previously discussed. The main reason for this is that both systems implement a domain-specific language that is embedded in Python, which itself has an automatic differentation backend that relies on side effects.

Both Pyro and Edward2 implement a messaging interface using composable request handlers, which in Pyro are known as ``messengers'' and in Edward2 are known as ``tracers''. These treat side effects as algebraic operations which allow the handlers to reason about a limited set of operations in a modular way (see \citet{pretnar2015introduction} for an introduction). This is precisely the case in probabilistic programming languages; the only operations that have side effects, in the sense that they may mutate some form of state in the inference computation, are \hop{sample} and \hop{observe}.

We will illustrate how we can design composable request handlers in the HOPPL, which is to say that we implement the handlers in the same language as the models. This allows us to simplify the messaging interface by assuming that a single process implements both program execution and inference. In the interface in Chapter~\ref{ch:eval-two}, the inference backend needed to handle 4 types of requests: \hop{"start"}, \hop{"sample"}, \hop{"observe"}, and \hop{"return"}. We will in this section do away with the \hop{"start"} request, which can be implemented by either calling the target program \hop{p} or the proposal program \hop{q}. For each the remaining requests, we will introduce a \hop{handler} function that dispatches requests
\begin{hoppl}
(declare handler)
(defn sample [$\alpha$ $d$]
  (handler ["sample" $\alpha$ $d$]))
(defn observe [$\alpha$ $d$ $c$]
  (handler ["observe" $\alpha$ $d$ $c$]))
(defn return [$\alpha$ $d$ $c$]
  (handler ["return" $\alpha$ $d$ $c$]))
\end{hoppl}
 Here we have used a special form \hop{(declare $v$)} to declare a variable whose binding will be specified at a later point in the program, allowing us to provide an implementation of the \hop{handler} function that tracks any state that will be needed in a specific inference computation, such as computing an importance weight.

We can now write programs exactly as we would have before, as long as we incorporate a call to \hop{return} at the end of the program, in order to ensure that the handler function can perform additional operations once execution has completed. In the program \hop{p}, this means that we replace the final expression \hop{e} with the expression \hop{(return e)}
\hop{(return e)}
\begin{hoppl}
(defn p [y $\eta$ $\theta$]
  (let [z (sample (multinomial 1 $\theta^z$))
        v (sample (normal ($\eta^v_\mu$ z $\theta^v$) ($\eta^v_\sigma$ z $\theta^v$)))]
    (observe (normal ($\eta^y_\mu$ v $\theta^y$) ($\eta^y_\sigma$ v $\theta^y$)) y)
    (return [z v])))
\end{hoppl}
To run this program, we need to implement a base case for the request handler. The minimal base case is to not perform inference, but simply generate samples from the prior when running the program
\begin{hoppl}
(defn default-handler [req]
  (let [req-type (first req)]
    (case req-type
      "sample" (let [[$\alpha$ $d$] (rest req)]
                 (dist-sample $d$)))
      "observe" (let [[$\alpha$ $d$ $c$] (rest req)]
                  $c$)
      "return" (let [$c$ (rest req)]
                 $c$)))
\end{hoppl}
Here we use a primitive function \hop{(dist-sample $d$)} to generate samples from a distribution $d$. This function corresponds to the mathematical function $\textsc{sample}(d)$ that we have used elsewhere in this introduction when generating samples in the inference backend.

With this base case in place, a call \hop{(p $y$ $\theta$)} will now run the program and return a sample $x$ from the prior. To perform inference, we can extend this base case to perform likelihood weighting. This requires that we compute the cumulative log probability of all \hop{observe} expressions the program. This means that we now need to somehow track the accumulated log probability across different calls to the handler. In Chapter~\ref{ch:eval-two}, we made use of a separate inference process for this purpose. This process can initialize a variable $\log w_\id$ each time we start a new execution with id $\id$, and update this variable each time the execution process issues an \hop{"observe"} request.

In the setup that we are defining here, we don't have a separate inference process. We simply have calls to the handler function. If this handler function is a pure function, then it cannot make any changes to variables such as an accumulated log probability. There are two solutions to this problem. We can either design handlers that are pure functions, in which case the handlers have to accept a state variable and return an updated state variable. This is, broadly speaking, the solution that was used in Probabilistic Torch. Alternatively, we can make use of mutable variables, which is the solution used in Pyro and Edward2. We will discuss both approaches.

\paragraph{Handlers with Immutable State.} In order for the handler to accept a state variable as input, we have to modify the requests that are passed to the handler to include a state variable $s$
\begin{hoppl}
(declare handler)
(defn sample [$s$ $\alpha$ $d$]
  (handler ["sample" $s$ $\alpha$ $d$]))
(defn observe [$s$ $\alpha$ $d$ $c$]
  (handler ["observe" $s$ $\alpha$ $d$ $c$]))
(defn return [$s$ $c$]
  (handler ["return" $s$ $c$]))
\end{hoppl}
We can then update the default handler to accept and return state
\begin{hoppl}
(defn default-handler [req]
  (let [req-type (first req)]
    (case req-type
      "sample" (let [[$s$ $\alpha$ $d$] (rest req)]
                 [$s$ (dist-sample $d$)]))
      "observe" (let [[$s$ $\alpha$ $d$ $c$] (rest req)]
                  [$s$ $c$])
      "return" (let [[$s$ $c$] (rest req)]
                 [$s$ $c$])))
\end{hoppl}
\cmt{Sam Stites}{Not worth adding because the focus on pure functions is good for this audience, but you can also ``nest'' Free Algebras / eDSLs which would let you get away with \textit{not} changing the syntax of these statements. This would be at the cost of needing additional ``wrapper'' and ``runner'' functions to initialize and extract state. That said, it might require a lot of boilerplate in a lisp.}
This implicitly changes the syntax of \hop{sample}, \hop{observe} and \hop{return}, since these functions now return a state and a value, rather than a value alone. This means that we also need to change our models in order to track changes in state as we perform our computation. To do so, we would modify the program \hop{p} to accept a state $s$ as its input, and update this state with each call to \hop{sample} and \hop{observe}
\begin{hoppl}
(defn p [$s$ $y$ $\eta_y$ $\eta_v$ $\theta$]
  (let [[$s$ z] (sample $s$ "z" (multinomial 1 $\theta_z$))
        [$s$ v] (sample $s$ "v" (normal ($\eta_{v, \mu}$ z $\theta$)
                                    ($\eta_{v, \sigma}$ z $\theta$)))
        [$s$ _] (observe "y" (normal ($\eta_y$ v $\theta$) 1.0) y)]
    (return s [z v])))
\end{hoppl}
With this modification in place, we can now make changes to the state variable inside the handler, which will then propagate through the computation.

To implement likelihood-weighting, all we need to do is create a handler that intercepts each \hop{"observe"} request, updates the state to increment a key \hop{"log-like"}, and then calls the base handler
\begin{hoppl}
(defn sum-log-observed [parent-handler]
  (fn [req]
    (if (= (first req) "observe")
      (let [[$s$ $\alpha$ $d$ $c$] (rest req)
            log-like (+ (get $s$ "log-like")
                       (dist-log-prob $d$ $c$))
            $s$ (put $s$ "log-like" log-like)]
        (parent-handler ["observe" $s$ $\alpha$ $d$ $c$]))
      (parent-handler req))))
\end{hoppl}
Here the primitive \hop{(dist-log-prob $d$ $c$)} corresponds to the mathematical function $\textsc{log-prob}(d, c)$ that we have previously used when describing inference algorithms in this introduction.

We can now use this handler to extend the base handler by defining
\begin{hoppl}
(def handler (sum-log-observed base-handler))
\end{hoppl}
This handler ensures that a program call \hop{(p s0 y theta)} based on an initial state \hop{s0} will return a pair \hop{[s x]} that comprises an updated state (which will contain a key \hop{"log-prob"}) and a return value \hop{x}. We can now perform likelihood weighting by simply repeatedly calling the program
\begin{hoppl}
(defn likelihood-weighting
  [p y theta]
  (let [s0 {"log-like" 0.0}
        [s x] (p s0 y theta)
        log-w (get s "log-like")]
    [log-w x]))
\end{hoppl}

\paragraph{Handlers with Mutable State.} A drawback of handlers with immutable state is that we have to rewrite our program to track this state. While doing so makes it explicit where changes in state can occur, it is somewhat cumbersome to have to store an updated state variable for each call to \hop{sample} and \hop{observe}. In the context of a language like the HOPPL, we could implement a source code transformation to rewrite the program automatically. An alternative, which is more practical in the context of languages that are embedded in Python, is to make use of mutable state.

To illustrate how this would work, we will consider a handler that updates a \hop{state} variable in the form of a mutable hash map, which we initialize using a constructor \hop{mutable}. There is in principle nothing that prevents us from including mutable data structures in the HOPPL. We need to tread carefully when using mutable data structures in a probabilistic program, because such data structures could lead to sequences of samples that are not identically distributed, which means that the posterior is not well-defined. However, there is nothing intrinsically problematic with using mutable data types in the inference backend.

Making use of a global mutable state allows us to  simplify the handler \hop{sum-log-observed}, which now can once again be written without including the state as part of the request.
\begin{hoppl}
(def state (mutable {"log-like" 0.0}))
(defn sum-log-observed [parent-handler]
  (fn [req]
    (if (= (first req) "observe")
      (let [[_ $d$ $c$] (rest req)
            log-like (+ (get state "log-like")
                       (dist-log-prob $d$ $c$))]
        (put! state "log-like" log-like)
        (parent-handler req))
      (parent-handler req))))
\end{hoppl}
Here, the exclamation mark in the primitive \hop{put!} signifies that this function modifies the variable \hop{state} in place. This is in contrast its counterpart \hop{put}, which accepts an immutable hashmap as its input, and returns an updated copy of the input while leaving the original unaffected.

A consequence of using handlers with mutable state is that we now have to explicitly manage changes to the global state. In particular, we will need to reset the state every time we start a new program execution. To do so, we would modify the \hop{likelihood-weighting} procedure
\begin{hoppl}
(defn likelihood-weighting
  [p y theta]
  (let [(set! state {"log-like" 0.0})
         x (p y theta)
         log-w (get state "log-like")]
    [log-w x]))
\end{hoppl}
This type of explicit management of global state is not uncommon in the context deep learning frameworks. When training a network in PyTorch, for example, the user needs to call a method \py{optimizer.zero_grad()} to reset the gradient computation at each step of gradient descent.

\subsection{Implementing Importance Sampling with Handlers}
\label{sec:effect-is}

Now that we have described the basics of implementing algebraic effects and their handlers, we are in a position to return to the implementation of the the importance sampling procedure that we outlined in Section~\ref{sec:aux-vars}. In the preceding section, we implemented a handler that performs additional operations before calling the base handler. A nice property of this design is that we can compose handlers, which is to say that we can employ a series of handlers that each perform operations before calling the parent handler, until the base case is reached.

To illustrate this design, we will stick with the mutable implementation, and consider how we might implement the importance weighting calculation from the preceding section, which computes the weight in Equation~\eqref{eq:gradient-importance-weight-aux}:
\begin{align*}
    w =
    \prod_{y \in \text{dom}(\mathcal{Y}_p)} \mathcal{P}_p(y)
    \prod_{x \in \text{dom}(\mathcal{X}_p) \cap \text{dom}(\mathcal{X}_q)}
    \frac{\mathcal{P}_p(x)}{\mathcal{P}_q(x)}.
\end{align*}
To compute this weight, we to evaluate the proposal \hop{q} to generate a trace $\mathcal{X}_q$ of values and their corresponding log probabilities $\log \mathcal{P}_q$. Then, we will have to evaluate the target \hop{p} whilst conditioning on $\mathcal{X}_q$ to generate a trace $\mathcal{X}_p$ of values and correponding log probabilities $\log \mathcal{P}_p$. This computation needs to store the following quantities
\begin{itemize}
\item We need to store the log probabilities $\log \mathcal{P}_p(x)$ and $\log \mathcal{P}_q(x)$ for all unobserved random variables $x$ in $\X_p$ and $\X_q$ respectively.
\item For \hop{p} we additionally need to computer the sum of log probabilities for all observed variables $\sum_{\alpha \in \text{dom}(\Y_p)} \log \mathcal{P}_p(y)$.
\end{itemize}
To implement these operations, we define handlers \hop{store-trace} and \hop{store-log-probs} which store sampled values and their log probabilities respectively. Implementations of these functions are shown in Figure~\ref{fig:importance-handlers}. To compute the sum of the log observed probabilities, we can use the \hop{sum-log-observed} handler.

When we use these three handlers,  we can compute the log weight from Equation~\eqref{eq:gradient-importance-weight-aux} from a target state \hop{sp} and proposal state \hop{sq}
\begin{hoppl}
(defn log-weight [sp sq]
 (let [log-ps (get sp "log-probs")
       log-qs (get sq "log-probs")
       addrs (intersection (keys log-ps)
                           (keys log-qs))]
  (+ (get sp "log-like")
     (- (sum (map (fn [a] (get log-ps a)) addrs))
        (sum (map (fn [a] (get log-qs a)) addrs))))))
\end{hoppl}


We can now implement an importance sampling procedure that first evaluates the proposal \hop{q} and stores the resulting inference state in a variable \hop{sq}, then runs the target program \hop{p} and stores the resulting inference state in a variable \hop{sp}. From these two states we can now compute all variables that we will need to construct variational objectives: the log weight \hop{log-w}, the joint probability of all variables in the target \hop{log-p}, and the joint probability of all variables in the proposal \hop{log-q}.

\begin{figure}[!t]
\begin{hoppl}
(defn importance-sample [$y$ p $\eta$ $\theta$ q $\lambda$ $\phi$]
  (let [base-state {"log-probs" {},
                    "trace" {}}
        base-handler (store-trace
                       (store-log-probs
                         default-handler))]
    (set! state base-state)
    (set! handler base-handler)
    (let [_ (q $y$ $\lambda$ $\phi$)
          sq (imutable state)])
      (set! state (put base-state
                    "conditioned" (get sq "trace")))
      (set! handler (sum-log-observed
                       base-handler))
      (let [v (p $y$ $\eta$ $\theta$)
            sp (immutable state)
            log-w (log-weight sp sq)
            log-p (+ (get sp "log-like")
                   (sum (vals (get sp "log-probs"))))
            log-q (sum (vals (get sq "log-probs")))]
        [log-w log-p log-q v])))
\end{hoppl}
\caption{
\label{fig:handler-importance-sample}
An implementation of importance sampling using handlers. The inputs to this computation are the observed data $y$, a target program \hop{p} that is parameterized by a neural network $\eta$ with weights $\theta$, and a proposal program \hop{q} which is parameterized by a network $\lambda$ with weights $\phi$. The computation begins by evaluating the proposal \hop{q} whilst storing the log probability of every variable in the inference state under the key \hop{"log-probs"} and storing all sampled values under the key \hop{"trace"}. We store the resulting inference state in a variable \hop{sq}. The second step of the computation evaluates the target program \hop{p} using a handler that additionally computes and stores the log-likelihood and stores the resulting state in a variable \hop{sp}. We then use the states \hop{sp} and \hop{sq} to compute the log importance weight $\log w$, the log joint probability for the target $\log p(Y, X ; \theta)$, and the log joint probability for the proposal $\log q(X \mid Y ; \phi)$, which together can be used to compute objectives for amortized stochastic variational inference.}
\end{figure}

\begin{figure}[!t]
\begin{hoppl}
(defn store-trace [parent-handler]
  (fn [req]
    (if (= (first req) "sample")
      (let [[$\alpha$ $d$] (rest req)
            $c$ (parent-handler req)
            trace (put (get state "trace") $\alpha$ $c$)]
        (put! state "trace" trace)
        $c$)
      (parent-handler req))))

(defn store-log-probs [parent-handler]
  (fn [req]
    (if (or (= (first req) "sample")
            (= (first req) "observe"))
      (let [[$\alpha$ $d$] (rest req)
            $c$ (parent-handler req)
            log-probs (put (get state "log-probs")
                           $\alpha$ (dist-log-prob $d$ $c$)]
        (put! state "log-probs" log-probs)
        $c$)
      (parent-handler req))))

(defn reuse-conditioned [parent-handler]
  (fn [req]
    (if (= (first req) "sample")
      (let [[$\alpha$ $d$] (rest req)
            contitioned (get state "conditioned")
            c (if (contains? conditioned $\alpha$)
                (get conditioned $\alpha$)
                (parent-handler req))
            trace (put (get state "trace") $\alpha$ $c$)]
        (put! state "trace" trace)
        $c$)
      (parent-handler req))))
\end{hoppl}
\vspace{-0.5\baselineskip}
\caption{
\label{fig:importance-handlers}
Composable functions for importance sampling. The handler \hop{store-trace} stores each sampled value in a \hop{"trace"} entry in the global \hop{state}. The handler \hop{store-log-prob} stores the log probability of each sampled or observed variable in a \hop{"log-probs"} entry. Finally, the handler \hop{reuse-conditioned} conditioned evaluation on a proposal trace by reusing previously sampled values when possible, and sampling from the prior in other cases.}
\end{figure}

We show the resulting program in Figure~\ref{fig:handler-importance-sample}. In this program we begin by evaluating the proposal. To do so we define a \hop{base-handler} that composes handlers to stack operations. The outer handler \hop{store-trace} ensures that all sampled values are stored in the \hop{"trace"} key in the inference state. In the process of doing so, it will call the parent handler, which will store the log probability of every variable in a key \hop{"log-probs"}. The parent, in turn, will call its parent, which is the base case that generates samples from the prior. We use the primitive \hop{immutable} to create an immutable copy of the state \hop{sq}, since the global state variable will be mutated when it is reinitialized.

To perform the second step of the computation, we need to condition the program \hop{p} on a proposed trace. For this purpose we define a handler \hop{reuse-conditioned}, which is also shown in Figure~\ref{fig:importance-handlers}. This handler once again intercepts \hop{"sample"} requests, reuses values from a trace that is stored in a \hop{"conditioned"} entry when available, and calls the parent handler to generate a sample when no stored value is available. This results in a state \hop{sp}, which we once again store as an immutable copy. We then use these states to compute \hop{log-w}, \hop{log-p}, and \hop{log-q}.

We see that we can use composable handlers to implement fundamental inference operations such as storing traces, log probabilities, and conditioning execution. By implementing each of these operations as individual handlers, we can stack side-effects as needed during each part of an inference computation, resulting in code that is more modular than the monolithic algorithms for PyProb-style and WebPPL-style proposals that we discussed in Section~\ref{sec:pyprob-proposals} and Section~\ref{sec:webppl-proposals}.

\subsection{Implementing Stochastic Variational Inference}

Now that we understand how to perform importance sampling using standalone programs as proposals, let us put all the pieces together to see how we can implement stochastic variational inference methods. From a technical perspective, the main thing to understand in this context is that programs in deep probabilistic programming systems are often designed to support vectorization, which is to say that we can generate multiple samples in parallel.

To understand how vectorization affects the underlying implementation, we will briefly discuss the basics of vectorization and broadcasting. We will then discuss how to implement handlers that generate tensors of samples rather than scalars of samples. With these building blocks in place, we can then explain how to implement the outer optimization loop for stochastic variational inference.

\paragraph{Vectorization and Broadcasting} An important consideration for computational efficiency is to vectorize computations when possible. Deep learning frameworks provide first-class support for vectorization, which facilitates efficient parallelized execution on the GPU.

Primitive functions in deep learning frameworks typically accept tensor-valued arguments as inputs and apply the function to each element in the tensor to construct a tensor of outputs. As an example, we can define a tensor-valued distribution over multinomial variables
\begin{hoppl}
(multinomial (ones $K$ $L$)
             (/ (ones $K$ $L$ $D$) $D$))
\end{hoppl}
A sample from this distribution will be tensor of dimension $K \times L \times D$, which comprises $K \times L$ samples that each have dimension $D$.

A second property of primitive functions in deep learning libraries is that they support broadcasting, which is a mechanism for aligning tensors that differ in size. We will illustrate broadcasting using a simple example. Suppose that we would like to implement the following generative model
\begin{align}
    \theta_k
    &\sim
    \text{Dirichlet}(\alpha_{k,1}, \dots, \alpha_{k,D}),
    &
    k & = 1, \dots, K,
    \\
    z_{l,k}
    &\sim
    \text{Multinomial}(n_l, \theta_k)
    &
    k &= 1, \dots, K, ~l = 1, \dots, L.
\end{align}
This model defines Dirichlet-Multinomial distribution. We first generate a tensor $\theta$ that has size $K \times D$. For each of the $K$ vectors $\theta_k$, we then generate $L$ samples from a Multinomial distribution, each with a different count $n_l$, resulting in a tensor of size $K \times L \times D$.

In a language that supports vectorization and broadcasting, we could implement this generative model as follows
\begin{hoppl}
(defn dir-mult [$n$ $\alpha$]
  (let [[K D] (shape $\alpha$)
        [L] (shape $n$)
        $\theta$ (sample "theta" (dirichlet $\alpha$))
        $z$ (sample "z" (multinomial
                          $n$ (reshape $\alpha$ [K 1 D])))]
    [$\theta$ $z$]))
\end{hoppl}
In this model, we generate $\theta$ by passing a tensor $\alpha$ of size $K \times D$ to the Dirichlet distribution. We then pass $n$ and $\theta$ to the Multinomial distribution. Here we have on argument of size $L$, and one argument of size $K \times D$, which we have reshaped into a tensor of size $K \times 1 \times D$. The language implementation will now evaluate this expression as follows:
\begin{itemize}
    \item Replicate the argument $n$ to create a $K \times L$ tensor in which $n$ is repeated $K$ times along the first dimension.
    \item Replicate the argument $\theta$ to create a $K \times L \times D$ tensor in which $\theta$ is repeated $L$ times along the second dimension.
    \item Use the replicated arguments to construct a vectorized distribution object that will generate $K \times L$ samples that each have the form of a $D$-dimensional vector.
\end{itemize}
This form of automated replication of arguments is known as broadcasting, and it is supported in most mainstream libraries for tensor computation. Broadcasting applies two rules to align inputs
\begin{itemize}
\item[1.] When inputs have incompatible dimensions, broadcasting prepends dimensions of size 1 to the lower-dimensional input. In the example above, we have a 3-dimensional argument $\theta$, which means that the language expects a 2-dimensional argument $n$. This means that broadcasting will reshape $n$ from size $L$ to size $1 \times L$.
\item[2.] Once inputs have compatible dimensions, broadcasting repeats inputs along any dimensions of size 1 to expand them to the size of other arguments. In the example above, broadcasting repeats $n$ along the first dimension, and $\theta$ along the second dimension.
\end{itemize}


\paragraph{Generating Vectorized Batches of Samples.} We performing amortized inference, each step of stochastic gradient descent involves the following computations (see Section~\ref{sec:grad-variational})
\begin{itemize}
    \item[-] Sample a mini-batch $\{Y^1, \dots, Y^B\}$ from the training data.
    \item[-] For each item $Y^b$, generate proposals $\{\mathcal{X}^{1,b}, \dots, \mathcal{X}^{L,b}\}$.
    \item[-] Average over the samples to compute losses $\hat{L}_q(\theta)$ and $\hat{L}_q(\phi)$.
    \item[-] Update $\theta$ and $\phi$ by computing $\nabla_\theta \hat{L}_q(\theta)$ and $\hat{L}_q(\phi)$.
\end{itemize}


To implement this procedure, we would like to provide the programs \hop{p} and \hop{q} with a mini-batch of inputs, and implement handlers that generate $L$ vectorized samples for every input. Here there is a subtlety that we need to deal with: we need to distinguish between dimensions that correspond to different samples of a program, and dimensions of an individual sample. For this purpose, deep probabilistic programming libraries differentiate between a \emph{sample size} and an \emph{event size}.

In general, it can be ambiguous which dimensions in a tensor correspond to distinct samples from a model, and which dimensions correspond to distinct variables within a model. One way to ensure that we can distinguish between these two is to explicitly annotate distribution objects with a keyword argument \hop{"event-size"}. For the program \hop{p} that we have considered throughout this chapter, we would annotate distributions as follows
\begin{hoppl}
(defn p [$y$ $\eta$ $\theta$]
  (let [z (sample "z" (multinomial 1 $\theta^z$))
        v (sample "v" (normal ($\eta^v_{\mu}$ z $\theta^v$) ($\eta^v_{\sigma}$ z $\theta$)
                              "event-size" $D^v$))]
    (observe "y" (normal ($\eta^y$ v $\theta^y$) 1.0
                         "event-size" $D^y$) $y$)
    (return [z v])))
\end{hoppl}
In this program, we annotate the variable \hop{"v"} with an event size to indicate that it contains samples of size $D^v$. Similarly, we define a variable \hop{"y"} which we once again annotate with the event size $D^y$. We can similarly annotate the proposal with event sizes
\begin{hoppl}
(defn q [$y$ $\lambda$ $\phi$ $\sigma^y$]
  (let [u (sample "u" (normal $y$ $\sigma^y$ "event-size" $D^y$))
        v (sample "v" (normal ($\lambda^v_{\mu}$ u $\phi$) ($\lambda^v_{\sigma}$ u $\phi$)
                              "event-size" $D^v$))]
    (return v)))
\end{hoppl}
However, annotating variables with event sizes alone is not sufficient. If we provide an input $y$ of size $B \times D^y$ to \hop{q}, then the variable \hop{"u"} by will have size $B \times D^u$, whereas we would like it to have size $L \times B \times D^u$. In other words, the batch size of the inputs is $B$, whereas the batch size for latent should be $L \times B$. To ensure that, at inference time, we generate a batch of samples of a specified size, we can implement handler that intercepts \hop{"sample"} and \hop{"observe"} requests, and modifies the distribution object before calling the parent handler.
\begin{hoppl}
(defn set-sample-size [parent-handler]
  (fn [req]
    (if (or (= (get req 0) "sample")
            (= (get req 0) "observe"))
      (let [$d$ (get req 2)
            sample-size (concat
                          (get state "batch-size")
                          (event-size $d$))
            $\tilde{d}$ (dist-broadcast $d$ sample-size)]
        (parent-handler (put req 2 $\tilde{d}$))))
    (parent-handler req)))
\end{hoppl}
In this handler, we begin by retrieving the distribution object $d$ from a request of the form \hop{["sample" $\alpha$ $d$]} or \hop{["observe" $\alpha$ $d$ $c$]}. To modify this distribution object, we first retrieve a variable at key \hop{"batch-size"} from the global inference state and use a primitive \hop{(event-size $d$)} to determine the event size of the distribution object. We concatenate these two sizes and define a transformed distribution object $\tilde{d}$ using a primitive \hop{(dist-broadcast $d$ size)}, which broadcasts arguments to distribution objects to enforce the size of the resulting sample. Finally we replace the distribution $d$ with there broadcasted distribution object $\tilde{d}$ in the original request, and pass this request to the parent handler to continue the inference computation.

\begin{figure}[!t]
\begin{hoppl}
(defn importance-sample [$y$ p $\eta$ $\theta$ q $\lambda$ $\phi$ $L$]
  (let [[$B$ _] (shape $y$)
        base-state {"log-probs" {},
                    "trace" {},
                    "batch-size" [$L$ $B$]}
        base-handler (store-trace
                       (store-log-probs
                         (set-sample-size
                           default-handler)))]
    (set! state base-state)
    (set! handler base-handler)
    (q $y$ $\lambda$ $\phi$)
    (let [sq (imutable state)])
      (set! state (put base-state
                    "conditioned"
                    (get sq "trace")))
      (set! handler (sum-log-observed
                       base-handler))
      (let [v (p $y$ $\theta$)
            sp (immutable state)
            log-w (log-weight sp sq)
            log-p (+ (get sp "log-like")
                   (sum (vals (get sp "log-probs"))))
            log-q (sum (vals (get sq "log-probs")))]
        [log-w log-p log-q v])))
\end{hoppl}
\label{fig:importance-handlers-vec}
\caption{An adaption of the importance sampling algorithm from Figure~\ref{fig:importance-handlers} that makes use of the \hop{set-sample-size} handler to generate batches of samples.}
\end{figure}

Now that we have determined how to generate vectorized batches of $L \times B$ by evaluating \hop{q} and \hop{p}, we we can modify they importance sampling procedure that we defined in Section~\ref{sec:effect-is} to generate batches of samples, rather than a single sample. To do so, we need to make two small modifications; we have to use the \hop{set-sample-size} handler to ensure that we generate batches of the correct size and we have to initialize an entry \hop{"batch-size"} in the inference state.

In deep probabilistic programming systems that are embedded in Python, this type of vectorized generation of proposals tends to be much more efficient than performing $L \times B$ individual calls to a non-vectorized \hop{importance-sample} method, because it is more amenable to parallelized execution on GPUs.

However, there are of course limitations to this vectorization strategy. In particular, it is not always possible to vectorize programs that incorporate stochastic control flow. Deep learning frameworks provide solutions for vectorization of common use cases, such as dealing with input sequences that vary in length. Moreover, systems like PyTorch and JAX provide support for just-in-time compilation techniques that can deal with certain forms of control flow. This means that vectorization of probabilistic programs with dynamic support is often possible, but may special considerations in the design of the program in order to ensure compatibility with the numerical backend.

\paragraph{Computing Objectives for Variational Inference} We have arrived at the end of our discussion of amortized inference with standalone programs as proposals. We have seen how we can deal with auxiliary variables and missing variables, how we can implement a messaging interface using composable effect handlers, and how we can vectorize execution of a target program and its proposal. All that remains is to put the pieces together to compute variational objectives from samples.

Computing the losses is straightforward once we have an implementation of importance sampling in place. If we are computing  self-normalized estimators for the log marginal likelihood (Section~\ref{sec:importance-weighted-model-learning}) and the inclusive KL divergence (Section~\ref{sec:inclusive-kl}), then the gradient estimates that we would like to compute will have the form that we derived in Equation~\eqref{eq:iw-grad-log-py} and Equation~\eqref{eq:is-grad-inc-kl}
\begin{equation}
\begin{split}
\nabla_\theta \log p(Y ; \theta)
\simeq
\sum_{l=1}^L
\frac{w^l}{\sum_{l'=1}^L w^l} \:
\nabla_\theta \: \log p(Y,X^l ; \theta),
\\
\nabla_\lambda \KL{p(X|Y ; \theta)}{q(X;\lambda)}
\simeq
\sum_{l=1}^L
\frac{w^l}{\sum_{l'=1}^L w^l} \:
\nabla_\lambda \: \log q(X^l ; \lambda).
\end{split}
\end{equation}
In Section~\ref{sec:grad-variational}, we derived that these gradient estimates can be computed by aggregating samples into an objectives $\hat{L}_p(\theta)$ for the generative model (Equation~\eqref{eq:Lp-rww}) and an objective $\hat{L}_q(\phi)$ for the inference model (Equation~\eqref{eq:Lq-rww}),
\begin{align}
\label{eq:Lp-rww}
\hat{L}_p(\theta)
&=
-
\frac{1}{B}
\sum_{l,b}
\frac{\textsc{unbox}(w^{l,b})}
     {\sum_{l'} \textsc{unbox}(w^{l',b})}
\:
\log p(Y^{l,b}, X^{l,b} ; \theta)
,
\\
\hat{L}_q(\phi)
&=
-
\frac{1}{B}
\sum_{l,b}
\frac{\textsc{unbox}(w^{b,l})}
     {\sum_{l'} \textsc{unbox}(w^{l',b})}
\log q(X^{l,b} \mid Y^b \,; \phi)
.
\end{align}
We implemented this objective calculation for PyProb-style proposals in Section~\ref{sec:grad-variational}. To perform the analogous computation with standalone proposals, we simply need to reimplement lines 27-29 of Algorithm~\ref{alg:gradient-reweighted-pyprob} to compute these self-normalized objectives from the log weights \hop{log-w}, the log joint probability for the generative model \hop{log-p} and the log joint probability for the proposal \hop{log-q}
\begin{hoppl}
(defn reweighted-loss [log-w log-p log-q sample-dim]
  (let [w (unbox (exp (softmax log-w sample-dim)))
        loss-p (sum (* -1 (* w log-p)))
        loss-q (sum (* -1 (* w log-q)))
    [loss-p loss-q]))
\end{hoppl}
If we would like to learn parameters for the generative model and inference model by maximizing the ELBO (Section~\ref{sec:advanced-variational-inference}), then can use the construction introduced by~\citet{ritchie2016deep} to approximate gradients in models with a combination reparameterized and reparameterized variables (Section~\ref{sec:grad-elbo-ritchie}, Equation~\ref{eq:grad-elbo-mixed})
\begin{align}
    \nabla_\theta \mathcal{L}
    &=
    \mathbb{E}_{\tilde{q}(\tilde{X} \mid Y ; \phi)}
    \left[
        \nabla_\theta \log p(Y, \tilde{X}, \theta)
    \right],
    \\
    \begin{split}
    \nabla_\phi \mathcal{L}
    &=
    \mathbb{E}_{\tilde{q}(\tilde{X} \mid Y ; \phi)}
    \left[
        \nabla_\phi \log w(Y, \tilde{X}, \phi)
    \right]
    \\
    &\qquad
    +
    \mathbb{E}_{\tilde{q}(\tilde{X} \mid Y ; \phi)}
    \left[
        \big(\nabla_\phi
        \log \tilde{q}(\tilde{X} \mid Y ; \phi)\big)
        \:
        \log w(Y, \tilde{X}, \phi)
    \right]
    .
    \end{split}
\end{align}
In Section~\ref{sec:aevi-ritchie}, we defined a single objective
$\hat{L}_p = \hat{L}_q = - \hat{L}$ from which we can compute these gradient estimates using reverse-mode automatic differentiation (Equation~\eqref{eq:elbo-loss-ritchie}),
\begin{align}
  \hat{\mathcal{L}}
  =
  \frac{1}{L} \frac{1}{B} \sum_{l,b}
  \log \tilde{q}(X^{l,b} \mid Y^b ; \phi) \: \textsc{unbox}(\log w^{l,b})
  +
  \log w^{l,b}
  .
\end{align}
We implemented this calculation for WebPPL-style proposals in Algorithm~\ref{alg:gradient-elbo-loss-webppl}, where line 24 performs the computation that we need
\begin{hoppl}
(defn elbo-loss [log-w log-p log-q sample-dim]
  (let [elbo (mean (+ (* log-q (unbox log-w))
                      log w)]
    [(- elbo) (- elbo)]))
\end{hoppl}
The only additional changes that are needed to the sampling procedure is that the handler \hop{store-log-prob} should store the probability of a sample according to the base distribution $\tilde{d}_q$.

	\section{Combining Differentiable and Probabilistic Programming}
	\label{sec:deep-outlook}

This brings us to the end of our discussion of model learning and amortized inference in deep probabilistic programs. As we have seen throughout this chapter, the computation of importance weights is in some sense the fundamental operation in stochastic-gradient methods for learning and inference. This holds for standard variational inference methods, which maximize an ELBO, and for reweighted ``wake-wake'' methods, which maximize the marginal likelihood and minimize the inclusive KL divergence. We have seen that we can compute this importance weight using generic recurrent neural proposals that are defined in the inference backend, by integrating the proposal into the generative model, or more generally by using a standalone program as a proposal. 

At the start of this book, we contrasted deep learning and probabilistic programming as two fundamentally different approaches to artificial intelligence research. What we have seen in this chapter, is that both lines of thinking are in many ways complementary and can be combined to take advantage of their respective strengths. We can use the scalability of stochastic gradient descent and the flexibility of neural proposals to approximate a wide range of program posteriors. At the same time, specifying (deep) generative models as programs allows us to incorporate inductive biases into unsupervised learning problems. These inductive biases can be mild, e.g.~they can encode a hypothesis that at set of images clusters into distinct modes, or that a single image contains distinct objects. However, they can also encode much stronger hypotheses, based on our knowledge of the physics or dynamics of the underlying domain, as is often the case in applications in science and engineering. By combining these two lines of thinking, deep probabilistic programming defines a path towards model designs that strike a balance between overspecification and underspecification, and that can be tailored to specific application domains.



\chapter{Conclusion}


Having made it this far (congratulations!), we can now summarize probabilistic programming relatively concisely and conclude with a few general remarks.

Probabilistic programming is largely about designing languages, interpreters, and compilers that translate inference problems denoted in programming language syntax into formal mathematical objects that allow and accommodate generic probabilistic inference, particularly Bayesian inference and conditioning.  In the same way that techniques in traditional compilation are largely independent of both the syntax of the source language and the peculiarities of the target language or machine architecture, the probabilistic programming ideas and techniques that we have presented are largely independent of both the source language and the underlying inference algorithm. 

While some might argue that knowing how a compiler works is not really a requirement for being a good programmer, we suggest that this is precisely the kind of deep knowledge that distinguishes truly excellent developers from the rest.  Furthermore, as traditional compilation and evaluation infrastructure has been around, at the time of this writing, for over half a century, the level of sophistication and reliability of implementations underlying abstractions like garbage collection are sufficiently high that, indeed, perhaps one can be a successful user of such a system without understanding deeply how it works.  However, at this point in time, probabilistic programming systems have not developed to such a level of maturity and as such knowing something about how they are implemented will help even those people who only wish to develop and use probabilistic programs rather than develop languages and evaluators.  

It may be that this state of affairs in probabilistic programming remains for comparatively longer time because of the fundamental computational characteristic of inference relative to forward computation.  We have not discussed computational complexity at all in this text largely because there is, effectively, no point in doing so.  It is well known that inference in even discrete-only random variable graphical models is NP-hard if no restrictions (e.g. bounding the maximum clique size) are placed on the graphical model itself.  As the language designs we have discussed do not easily allow denoting or enforcing such restrictions, and, worse, allow continuous random variables, and in the case of HOPPLs, a potentially infinite collection of the same, inference is even harder.  This means that probabilistic programming evaluators have to be founded on approximate algorithms that work well some of the time and for some problem types, rather than in the traditional programming language setting where the usual case is that exact computation works most of the time even though it might be prohibitively slow on some inputs.  This is all to say that knowing intimately how a probabilistic programming system works will be, for the time being, necessary to be even a proficient power user.

These being early days in the development of probabilistic programming languages and systems means that there exist multiple opportunities to contribute to the foundational infrastructure, particularly on the approximate inference algorithm side of things.  While the correspondence between first-order probabilistic programming languages and graphical models  means that research to improve general-purpose inference algorithms for graphical models applies more-or-less directly to probabilistic programming systems, the same is not quite as true for HOPPLs.  The primary challenge in HOPPLs, the infinite-dimensional parameter space, is effectively unavoidable if one is to use a ``standard'' programming language as the model denotation language.  This opens challenges related to inference that have not yet been entirely resolved and suggests a research quest towards developing a truly general-purpose inference algorithm.  

In either case it should be clear at this point that not all inference algorithm research and development is equally useful in the probabilistic programming context.  In particular developing a special-purpose inference algorithm designed to work well for exactly one model is, from the programming languages perspective, like developing a compiler optimization for a single program -- not a good idea unless that one program is very important.  There are indeed individual models that are that important, but our experience suggests that the amount of time one might spend on an optimized inference algorithm will typically be more than the total time accumulated from writing a probabilistic program once, right away, a simply letting a potentially slower inference algorithm proceed towards convergence.  

Of course there also will be generations and iterations of probabilistic programming language designs with technical debt in terms of programs written accruing along with each successive iterate.  What we have highlighted is the inherent tension between flexible language design, the phase transition in model parameter count, and the difficulty of the associated underlying inference problem.  We have, as individual researchers, generally striven to make probabilistic programming work even with richly expressive modeling languages (i.e. ``regular'' programming languages) for two reasons.  One, the accrued technical debt of simulators written in traditional programming languages should be elegantly repurposable as generative models.  The other is simply aesthetic.  There is much to be said for avoiding the complications that come along with such a decision and this presents an interesting language design challenge: how to make the biggest, finite-variable cardinality language that allows natural model denotation, efficient forward calculation, and minimizes surprises about what will ``compile'' and what will not.  And if modeling language flexibility is desired, our thinking has been, why not use as much existing language design and infrastructure as possible?  

Our focus throughout this text has mostly been on automating inference in known and fixed models, and reporting state of the art techniques for such one-shot inference, however we believe that the challenge of model learning and rapid, approximate, repeated inference are both of paramount importance, particularly for artificial intelligence applications.  Our belief is that probabilistic programming techniques, and really more the practice of paying close attention to how language design choices impact both what the end user can do easily and what the evaluator can compute easily, should be considered throughout the evolution of the next toolchain for artificial intelligence operations.



%
%
%
%
%
%
%
%
%
%
%
%
%
%
%
%
%



\backmatter

\bibliography{pp-tutorial-paper}

\end{document}